\documentclass[times,twocolumn,final]{elsarticle}

\usepackage[margin=0.7in]{geometry}
\usepackage{framed,multirow}
\usepackage{booktabs}
\usepackage{tabularray}
\UseTblrLibrary{booktabs}
\usepackage{boldline} %$Bold lines for tables
\usepackage{fontawesome5}
\usepackage{amssymb,amsmath}
\usepackage{romannum}
\AtBeginDocument{\pagenumbering{arabic}}
\usepackage{pifont}
\newcommand{\xmark}{\ding{55}}
\usepackage{latexsym}
\usepackage{mathrsfs} 
\usepackage{soul}
\usepackage{url}
\usepackage[dvipsnames,table,xcdraw, svgnames,x11names]{xcolor}
\usepackage{makecell}

\usepackage{blindtext}
\usepackage{graphicx}

\usepackage[listings,skins,breakable]{tcolorbox}
\usepackage[colorlinks]{hyperref}
\usepackage[printonlyused,withpage]{acronym}
\usepackage[nameinlink]{cleveref}
\usepackage{caption}
\usepackage{subcaption}

\definecolor{newcolor}{rgb}{.8,.349,.1}
\definecolor{maroon}{cmyk}{0,0.87,0.68,0.32}
\definecolor{lightorange}{rgb}{1,0.753,0.478}

\graphicspath{{./images/}}  
\begin{document}

\begin{frontmatter}

\title{Advances in Medical Image Analysis with Vision Transformers: A Comprehensive Review}%
% \tnotetext[tnote1]{This project if funded by the LFB research grant, RWTH.}

\author{Reza Azad\textsuperscript{1}, Amirhossein Kazerouni\textsuperscript{2}, Moein Heidari\textsuperscript{2}, Ehsan Khodapanah Aghdam\textsuperscript{3}, Amirali Molaei\textsuperscript{4}, Yiwei Jia\textsuperscript{1}, Abin Jose\textsuperscript{1}, Rijo Roy\textsuperscript{1}, Dorit Merhof\textsuperscript{$\dagger$,5,6}\corref{cor1}}

\cortext[cor1]{Corresponding author: Dorit Merhof,$\:$%
  Tel.: +49 (941) 943-68509,$\:$% 
  E-mail: dorit.merhof@ur.de.}

\address[1]{Faculty of Electrical Engineering and Information Technology, RWTH Aachen University, Aachen, Germany}
\address[2]{School of Electrical Engineering, Iran University of Science and Technology, Tehran, Iran}
\address[3]{Department of Electrical Engineering, Shahid Beheshti University, Tehran, Iran}
\address[4]{School of Computer Engineering, Iran University of Science and Technology, Tehran, Iran}
\address[5]{Faculty of Informatics and Data Science, University of Regensburg, Regensburg, Germany}
\address[6]{Fraunhofer Institute for Digital Medicine MEVIS, Bremen, Germany}

% \received{1 May 2013}
% \finalform{10 May 2013}
% \accepted{13 May 2013}
% \availableonline{15 May 2013}
% \communicated{S. Sarkar}

\begin{abstract} The remarkable performance of the Transformer architecture in natural language processing has recently also triggered broad interest in Computer Vision. Among other merits, Transformers are witnessed as capable of learning long-range dependencies and spatial correlations, which is a clear advantage over convolutional neural networks (CNNs), which have been the de facto standard in Computer Vision problems so far. Thus, Transformers have become an integral part of modern medical image analysis. In this review, we provide an encyclopedic review of the applications of Transformers in medical imaging. Specifically, we present a systematic and thorough review of relevant recent Transformer literature for different medical image analysis tasks, including classification, segmentation, detection, registration, synthesis, and clinical report generation. For each of these applications, we investigate the novelty, strengths and weaknesses of the different proposed strategies and develop taxonomies highlighting key properties and contributions. Further, if applicable, we outline current benchmarks on different datasets. Finally, we summarize key challenges and discuss different future research directions. In addition, we have provided cited papers with their corresponding implementations in \url{https://github.com/mindflow-institue/Awesome-Transformer}.

\end{abstract}

\end{frontmatter}

%\linenumbers

%%%%%%%%% Introduction %%%%%%%%% 
\section{Introduction} \label{sec:intro}
\begin{figure*}[t]
 \centering
 \includegraphics[width=\textwidth]{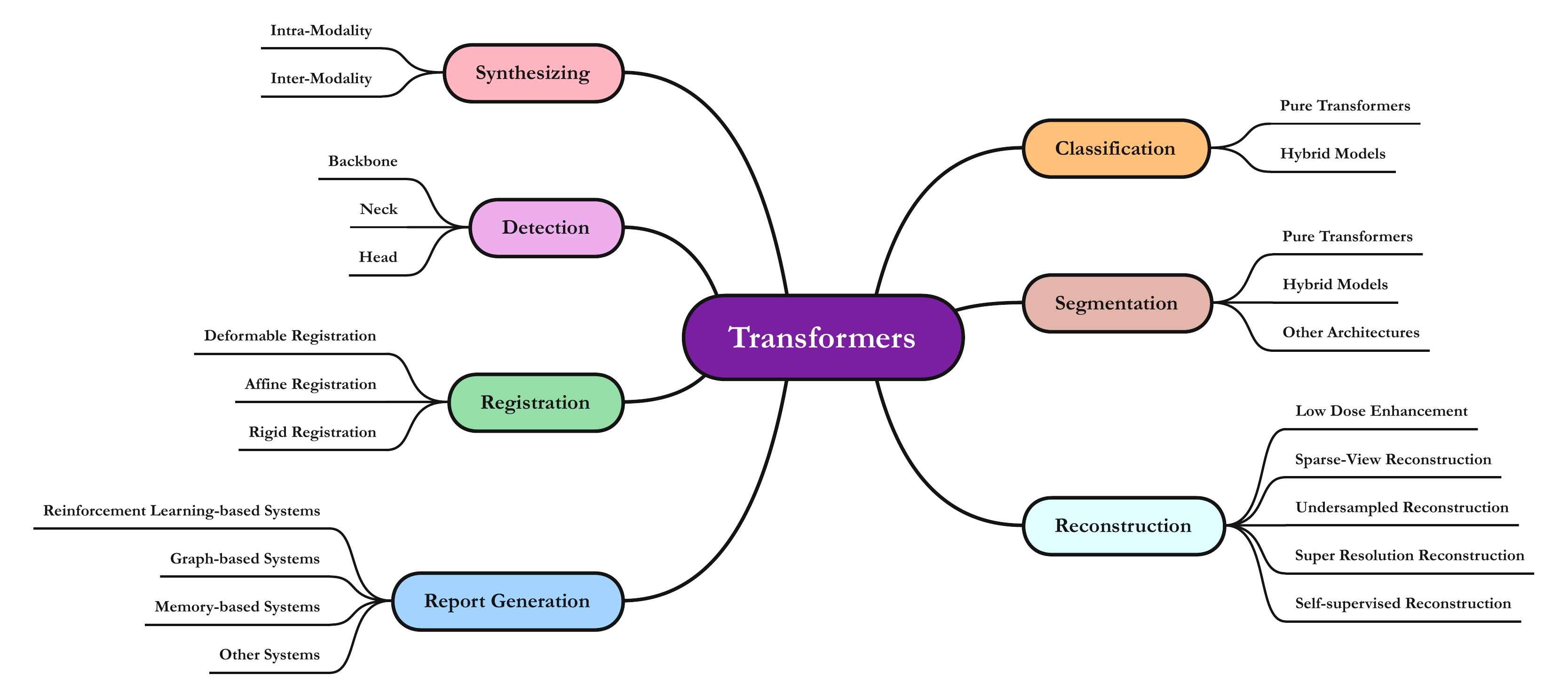}
 \caption{Overview of the applications covered in this review.}
 \label{fig:template}
\end{figure*}
Convolutional neural networks (CNNs) have been an integral part of research in the field of medical image analysis for many years. By virtue of convolutional filters whose primary function is to learn and extract necessary features from medical images, a wealth of research has been dedicated to CNNs ranging from tumor detection and classification \citep{arevalo2016representation}, detection of skin lesions \citep{karimijafarbigloo2023ms,azad2019bi} to brain tumor segmentation \citep{azad2022smu}, to name only a few. CNNs have also contributed significantly to the analysis of different imaging modalities in clinical medicine, including X-ray radiography, computed tomography (CT), magnetic resonance imaging (MRI), ultrasound (US), and digital pathology. Despite their outstanding performance, CNNs suffer from conceptual limitations and are innately unable  to model explicit long-distance dependencies due to the limited receptive field of convolution kernels. Moreover, the convolutional operator suffers from the fact that at inference time, it applies fixed weights regardless of any changes to the visual input. To mitigate the aforementioned problems, there have been great research efforts to integrate attention mechanisms, which can be regarded as a dynamic weight adjustment process based on input features to the seminal CNN-based structures to improve the non-local modeling capability \citep{ramachandran2019stand,bello2019attention,vaswani2021scaling}.

To this end, Wang et al. \citep{wang2018non} designed a non-local flexible building block, which can be plugged into multiple intermediate convolution layers. SENet \citep{hu2018squeeze} suggested a channel attention squeeze-and-excitation (SE) block, which collects global information in order to recalibrate each channel accordingly to create a more robust representation. Inspired by this line of research, there has been an overwhelming influx of models with attention variants proposed in the medical imaging field \citep{al2022procan,sang2021ag,yao2021claw}. Although these attention mechanisms allow the modeling of full image contextual information, as the computational complexity of these approaches typically grows quadratically with respect to spatial size, they imply an intensive computational burden, thus making them inefficient in the case of medical images that are dense in pixel resolution \citep{gonccalves2022survey}. Moreover, despite the fact that the combination of the attention mechanism with the convolutional operation leads to systematic performance gains, these models inevitably suffer from constraints in learning long-range interactions. Transformers \citep{vaswani2017attention} have demonstrated exemplary performance on a broad range of natural language processing (NLP) tasks, e.g., machine translation, text classification, and question answering. Inspired by the eminent success of Transformer architectures in the field of NLP, they have become a widely applied technique in modern Computer Vision (CV) models. Since the establishment of Vision-Transformers (ViTs) \citep{dosovitskiy2020image}, Transformers proved to be valid alternatives to CNNs in diverse tasks ranging from image recognition \citep{dosovitskiy2020image}, object detection \citep{zhu2021deformable}, image segmentation \citep{chen2021transunet} to video understanding \citep{arnab2021vivit} and image super-resolution \citep{chen2023activating}. As a central piece of the Transformer, the self-attention mechanism comes with the ability to model relationships between elements of a sequence, thereby learning long-range interactions \citep{azad2023foundational}. Moreover, Transformers allow for large-scale pre-training for specific downstream tasks and applications and are capable of dealing with variable-length inputs. The immense interest in Transformers has also spurred research into medical imaging applications (see \Cref{fig:template}). Being dominant in reputable top-tier medical imaging conferences and journals, it is extremely challenging for researchers and practitioners to keep up with the rate of innovation. The rapid adoption of Transformers in the medical imaging field necessitates a comprehensive summary and outlook, which is the main scope of this review. Specifically, this review provides a holistic overview of the Transformer models developed for medical imaging and image analysis applications. We provide a taxonomy of the network design, highlight the major strengths and deficiencies of the existing approaches, and introduce the current benchmarks in each task. We inspect several key technologies that arise from the various medical imaging applications, including medical image segmentation, medical image registration, medical image reconstruction, and medical image classification. So far, review papers related to Transformers do not concentrate on applications of Transformers in the medical imaging and image analysis domain \citep{khan2022Transformers}. The few literature reviews that do focus on the medical domain \citep{shamshad2022Transformers,he2022Transformers}, despite being very comprehensive, do not necessarily discuss the drawbacks and merits of each method. In our work, we explicitly cover this aspect and also provide a taxonomy that comprises the imaging modality, organ of interest, and type of training procedure each paper has selected. More specifically, in \Cref{sec:classification} (Medical Image Classification), we comprehensively elaborate on the most promising networks along with their key ideas, limitations, the number of parameters, and the specific classification task they are addressing. In \Cref{sec:segmentation} (Medical Image Segmentation), we analyze network architectures in terms of their design choice and propose a detailed taxonomy to categorize each network to provide insight for the reader to understand the current limitations and progress in segmentation networks based on the Transformer architecture. In \Cref{sec:reconstruction} (Medical Image Reconstruction), we take a different perspective to categorize networks based on their network structure and the imaging modality they are built upon. We categorize the synthesis methods in \Cref{sec:synthesize} based on their objective (intra-modality or inter-modality) and then provide detailed information regarding the network architecture, parameters, motivations, and highlights. In the sections related to detection (\Cref{sec:detection}), registration (\Cref{sec:registration}), and report generation (\Cref{sec:report}) we review in detail the state-of-the-art (SOTA) networks and provide detailed information regarding the network architectures, advantages, and drawbacks. Moreover, due to the swift development of the field, we believe that the community requires a more recent overview of the literature.

\begin{figure*}[!th]
 \centering
 \includegraphics[width=\textwidth]{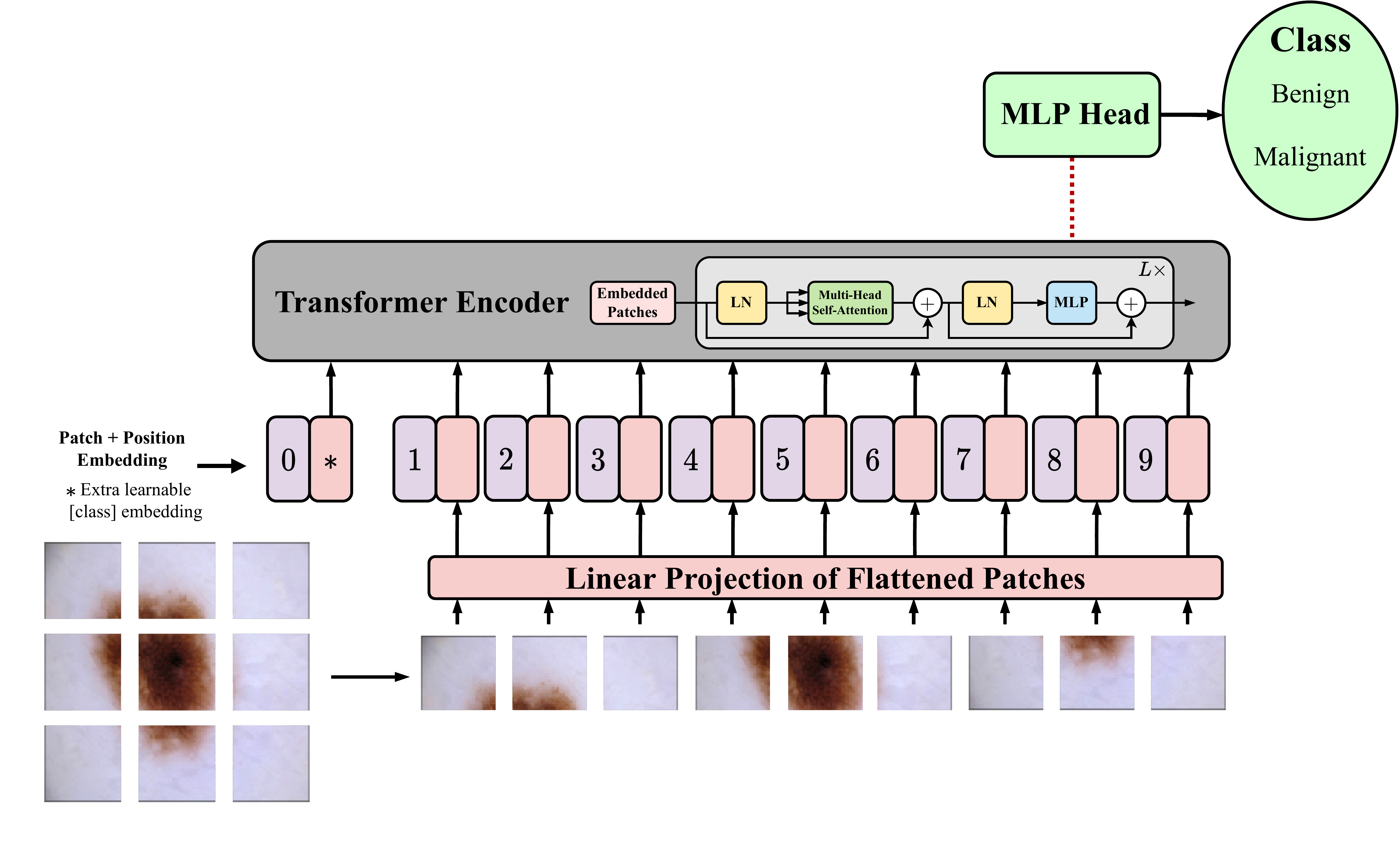}
 \caption{Architecture of the Vision Transformer as proposed in \citep{dosovitskiy2020image} and the detailed structure of the Vision Transformer encoder block.
In the Vision Transformer, sequential image patches are used as the input and processed using a
Transformer encoder to produce the final classification output.}
 \label{fig:vitbase}
\end{figure*}

We hope this work will point out new research options and  provide a guideline for researchers and initiate further interest in the vision community to leverage the potential of Transformer models in the medical domain.
Our major contributions are as follows:
\begin{itemize}

\item We systematically and comprehensively review the applications of Transformers in the medical imaging domain and provide a comparison and analysis of SOTA approaches for each task. Specifically, more than 200 papers are covered in a hierarchical and structured manner.

\item Our work provides a taxonomized (\Cref{fig:template}), in-depth analysis (e.g. task-specific research progress and limitations), as well as a discussion of various aspects.

\item Finally, We discuss challenges and open issues and also identify new trends, raise open questions and identify future directions.

\end{itemize}

\textbf{\emph{Search Strategy}}. We conducted a thorough search using DBLP, Google Scholar, and Arxiv Sanity Preserver, utilizing customized search queries that allowed us to obtain lists of scholarly publications. These publications included peer-reviewed journal papers, conference or workshop papers, non-peer-reviewed papers, and preprints. Our search queries consisted of keywords \texttt{(transformer* $|$ deep* $|$ medical* $|$ \textbf{\{Task\}}*), (transformer $|$ medical*), (transformer* $|$ medical* $|$ image* $|$ model*), (convolution* $|$ vision* $|$ transformer* $|$ medical*)}, where \textbf{\{Task\}} refers to one application covered in this review (see \Cref{fig:template}).
To ensure the selection of relevant papers, we meticulously evaluated their novelty, contribution, and significance, and prioritized those that were the first of their kind in the field of medical imaging. Following these criteria, we chose papers with the highest rankings for further examination. It is worth noting that our review may have excluded other significant papers in the field, but our goal was to provide a comprehensive overview of the most important and impactful ones.

\emph{\textbf{Paper Organizations.}}
The remaining sections of the paper are organized as
follows. In \Cref{sec:background}, we provide an overview of the key components of the well-established Transformer architecture and its clinical importance. Moreover, this section clarifies the categorization of neural network variants in terms of the position where the Transformer is located. \Cref{sec:classification} to \Cref{sec:report} comprehensively review the applications of Transformers in diverse medical imaging tasks as depicted in \Cref{fig:template}. For each task, we propose a taxonomy to characterize technical innovations and major use cases. \Cref{sec:challenges} presents open challenges and future perspectives of the field as a whole, while finally, \Cref{sec:discussion} concludes this work.

%%%%%%%%% Background %%%%%%%%% 
\section{Background}\label{sec:background}
In this section, we first provide an overview of the Transformer module and the key ideas behind its feasible design. Then, we outline a general taxonomy of Transformer-based models, characterized by their core techniques of using Transformers, i.e., whether they are purely Transformer-based, or whether the Transformer module is either used in the encoder, decoder, bottleneck, or skip connection, respectively.

\subsection{Transformers}
The original Transformer \citep{vaswani2017attention} was first applied to the task for machine translation as a new attention-driven building block. The vanilla Transformer consists of an encoder and a decoder, each of which is a stack of \emph{L} tandem of consecutive identical blocks. The Transformer module is convolutional-free and solely based on the self-attention mechanism or attention mechanism in short. Specifically, these attention blocks are neural network layers that relate different positions of a single sequence to compute the sequence’s representation. Since the establishment of Transformer models, they have attained remarkable performance in diverse natural language processing tasks \citep{kalyan2021ammus}. Inspired by this, Dosovitskiy et al. proposed the Vision Transformer (ViT) \citep{dosovitskiy2020image} model as illustrated in \Cref{fig:vitbase}. When trained on large datasets, for instance, JFT-300M, ViT outperforms the then state-of-the-art, namely ResNet-based models like BiT \citep{BiT}. In their approach, an image is turned into fixed-sized patches before being flattened into vectors. These vectors are then passed through a trainable linear projection layer that maps them into $N$ vectors with the dimensionality of $D$ $\times$ $N$ is the number of patches. The outputs of this stage are referred to as patch embeddings. To preserve the positional information present within each patch, they add positional embeddings to the patch embeddings. In addition to this, a trainable class embedding is also appended to the patch embeddings before going through the Transformer encoder. The Transformer encoder is comprised of multiple Transformer encoder blocks. There is one multi-head self-attention (MSA) block and an MLP block in each Transformer encoder block. The activations are first normalized using LayerNorm (LN) before going into these blocks in the Transformer encoder block. Furthermore, there are skip connections before the LN that add a copy of these activations to the corresponding MSA or MLP block outputs. In the end, there is an MLP block used as a classification head that maps the output to class predictions. The self-attention mechanism is a key defining characteristic of Transformer models. Hence, we start by introducing the core principle of the attention mechanism.
\subsubsection{Self-Attention}
In a self-attention layer (\Cref{fig:SA}), the input vector is firstly transformed into three separate vectors, i.e., the query vector \emph{q}, the key vector \emph{k}, and the value vector \emph{v} with a fixed dimension. These vectors are then packed together into three different weight matrices, namely $W^{Q}$, $W^{K}$, and $W^{V}$. A common form of $Q$, $K$, and $V$ can be formulated as \Cref{eq: eq1} for an input $X$
\begin{equation}
K=W^K X, Q=W^Q X, V=W^V X, \label{eq: eq1}
\end{equation}
where $W^{K}$, $W^{Q}$, and $W^{V}$ refers to the learnable parameters. The scaled dot-product attention mechanism is then formulated as
\begin{equation}
\text{Attention}(Q, K, V)=\text{Softmax}\left(\frac{Q K^T}{\sqrt{d_k}}\right) V,
\end{equation}
where $\sqrt{d_k}$ is a scaling factor, and a softmax operation is applied to the generated attention weights to translate them into a normalized distribution.
\subsubsection{Multi-Head Self-Attention}
\begin{figure}[h]
		\centering
		\begin{subfigure}[t]{0.47\columnwidth}
			\includegraphics[width=\textwidth]{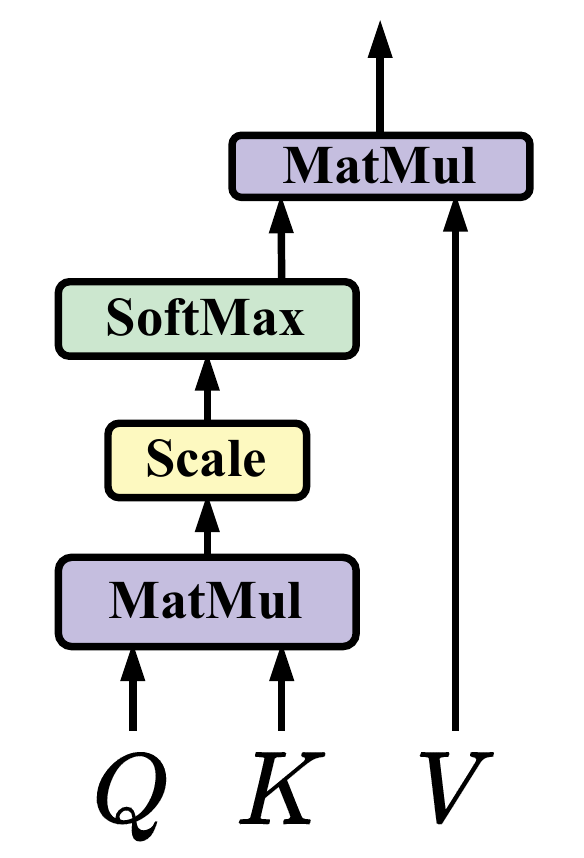}
			\caption{Self-Attention}
			\label{fig:SA}
		\end{subfigure}
	\hfill
		\begin{subfigure}[t]{0.47\columnwidth}
			\includegraphics[width=\textwidth]{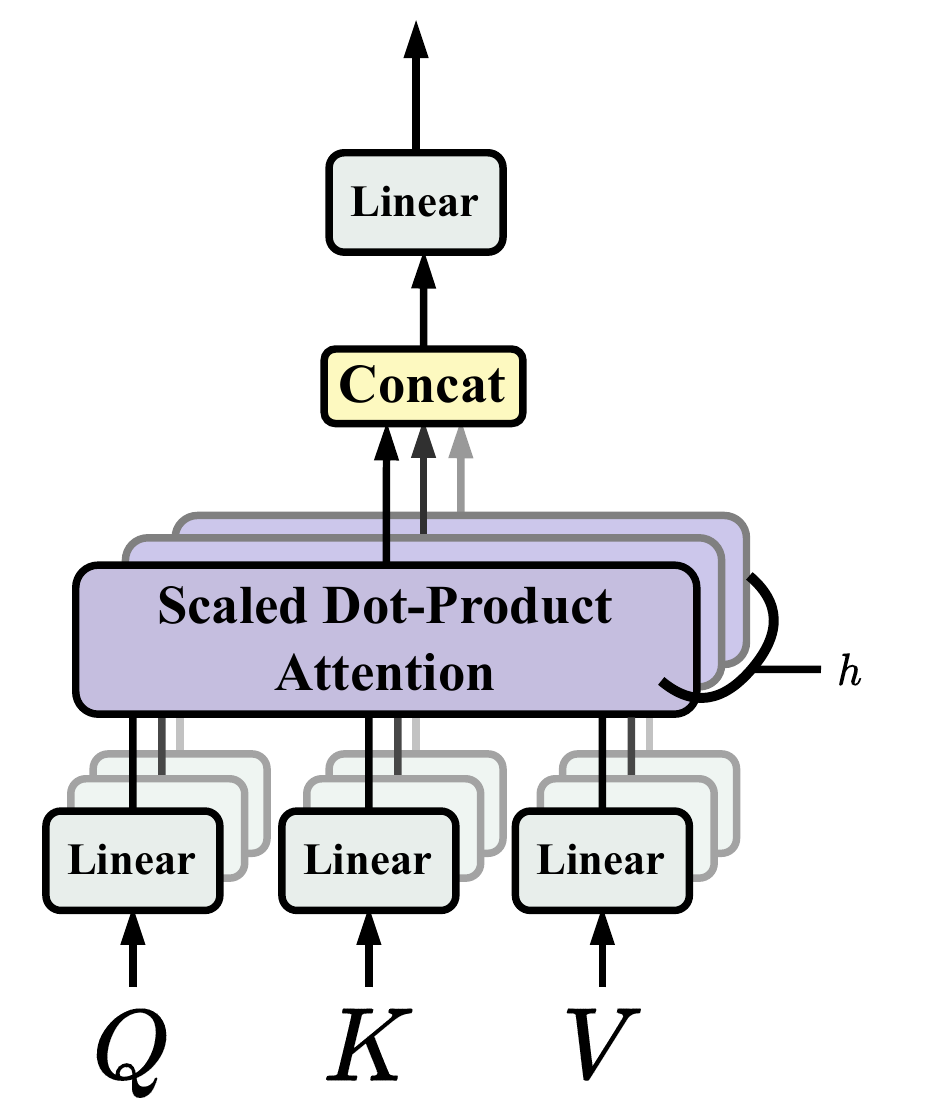}
			\caption{Multi-Head Self-Attention}
			\label{fig:MHSA}
		\end{subfigure}
		\caption{(\subref{fig:SA}) The process of self-attention. (\subref{fig:MHSA}) Multi-head attention. The MSA consists of multiple SA blocks (heads) concatenated together channel-wise as proposed in \citep{vaswani2017attention}.}
		\label{fig:SAMSA}
\end{figure}

The multi-head self-attention (MHSA) mechanism (\Cref{fig:MHSA}) has been proposed \citep{vaswani2017attention} to model the complex relationships of token entities from different aspects. Specifically, the MHSA block helps the model to jointly attend to information from multiple representation sub-spaces, as the modeling capability of the single-head attention block is quite coarse.
The process of MHSA can be formulated as
\begin{align}
\text {MultiHead} (Q, K, V)=\text [{Concat}\left(\text {head}_1, \ldots, \text {head}_h\right)] W^O,
\end{align}
where $\text{head}_i=$ $\operatorname{Attention}\left(Q W_i^Q, K W_i^K, V W_i^V\right)$, and $W^O$ indicates a linear mapping function to combine multi-head representation. Note that $h$ is a hyper-parameter set to $h = 8$ in the original paper.

\subsection{Transformer modes}
While the Transformer was originally introduced with an encoder-decoder pipeline, many
modern architectures generally exploit the Transformer architecture in different fashions, which generally depend on the target application. The usage of Transformers in vision tasks can broadly be classified into pure and hybrid designs.
% five groups, namely: (1) model-encoder, (2)
% model-decoder, (3) bottleneck of a model (e.g., for machine translation), (4) skip-connection of the encoding-decoding criteria, and (5) encoder-decoder (pure Transformer).

\subsubsection{Pure Transformers} 
Due to the deficiency of CNN-based architectures in learning global and long-range semantic information interactions, which stems from the locality of convolution operation, a cohort study has investigated the purely Transformer-based models without any convolution layer. These models usually consist of encoder, bottleneck, decoder, and skip
connections directly built upon the ViT or its variants. In this criteria, there are usually multiple multi-head self-attention modules in both encoding and decoding sections that allow the decoder to utilize information from the encoder. Examples of such methods are the Swin-Unet \citep{cao2021swin} and the TransDeepLab \citep{azad2022transdeeplab} networks which, as their name suggests, try to model the seminal U-Net \citep{ronneberger2015unet}, and DeepLab \citep{chen2018encoder} architectures.

\subsubsection{Transformer: Hybrid}
The hybrid Transformer models usually modify the base CNN structure by replacing the encoder or decoder modules.

\textbf{Encoder}: Encoder-only models such as the seminal BERT \citep{devlin2018bert} are designed to make a single prediction per input or a single prediction for an entire input sequence. In the computer vision era, these models are applicable for classification tasks. Moreover, as utilizing a pure Transformer can result in limited localization capacity stemming from inadequate low-level features, many cohort studies try to combine CNN and Transformer in the encoding section \citep{chen2021transunet}. Such a design can enhance finer details by recovering localized spatial information.
 
\textbf{Decoder:}
Transformers can also be used in a decoding fashion. Such a causal model is typically used for generation tasks such as language modeling. Besides that, the modification can apply to the skip connections of the decoder module. Skip connection is a widely used technique to improve the performance and the convergence of deep neural networks. It can also serve as a modulating mechanism between the encoder and the decoder. To effectively provide low-level spatial information for the decoding path, the idea of exploiting Transformers in designing skip connections has emerged. This notable idea can lead to finer feature fusion and recalibration while guaranteeing the aggregation scheme of using both high-level and low-level features \citep{chen2021transunet,heidari2022hiformer}.

\subsection{Clinical Importance}
In practice, medical professionals conduct medical image analysis qualitatively in real-world scenarios. This can lead to different understandings and levels of precision due to variations in the reader's expertise or differences in image quality, as well as being timely and labor-expensive. Therefore, deep learning techniques have gained extensive attraction in medical image analysis aiming to reduce inter-reader variation and decrease the expenses associated with time and workforce \citep{shamshad2022Transformers}. 

The fundamental question that emerges is \textbf{What are the motivations behind using Transformers in the medical domain?}

Advances in adversarial attacks create an unavoidable danger in which potential attackers might attempt to gain benefits by exploiting vulnerabilities in the healthcare system. While there is substantial literature available concerning the robustness of CNNs in the field of medical imaging, it is still a challenging direction to explore. Recent studies \citep{benz2021adversarial} prove that ViTs are more robust to adversarial attacks than CNNs. Specifically, ViT is significantly more robust than CNN in a wide range of white-box attacks. A similar trend is also observed in the query-based and transfer-based black-box attacks which can be attributed to the fact that ViTs are more reliant on low-frequency (robust) features while CNNs are more sensitive to high-frequency features.

Besides, due to strict privacy rules, low occurrence rates of certain diseases, concerns about data ownership, and a limited number of patients, providing suitable data have always been a paramount concern in the medical domain. Recently, the idea of exploiting the inherent structure of ViT in distributed medical imaging applications has emerged.  Owning to the inherent structure of ViT (as opposed to CNN), it can be split into shared body and task-specific heads which demonstrate that a ViT body with sufficient capacity can be shared across relevant tasks by leveraging different strategies such as multi-task learning \citep{park2021federated}. This way, ViTs can be a paramount application in decentralized medical imaging solutions.

Moreover, Transformers excel at capturing long-range dependencies in data, making them well-suited for analyzing complex medical images where contextual information plays a crucial role. 
Transformers offer greater interpretability than CNNs, as they provide attention maps that highlight the regions of an image that contribute most to the model's decision-making. This transparency is crucial in medical settings where understanding the reasoning behind a diagnosis is important.
The flexible design of ViTs can be used on edge devices to speed diagnosis, improve clinician workflows, enhance patient privacy, and save time. Below, we provide recent examples of clinical use cases of Transformers in the medical domain. 

 \textbf{Med-PaLM 2} \citep{singhal2023towards}, a Transformer-based model, by \textit{Google} demonstrated a feasible comparison between physicians and the application of Transformers in the medical domain questioning. The answers provided by Med-PaLM 2 are pretty impressive, and the authors provided a metric to assess model performance by exposing the clinician answers and the model's output to other physicians and lay-persons to rate them in nine categories such as factuality, medical reasoning capability, etc. As a result, the quantitative results endorse that 72.9\% of the time, answers provided by Med-PaLM 2 are preferable regarding reporting the medical consensus.

Seenivasan et al. \citep{seenivasan2023surgicalgpt} addressed the knowledge transfer's steep learning curve from surgical experts to medical trainees and patients and proposed the \textbf{SurgicalGPT}, a Transformer-based multi-modal visual question-answering framework that utilizes the large language models with visual cues. SurgicalGPT demonstrated outperforming results over several (robotic) surgical datasets in terms of accuracy compared to the uni-modality text generation models.

Consequently, the clinical use cases of Med-PaLM 2 and SurgicalGPT have demonstrated the feasibility and benefits of employing Transformers in the medical domain, showing their effectiveness and potential applications.

%%%%%% Application %%%%%%
%%%%%%%%% Classification %%%%%%%%% 
\begin{figure*}[t]
 \centering
 \includegraphics[width=0.99\textwidth]{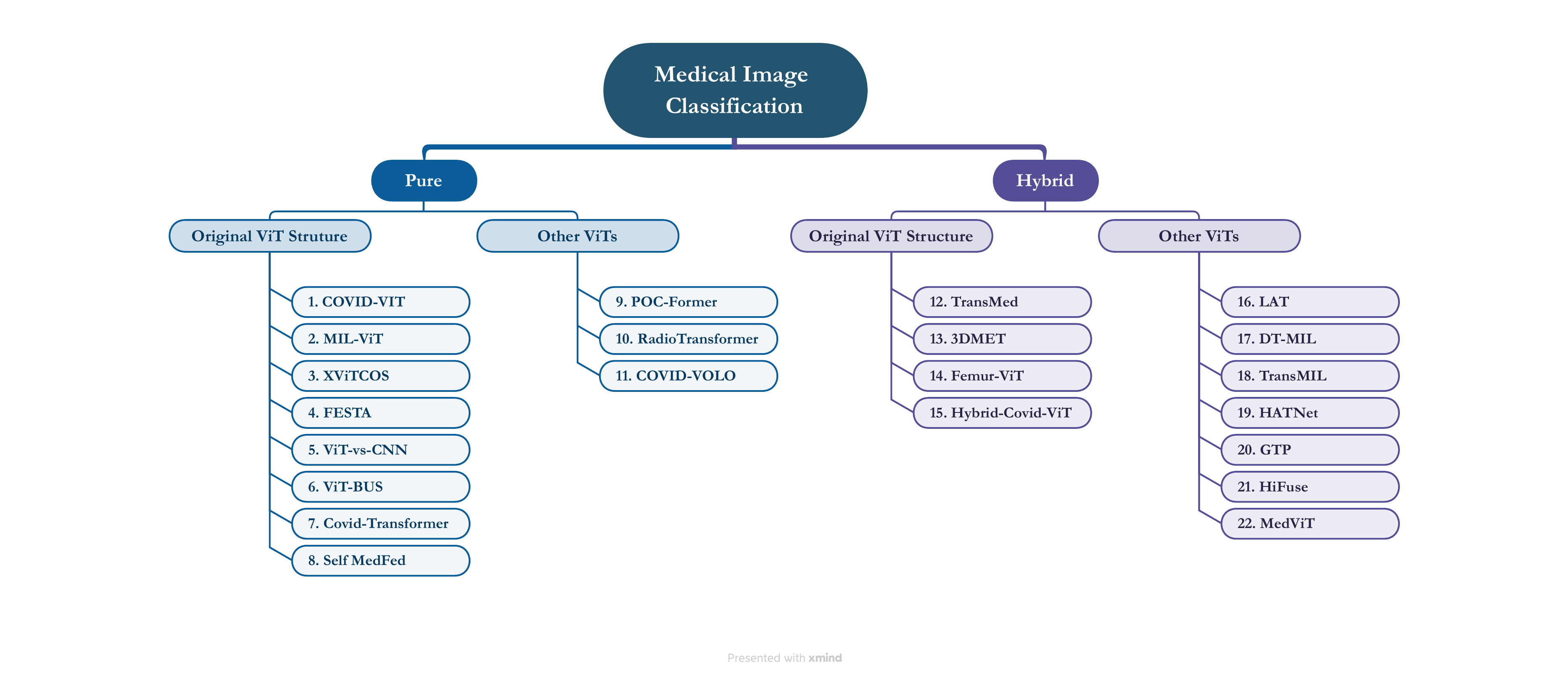}
 \caption{Taxonomy of ViT-based approaches in medical image classification. Methods are categorized based on their proposed architecture into pure and hybrid methods, in which they adopt the vanilla ViT or present a new type of vision Transformer in medical image classification. Notably, we utilize the prefix numbers in the paper’s name in ascending order and denote the reference for each study as follows: 1. \citep{gao2021covid}, 2. \citep{yu2021mil}, 3. \citep{mondal2021xvitcos}, 4. \citep{park2021federated}, 5. \citep{matsoukas2021time}, 6. \citep{gheflati2022vision}, 7. \citep{shome2021covid}, 8. \citep{yan2023label}, 9. \citep{perera2021pocformer}, 10. \citep{radioTransformer}, 11. \citep{liu2021automatic}, 12. \citep{transmed}, 13. \citep{wang20213dmet}, 14. \citep{tanzi2022vision}, 15. \citep{park2021vision}, 16. \citep{sun2021lesion}, 17. \citep{li2021dt}, 18. \citep{transmil}, 19. \citep{mehta2022end}, 20. \citep{zheng2022graph}, 21. \citep{huo2022hifuse}, 22.\citep{manzari2023medvit}.}
 \label{fig:classification}
\end{figure*}

\section{Medical Image Classification} \label{sec:classification}
Image classification is still one of the challenging problems in computer vision, which aids in segregating extensive quantities of data into meaningful categories \citep{aminimehr2023entri}. Vision Transformers (ViT) have recently demonstrated outstanding results in various image classification tasks and offer significant advantages over conventional CNNs \citep{liu2021swin,xia2022vision,fayyaz2021ats,dong2022cswin,li2022sepvit,yao2022dual}. These advantages include long-range relationships, adaptive modeling, and attention maps that yield intuition on what the model deems more important inside an image \citep{matsoukas2021time}. Due to these alluring advantages, there is rising interest in building Transformer-based models for medical image classification. Therefore, highly precise classification is becoming increasingly vital for facilitating clinical care.

In this section, we exhaustively examine ViTs in medical image classification. As illustrated in \Cref{fig:classification}, we have broadly classified these methods based on the role ViT plays in their architecture. These categories include pure Transformers and Hybrid Models. Generally, a vision Transformer-based classification architecture consists of three modules: (1) a backbone for capturing input features, (2) an encoder for modeling the information, and (3) a classification head for generating output based on the specified task. Therefore, the Transformer can be adopted in each module. However, some works, including Lesion Aware Transformer (LAT) \citep{sun2021lesion} and Deformable Transformer for Multi-Instance Learning (DT-MIL) \citep{li2021dt}, take a different approach and utilize encoder-decoder structures. LAT proposes a unified encoder-decoder system for Diabetic Retinopathy (DR) grading, and DT-MIL introduces a Transformer-based encoder-decoder architecture for classifying histopathological images, where the deformable Transformer was embraced for the encoder part. In the following, we will go into great depth on both hybrid and pure models.

\subsection{Pure Transformers}

Since the emergence of Transformers, there has been a growing debate regarding whether it is time to switch entirely from CNNs to Transformers. Matsoukas et al. \citep{matsoukas2021time} conduct a series of experiments to answer this critical question. They take ResNet50 \citep{resnet} and the DeiT-S \citep{DeiT} models to represent CNN and ViT models, respectively. They train each of these two models in 3 different fashions: a) randomly initialized weights, b) pre-trained on ImageNet (transfer learning), and c) pre-training on the target dataset in a self-supervised learning (SSL) scheme using DINO \citep{caron2021emerging}. Their findings show that when utilizing random initialization, ViTs are inferior to CNNs. In the case of transfer learning, the results are similar for both models, with ViT being superior for two out of three datasets. Additionally, ViT performs better when self-supervision on the target data is applied. They conclude that Vision Transformers, indeed, are suitable replacements for CNNs.

Transformers have had a profound effect on medical development. Researchers have thoroughly investigated adopting the ViT in medical image classification tasks since its introduction. However, the limited number of medical images has hindered Transformers from replicating their success in medical image classification. \textbf{ViT-BUS} \citep{gheflati2022vision} studies the use of ViTs in medical ultrasound (US) image classification for the first time. They propose to transfer pre-trained ViT models based on the breast US dataset to compensate for the data-hunger of ViTs. Evaluated results on B \citep{yap2017automated}, BUSI \citep{al2020dataset}, and B+BUSI datasets indicate the predominance of attention-based ViT models over CNNs on US datasets. Likewise, \textbf{COVID-Transformer}~\citep{shome2021covid} utilizes ViT‑L/16 to detect COVID from Non-COVID based on CXR images. Due to the limitation of sufficient data, they introduce a balanced dataset containing 30K chest X-ray images for multi-class classification and 20K images for binary classification. The published dataset is created by merging datasets~\citep{qi2021chest}, \citep{el2020extensive}, and \citep{sait2020curated}. They fine-tune the model on the dataset with a custom MLP block on top of ViT to classify chest X-ray (CXR) images. Moreover, COVID-Transformer exploits the GradCAM Map~\citep{selvaraju2017grad} to visualize affected lung areas that are significant for disease prediction and progression to display the model interpretability. Similarly, Mondal et al.~\citep{mondal2021xvitcos} present \textbf{xViTCOS} for detecting COVID-19 in CTs and CXRs. xViTCOS employs a model that has been pre-trained on ImageNet-21k~\citep{deng2009large}. Nevertheless, the training data capacity might overshadow the generalization performance of the pre-trained ViT to transfer the knowledge from the learned domain to the target domain. By training the model on the COVIDx-CT-2A dataset \citep{gunraj2021covidx}, a moderately-sized dataset, xViTCOS overcomes this problem. However, due to the shortage of the insufficient amount of CXR images, the pre-trained ViT model is fine-tuned using the CheXpert dataset \citep{irvin2019chexpert}. In addition, xViTCOS leverages the Gradient Attention Rollout algorithm \citep{chefer2021Transformer} to visually demonstrate the model's prediction on the input image for clinically interpretable and explainable visualization.
% (see \Cref{fig:XViTCOS-results}). 
In experiments using COVID CT-2A and their custom-collected Chest X-ray dataset, xViTCOS significantly outperforms conventional COVID-19 detection approaches. \textbf{MIL-VT} \citep{yu2021mil} similarly suggests pre-training the Transformer on a fundus image large dataset beforehand, initialized by the pre-trained weight of ImageNet, then fine-tuning it on the downstream retinal disease classification task in order to encourage the model to learn global information and achieve generalization. Unlike previous approaches, they apply some modifications to the vanilla ViT structure. In the classic ViT, embedded features are neglected for classification; instead, only the class token, which retains the summarization of embedded features' information, is used. Yu et al. \citep{yu2021mil} propose a novel multiple-instance learning (MIL)-head module to exploit those embedded features to complement the class token. This head comprises three submodules that attach to the ViT in a plug-and-play manner: 1) the MIL embedding submodule that maps the feature embeddings to a low-dimensional embedding vector, 2) the attention aggregation submodule that outputs a spatial weight matrix for the low-dimensional patch embeddings; this weight matrix is then applied to the low-dimensional embeddings to ascertain each instance's importance, 3) the MIL classifier submodule that determines the probability of each class through aggregated features. In the downstream task, both MLP and MIL heads use the weighted cross-entropy loss function for training. The outputs of both heads are then weight-averaged for the inference time. Results indicate the effectiveness of the proposed training strategy and the MIL-head module by dramatically boosting the performance over APTOS2019 \citep{APTOS2019} and RFMiD2020 \citep{pachade2021retinal} datasets when compared to CNN-based baselines. In contrast to the previous 2D-based methods that employ transfer learning, \textbf{COVID-ViT} \citep{gao2021covid} proposes training ViT to classify COVID and non-COVID cases using 3D CT lung images. Given that a COVID volume may contain non-COVID 2D slices, COVID-ViT applies a slice voting mechanism after the ViT classification result in which the subject is categorized as having COVID if more than a certain percentage of slices (e.g., $25\%$) are predicted to be COVID. The findings reported for the MIA-COVID19 competition \citep{kollias2021mia} confirm that ViT outperforms CNN-based approaches such as DenseNet \citep{iandola2014densenet} in identifying COVID from CT images.

% \begin{figure}[h]
%  \centering
%  \includegraphics[width=0.48\textwidth]{images/XViTCOS_results.pdf}
%  \caption{xViTCOS results on CT (a) and X-ray (b) images for Normal, Pneumonia, and COVID-19 classes alongside their ground truth and saliency maps \citep{mondal2021xvitcos}.}
%  \label{fig:XViTCOS-results}
% \end{figure}

Besides the remarkable accuracy of Transformers compared to CNNs, one of their major drawbacks is their high computational cost, thereby making them less effective for real-world applications, such as detecting COVID-19 in real time. In light of the prevalence of COVID-19, rapid diagnosis will be beneficial for starting the proper course of medical treatment. CXR and lung CT scans are the most common imaging techniques employed. However, CT imaging is a time-consuming process, and using CXR images is unreliable in identifying COVID-19 in the early stage. In addition, vision Transformers are computationally expensive to deploy on mobile devices for real-time COVID-19 classification. Therefore, Perera et al. \citep{perera2021pocformer} present a lightweight \textbf{Point-of-Care Transformer (POCFormer)}. The compactness of POCFormer allows for real-time diagnosis of COVID-19 utilizing commercially accessible POC ultrasound devices. POCFormer reduces the complexity of the vanilla ViT self-attention mechanism from quadratic to linear using Linformer \citep{wang2020linformer}. The results display the superiority of POCFormer in the real-time detection of COVID-19 over the CNN-based SOTAs on the POCUS dataset \citep{born2020pocovid}.

In addition, despite the great potential shown by ViTs in ImageNet classification, their performance is still lower than the latest SOTA CNNs without additional data. These Transformers mainly focus on a coarse level by adopting a self-attention mechanism to establish global dependency between input tokens. However, relying only on a coarse level restricts the Transformer's ability to achieve higher performance. Thus, Liu et al. \citep{liu2021automatic} leverage a pre-trained version of \textbf{VOLO} for an X-ray COVID-19 classification. VOLO \citep{yuan2022volo} first encodes fine-level information into the token representations through proposed outlook attention, alleviating the limitations of Transformers that require a large amount of data for training, and second aggregates the global features via self-attention at the coarse level. Through the outlook attention mechanism, VOLO dynamically combines fine-level features by treating each spatial coordinate $(i,j)$ as the center of a $K\times K$ local window and calculating its similarity with all its neighbors. The findings indicate that fine-tuning VOLO on Dataset-1 \citep{chowdhury2020can} leads to $99.67\%$ top1 accuracy on Dataset-1 test cases and $98.98\%$ top1 accuracy on unseen Dataset-2 \citep{cohen2020covid}, which demonstrates the generality of the approach.

\begin{figure}[h]
 \centering
 \includegraphics[width=0.48\textwidth]{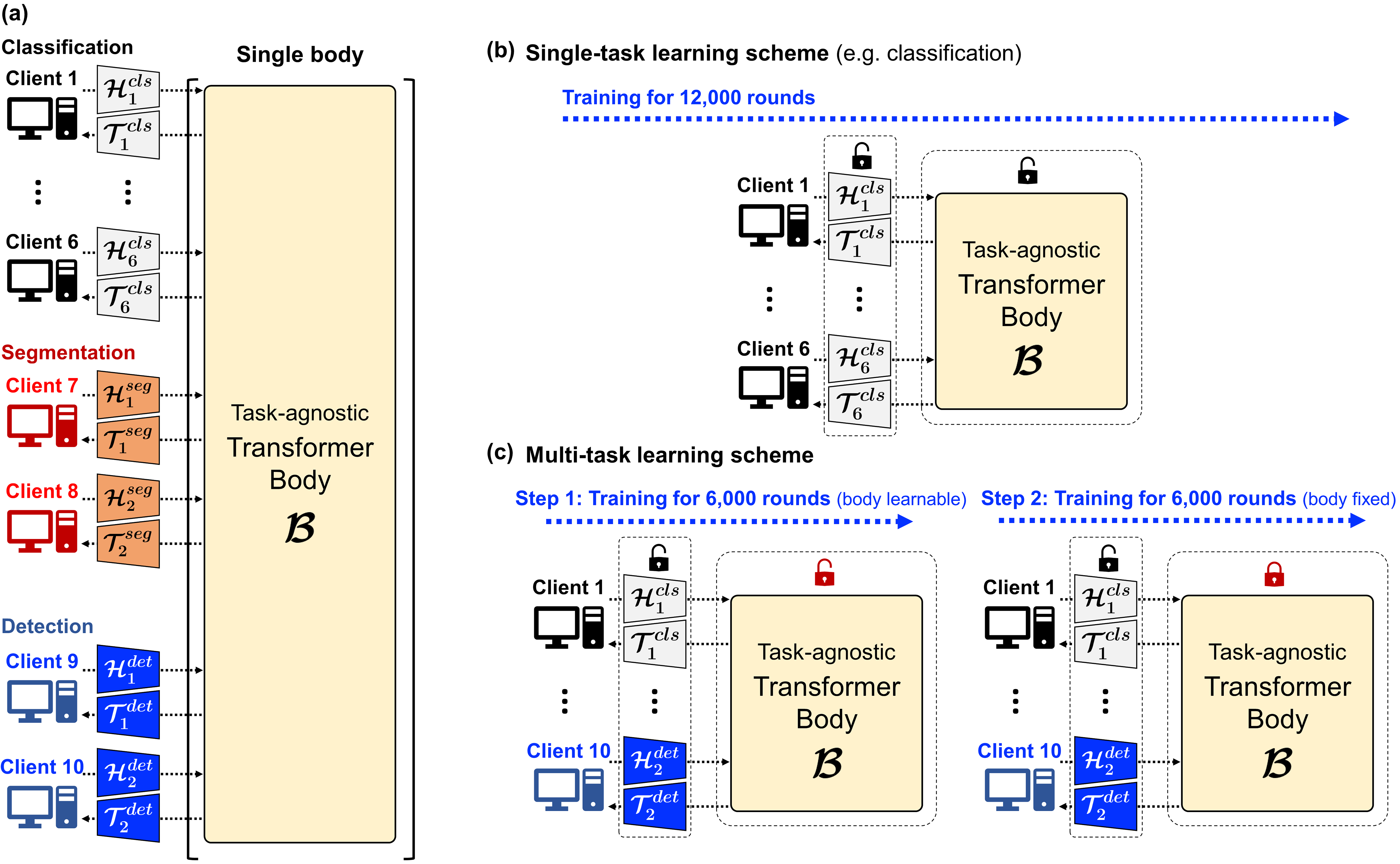}
 \caption{ Overview of the FESTA framework \citep{park2021federated}, which utilizes ViT for multi-task learning of COVID-19 CXR classification, detection, and segmentation. (a) FESTA leverages ViT's decomposable modular design to train heads ($\mathcal{H}$) and tails ($\mathcal{T}$) via clients while sharing the server-side Transformer body ($\mathcal{B}$) between clients to integrate retrieved features. Final predictions are then derived by feeding embedded features to their task-specific tails on the client side. (b) illustrates the single-task learning scheme, and (c) two steps multi-task learning scheme. The former is trained for 12000 rounds, while the latter undergoes two training steps. First, the whole parts train in 6000 rounds. Then by freezing the weights of the Transformer body, the heads and tails are fine-tuned for 6000 steps based on the desired specific task.}
 \label{fig:FESTA}
\end{figure}

Furthermore, accessible labeled images have considerably influenced research on the use of Transformers to diagnose COVID-19. Considering the shortage of labeled data, data sharing between hospitals is needed so as to create a viable centralized dataset. However, such collaboration is challenging due to privacy concerns and patient permission. Motivated by Federated Learning (FL) and Split Learning (SL), Park et al. \citep{park2021federated} present a \textbf{Federated Split Task-Agnostic (FESTA)} framework that uses ViT for multi-task learning of classification, detection, and segmentation of COVID-19 CXR images. FESTA benefits from the decomposable modular design of ViT to train heads and tails via clients and share the server-side Transformer body across clients to aggregate extracted features and process each task. The embedded features from the body Transformer are then passed to their task-specific tail on the client side to produce the final prediction (\Cref{fig:FESTA}(a)). \Cref{fig:FESTA}(b) illustrates the single-task learning scheme and (c) the multi-task learning scheme. In multi-task learning, heads, tails, and a task-agnostic Transformer body are first jointly trained for 6000 rounds (see \Cref{fig:FESTA}(c)). Then, heads and tails are fine-tuned according to the desired specific task while freezing the weights of the Transformer body. FESTA merits from 220000 decentralized CXR images and attains competitive results compared to the data-centralized training approaches. The experimental results also demonstrate the stable generalization performance of FESTA, where multi-task learning enhances the performance of the individual tasks through their mutual effect during training.

\begin{figure}[h]
 \centering
 \includegraphics[width=0.48\textwidth]{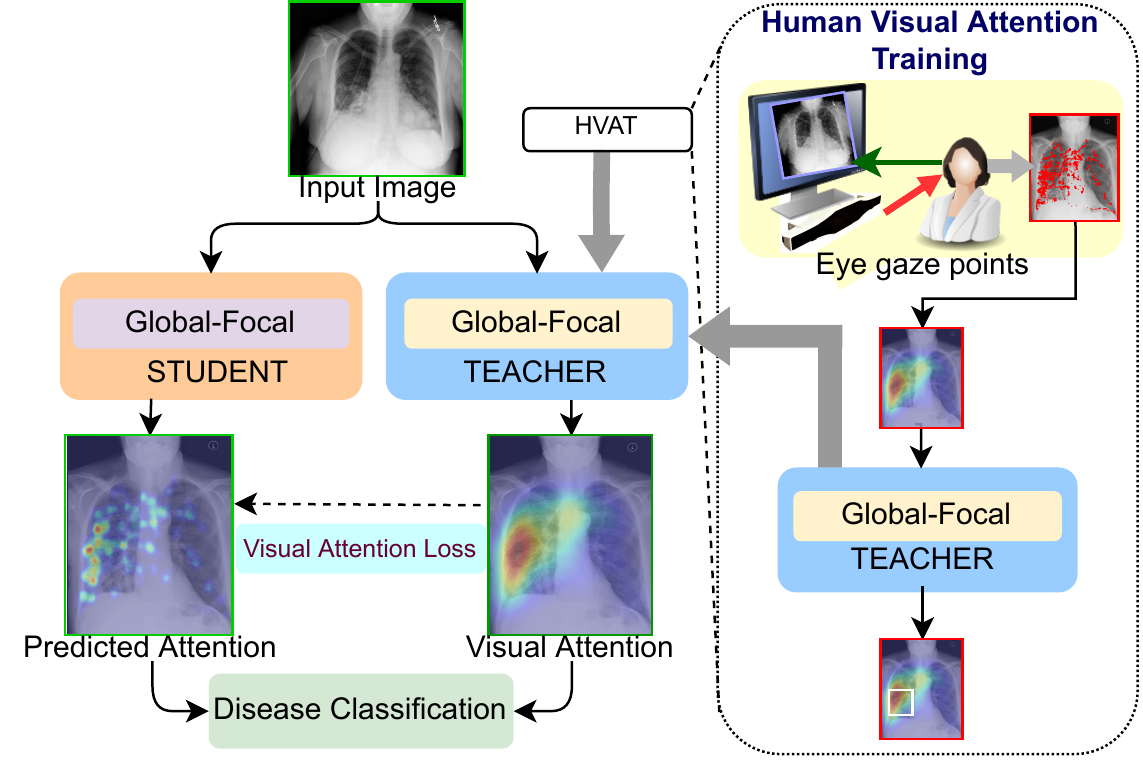}
 \caption{Overview of RadioTransformer \citep{radioTransformer}. Human Visual Attention Training (HVAT) block first uses radiologists' visual observations of chest radiographs to train a global-focal teacher network. The pre-trained teacher network is then utilized to distill the teacher's knowledge to a global-focal student network through visual attention loss, enabling the student to learn visual information. Following the teacher-student strategy and incorporating radiologist visual examinations leads to an improvement in the classification of disease on chest radiographs.}
 \label{fig:RadioTransformer}
\end{figure}

Most attention-based networks utilized for detection and classification rely on the neural network to learn the necessary regions of interest. Bhattacharya et al. \citep{radioTransformer} in \textbf{RadioTransformer} argue that in certain applications, utilizing experts' opinions can prove beneficial. Specifically, they apply this notion to leverage radiologists' gaze patterns while diagnosing different diseases on medical images; then, using a teacher-student architecture, they teach a model to pay attention to regions of an image that a specialist is most likely to examine. As illustrated in \Cref{fig:RadioTransformer}, The teacher and the student networks consist of two main components: global and focal. The global component learns coarse representation while the focal module works on low-level features, and both these segments are comprised of Transformer blocks with shifting windows. In addition, the global and focal components are interconnected using two-way lateral connections to form the global-focal module; this is to address the inherent attention gap between the two. The teacher network is first directly pre-trained on human visual attention maps. Then, the entire model is trained for different downstream tasks, e.g., object detection and classification. Furthermore, the authors propose a self-supervised Visual Attention Loss (VAL) that incorporates both GIoU and MSE loss. The student network is trained to predict probability values for different classes and attention regions. These attention regions are then compared to those obtained from the teacher model, and the weights are optimized using VAL.

\subsection{Hybrid Models}

In spite of the vision Transformers' ability to model global contextual representations, the self-attention mechanism undermines the representation of low-level details. CNN-Transformer hybrid approaches have been proposed to ameliorate the problem above by encoding both global and local features using the locality of CNNs and the long-range dependency of Transformers.

\begin{figure*}[t]
 \centering
 \includegraphics[width=0.99\textwidth]{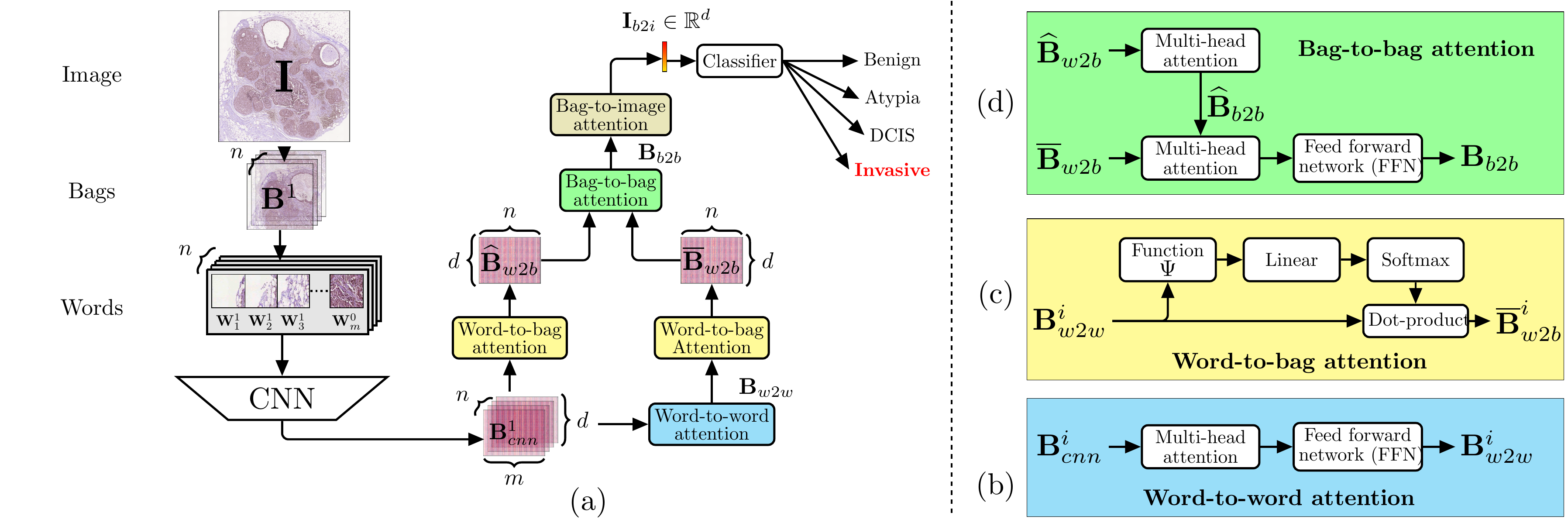}
 \caption{The overall architecture of \citep{mehta2020hatnet}. HATNet hierarchically divides an input image into $n\times m$ words, which are then fed into the CNN encoder to provide word-level representations for each bag. Then by performing a bottom-up decoding strategy and applying a linear classifier, breast biopsy classification results are obtained. Notably, bag-to-image attention has the same procedure as word-to-bag attention, shown in (c).} 
 \label{fig:HATNet}
\end{figure*}

\textbf{TransMed} \citep{transmed} proposes a hybrid CNN-Transformer network that leverages the locality of CNNs and the long-range dependency character of Transformers for parotid gland tumor and knee injury classification. Multimodal medical images primarily have long-range interdependencies, and improving performance requires an effective fusion strategy. TransMed proposes a novel image fusion strategy. Firstly, three neighboring 2D slices of a multimodal image are overlaid to create three-channel images. Then, each image is partitioned into $K\times K$ patches. This fusion approach allows the following network to learn mutual information from images of different modalities. Patch tokens are fed into a CNN network to capture their low-level features and generate patch embeddings. The classic ViT is then used to determine the relationship between patch sequences. TransMed's final results verify the effectiveness of hybrid models in classifying multimodal medical images by outperforming all its counterparts by a large margin. TransMed-S enhances average accuracy on the PGT dataset by about $10.1\%$ over its nearest counterpart, BoTNet \citep{Botnet}, while requiring fewer parameters and FLOP count. Comparably, Tanzi et al. \citep{tanzi2022vision} develop a new CAD system \textbf{(Femur-ViT)} based on Vision Transformers for diagnosing femoral fractures. First, YOLOv3 \citep{redmon2018yolov3} is utilized to detect and crop the left and right femur regions. Afterward, a CNN (InceptionV3 \citep{szegedy2016rethinking}) and a hierarchical CNN (different InceptionV3 networks in cascade) \citep{tanzi2020hierarchical} are applied to the dataset, and the results serve as baselines for the classification. Then, they use a modified ViT to classify seven different fracture types. Finally, a clustering approach is proposed as an evaluation technique for the ViT encoder. This study highlights the power of using ViT models for medical image classification and the ability of the proposed CAD system to significantly increase clinicians' diagnostic accuracy. \textbf{3DMeT} \citep{wang20213dmet} proposes applying a 3D medical image Transformer for assessing knee cartilage defects in three grades: grade 0 (no defect), grade 1 (mild defect), and grade 2 (severe defect). Primarily, using medical 3D volumes as an input to the Transformer is computationally expensive, thereby making it impractical. 3DMeT resolves the high computational cost problem by replacing conventional linear embedding with 3D convolutional layers. The weights of convolutional layers are adopted using the teacher-student training strategy. 3DMeT takes an exponential moving average from the first one/few-layer(s) of the CNN teacher's weights and uses it as convolutional layers' weights. This method enables Transformers to be compatible with small medical datasets and to benefit from CNNs' spatial inductive biases. Lastly, the Transformer and CNN teacher's outputs are combined in order to derive the classification results.

Operating Transformers over Whole Slide Images (WSIs) is computationally challenging since WSI is a gigapixel image that retains the original structure of the tissue. MIL and CNN backbones have demonstrated practical tools for acting on WSI. MIL is a weakly supervised learning approach that enables deep learning methods to train high-resolution images like WSI. Since annotating such images at the pixel level is impractical, MIL proposes to divide an input WSI into a bag of instances and assign a single label to the bag of each image based on pathology diagnosis. The bag has a positive label if it contains at least one positive instance, and it is considered negative if all the instances in the bag are negative. Then CNN backbones are employed to down-sample and extract the features of each instance and allow Transformers to operate according to the generated feature maps and currently available hardware. Therefore, \textbf{DT-MIL} \citep{li2021dt} proposes to compress WSIs into compact feature images by embedding each patch of the original WSI into a super-pixel at its corresponding position using EfficientNet-B0 \citep{tan2019efficientnet}. The resulting thumbnail image feed into a $1\times1$ Conv for feature reduction, followed by a deformable Transformer encoder that aggregates instance representations globally. A similar approach is adopted by \textbf{H}olistic \textbf{AT}tention \textbf{Net}work \textbf{(HATNet)}  \citep{mehta2022end}, where they first divide an input image into $n$ non-overlapping bags, each broken down into $m$ non-overlapping words (or patches). $n\times m$ words are fed into the CNN encoder to obtain word-level representations for each bag. HATNet aims to develop a computer-aided diagnosis system to help pathologists in reducing breast cancer detection errors. According to the World Health Organization (WHO), breast cancer is the most frequent non-skin cancer in women, accounting for one out of every four new female cancers annually \citep{WHO2021}. As illustrated in \Cref{fig:HATNet}, HATNet follows a bottom-up decoding strategy such that it first performs multi-head attention to words in a \textit{word-to-word attention} block, then considers the relationship between words and bags in \textit{word-to-bag attention}, followed by \textit{bag-to-bag attention} to attain inter-bag representations. The acquired bag features are then aggregated in \textit{bag-to-image attention} to build image-level representations. A linear classifier is ultimately applied to achieve the final results. Furthermore, unlike most MIL methods that take all the instances in each bag independent and identically distributed \citep{lu2021data, sharma2021cluster, naik2020deep}, \textbf{TransMIL} \citep{transmil} suggests that it is essential to consider the correlation between different instances and explore both morphological and spatial information. Two Transformer layers address the morphological information, and a conditional position encoding layer named Pyramid Position Encoding Generator (PPEG) addresses the spatial information. The proposed PPEG module has two merits: 1) It handles positional encoding of sequences with a variant number of instances by using group convolution over the 2D reshaped patch tokens, and 2) It enriches the features of tokens by capturing more context information through convolutions. In contrast to conventional IID-based MIL methods requiring many epochs to converge, TransMIL converges two to three times faster by using morphological and spatial information. TransMIL also outperforms all the latest MIL methods \citep{ilse2018attention, li2021dual, campanella2019clinical, li2019patch, lu2021data} in terms of accuracy and AUC by a significant margin in binary and multiple classification tasks and exhibits the superiority of taking the correlation between different instances into account and considering both morphological and spatial information.

\begin{table*}[!ht]
    \centering
    \caption{An overview of the reviewed Transformer-based medical image classification models.}
    \label{tab:classification}
    \resizebox{\textwidth}{!}{
    \begin{tabular}{lccccccc}  
    \toprule
    \textbf{Method} & \textbf{Modality} & \textbf{Organ} & \textbf{Type} & \textbf{Pre-trained Module: Type} & \textbf{Datasets} & \textbf{Metrics} & \textbf{Year}\\ 

    \multicolumn{8}{c}{{\cellcolor[rgb]{1,0.753,0.478}}\textbf{Pure}} \\
     \makecell[l]{ViT-vs-CNN \citep{matsoukas2021time}} & \makecell{Fundus\\ Dermoscopy \\ X-ray} & \makecell{Eye \\ Skin \\ Breast} & 2D & ViT: Self-supervised \& Supervised &  \makecell{$^1$APTOS-2019 \citep{APTOS2019}\\$^2$ISIC-2019 \citep{combalia2019bcn20000} \\$^3$CBIS-DDSM \citep{lee2017curated}} & \makecell{Kappa\\Recall \\ROC-AUC}   &  2021 
     \\
     \midrule
     \makecell[l]{ViT-BUS \citep{gheflati2022vision}} & Ultrasound & Breast & 2D & ViT: Supervised & \makecell{ $^1$B \citep{yap2017automated} \\$^2$BUSI \citep{al2020dataset}} & \makecell{ACC \\AUC} &  2022 
     \\
     \midrule
     \makecell[l]{POCFormer \citep{perera2021pocformer}} & Ultrasound & Chest & 2D & \xmark & POCUS \citep{born2020pocovid} &  \makecell{ Recall, F1\\SP, SE, ACC} & 2021 
     \\
     \midrule
     \makecell[l]{MIL-VT \citep{yu2021mil}} & Fundus & Eye & 2D &  ViT: Supervised & \makecell{$^1$Private Dataset\\$^2$APTOS-2019 \citep{APTOS2019} \\$^3$RFMiD-2020 \citep{pachade2021retinal}} &
     \begin{tabular}[c]{@{}c@{}} Recall, F1\\ACC, AUC \\ Precision \end{tabular} & 2021 
     \\
     \midrule
     \makecell[l]{COVID-VIT \citep{gao2021covid}} & CT & Chest & 3D & \xmark & MIA-COV19 \citep{kollias2021mia} & ACC, F1 & 2021 
     \\
     \midrule
     \makecell[l]{xViTCOS \citep{mondal2021xvitcos}} & \makecell{X-ray \\ CT}& Chest & 2D & ViT: Supervised &  \makecell{$^1$COVID CT-2A \citep{gunraj2021covidx} \\ $^2$CheXpert \citep{irvin2019chexpert}} & \makecell{Recall, F1 \\ Precision \\ SP, NPV} & 2021 
     \\
     \midrule
     \makecell[l]{FESTA \citep{park2021federated}} & X-ray & Chest & 2D & ViT: Supervised & \makecell{$^1$Four Private Datasets\\$^2$CheXpert \citep{irvin2019chexpert}, $^3$BIMCV \citep{vaya2020bimcv} \\ $^4$Brixia \citep{signoroni2021bs}, $^5$NIH \citep{wang2017chestx} \\$^6$SIIM-ACR \citep{siim2018pneumothorax}, $^7$RSNA \citep{rsna2018pneumonia}} & \makecell{Recall, F1\\SP, SE, AUC} & 2021 
     \\  
     \midrule
     \makecell[l]{COVID-Transformer \citep{shome2021covid}} & X-ray & Chest & 2D & ViT: Supervised & \makecell{$^1$\citep{qi2021chest}, $^2$\citep{sait2020curated}\\$^3$\citep{el2020extensive}} & \makecell{Recall, F1\\ACC, AUC \\Precision} & 2021 
     \\
     \midrule
     \makecell[l]{COVID-VOLO \citep{liu2021automatic}} & X-ray & Chest & 2D & ViT: Supervised & \makecell{$^1$\citep{chowdhury2020can}\\$^2$\citep{cohen2020covid}} & ACC & 2021 
     \\  
     \midrule
     \makecell[l]{RadioTransformer \citep{radioTransformer}} &  X-ray & Chest & 2D & ViT: Supervised & \makecell{$^1$RSNA \citep{RSNA-Pneumonia}, $^2$Cell Pneumonia \citep{kermany2018identifying} \\ $^3$COVID-19 Radiography \citep{chowdhury2020can,rahman2021exploring} \\ $^4$NIH \citep{wang2017chestx}, $^5$VinBigData \citep{nguyen2022vindr} \\ $^6$SIIM-FISABIO-RSNA \citep{lakhani20212021}\\ $^7$RSNA-MIDRC \citep{midrc,midrc_1}\\$^8$TCIA-SBU COVID-19 \citep{clark2013cancer,sbu}}
      & \begin{tabular}[c]{@{}c@{}} Recall, F1\\ACC, AUC \\Precision \end{tabular} & 2022 
      \\
     \midrule
     \makecell[l]{Self MedFed \citep{yan2023label}} &  \makecell{X-ray \\ Dermoscopy \\ Fundus} &  \makecell{Chest \\ Skin \\ Eye} & 2D & ViT: Self-supervised & \makecell{$^1$EyePACKS \citep{EyePACKS} \\ $^2$ISIC 2017 \citep{codella2018skin}, $^3$ISIC 2018 \citep{codella2019skin} \\ $^4$ISIC 2020 \citep{rotemberg2021patient} \\ $^5$COVID-FL \citep{yan2023label}}
      & ACC & 2023
      \\
    \multicolumn{8}{c}{{\cellcolor[rgb]{1,0.753,0.478}}\textbf{Hybrid}} 
    \\
     \makecell[l]{TransMIL \citep{transmil}} & Microscopy & Multi-organ & 2D & CNN: Supervised &
    \makecell{$^1$Camelyon16 \citep{bejnordi2017diagnostic}\\$^2$TCGA-NSCLC \citep{TCGA-LUAD,TCGA-LUSC} \\$^3$TCGA-RCC \citep{TCGA-RCC}} & \makecell{ACC \\ AUC} & 2021  
     \\
     \midrule
     \makecell[l]{LAT \citep{sun2021lesion}} & Fundus & Eye & 2D & CNN: Supervised & \makecell{$^1$Messidor-1 \citep{decenciere2014feedback}\\$^2$Messidor-2 \citep{krause2018grader} \\$^3$EyePACKS \citep{EyePACKS}} & AUC \& Kappa &  2021  
     \\
     \midrule
    \makecell[l]{TransMed \citep{transmed}} & MRI & \makecell{Ear \\Knee}& 3D & \makecell{ViT: Supervised \\CNN: Supervised} & \makecell{$^1$PGT \citep{transmed} \\ $^2$MRNET \citep{bien2018deep}} & \makecell{Precision \\ ACC} &  2021 
     \\
    \midrule
     \makecell[l]{3DMeT \citep{wang20213dmet}} & MRI & Knee & 3D & CNN: Supervised & Private dataset & ACC, Recall, F1  &  2021 
     \\
     \midrule
      \makecell[l]{Hybrid-COVID-ViT \citep{park2021vision}} & X-ray & Chest & 2D & CNN: Supervised & CheXpert \citep{irvin2019chexpert} & \makecell{AUC, ACC\\SP, SE} & 2021 
      \\
     \midrule
     \makecell[l]{Femur-ViT \citep{tanzi2022vision}} & X-ray & Femur & 2D & \makecell{ViT: Supervised \\ CNN: Unsupervised} & Private dataset & \makecell{Recall, F1 \\ Precision, ACC} & 2022 
     \\
     \midrule
     \makecell[l]{DT-MIL \citep{li2021dt}} & Microscopy & \makecell{Lung \\ Breast} & 2D & CNN: Supervised & \makecell{$^1$CPTAC-LUAD \citep{clark2013cancer}\\ $^2$BREAST-LNM \citep{li2021dt}} & \makecell{Recall, F1\\ AUC, precision } & 2021  
     \\
     \midrule
      \makecell[l]{GTP \citep{zheng2022graph}} & Microscopy & Lung & 2D & CNN: Self-supervised & \makecell{$^1$NLST \citep{national2011reduced}, $^2$CPTAC \citep{edwards2015cptac} \\ $^3$TCGA \citep{tcga}} & \makecell{Precision, Recall\\SP, SE, ACC, AUC} & 2022  
     \\
     \midrule
     \makecell[l]{HATNet \citep{mehta2022end}} & Microscopy & Breast & 2D & CNN: Supervised & Breast Biopsy WSI Dataset \citep{elmore2015diagnostic} & \makecell{ACC, ROC-AUC\\F1, SP, SE} &  2022 
     \\
     \midrule
     \makecell[l]{HiFuse \citep{huo2022hifuse}} & \makecell{CT \\ Dermoscopy \\ Endoscopy} & \makecell{Skin \\ Chest \\ Stomach} & 2D & \xmark & \makecell{$^1$ISIC 2018 \citep{codella2019skin} \\ $^2$COVID19-CT \citep{he2020sample} \\ $^3$Kvasir \citep{pogorelov2017kvasir}} & \makecell{Recall, ACC \\ F1, Precision} & 2022
    \\
     \midrule
     \makecell[l]{MedViT \citep{manzari2023medvit}} & \makecell{CT \\ X-ray \\ ultrasound \\ OCT} & Multi-organ & 2D & \xmark & MedMNIST \citep{yang2021medmnist} & \makecell{AUC, ACC \\ Top-1} & 2023
    \\
    \bottomrule
    \end{tabular}
    }
\end{table*}

\begin{table*}[!t]
    \centering
    \caption{A brief description of the reviewed Transformer-based medical image classification models. The unreported number of parameters indicates that the value was not mentioned in the paper, and the code was unavailable.}
    \label{tab:classification-highlight}
    \resizebox{\textwidth}{!}{
    \begin{tabular}{llp{15cm}p{15cm}} 
    \toprule
    \textbf{Method} & \textbf{\# Params} & \textbf{Contributions} & \textbf{Highlights}\\ 
    \multicolumn{4}{c}{{\cellcolor[rgb]{1,0.753,0.478}}\textbf{Pure}} 
    \\
     \makecell[l]{ViT-vs-CNN \citep{matsoukas2021time}} 
     & 
     22M 
     & 
     $\bullet$ They investigate three different weight initialization approaches on three medical datasets: (1) random initialization, (2) transfer learning using supervised ImageNet pre-trained weights, and (3) self-supervised pretraining on the target dataset using DINO \citep{caron2021emerging}. Their final verdict is that ViTs can replace CNNs.
     &  
     $\bullet$ Utilize three different training schemes on three different datasets to conclude whether ViT can replace CNN. \newline
     $\bullet$ They repeat their processes five times to be certain of the outcome. \newline
     $\bullet$ Comparing only two models, DeiT-S and Resnet-50, on only three datasets cannot generalize the conclusion of the superiority of each of the Transformer and CNN.
     \\
     \midrule
     {\makecell[l]{ViT-BUS \citep{gheflati2022vision}}}
     &
     \makecell[l]{ViT-Ti/16: 5.53M \\
     ViT-S/32: 22.87M \\
     ViT-B/32: 87.44M}
     &
     $\bullet$ Proposes the use of ViT for the classification of breast ultrasound images for the first time.
     & 
     $\bullet$ Transferring pre-trained ViT models based on small ultrasound datasets yields much higher accuracy than CNN models.
     \\
     \midrule
     \makecell[l]{POCFormer \citep{perera2021pocformer}} 
     &
     \makecell[l]{Binary CLS: 2.8M \\ Multiclass CLS: 6.9M} 
     &
     $\bullet$ Proposes a lightweight Transformer architecture that uses lung ultrasound images for real-time detection of COVID-19.
     &
     $\bullet$ POCFormer can perform in real-time and be deployed to portable devices. \newline
     $\bullet$ POCFormer can be used for rapid mass testing of COVID-19 due to its compactness.
     \\
     \midrule
     \makecell[l]{MIL-VT \citep{yu2021mil}}  
     & 
     22.12M
     & 
     $\bullet$ Proposes to first pre-train the Vision Transformer on a fundus image large dataset and then fine-tune it on the downstream task of the retinal disease classification. \newline
     $\bullet$ Introduces the MIL-VT framework with a novel Multiple Instance Learning(MIL)-head to effectively utilize embedding features to improve the ViT performance.
     &
     $\bullet$ The MIL head can significantly enhance the performance by easily attaching to the Transformer in a plug-and-play manner. \newline
     $\bullet$ MIL-VT efficiently exploits the embedding features overlooked in the final prediction of the ViT.
     \\
     \midrule
     \makecell[l]{COVID-VIT \citep{gao2021covid}} 
     &
     52.81M
     &
     $\bullet$ Offers to utilize ViT to classify COVID and non-COVID patients using 3D CT lung images.
     &
     $\bullet$ COVID-ViT performs better in classifying COVID from Non-COVID compared to DenseNet.\newline
     $\bullet$ The reported result is not enough to conclude. They only compare ViT with DenseNet.
     \\
     \midrule
     \makecell[l]{xViTCOS \citep{mondal2021xvitcos}} 
     &  
     85.99
     &
     $\bullet$ Proposes and explores using ViT for detecting COVID-19 from CXR and CT images. \newline
     $\bullet$ xViTCOS makes use of the Gradient Attention Rollout algorithm \citep{chefer2021Transformer} for visualization and clinical interpretability of the output.
     & 
     $\bullet$ Uses a heatmap plot to demonstrate the model's explainability.
     \\
     \midrule
     \makecell[l]{FESTA \citep{park2021federated}} 
     &
     \makecell[l]{Body: 66.37M \\
     (CLS) Head: 13.31M, Tail: 2k \\
     (SEG) Head: 15.04M, Tail: 7.39M \\
     (DET) Head: 25.09M, Tail: 19.77M}
     &
     $\bullet$ Proposes a Federated Split Task-Agnostic (FESTA) framework that leverages ViT to merit from federated learning and split learning. \newline
     $\bullet$ They use multi-task learning to classify, detect, and segment COVID-19 CXR images.
     &
     $\bullet$ The proposed FESTA Transformer improved individual task performance when combined with multi-task learning. \newline
     $\bullet$ Experimental results demonstrate stable generalization and SOTA performance of FESTA in the external test dataset even under non-independent and non-identically distributed (non-IID) settings. \newline
     $\bullet$ FESTA eliminates the need for data exchange between health centers while maintaining data privacy. \newline
     $\bullet$ Using FESTA in the industry may not be safe because it may encounter external attacks on the server that may lose the parameters of the entire network. In addition, using this method may jeopardize patient information through privacy attacks. \newline
     $\bullet$ Authors did not evaluate the robustness of their approach to difficulties through communication, stragglers, and fault tolerance.
     \\
     \midrule
     \makecell[l]{COVID-Transformer \citep{shome2021covid}} 
     &
     ViT-L/16: 307M
     &
     $\bullet$ COVID-Transformer investigates using ViT for detecting COVID-19 from CXR images. \newline
     $\bullet$ COVID-Transformer introduces a new balanced chest X-ray dataset containing 30K images for multi-class classification and 20K for binary classification.
     &
     $\bullet$ Uses a heatmap plot to demonstrate the model's explainability.
     \\
     \midrule
     \makecell[l]{COVID-VOLO \citep{liu2021automatic}} 
     & 
     86.3M
     &  
     $\bullet$ Proposes fine-tuning the pre-trained VOLO \citep{yuan2022volo} model for COVID-19 diagnosis.
     &
     $\bullet$ Using VOLO enables capturing both fine-level and coarse-level features resulting in higher performance in COVID-19 binary classification.
     \\
     \midrule
     \makecell[l]{RadioTransformer \citep{radioTransformer}} 
     &
     3.93M
     &
     $\bullet$ Presents a novel global-focal RadioTransformer architecture, including Transformer blocks with shifting windows, to improve diagnosis accuracy by leveraging the knowledge of experts. \newline
     $\bullet$ Introduces an innovative technique for training student networks by utilizing visual attention regions generated by teacher networks.
     &
     $\bullet$ outperforms counterpart backbones on multiple datasets. \newline
     $\bullet$ Model's explainability
     \\
    \midrule
     Self MedFed \citep{yan2023label} 
     &
     85.2M
     &
     $\bullet$ Self MedFed is a novel privacy-preserving federated self-supervised pre-training framework designed to learn visual representations from decentralized data by leveraging masked image modeling.  \newline
     $\bullet$ Introduces a benchmark dataset called COVID-FL, consisting of federated chest X-ray data from eight different medical datasets.
     &
     $\bullet$ The proposed network simultaneously tackles the dual challenges of data heterogeneity and label deficiency. \newline
     $\bullet$ Experimental results on various medical datasets validate the superiority of the proposed method compared to ImageNet-supervised baselines and existing federated learning algorithms in terms of label efficiency and robustness to non-IID data.
     \\
    \multicolumn{4}{c}{{\cellcolor[rgb]{1,0.753,0.478}}\textbf{Hybrid}} 
     \\
     \makecell[l]{TransMIL \citep{transmil}} 
     &
     2.67M
     &
     $\bullet$ Presents a Transformer-based Multiple Instance Learning (MIL) approach that uses both morphological and spatial information for weakly supervised WSI classification. \newline
     $\bullet$ Proposes to consider the correlation between different instances of WSI instead of assuming them independently and identically distributed \newline
     $\bullet$ Proposes a CNN-based PPEG module for conditional position encoding, which is adaptive to the number of tokens in the corresponding sequence
     &
     $\bullet$ Converges two to three times faster than SOTA MIL methods. \newline
     $\bullet$ The proposed method can be employed for unbalanced/balanced and binary/multiple classification with great visualization and interpretability. \newline
     $\bullet$ TransMIL is adaptive for positional encoding as token numbers in the sequences changes. \newline
     $\bullet$ It needs further improvement to handle higher magnification than $\times 20$ of WSIs - Higher magnification means longer sequences, which in turn require more memory and computational costs to process.
     \\
     \midrule
     \makecell[l]{LAT \citep{sun2021lesion}} 
     &
     -
     &
     $\bullet$ Proposes a unified Transformer-based encoder-decoder structure capable of DR grading and lesion detection simultaneously. \newline
     $\bullet$ Proposes a Transformer-based decoder to formulate lesion discovery as a weakly supervised lesion localization problem. \newline
     $\bullet$ Proposes lesion region importance mechanism to determine the importance of lesion-aware features. \newline
     $\bullet$ Proposes lesion region diversity mechanism to diversify and compact lesion-aware features.
     &
     $\bullet$ Unlike most approaches that confront lesion discovery and diabetic retinopathy grading tasks independently, which may generate suboptimal results, the proposed encoder-decoder structure is jointly optimized for both tasks. \newline
     $\bullet$ Despite existing methods that only perform well for discovering explicit lesion regions, LAT can also detect less dense lesion areas. \newline
     $\bullet$ The proposed LAT is capable of identifying Grades 0 and 1, which are hard to distinguish.
     \\
     \midrule
     \makecell[l]{TransMed \citep{transmed}} 
     &
     \makecell[l]{TransMed-Tu: 17M \\
     TransMed-S: 43M \\
     TransMed-B: 110M \\
     TransMed-L: 145M}
     &
     $\bullet$ Proposes a hybrid CNN-Transformer network for multimodal medical image classification. \newline
     $\bullet$ Proposes a novel image fusion strategy for 3D MRI data.
     &
     $\bullet$ TransMed achieves much higher accuracy in classifying parotid tumors and knee injuries than CNN models. \newline
     $\bullet$ Requiring fewer computational resources compared to SOTA CNNs.
     \\
     \midrule
     \makecell[l]{3DMeT \citep{wang20213dmet}} 
     & 
     -
     &  
     $\bullet$ Replaces conventional linear embedding with 3D convolution layers to reduce the computational cost of using 3D volumes as the Transformer's inputs. \newline
     $\bullet$ Obtains weights for 3D convolution layers by using a teacher-student training strategy.
     &  
     $\bullet$ The proposed method makes the Transformers capable of using 3D medical images as input. \newline
     $\bullet$ 3DMeT Uses significantly fewer computational resources. \newline
     $\bullet$ Adopting CNN as a teacher assists in inheriting CNN's spatial inductive biases.
     \\
     \midrule
     \makecell[l]{Hybrid-COVID-ViT \citep{park2021vision}} 
     &
     -
     &
     $\bullet$ Proposes a vision Transformer that embeds features for high-level COVID-19 diagnosis classification using a backbone trained to spot low-level abnormalities in CXR images.
     &
     $\bullet$ Different from SOTA models, the proposed model does not use the ImageNet pre-trained weights while archiving significantly better results. \newline
     $\bullet$ They examine the interpretability of the proposed model.
     \\
     \midrule
     \makecell[l]{Femur-ViT \citep{tanzi2022vision}} 
     &
     -
     &
     $\bullet$ Investigates using ViT for classifying femur fractures. \newline
     $\bullet$ proposes using unsupervised learning to evaluate the ViT results.
     &
     $\bullet$ Achieves SOTA results compared to the CNN models.
     \\
     \midrule
     \makecell[l]{GTP \citep{zheng2022graph}} 
     & 
     -
     & 
     $\bullet$ Proposes a graph-based vision Transformer (GTP) framework for predicting disease grade using both morphological and spatial information at the WSIs. \newline
     $\bullet$ Proposes a graph-based class activation mapping (GraphCAM) method that captures regional and contextual information and highlights the class-specific regions. \newline
     $\bullet$ They use a self-supervised contrastive learning approach to extract more robust and richer patch features. \newline
     $\bullet$ They exploit a mincut pooling layer  \citep{bianchi2020spectral} before the vision Transformer layer to lessen the number of Transformer input tokens and reduce the model complexity. \newline
     &
     $\bullet$ In contrast to SOTA approaches, the proposed GTP can operate on the entire WSI by taking advantage of graph representations. In addition, GTP can efficiently classify disease gade by leveraging a vision Transformer. \newline
     $\bullet$ The proposed GTP is interpretable so that it can identify salient WSI areas associated with the Transformer output class. \newline
     $\bullet$ GTP obviates the need for adding extra learnable positional embeddings to nodes by using the graph adjacency matrix. It enables diminishing the complexity of the model. \newline
     $\bullet$ The proposed GTP takes both morphological and spatial information into account.
     \\
     \midrule
    \makecell[l]{DT-MIL \citep{li2021dt}} 
     &
     10.88M
     &
     $\bullet$ Presents a novel embedded-space MIL approach incorporated with an encoder-decoder Transformer for histopathological image analysis. Encoding is done with a deformable Transformer, and decoding with a classic ViT. \newline
     $\bullet$ An efficient method to render a huge WSI is proposed, which encodes the WSI into a position-encoded feature image. \newline
     $\bullet$ The proposed method selects the most discriminative instances simultaneously by utilizing associated attention weights and calibrating instance features using the deformable self-attention.
     &
     $\bullet$ The proposed method efficiently embeds instances' position relationships and context information into bag embedding. \newline
     An extensive analysis of four different bag-embedding modules is presented on two datasets.
     \\
     \midrule
     \makecell[l]{HATNet \citep{mehta2022end}} 
     &
     \makecell[l]{(w/ MobileNetv2): 5.59M \\
     (w/ ESPNetv2): 5.58M \\
     (w/ MNASNet): 5.47M}
     &
     $\bullet$ Presents a novel end-to-end hybrid method for classifying histopathological images.
     &
     $\bullet$ HATNet surpasses the bag-of-words models by following a bottom-up strategy and taking into account inter-word, word-to-bag, inter-bag, and bag-to-image representations, respectively. (word $\rightarrow$ bag $\rightarrow$ image)
     \\
     \midrule
     \makecell[l]{HiFuse \citep{mehta2022end}}
     &
     \makecell[l]{HiFuse-Ti: 82.49M \\
     HiFuse-S: 93.82M \\
     HiFuse-B: 127.80M}
     &
     $\bullet$ A parallel framework is developed to effectively capture local spatial context features and global semantic information representation at various scales. \newline
     $\bullet$ An adaptive hierarchical feature fusion (HFF) block is created to selectively blend semantic information from various scale features of each branch using spatial attention, channel attention, residual inverted MLP, and shortcut connection.
     &
     $\bullet$ The HiFuse model yields favorable outcomes on the three different ISIC 2018 \citep{codella2018skin}, Covid19-CT \citep{he2020sample}, and Kvasir \citep{pogorelov2017kvasir} datasets. \newline
     $\bullet$ HiFuse incorporates a modular design that offers rich scalability and linear computational complexity.
     \\
    \midrule
     \makecell[l]{MedViT \citep{manzari2023medvit}} 
     &
     \makecell[l]{MedViT-Ti: 10.2M \\
     MedViT-S: 23M \\
     MedViT-B: 45M}
     &
     $\bullet$ Proposes a hybrid model that combines the locality of CNNs and the global connectivity of vision Transformers, aiming to improve the reliability of deep medical diagnosis systems. \newline
     $\bullet$ By utilizing an efficient convolution operation, the proposed attention mechanism enables joint attention to the information in various representation subspaces, thereby reducing computational complexity while maintaining performance. \newline
     $\bullet$ Introduces modifying the shape information in the high-level feature space by permuting the feature mean and variance within mini-batches to improve the model's resilience against adversarial attacks.
     &
     $\bullet$ The proposed attention mechanism is efficient.\newline
     $\bullet$ MedViT enhances the robustness of the Transformer models against adversarial attacks.
     \\
    \bottomrule
    \end{tabular}
    }
\end{table*}

Previous methods mainly rely on weakly supervised learning or dividing WSIs into image patches and using supervised learning to assess the overall disease grade. Nevertheless, these approaches overlook WSI contextual information. Thus, Zheng et al. \citep{zheng2022graph} propose a \textbf{Graph-based Vision Transformer (GTP)} framework for predicting disease grade using both morphological and spatial information at the WSIs. The graph term allows for the representation of the entire WSI, and the Transformer term allows for computationally efficient WSI-level analysis. The input WSI is first divided into patches, and those that contain more than $50\%$ of the background are eliminated and not considered for further processing. Selected patches are fed forward through a contrastive learning-based patch embedding module for feature extraction. A graph is then built via a graph construction module utilizing patch embeddings as nodes of the graph. In the graph Transformer section, a graph convolution layer followed by a mincut pooling layer \citep{bianchi2020spectral} is applied first to learn and enrich node embeddings and then lessen the number of Transformer input tokens. Since the graph adjacency matrix contains spatial information of nodes, by adding an adjacency matrix to node features, GTP obviates the need for adding extra learnable positional embeddings to nodes. The final Transformer layer predicts the WSI-level class label for three lung tumor classes: Normal, LUAD, and LSCC. GTP also introduces a graph-based class activation mapping (GraphCAM) technique that highlights the class-specific regions. GraphCAM exploits attention maps from multi-head self-attention (MHSA) blocks in the Transformer layer and maps them to the graph space to create a heatmap for the predicted class. The experiments show that GTP performs as a superior interpretable and efficient framework for classifying WSI images while considering morphological and spatial information.

\begin{figure}[!th]
 \centering
 \includegraphics[width=0.48\textwidth]{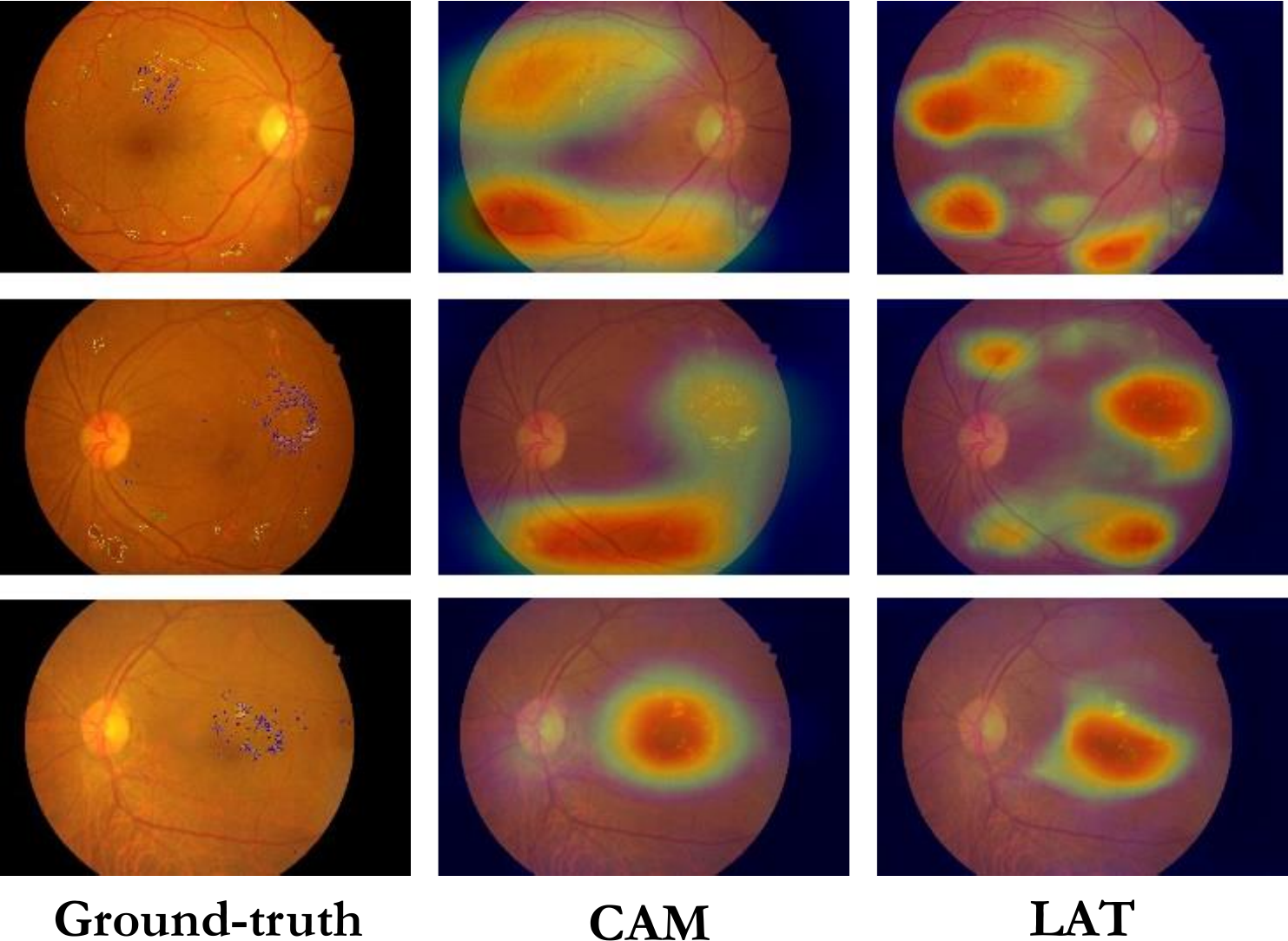}
 \caption{LAT \citep{sun2021lesion} vs. CAM \citep{zhou2016learning} visual comparison. The ground truth consists of microaneurysms, hemorrhages, soft exudates, and hard exudates, which are colored as green, yellow, green, and blue dots, respectively (zoom in on the ground truth image for better clarity)}.
 \vspace{-1em}
 \label{fig:LAT-results}
\end{figure}

Diabetic Retinopathy (DR) is an eye disorder that can cause impaired vision and sightlessness by damaging blood vessels in the retina. Most deep-learning approaches view lesion discovery and DR grading as independent tasks that may produce suboptimal results. In contrast to conventional methods, LAT \citep{sun2021lesion} proposes a unified encoder-decoder structure that comprises a pixel relation-based encoder to capture the image context information and a lesion filter-based decoder to discover lesion locations, which the whole network jointly optimized and complemented during training. The encoder is particularly in charge of modeling the pixel correlations, and the Transformer-based decoder part is formulated as a weakly supervised localization problem to detect lesion regions and categories with only DR severity level labels. 
% \begin{figure}[h]
%  \centering
%  \includegraphics[width=0.48\textwidth]{images/LAT.png}
%  \caption{The overall pipeline of LAT \citep{sun2021lesion} comprises a unified encoder-decoder structure and a classification module, where all are processed simultaneously and output DR grading and lesion discovery results.}
%  \label{fig:LAT structure}
% \end{figure}
In addition, LAT proposes two novel mechanisms to improve the effectiveness of lesion-aware filters: 1) Lesion region importance mechanism, $g(\cdot|\Phi)$, to determine the contribution of each lesion-aware feature, and 2) Lesion region diversity mechanism to diversify and compact lesion-aware features. The former is a linear layer followed by a sigmoid activation function that generates importance weights for lesion-aware features, and the latter adopts a triplet loss \citep{zhong2018generalizing} to encourage lesion filters to find diverse lesion regions. In the DR grading branch, LAT presents a DR grading classification module that calculates a global consistency loss based on the lesion-aware features, indicated as $h(\cdot|\sigma)$. Eventually, the final DR grading prediction is achieved by calculating the cross-entropy loss between the predicted labels obtained from the fusion of $g(\cdot|\Phi)$ and  $h(\cdot|\sigma)$ and the ground truth. The total loss is the aggregation of cross-entropy loss, global consistency loss, and triplet loss. Visual results of LAT regarding the lesion discovery are depicted in \Cref{fig:LAT-results}.

The current success of deep learning in medical image interpretation relies heavily on supervised learning, which requires large labeled datasets. However, annotating medical imaging data is costly and time-consuming. Additionally, models struggle to generalize across external institutions or different tasks. Self-supervised learning offers a solution by leveraging unlabeled medical data to develop robust models without costly annotations \citep{huang2023self}. Chen et al. \citep{chen2021empirical} propose a valuable empirical study of training self-supervised vision Transformers, examining their effectiveness and performance. According to their findings, self-supervised Transformers demonstrate strong performance when trained using contrastive learning, outperforming ImageNet-supervised ViT models as their size increases. For example, self-supervised pre-training surpasses supervised pre-training in certain cases, particularly with the very large ViT-Large. Furthermore, self-supervised ViT models achieve competitive results compared to prominent convolutional ResNets in prior studies \citep{chen2020big,grill2020bootstrap}, showcasing the potential of ViT with relatively fewer inductive biases \citep{dosovitskiy2020image}. Notably, they discovered that removing the position embedding in ViT only has a slight negative impact on accuracy, indicating that self-supervised ViT can learn powerful representations without relying heavily on positional information, but also suggesting that the positional information may not have been fully utilized. Therefore, these findings indicate that the combination of self-supervised learning and vision Transformers can be beneficial to adopt for different medical image analysis tasks. For example, Yan et al. \citep{yan2023label} propose a robust self-supervised federated learning framework, \textbf{Self MedFed}, for medical image classification. Their approach combines masked image modeling with Transformers to pre-train models directly on decentralized target task datasets. This enables more robust representation learning on heterogeneous data and effective knowledge transfer. Experimental results on public medical imaging datasets \citep{EyePACKS,codella2018skin,codella2019skin} and their created COVID-FL dataset demonstrate the framework's effectiveness in improving model robustness and learning visual representations across non-IID clients. The research addresses data heterogeneity and label deficiency, showcasing the broad applicability of their self-supervised federated learning framework in medical image analysis.

\begin{table*}
    \centering
    \caption{Comparison of Transformer-based models on different medical image classification datasets. ($\dagger$) and ($\star$) indicate that results are adopted from \citep{huo2022hifuse} and \citep{manzari2023medvit}.}
    \label{tab:performance-classification}
    \resizebox{\textwidth}{!}{
    \begin{tblr}{
      row{1} = {lightorange,c},
      column{2} = {c},
      column{3} = {c},
      column{4} = {c},
      column{5} = {c},
      column{6} = {c},
      column{7} = {c},
      column{9} = {c},
      column{10} = {c},
      column{11} = {c},
      cell{1}{1} = {c=7}{},
      cell{1}{8} = {c=4}{},
      cell{7}{1} = {c=7}{lightorange,c},
      cell{12}{1} = {c=7}{lightorange,c},
      hlines,
      vline{2} = {1}{},
      vline{2-11} = {2-6,8-11,13-16}{},
      vline{2,9-11} = {7,12}{},
      vline{1,12} = {3pt},
      hline{1,17} = {3pt},
      vline{8} = {1.5pt},
    }
    \textbf{ISIC 2018$\dagger$ \citep{codella2019skin}} &  &  &  &  &  &  & \textbf{TissueMNIST$\star$ \citep{ljosa2012annotated}} &  &  & \\
    \textbf{Method} & \textbf{Params (M)} & \textbf{FLOPs (G)} & \textbf{Accuracy \%} & \textbf{F1 \%} & \textbf{Precision \%} & \textbf{Recall \%} & \textbf{Method} & \textbf{Params (M)} & \textbf{FLOPs (G)} & \textbf{Top-1\%} 
    \\
    \textbf{HiFuse-S \citep{huo2022hifuse}} & 93.82 & 8.84 & \textbf{83.59} & \textbf{72.70} & 72.70 & \textbf{73.14} &
    \textbf{MedViT-S \citep{manzari2023medvit}} & 23.6 & 4.9 & \textbf{73.1}
    \\
    \textbf{Conformer-B/16 \citep{peng2021conformer}} & \textbf{83.29} & 22.89 & 82.66 & 72.44 & \textbf{73.31} & 71.66 & 
    \textbf{Twins-SVT-S \citep{chu2021twins}} & 24.0 & 2.9 & 72.1
    \\
    \textbf{Swin-B \citep{liu2021swin}} & 87.77 & 15.14 & 79.79 & 63.95 & 65.09 & 63.65 &
    \textbf{PoolFormer-S36 \citep{yu2022metaformer} } & 31.2 & 5.0 & 71.8
    \\
    \textbf{ViT-B/32 \citep{dosovitskiy2020image}} & 88.30 & \textbf{8.56} & 77.92 & 57.52 & 58.74 & 56.90 & 
    \textbf{Swin-T \citep{liu2021swin}} & 29.0 & 4.5 & 71.7
    \\
    \textbf{COVID19-CT$\dagger$ \citep{he2020sample}} &  &  &  &  &  &  & 
    \textbf{CvT-13 \citep{wu2021cvt}} & 20.1 & 4.5 & 71.6
    \\
    \textbf{HiFuse-S  \citep{huo2022hifuse}} & 93.82 & 8.84 & \textbf{76.88} & \textbf{76.31} & \textbf{77.78} & 76.19 & 
    \textbf{RVT-S \citep{mao2022towards}} & 22.1 & 4.7 & 71.2
    \\
    \textbf{Conformer-B/16 \citep{peng2021conformer}} & \textbf{83.29} & 22.89 & 75.81 & 75.60 & 76.81 & \textbf{77.81} &
    \textbf{MedViT-T \citep{manzari2023medvit}} & 10.8 & \textbf{1.3} & 70.3
    \\
    \textbf{ViT-B/32 \citep{dosovitskiy2020image}} & 88.30 & \textbf{8.56} & 61.83 & 60.59 & 61.89 & 60.94 & 
    \textbf{MedViT-L \citep{manzari2023medvit}} & 45.8 & 13.4 & 69.9
    \\
    \textbf{Swin-B \citep{liu2021swin}} & 87.77 & 15.14 & 60.75 & 56.36 & 63.20 & 58.95 & 
    \textbf{RVT-Ti \citep{mao2022towards}} & 8.6 & \textbf{1.3} & 69.6
    \\
    \textbf{Kvasir$\dagger$ \citep{pogorelov2017kvasir}} &  &  &  &  &  &  & 
    \textbf{CoaT Tiny \citep{xu2021co}} & \textbf{5.5} & 4.4 & 69.3
    \\
    \textbf{HiFuse-S \citep{huo2022hifuse}} & 93.82 & 8.84 & \textbf{85.00} & \textbf{84.96} & \textbf{85.08} & \textbf{85.00} & 
    \textbf{RVT-B \citep{mao2022towards}} & 86.2 & 17.7 & 69.3
    \\
    \textbf{Conformer-B/16 \citep{peng2021conformer}} & \textbf{83.29} & 22.89 & 84.25 & 84.27 & 84.45 & 84.37 & 
    \textbf{DeiT-S  \citep{DeiT}} & 22.0 & 4.6 & 67.0
    \\
    \textbf{Swin-B \citep{liu2021swin}} & 87.77 & 15.14 & 77.30 & 77.29 & 77.74 & 77.44 & 
    \textbf{PiT-S \citep{heo2021rethinking}} & 23.5 & 2.9 & 66.9
    \\
    \textbf{ViT-B/32 \citep{dosovitskiy2020image}} & 88.30 & \textbf{8.56} & 73.80 & 73.50 & 74.24 & 73.72 & 
    \textbf{PVT-S \citep{wang2021pyramid}} & 25.4 & 4.0 & 66.7
    \end{tblr}
    }
\end{table*}

\begin{tcolorbox}[breakable ,colback={lightorange!60},title={\subsection{Discussion and Conclusion}},colbacktitle=lightorange!60,coltitle=black , left=2pt , right =2pt]
\label{class_discuss}	
\Cref{sec:classification} thoroughly outlines 22 distinctive Transformer-based models in medical image classification. We have categorized the introduced models based on their architectures into hybrid and pure. These approaches differ according to whether they adhere to the original structure of the vanilla ViT or provide a new variant of the vision Transformer that can be applied to medical applications. In addition, we have presented details on the studied classification methods regarding their architecture type, modality, organ, pre-trained strategy, datasets, metrics, and the year of publication in \Cref{tab:classification}. Additional descriptions of the methods, including their model size, contributions, and highlights, are described in \Cref{tab:classification-highlight}. \Cref{tab:performance-classification} presents a comparison of Transformer-based models across four distinct medical image classification datasets. The evaluation primarily focuses on assessing the accuracies achieved by various models, as well as analyzing their Flop count and number of parameters. 

Additionally, time inference and GPU usage are pivotal factors when evaluating method performance. Inference time is a crucial metric for real-time or time-critical applications, and efficient GPU utilization ensures optimal resource allocation, preventing undue stress on the hardware. We conducted experiments using different MedViT models (small, base, and large) on an NVIDIA Tesla T4 GPU, assessing GPU usage and inference times for input images of size $3\times224\times224$. We performed 1000 model inferences after a warm-up period of 50 iterations to gauge performance. For a batch size of 8, MedViT-S consumes 3.92 GB of GPU memory, while MedViT-B and MedViT-L require 5.04 GB and 6.06 GB, respectively. In terms of inference time (batch size 1), MedViT-S, -B, and -L averaged 49.60 ms, 66.89 ms, and 87.95 ms, respectively. These insights provide valuable information for choosing the right model for specific applications, balancing performance with GPU resource constraints, and enabling informed comparisons with other models when considering factors like the number of parameters and FLOPs. By examining these metrics, researchers can gain insights into the performance and efficiency of different models in the context of medical image classification tasks.

As is evident in the storyline of this section, we have discussed methods in each paragraph regarding the underlying problems in medical image classification and introduced solutions and how they address such issues. However, the need for more research on these problems is crucial to making such approaches widely applicable.

Data availability in the medical domain is one of the most challenging aspects of developing Transformer-based models since Transformer models are known for being data-hungry to generalize. Reasons for data scarcity in the medical field can be referred to as privacy concerns of patients, the time-consuming and costly process of annotation, and the need for expert staff. To this end, the use of generative models \citep{pinaya2022brain,moghadam2022morphology,kazerouni2022diffusion} and their integration with Transformer models can become prominent since they are capable of creating synthetic data that is comparable to genuine data. In addition, another way to attack this problem is by utilizing federated learning, such as \citep{park2021federated}. Nevertheless, there is still room for improvement when it comes to privacy concerns since, in federated learning, communication between the client and server is required. 

Despite their SOTA performance, Transformer-based networks still face challenges in deploying their models in the real world due to computational limitations. As shown in \Cref{tab:classification-highlight}, most approaches have a high number of parameters which provokes a serious problem. Different novel approaches have been introduced to reduce the quadratic complexity of self-attention, which can be leveraged in the medical domain. Furthermore, though ViTs have shown impressive capabilities in ImageNet classification, their performance is still lower than the latest SOTA CNNs without additional data \citep{liu2021automatic}. Hence, existing methods mostly follow pre-training strategies on the ImageNet dataset to build the pre-trained weights for the subsequent downstream tasks. However, despite the enhancement, the domain of natural images is significantly different from medical data, thereby restricting the performance of further improvement. Therefore, we believe efficient Transformers will considerably influence the future research of Transformer-based models.
\end{tcolorbox}

\begin{figure*}[t]
    \centering
    \includegraphics[width=0.99\textwidth]{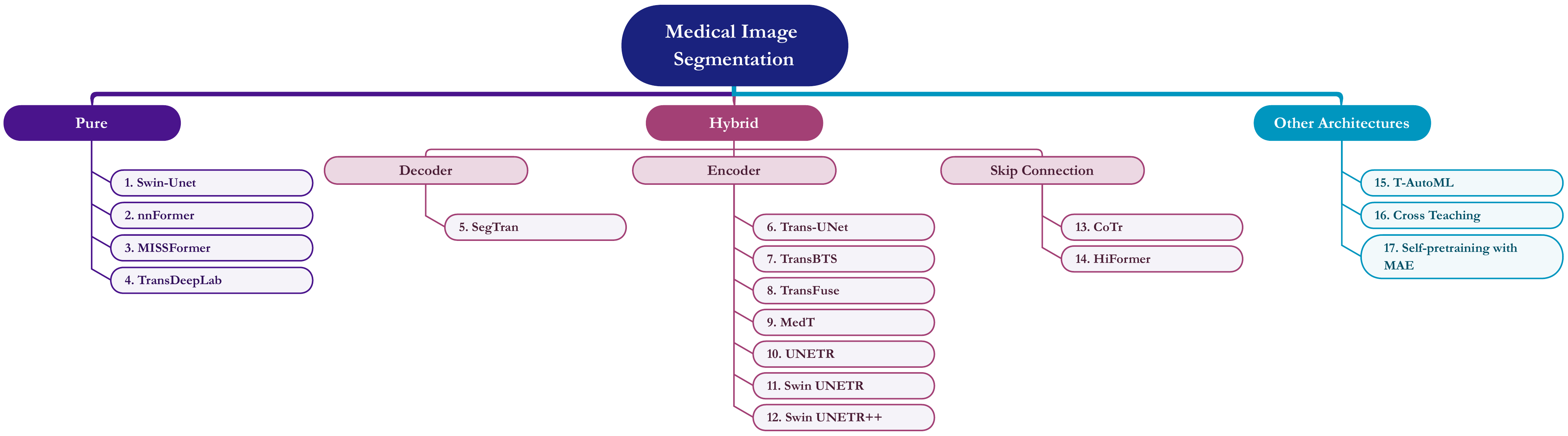}
    \caption{An overview of ViTs in medical image segmentation. Methods are classified into the pure Transformer, hybrid Transformer, and other architectures according to the positions of the Transformers in the entire architecture. The prefix numbers of the methods denote 1. \citep{cao2021swin}, 2. \citep{zhou2021nnformer}, 3. \citep{huang2021missformer}, 4. \citep{azad2022transdeeplab}, 5. \citep{li2021segtran}, 6. \citep{chen2021transunet}, 7. \citep{wang2021transbts}, 8. \citep{zhang2021transfuse}, 9. \citep{valanarasu2021medt}, 10. \citep{hatamizadeh2021unetr}, 11. \citep{hatamizadeh2022swinunetr}, 12. \citep{tang2022selfswinunetr}, 13. \citep{xie2021cotr}, 14. \citep{heidari2022hiformer}, 15. \citep{yang2021TAutoML}, 16. \citep{luo2021semi}, 17. \citep{zhou2022MAE}.}
    \label{fig:segmentationtaxonomy}.
\end{figure*}

%%Jia------------------------------------------
%%-----------------------------------------------
\section{Medical Image Segmentation} \label{sec:segmentation}
\begin{figure}[!tbh]
    \centering
    \includegraphics[width=0.4\textwidth]{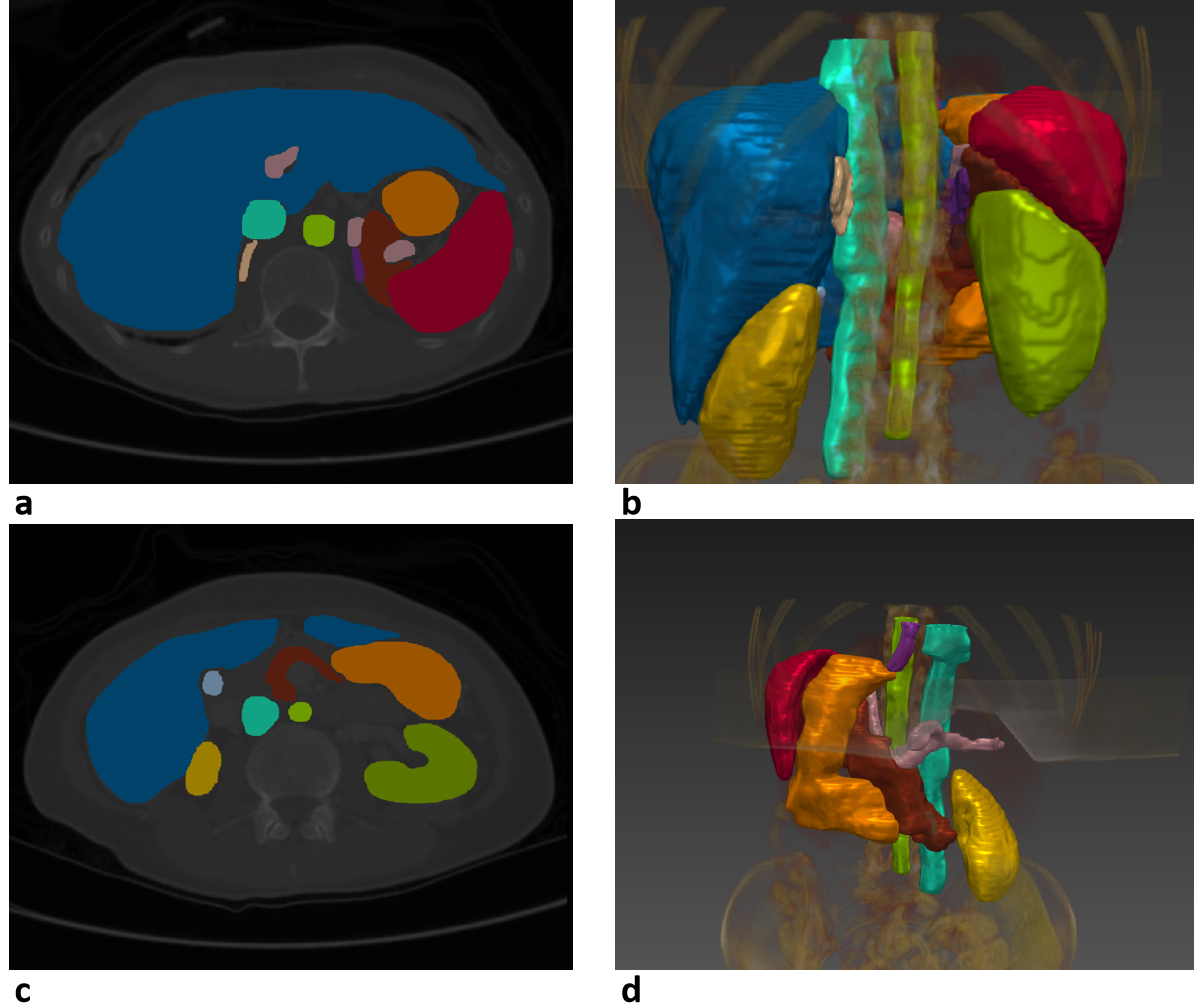}
    \caption{Transformer-based models can perform image segmentation on medical image datasets. Figure \textbf{a} and \textbf{c} illustrate two 2D slices of raw images with the labels from Synapse dataset \citep{Synapse}. Figure \textbf{b} and \textbf{d} show the 3D visualization of the labeled organs from different angles. These images were generated with MITK Workbench \citep{nolden2013mitk}.}
    \label{fig:3d_vis_synapse}
\end{figure}

Medical segmentation is a significant sub-field of image segmentation in digital image processing. It aims to extract features from a set of regions partitioned from the entire image and segment the key organs simultaneously, which can assist physicians in making an accurate diagnosis in practice. X-ray, positron emission tomography (PET), computed tomography (CT), magnetic resonance imaging (MRI), and ultrasound are common imaging modalities used to collect data. The CNN-based U-Net \citep{ronneberger2015unet,azad2022smu,azad2022medicalunet,asadi2020multi} has been the main choice in this field due to its effective performance and high accuracy. Nevertheless, it cannot extract long-range dependencies in high-dimensional and high-resolution medical images 
\citep{karimijafarbigloo2023mmcformer,aghdam2022attention}. Therefore, the flexible combination of the U-Net structure with Transformers become a prevalent solution to the segmentation problem at present. Take the multi-organ segmentation task as an example: some networks can achieve state-of-the-art multi-organ segmentation performance on the Synapse dataset (as shown in \Cref{fig:3d_vis_synapse}) for abdominal images.

In this section, we present the application of ViTs in segmentation tasks. First, we divide the approaches into two categories: \textit{pure Transformers} and \textit{hybrid Transformers}, where the \textit{pure Transformer} denotes the use of the multiple multi-head self-attention modules in both the encoder and decoder. Hybrid architecture-based approaches fuse the ViTs with convolution modules as the encoder, bottleneck, decoder, or skip connection part to leverage information about the global context and local details. Furthermore, we review some methods with other architectures that propose several novel manners for self-supervised learning.
\Cref{fig:segmentationtaxonomy} demonstrates the different directions of the methods employing Transformers in the U-Net architecture.

\subsection{Pure Transformers} \label{sec:seg_pure_Transformer}
In this section, we review several networks referred to as \textit{pure Transformers}, which employ Transformer blocks in both the encoding and the decoding paths.
Despite the great success of CNN-based approaches in medical segmentation tasks, these models still have limitations in learning long-range semantic information of medical images. The authors proposed \textbf{Swin-Unet}, a symmetric Encoder-Decoder architecture motivated by the hierarchical Swin Transformer \citep{liu2021swin}, to improve segmentation accuracy and robust generalization capability. In contrast to the closest approaches \citep{valanarasu2021medt,zhang2021transfuse,wang2021transbts,azad2022transnorm} using integrations of CNN with Transformer, Swin-Unet explores the possibility of pure Transformer applied to medical image segmentation.

\begin{figure}[h]
 \centering
 \includegraphics[width=0.5\textwidth]{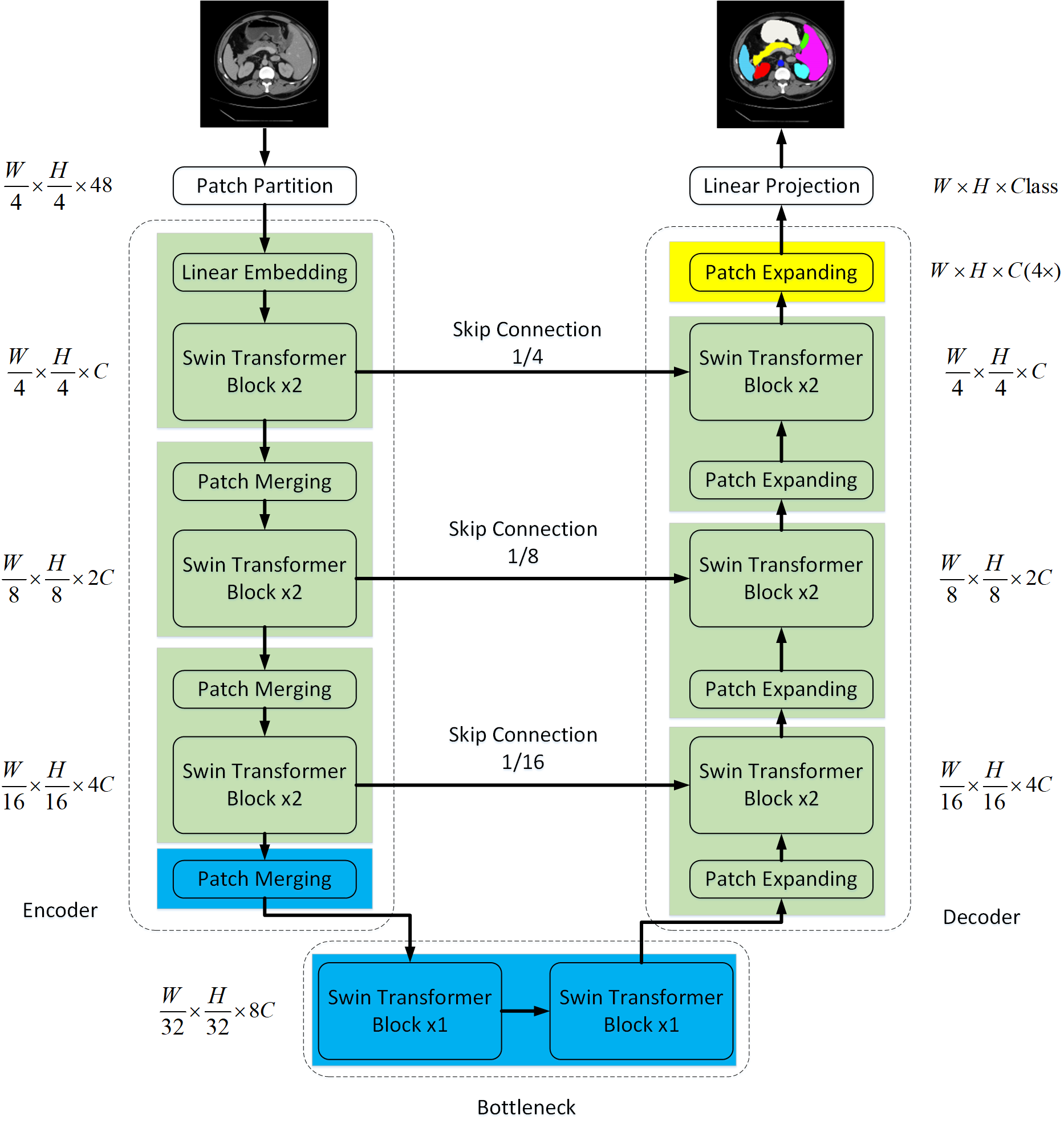}
 \caption{The architecture of the Swin-Unet \citep{cao2021swin} which follows the U-Shape structure. It contains the encoder, the bottleneck, and the decoder part, which are built based on the Swin Transformer block. The encoder and the decoder are connected with skip connections.}
 \label{fig:swinunet structure}
\end{figure}

As shown in \Cref{fig:swinunet structure}, Swin-Unet consists of encoder, bottleneck, decoder, and skip connections utilizing the Swin Transformer block with shifted windows as the basic unit. 
For the encoder, the sequence of embeddings transformed from image patches is fed into multiple Swin Transformer blocks and patch merging layers, with Swin Transformer blocks performing feature learning, and patch merging layers downsampling the feature resolution and unifying the feature dimension.
The designed bottleneck comprises two consecutive Swin Transformer blocks to learn the hierarchical representation from the encoder with feature resolution and dimension unchanged.

Swin Transformer blocks and patch-expanding layers construct the symmetric Transformer-based decoder. In contrast to the patch merging layers in the encoder, each patch expanding layer is responsible for upsampling the feature maps into double resolutions and halving the corresponding feature dimension. The final reshaped feature maps pass through a linear projection to produce the pixel-wise segmentation outputs.
Inspired by the U-Net, the framework also employs skip connections to combine multi-scale features with the upsampled features at various resolution levels to reduce the loss of fine-grained contextual information caused by down-sampling.

In contrast to the CNN-based methods showing over-segmentation issues, the proposed U-shape pure Transformer presents better segmentation performance resulting from learning both local and long-range dependencies. Compared to the previous methods \citep{oktay2018attention,chen2021transunet}, the HD evaluation metric of Swin-Unet shows an improvement in accuracy for better edge prediction. The experiments on the Synapse multi-organ CT dataset and ACDC dataset from MRI scanners also demonstrate the robustness and generalization ability of the method.

%nnFormer
Compared to Swin-Unet and DS-TransUNet, \textbf{nnFormer} \citep{zhou2021nnformer} proposed by Zhou et al. preserves the superior performance of convolution layers for local detail extraction and employs a hierarchical structure to model multi-scale features. It utilizes the volume-based multi-head self-attention (V-MSA) and the shifted version (SV-MSA) in the Transformer blocks instead of processing 2D slices of the volume. The overall architecture of nnFormer is composed of an encoder and a decoder. Each stage in the encoder and decoder consists of a Transformer block applying V-MSA and SV-MSA and a successive upsampling or downsampling block built upon convolution layers, which is referred to as the interleaved architecture. V-MSA conducts self-attention within 3D local volumes instead of 2D local windows to reduce the computational complexity by approximately 98\% and 99.5\% on the Synapse and ACDC datasets, respectively. nnFormer is first pre-trained on the ImageNet dataset~\citep{deng2009large} and utilizes symmetrical initialization to reuse the pre-trained weights of the encoder in the decoder. It employs the self-supervised learning strategy, deep supervision, to train the network. Specifically, the output feature maps of all resolutions in the decoder are passed to the corresponding expanding block. It applies cross-entropy loss and dice loss to evaluate the loss for each stage. Accordingly, the resolution of ground truth is downsampled to match the predictions of all the resolutions. The weighted losses of all three stages comprise the final training objective function. The results of experiments that compare nnFormer with prior Transformer-based \citep{chen2021transunet,cao2021swin} and CNN-based arts \citep{isensee2020nnunet} illustrate nnFormer makes significant progress on the segmentation task.

% \begin{figure}[h]
%  \centering
%  \includegraphics[width=0.4\textwidth]{images/nnFormer_overview.pdf}
%  \caption{Illustration of the U-shape architecture of nnFormer. The 3D medical images are fed into the encoder of the network. The lower part demonstrates the details of the embedding part, upsampling, and expanding part. The $M$ in the decoder denotes the number of classes \citep{zhou2021nnformer}.}
%  \label{fig:nnformer}
% \end{figure}

%MISSFormer
Although the recent Transformer-based methods improve the problem that CNN methods cannot capture long-range dependencies, they show the limitation of the capability of modeling local details. Some methods directly embedded the convolution layers between fully connected layers in the feed-forward network. Such structure supplements the low-level information but limits the discrimination of features. Huang et al. propose \textbf{MISSFormer} \citep{huang2021missformer}, a hierarchical encoder-decoder network, which employs the Transformer block named Enhanced Transformer Block and equips the Enhanced Transformer Context Bridge.

The Enhanced Transformer Block utilizes a novel efficient self-attention module that illustrates the effectiveness of spatial reduction for better usage of the high-resolution map. The original multi-head self-attention can be formulated as follows:
\begin{equation}
Attention(Q,K,V)=Softmax(\frac{QK^T}{\sqrt{d_{head}}})V,
\end{equation}
where $Q$, $K$, and $V$ refer to query, key, and value, respectively, and have the same shape of $N \times C$, $d_{head}$ denotes the number of heads. The computational complexity is $\mathcal{O}(N^2)$.
In efficient self-attention, the $K$ and $V$ are reshaped by a spatial reduction ratio $R$. Take $K$, for example:
\begin{equation}
new\_K = Reshape(\frac{N}{R},C \cdot R)W(C \cdot R, C)
\end{equation}

$K$ is first resized from $N \times C$ to $\frac{N}{R} \times (C \cdot R)$ and then projected linearly to restore the channel depth from $C \cdot R$ to $C$. The computational cost reduces to $\mathcal{O}(\frac{N^2}{R})$ accordingly.
Furthermore, the structure of the Enhanced Mix Feed-forward network (Mix-FFN) extended from \citep{liu2021rethinking} introduces recursive skip connections to make the model more expressive and consistent with each recursive step.

The U-shaped architecture of the MISSFormer contains the encoder and decoder built on the Enhanced Transformer blocks connected with an enhanced Transformer context bridge. Multi-scale features produced from the encoder are flattened and concatenated together and passed through the Enhanced Transformer Context Bridge.
The pipeline of the Enhanced Transformer Context Bridge is based on the Enhanced Transformer Block to fuse the hierarchical features. The output of the bridge is split and recovered to each original spatial dimension to pass through the corresponding stage in the decoder.
The results of experiments show a robust capacity of the method to capture more discriminative details in medical image segmentation. It is worth mentioning that MISSFormer trained from scratch even outperforms state-of-the-art methods pre-trained on ImageNet.

% \begin{figure}[h]
%     \centering
%     \includegraphics[width=0.4\textwidth]{./images/MISSFormer_overall.pdf}
%     \caption{Overview of the MISSFormer \citep{huang2021missformer} architecture.
% 	The network is composed of a hierarchical encoder, a decoder, and an Enhanced Transformer Context Bridge. The encoder and decoder are constructed based on the enhanced Transformer blocks and modules for patch processing. The outputs of each stage within the encoder are fused and passed through the bridge to model the local and global dependencies of different scales \citep{huang2021missformer}.}
%     \label{fig:MISSFormer_overall}
% \end{figure}

\begin{figure}[h]
    \centering
    \includegraphics[width=0.5\textwidth]{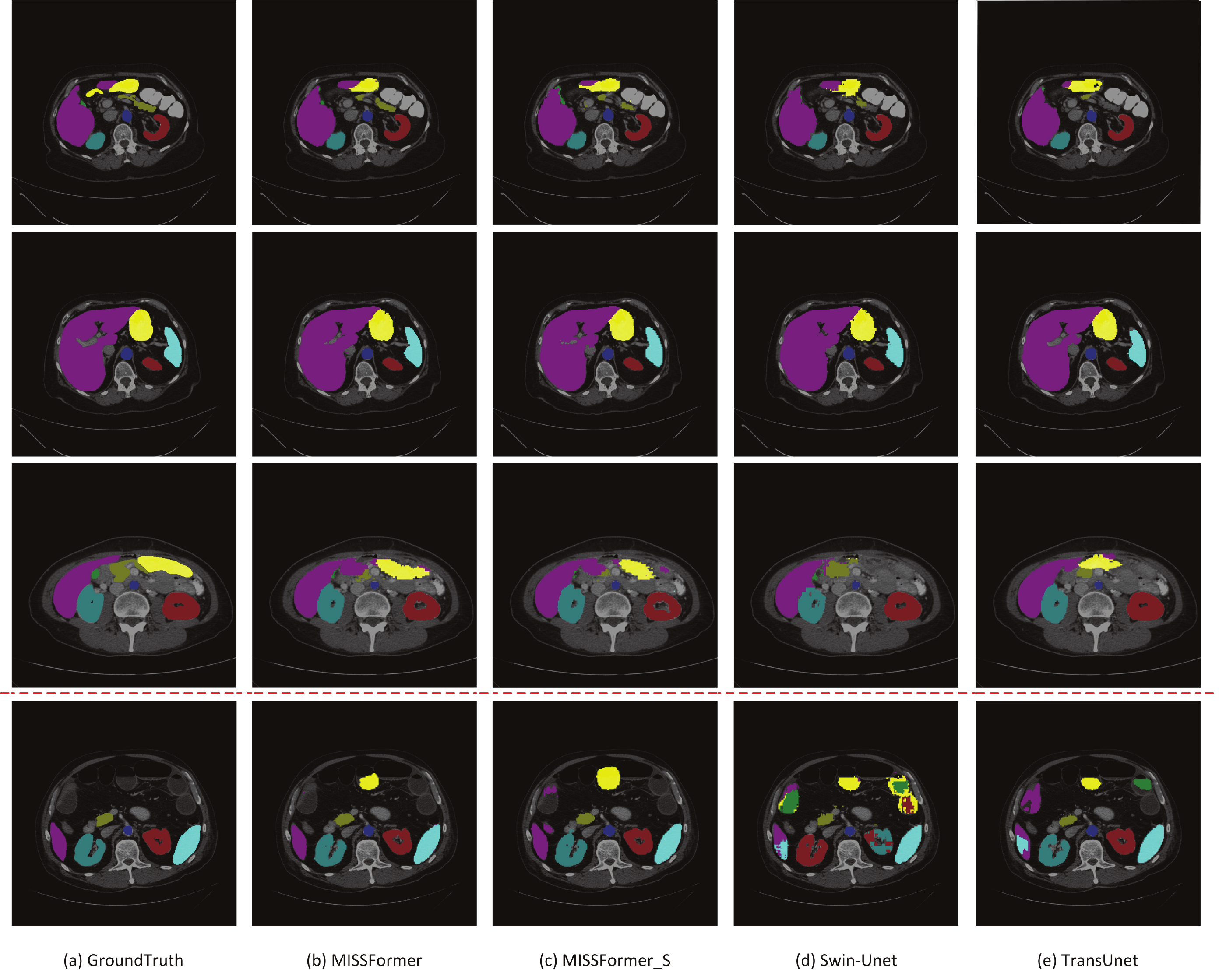}
    \caption{A visual comparison with the state-of-the-art approaches on Synapse dataset. Above the red line shows the successful cases of segmentation, and below the red line are the failed cases with relatively large errors \citep{huang2021missformer}}.
    \label{fig:synapse_vis}
\end{figure}

The results in \Cref{fig:synapse_vis} show that the performance of MISSFormer for prediction and segmentation of edges in pure Transformer network structures is more accurate compared to TransUNet and Swin-Unet. Comparing MISSFormer and MISSFormer-S (MISSFormer without bridge), MISSFormer has fewer segmentation errors because the bridge is effective for integrating multi-scale information.

%  \subsubsection{TransDeepLab}
Inspired by the notable DeepLabv3 \citep{chen2017deeplabv3} which utilizes the Atrous Spatial Pyramid Pooling (ASPP) to learn multi-scale feature representations and depth-wise separable convolution to reduce the computational burden, the authors propose \textbf{TransDeepLab} \citep{azad2022transdeeplab} to combine the DeepLab network with the Swin-Transformer blocks. Applying the Swin-Transformer module with windows of multiple sizes enables the fusion of multi-scale information with a lightweight model.

\begin{figure}[h]
    \centering
    \includegraphics[width=0.49\textwidth]{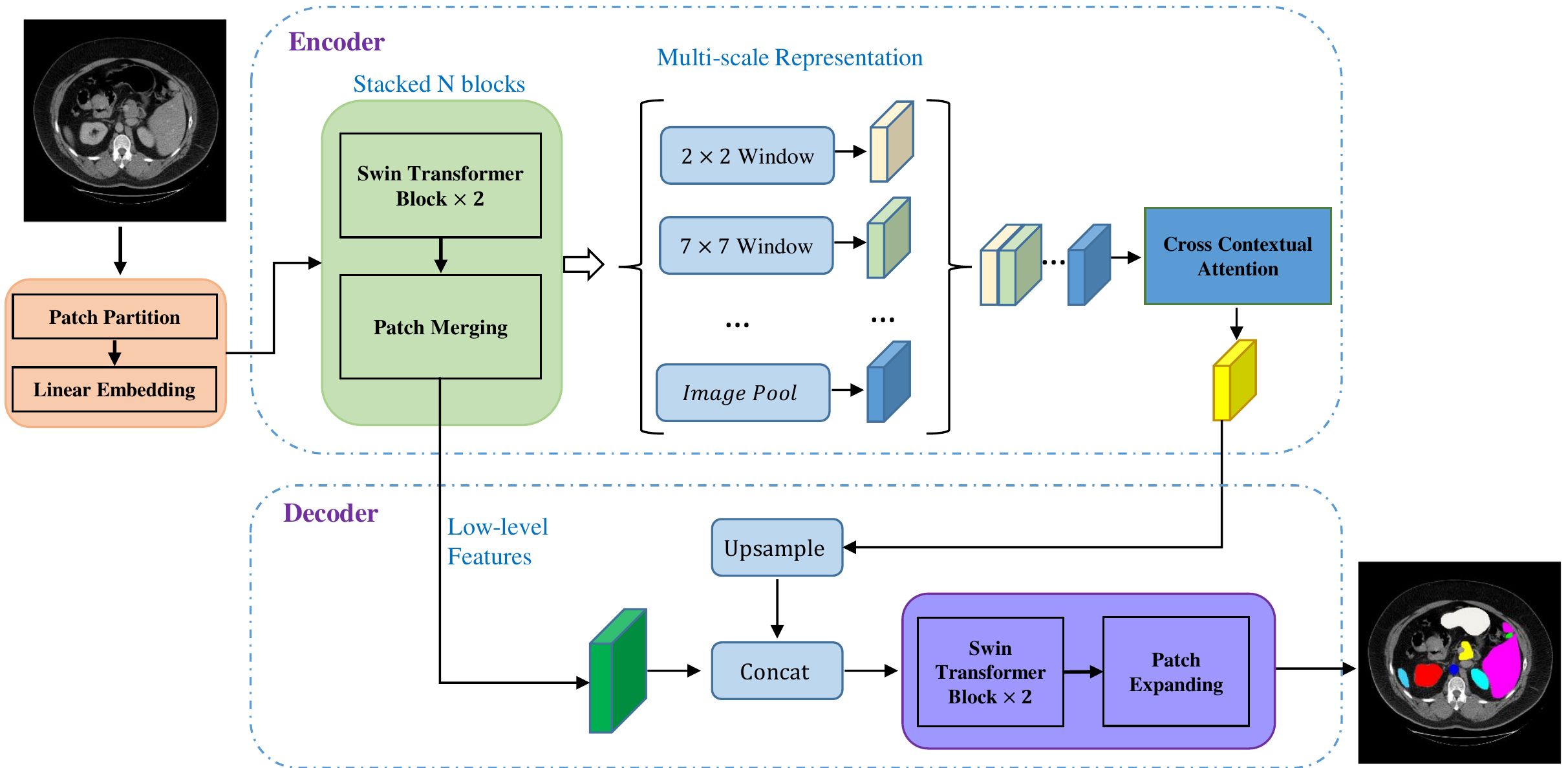}
    \caption{The overview architecture of TransDeepLab, which comprises encoder and decoder built on Swin-Transformer blocks. It is the pure Transformer-based extension of DeepLabv3++ \citep{azad2022transdeeplab}. }
    \label{fig:TransDeepLab}
\end{figure}

TransDeepLab is a pure Transformer-based DeepLabv3+ architecture, as shown in \Cref{fig:TransDeepLab}. The model builds a hierarchical architecture based on the Swin-Transformer modules. TransDeepLab first employs $N$ stacked Swin-Transformer blocks to model the input embedded images into a deep-level feature space. 2D medical images are first to split into non-overlapping patches of dimension $C$ and size $4 \times 4$. The ensuing Swin-Transformer block learns local semantic information and global contextual dependencies of the sequence of patches.
Then, the authors introduce windows with different sizes to process the output of the Swin-Transformer and fuse the resulting multi-scale feature layers, which are then passed through Cross Contextual attention. This design, referred to as Spatial Pyramid Pooling (SSPP) block, replaces the original Atrous Spatial Pyramid Pooling (ASPP) module exploited in DeepLabV3. A cross-contextual attention mechanism is utilized to explore the multi-scale representation after fusion. This attention module applies channel attention and spatial attention to the output from windows of each size (from each layer of the spatial pyramid). Finally, in the decoder part, the low-level features from the encoder are concatenated with the multi-scale features extracted by Cross Contextual Attention after bilinear upsampling. The last two Swin-Transformer blocks and the patch expanding module generate the final prediction masks.

\subsection{Hybrid Models} \label{sec:seg_Hybrid_Models}
Hybrid Transformers concatenate Transformer blocks with convolution layers to extract local details and long-range dependencies. We further classify this category into \textit{Transformer: Encoder}, \textit{Transformer: Decoder} and \textit{Transformer: skip connection} according to the position of the combined module in the U-Net architecture.

\subsubsection{Transformer: Encoder}
Starting with TransUNet \citep{chen2021transunet}, multiple methods in the medical image segmentation field adopt the self-attention mechanism in the encoder.

% \subsubsection{TransUNet: Transformers Make Strong Encoders for Medical Image Segmentation}
Transformers have developed as an alternative architecture for modeling global context that exclusively relies on attention mechanisms instead of convolution operators. However, its inner global self-attention mechanism induces missing low-level details. Direct upsampling cannot retrieve the local information, which results in inaccurate segmentation results. The authors propose the \textbf{TransUNet} architecture, a hybrid approach that integrates CNN-Transformer hybrid as the encoder and cascaded upsampler as the decoder, combining the advantages of Transformer and U-Net to boost the segmentation performance by recovering localized spatial information.

\begin{figure}[h]
 \centering
 \includegraphics[width=0.5\textwidth]{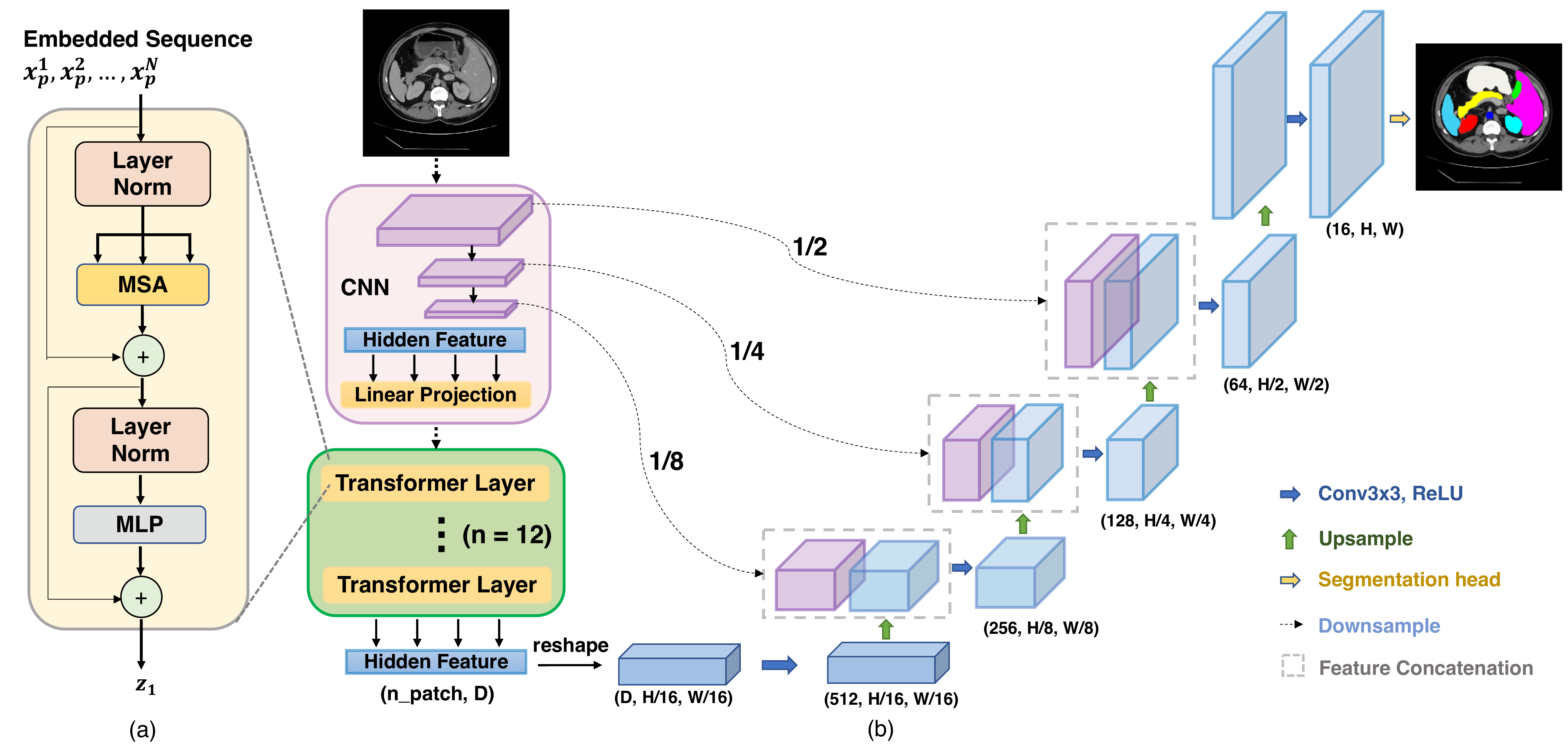}
 \caption{The overview architecture of the TransUNet \citep{chen2021transunet}. The Transformer layers are employed in the encoder part. The schematic of the Transformer is shown on the left.}
 \label{fig:TransUnet structure}
\end{figure}

The framework of the TransUNet is illustrated in \Cref{fig:TransUnet structure}. The proposed encoder initially employs CNN as a feature extractor to build a feature map for the Transformer input layer, rather than the Transformer directly projecting the raw tokenized picture patches to latent embedding space. In this way, the intermediate CNN feature maps of different resolutions can be saved and utilized in the following process.

For the decoder, the Cascaded Upsampler (CUP) is proposed to replace naive bilinear upsampling, applying several upsampling blocks to decode the hidden feature and output the final segmentation result.
Finally, the hybrid encoder and the CUP constitute the overall architecture with skip connections to facilitate feature aggregation at different resolution levels. This strategy can compensate for the loss of local fine-grained details caused by the Transformer encoder and merge the encoded global information with the local information contained in intermediate CNN feature maps.

The experiments show that TransUNet significantly outperforms the model consisting of pure Transformer encoder and naive upsampling, as well as the ViT-hybrid model without skip connections \citep{chen2021transunet}. Comparisons with prior work \citep{ronneberger2015unet,oktay2018attunet} also demonstrate the superiority of TransUNet over competing CNN-based approaches in terms of both qualitative visualization and the quantitative evaluation criteria (i.e.average DSC and HD). TransUNet integrates the benefits of both high-level global contextual information and low-level details as an alternative approach for medical image segmentation.

% \subsubsection{TransBTS: Multimodal Brain Tumor Segmentation Using Transformer}

Wang et al. \citep{wang2021transbts} propose the encoder-decoder architecture, \textbf{TransBTS}, which leverages Transformer on learning global contextual information and merits the 3D CNN for modeling local details. In contrast to the concurrent Transformer-based model \citep{chen2021transunet}, which analyzes 3D medical volumetric data in a slice-by-slice manner, TransBTS also explores the local features along the depth dimension by processing all the image slices at once.

The network encoder initially employs a 3D CNN to capture volumetric spatial features, simultaneously downsampling the input 3D images, yielding compact volumetric feature maps. Each feature map is projected into a token and fed into the Transformer encoder to investigate the global relationships. The full-resolution segmentation maps are generated by the 3D CNN decoder after the progressive upsampling while using the feature embedding from the Transformer.
 For the encoder part, TransBTS first utilizes the $3\times3\times3$ convolution blocks with downsampling to process the 3D input medical image data, which boosts the effective embedding of rich local 3D features a cross spatial and depth dimensions into the low-resolution feature representation $F$. They apply a linear projection to the feature representation $F$ to obtain the sequence $f$, which is then integrated with position embeddings, as the input for the Transformer encoder. The Transformer encoder consists of multiple Transformer layers, each of which comprises a Multi-Head Attention(MHA) block and a Feed-Forward Network(FFN). The output sequence of the Transformer encoder passes through the feature mapping module to be reshaped to a 4D feature map $Z$ of the same dimension as $F$. The approach employs cascaded upsampling and convolution blocks to progressively restore the segmentation predictions at the original resolution. Furthermore, skip connections combine the fine-grained details of local information with the decoder modules, resulting in more accurate segmentation masks.

The authors conduct comparisons between the proposed TransBTS and the closest method TransUNet \citep{chen2021transunet}. TransUNet essentially processes 3D medical images slice by slice, while TransBTS is a 3D model that explores the continuous interaction through the depth dimension by processing a 3D medical image in a single pass. In contrast to TransUNet, which adopts pre-trained ViT models on other large-scale datasets, TransBTS is trained on the dataset for the specified task without relying on pre-trained weights.

The framework is evaluated on the Brain Tumor Segmentation (BraTS) 2019 challenge and 2020 challenge. Compared to the 3D U-Net baseline, TransBTS achieves a significant enhancement in segmentation. The prediction results indicate the improved accuracy and the superiority of modeling long-range dependencies.

% \begin{figure}[h]
%  \centering
%  \includegraphics[width=0.4\textwidth]{images/TransBTS.pdf}
%  \caption{Overview of the TransBTS \citep{wang2021transbts} architecture. The output of the 3D CNN Encoder is fed into the linear projection. Then the Transformer encoder models the long-range dependency in the global space. The 3D CNN decoder performs upsampling and generates pixel-level prediction results \citep{wang2021transbts}.}
%  \label{fig:TransBTS structure}
% \end{figure}

% \subsubsection{TransFuse: Fusing Transformers and CNNs for Medical Image Segmentation}

Previous approaches \citep{zhang2021transfuse} primarily focus on replacing convolution operation with Transformer layers or consecutively stacking the two together to address the inherent lack of pure Transformer-based models to learn local information. In this study, the authors propose a new strategy-\textbf{TransFuse} which consists of the CNN-based encoder branch and the Transformer-based branch in parallel fused with the proposed BiFusion module, thus further exploring the benefits of CNNs and Transformers.
The construction of the Transformer branch is designed in the typical encoder-decoder manner. The input images are first split into non-overlapped patches. The linear embedding layer then projects the flattened patches into the raw embedding sequence which is added to the learnable position embedding of the same dimension. The obtained embeddings are fed into the Transformer encoder, which comprises $L$ layers of MSA and MLP. The output of the last layer of the Transformer encoder is passed through layer normalization to obtain the encoded sequence.

The decoder part utilizes the same progressive upsampling (PUP) approach as SETR \citep{zheng2021rethinking}. The encoded sequence is first reshaped back to a sequence of 2D feature maps. Then they employ two stacked upsampling-convolution layers to restore the feature scales.
The feature maps with different spatial resolutions generated by each upsampling-convolution layer are retained for the subsequent fusion operation.
For the CNN branch, the approach discards the last layer of the traditional CNNs architecture and combines the information extracted from the CNNs with the global contextual features obtained from the Transformer branch. A shallower model is yielded as a result of this design, avoiding the requirement for extremely deep models that exhaust resources to get long-range dependencies. For instance, there are five blocks in a typical ResNet-based network where only the outputs of the 4th, 3rd, and 2nd layers are saved for the following fusion with the feature maps from the Transformer branch.

The BiFusion module is proposed to fuse the features extracted from the two branches mentioned above to predict the segmentation results of medical images. The global features from the Transformer branch are boosted by the channel attention proposed in SE-Block \citep{hu2018squeeze}. Meanwhile, the feature maps from the CNN branch are filtered by the spatial attention which is adopted in the CBAM \citep{woo2018cbam} block to suppress the irrelevant and noisy part and highlight local interaction. Then the Hadamard product is applied to the features from the two branches to learn the interaction between them. They concatenate the interaction feature $b^i$ with attended features $\hat{t}^i$ and $\hat{g}^i$ and feed the results through a residual block to produce the feature $f^i$, which successfully models both the global and local features at the original resolution. Finally, the segmentation prediction is generated by integrating the $f^i$ from different BiFusion modules via the attention-gated (AG) \citep{schlemper2019attention} skip connection.

They evaluate the performance of three variants of TransFuse on four segmentation tasks with different imaging modalities and target sizes. TransFuse-S is constructed with ResNet-34 (R34) and 8-layer DeiT-Small (DeiT-S)  \citep{DeiT}. Besides, TransFuse-L is composed of Res2Net-50 and 10-layer DeiT-Base (DeiT-B). TransFuse-L* is implemented based on ResNetV2-50 and ViT-B \citep{dosovitskiy2020image}. For polyp segmentation, Transfuse-S/L outperforms significantly the CNN baseline models with fewer parameters and faster running time. TransFuse-L* also achieves the best performance among the previous SOTA Transformer-based methods with a faster speed for inference. It runs at 45.3 FPS and about 12\% faster than TransUNet. The experiments for other segmentation tasks also show the superiority of the segmentation performance.

% \begin{figure}[h]
% 		\centering
% 		\includegraphics[width=0.4\textwidth]{images/TransFuse.pdf}
% 	\caption{The overview architecture of TransFuse is composed of two parallel branches: Transformer branch (left) and CNN branch (right). The feature outputs from the two branches are fused by BiFusion \citep{zhang2021transfuse}.}
% 	\label{fig:TransFuse structure}
% \end{figure}

Despite the powerful results of applying Transformers to segmentation tasks \citep{wang2020axial, zheng2021rethinking}, the dilemma is that properly training existing Transformer-based models requires large-scale datasets, whereas the number of images and labels available for medical image segmentation is relatively limited. To overcome the difficulty, \textbf{MedT} \citep{valanarasu2021medt} proposes a gated position-sensitive axial attention mechanism where the introduced gates are learnable parameters to enable the model to be applied to a dataset of arbitrary size. Furthermore, they suggested a Local-Global(LoGo) training strategy to improve the segmentation performance by operating on both the original image and the local patches.

\begin{figure}[t]
		\centering
		\includegraphics[width=0.5\textwidth]{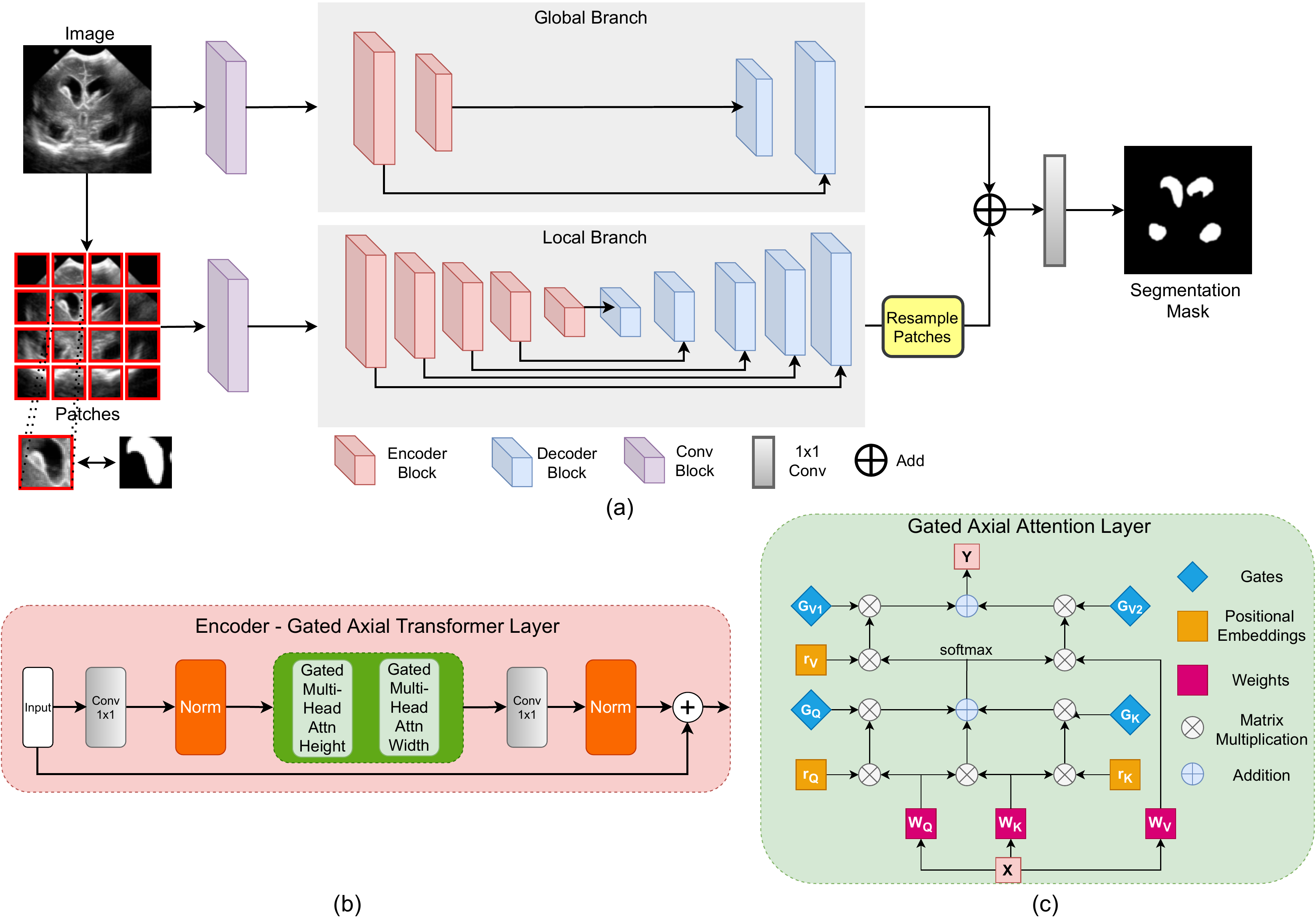}
	\caption{Overview of the MedT \citep{valanarasu2021medt} architecture. The network uses the LoGo strategy for training. The upper global branch utilizes the first fewer blocks of the Transformer layers to encode the long-range dependency of the original image. In the local branch, the images are converted into small patches and then fed into the network to model the local details within each patch. The output of the local branch is re-sampled relying on the location information. Finally, a $1 \times 1$ convolution layer fuses the output feature maps from the two branches to generate the final segmentation mask.}
	\label{fig:MedT structure}
\end{figure}

The main architecture of MedT, as shown in \Cref{fig:MedT structure} (a), is composed of 2 branches: a shallow global branch that works on the original resolution of the entire image, and a deep local branch that acts on the image patches. Two encoder blocks and two decoder blocks comprise the global branch, which is sufficient to model long-range dependencies. In the local branch, the original image is partitioned into 16 patches and each patch is feed-forwarded through the network. The output feature maps are re-sampled based on their locations to obtain the output feature maps of the branch. Then the results generated from both branches are added and fed into a $1\times1$ convolution layer to produce the output segmentation mask. The LoGo training strategy enables the global branch to concentrate on high-level information and allows the local branch to learn the finer interactions between pixels within the patch, resulting in improved segmentation performance.

\Cref{fig:MedT structure} (b) and (c) illustrate the gated axial Transformer layer, which is used as the main building block in MedT, and the feed-forward structure in it. They introduced four learnable gates $G_{V1}, G_{V2}, G_{Q}, G_{K}\in \mathbb{R}$ that control the amount of information the positional embeddings supply to key, query, and value. Based on whether a relative positional encoding is learned accurately or not, the gate parameters will be assigned weights either converging to 1 or some lower value. The gated mechanism can control the impact of relative positional encodings on the encoding of non-local context and allows the model to work well on any dataset regardless of size.

Unlike the fully-attended baseline \citep{wang2020axial}, MedT trained on even smaller datasets outperforms the convolutional baseline and other Transformer-based methods. In addition, improvements in medical segmentation are also observed since the proposed method takes into account pixel-level dependencies.

% \subsubsection{UNETR: Transformers for 3D Medical Image Segmentation}
In contrast to multiple proposed methods \citep{zheng2021rethinking, chen2021transunet, valanarasu2021medt, zhang2021transfuse} that investigate the task of 2D medical image segmentation, \textbf{UNETR} \citep{hatamizadeh2021unetr} proposes a novel Transformer-based architecture for 3D segmentation which employs the Transformer as the encoder to learn global contextual information from the volumetric data. In addition, unlike the previous frameworks proposed for 3D medical image segmentation \citep{xie2021cotr,wang2021transbts}, the encoded feature from the Transformer of this proposed model is directly connected to a CNN-based decoder via skip connections at different resolution levels.
The U-shaped UNETR comprises a stack of Transformers as the encoder and a decoder coupling with it by skip connections. They begin by generating the 1D sequence of patches by splitting the 3D input volume in a non-overlapping manner. The flattened input patches are then passed through a linear projection layer to yield $K$ dimensional patch embeddings. They attach a 1D learnable positional embedding to each patch embedding taking into account the spatial information of the extracted patches. After the embedding layer, the global multi-scale representation is captured using Transformer blocks composed of multi-head self-attention modules and multilayer perceptron layers. They resize and project the sequence representation extracted from the Transformer at different resolutions for use in the decoder in order to retrieve spatial information of the low-level details.

In the expanding pattern of the framework, the proposed CNN-based decoder combines the output feature of different resolutions from the Transformer with upsampled feature maps to properly predict the voxel-wise segmentation mask at the original input resolution.

The paper claims UNETR achieves new state-of-the-art performance on all organs compared against CNN \citep{isensee2021nnunet,chen2018encoder,tang2021high,zhou2019prior} and competing for Transformer-based \citep{xie2021cotr,chen2021transunet,zheng2021rethinking} baselines on BTCV dataset, with significant improvement in performance on small organs in particular. In addition, it outperforms the closest methodologies on brain tumor and spleen segmentation tasks in the MSD dataset. UNETR shows the superiority of learning both global dependencies and fine-grained local relationships in medical images.

% \begin{figure}[h]
% 		\centering
% 		\includegraphics[width=0.4\textwidth]{images/unetr.pdf}
% 	\caption{Overview of the UNETR \citep{hatamizadeh2021unetr} architecture. The 3D input image is partitioned into a set of patches and transformed into the embedding space via the linear projection layer. Then the sequence is fed into the Transformer encoder including 12 layers with multi-head attention modules. The encoded features of Layer 3, Layer 6, Layer 9 and Layer 12 are concatenated with the decoder to fuse the multi-scale information.}
% 	\label{fig:unetr structure}
% \end{figure}

\begin{figure}[h]
 \centering
 \includegraphics[width=0.5\textwidth]{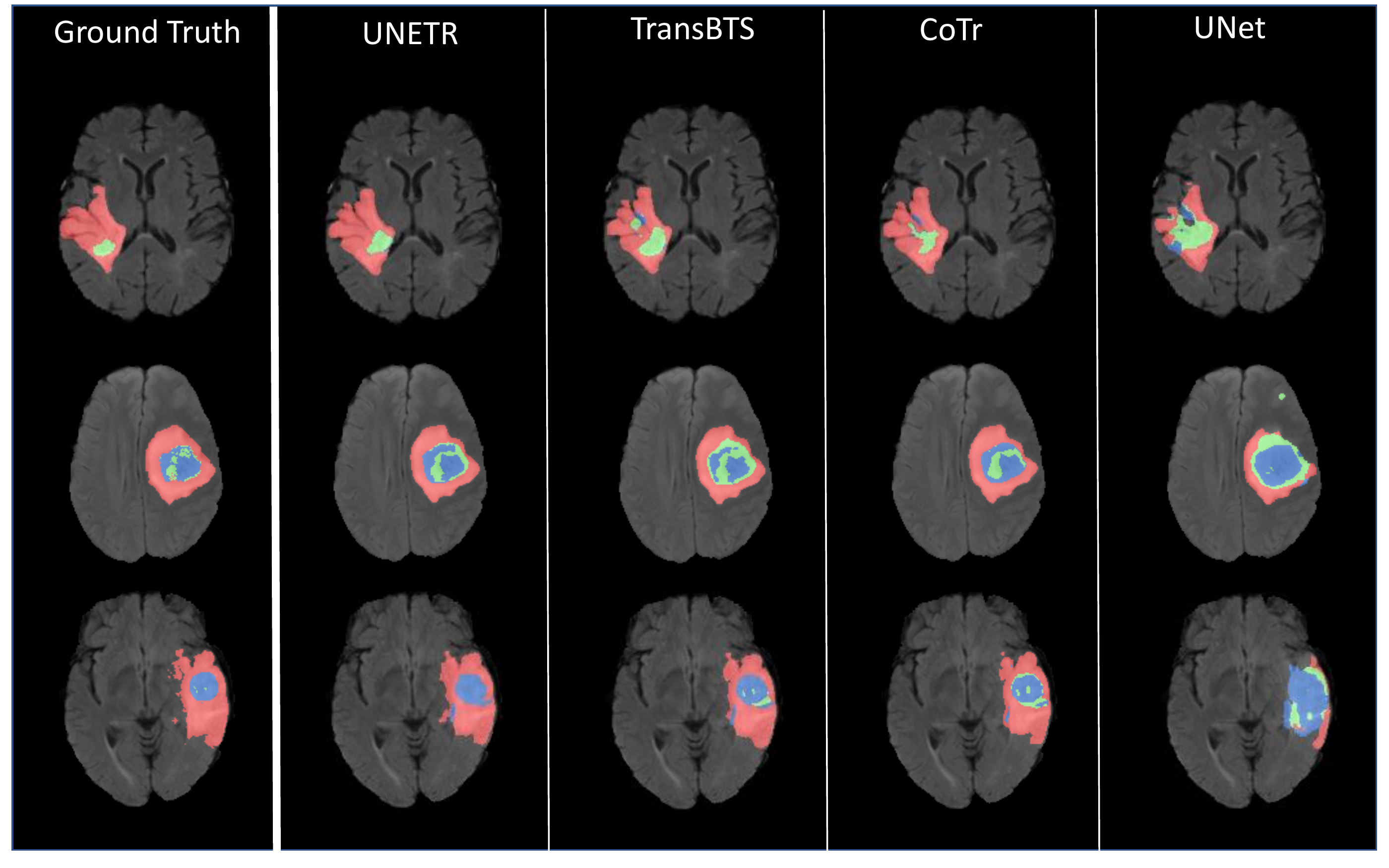}
 \caption{Comparison of visualization of brain tumor segmentation on the MSD dataset. The whole tumor (WT) includes a combination of red, blue, and green regions. The union of red and blue regions demonstrates the tumor core (TC). The green regions indicate the enhanced tumor core (ET) \citep{hatamizadeh2021unetr}.}
 \label{fig:unetr_overlay}
\end{figure}

\Cref{fig:unetr_overlay} presents qualitative
segmentation comparisons for brain tumor segmentation on
the MSD dataset between UNETR \citep{hatamizadeh2021unetr}, TransBTS \citep{wang2021transbts}, CoTr \citep{xie2021cotr} and U-Net \citep{ronneberger2015unet}. It can be seen that the details of the brain tumor are captured well by UNETR \citep{hatamizadeh2021unetr}.

% \subsubsection{Swin UNETR: Swin Transformers for Semantic Segmentation of Brain Tumors in MRI Images}

As opposed to other methods that attempted to utilize the Transformer module as an additional block beside the CNN-based components in the architectures, UNETR \citep{hatamizadeh2021unetr} leverages the Transformer as the encoder instead of the CNN-based encoder. The Swin Transformer \citep{liu2021swin} is a hierarchical visual Transformer featuring an efficient shift-window partitioning scheme for computing self-attention. Inspired by these two approaches, a novel model termed \textbf{Swin} \textbf{Unet} \textbf{Tr}ansformer (\textbf{Swin UNETR}) \citep{hatamizadeh2022swinunetr} is proposed for brain tumor segmentation in this work.

The proposed framework applies a U-shape architecture with the Swin Transformers as the encoder and a CNN-based module as the decoder connected to the encoder via skip connections at different resolution levels. The model initially converts 3D MRI images with four channels to non-overlapping patches and creates windows of a specific size with a patch partition layer.

The Swin UNETR encoder is composed of 4 stages. Each stage comprises 2 Transformer blocks and a patch merging layer. In the Transformer blocks, the self-attention is computed with a shifted windowing mechanism. Swin UNETR employs the windows of size $M \times M \times M$ to partition the 3D token with resolution of $H' \times W' \times D' $ into regions of $\lceil \frac{H'}{M} \times \frac{W'}{M} \times \frac{D'}{M} \rceil$ at layer $l$. The partitioned window regions are then shifted by $(\lfloor \frac{M}{2} \rfloor,\lfloor \frac{M}{2} \rfloor, \lfloor \frac{M}{2} \rfloor )$ voxels at the following $l+1$ layer. The patch merging layer after the Transformer components reduces the resolution of feature maps by a factor of two and concatenates them to form a feature embedding with the doubled dimensionality of the embedding space.

For the decoder of the architecture, the output feature representations of the bottleneck are reshaped and passed through the residual block containing two convolutional layers. The subsequent deconvolutional layer increases the resolution of feature maps by a factor of 2. The outputs are then concatenated with the outputs of the previous stage and fed into another residual block. After the resolutions of the feature maps are restored to the original $H' \times W' \times D'$, a head is utilized to generate the final segmentation predictions.

The authors conduct the experiments to compare Swin UNETR against the previous methodologies SegResNet \citep{myronenko2018segresnet}, nn-UNet \citep{isensee2020nnunet}and TransBTS \citep{wang2021transbts} in this work. The results demonstrate that the proposed model has prominence as one of the top-ranking approaches in the BraTS 2021 challenge. It is due to the better capability of learning multi-scale contextual information and modeling long-range dependencies by Swin Transformers in comparison to regular Transformers with a fixed resolution of windows.

% \begin{figure}[h]
%  \centering
%  \includegraphics[width=0.5\textwidth]{images/SwinUNETR.pdf}
%  \caption{ Illustration of the Swin UNETR architecture. The model processes 3D MRI images with 4 channels. The encoder employs Swin Transformer blocks for computing self-attention. The feature maps at different scales are connected to the corresponding stage in the decoder. The 3 channels of the segmentation output refer to ET, WT, and TC sub-regions respectively \citep{hatamizadeh2022swinunetr}.}
%  \label{fig:SwinUNETR overview}
% \end{figure}

Apart from the architecture of Swin UNETR \citep{hatamizadeh2022swinunetr}, they also proposed a hierarchical encoder for pre-training the encoder of Swin UNETR \citep{tang2022selfswinunetr}, called \textbf{Swin UNETR++}. The pre-training tasks are tailored to learning the underlying pattern of human anatomy. Various proxy tasks such as image inpainting, 3D rotation prediction, and contrastive learning are leveraged to extract the features from the naturally consistent contextual information in 3D radiographic images such as CT. Specifically, inpainting enables learning the texture, structure, and relationship of masked regions to their surrounding context. Contrastive learning is utilized to recognize separate regions of interest (ROIs) of different body compositions. The rotation task creates a variety of sub-volumes that can be used for contrastive learning. To validate the effectiveness of the pre-trained encoder, they fine-tuned the pre-trained Swin UNETR on the downstream segmentation task and achieved SOTA performance on BTCV dataset \citep{Synapse}.

\subsubsection{Transformer: Decoder}
Another direction is to modify the decoder of the U-shape structure to aggregate the Transformer-CNN-based modules. 
%\subsubsection{SegTran: Medical Image Segmentation Using Squeeze-and-Expansion Transformers}

\begin{figure}[h]
 \centering
 \includegraphics[width=0.5\textwidth]{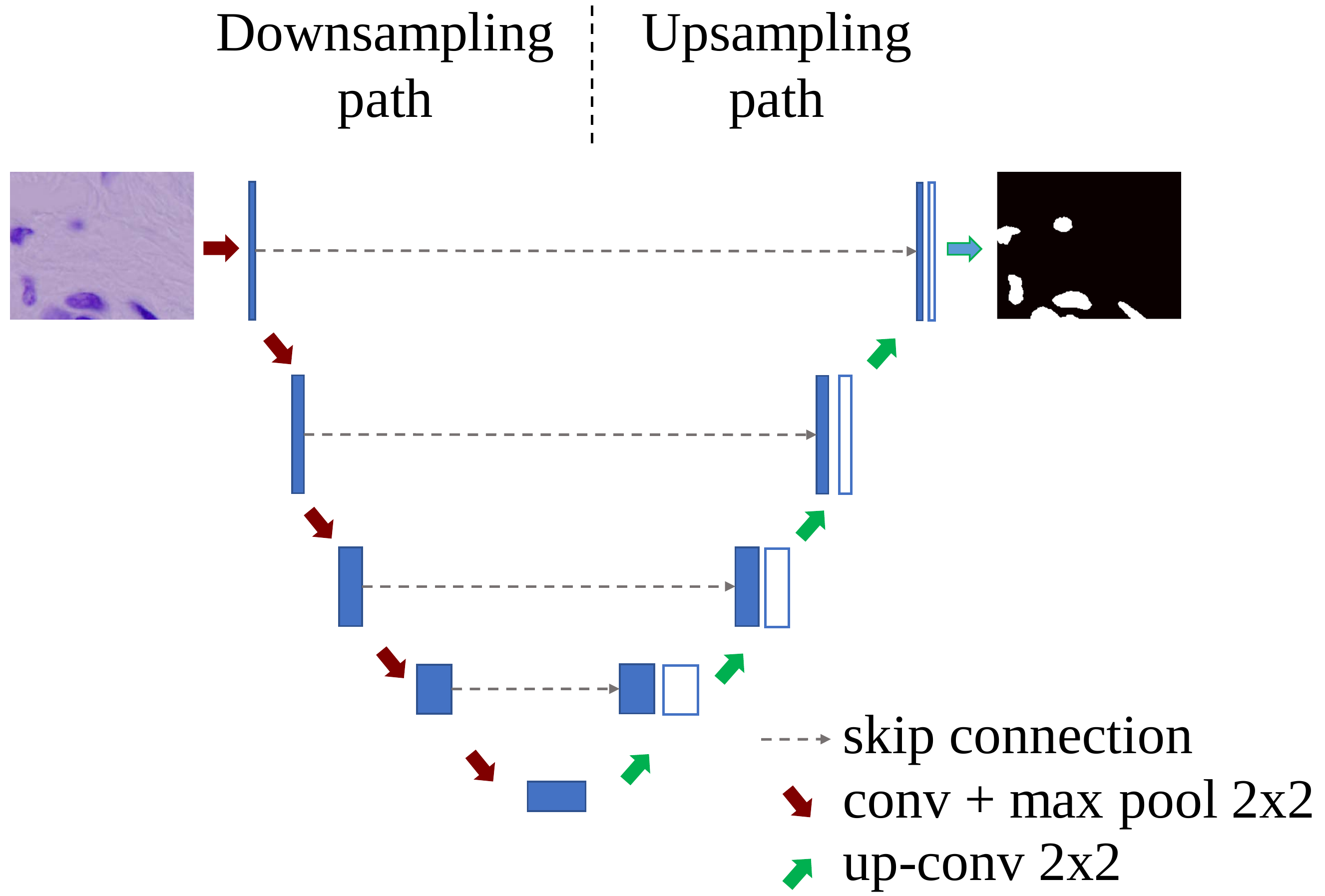}
 \caption{Segtran network extracts image features using a CNN backbone and combines the features with the position encoding of pixels flattened into a series of local feature vectors. Multiple squeezed and extended transform layers are stacked to process the local feature vectors. Finally, an output FPN after the Transformer upsamples the features to generate the final prediction \citep{li2021segtran}.}
 \label{fig:SegTran structure}
\end{figure}

% As Transformers were originally designed for Natural Language Processing (NLP) tasks, several aspects could be adapted to perform medical image segmentation better. First, The $N \times N$ attention matrix obtained by combining input features is large, which makes it vulnerable to noises and overfitting. Second, the multi-head Transformer only has one set of transformations after concatenation, which may not have sufficient capacity to extract features from data variation.
In the \textbf{Segtran} framework \citep{li2021segtran}, as illustrated in \Cref{fig:SegTran structure}, Squeeze-and-Expansion Transformer is proposed to "squeeze" the attention matrix and aggregate multiple sets of contextualized features from the output.
A novel Learnable Sinusoidal Position Encoding is also employed to impose the continuity inductive bias for images.
The Segtran consists of five components: a CNN backbone to extract image features, 2) input/output feature pyramids to do upsampling, 3) the Learnable Sinusoidal Positional Encoding, 4) Squeeze-and-Expansion Transformer layers to contextualize features, and 5) a segmentation head.
The pretrained CNN backbone is first utilized to learn feature maps from the input medical images.
Since the input features to Transformers are of a low spatial resolution, the authors increase their spatial resolutions with an input Feature Pyramid Network (FPN) \citep{liu2018path} to upsample the feature maps by bilinear interpolation.
Then the proposed Learnable Sinusoidal Positional Encoding is added to the visual features to inject spatial information. In contrast to the previous two mainstream PE schemes \citep{carion2020end,dosovitskiy2020image}, the new positional embedding vector, a combination of sine and cosine functions of linear transformations of $(x, y)$, brings in the continuity bias with adaptability.
The equation of the encoding strategy varies gradually with pixel coordinates. Thus, close units receive similar positional encodings, increasing the attention weights between them towards higher values. The encoding vectors generated from the addition of positional encodings and visual features are then fed into the Transformer.

The novel Transformer architecture combines Squeezed Attention Block (SAB) \citep{lee2019set} with an Expanded Attention Block. Here this method employs the Induced Set Attention Block (ISAB) proposed by \citep{lee2019set} as a squeezed attention block. The Squeezed Attention Block computes attention between the input and inducing points and compresses the attention matrices to lower rank matrices, reducing noises and overfitting.
The Expanded Attention Block (EAB), a mixture-of-experts model, outputs $N_m$ sets of complete contextualized features from $N_m$ modes. Each mode is an individual single-head Transformer and shares the same feature space with each other. That is as opposed to multi-head attention in which each head outputs an exclusive feature subset. All features are then aggregated into one set using dynamic mode attention. The dynamic mode attention can be obtained by doing a linear transformation of each mode feature and taking softmax over all the modes.

Compared with representative existing methods in the experiments, Segtran consistently achieved the highest segmentation accuracy and exhibited good cross-domain generalization capabilities.

\subsubsection{Transformer: Skip Connection}
In this section, Transformer blocks are incorporated into the skip connections to facilitate the transmission of detailed information from the encoder to the decoder.
% \subsubsection{CoTr: Efficiently Bridging CNN and Transformer for 3D Medical Image Segmentation}

Although Transformer-based methods overcome the limitation of capturing long-range dependency, they present extreme computational and spatial complexity in analyzing high-resolution volumetric image data. Some studies \citep{carion2020end, chen2021transunet} employ hybrid structures, fusing CNN with Transformer in an attempt to reduce the training requirement on huge datasets. The recent approach, TransUNet \citep{chen2021transunet}, shows good performance. However, it is difficult to optimize the model due to the inner self-attention mechanism of the vanilla Transformer. First, it takes a long time to train the attention, which is caused by initially distributing attention uniformly to each pixel within the salient regions \citep{vaswani2017attention}. Second, a vanilla Transformer struggles to handle multi-scale and high-resolution feature maps due to its high computational cost.

Motivated by this, \citep{xie2021cotr} proposes a novel encoder-decoder framework, \textbf{CoTr}, which bridges CNN and Transformer. The architecture exploits CNN to learn feature representations. An efficient deformable self-attention mechanism in the Transformer is designed to model the global context from the extracted feature maps, which reduces the computational complexity and enables the model to process high-resolution features. The final segmentation results are generated by the decoder.

\begin{figure}[h]
 \centering
 \includegraphics[width=0.5\textwidth]{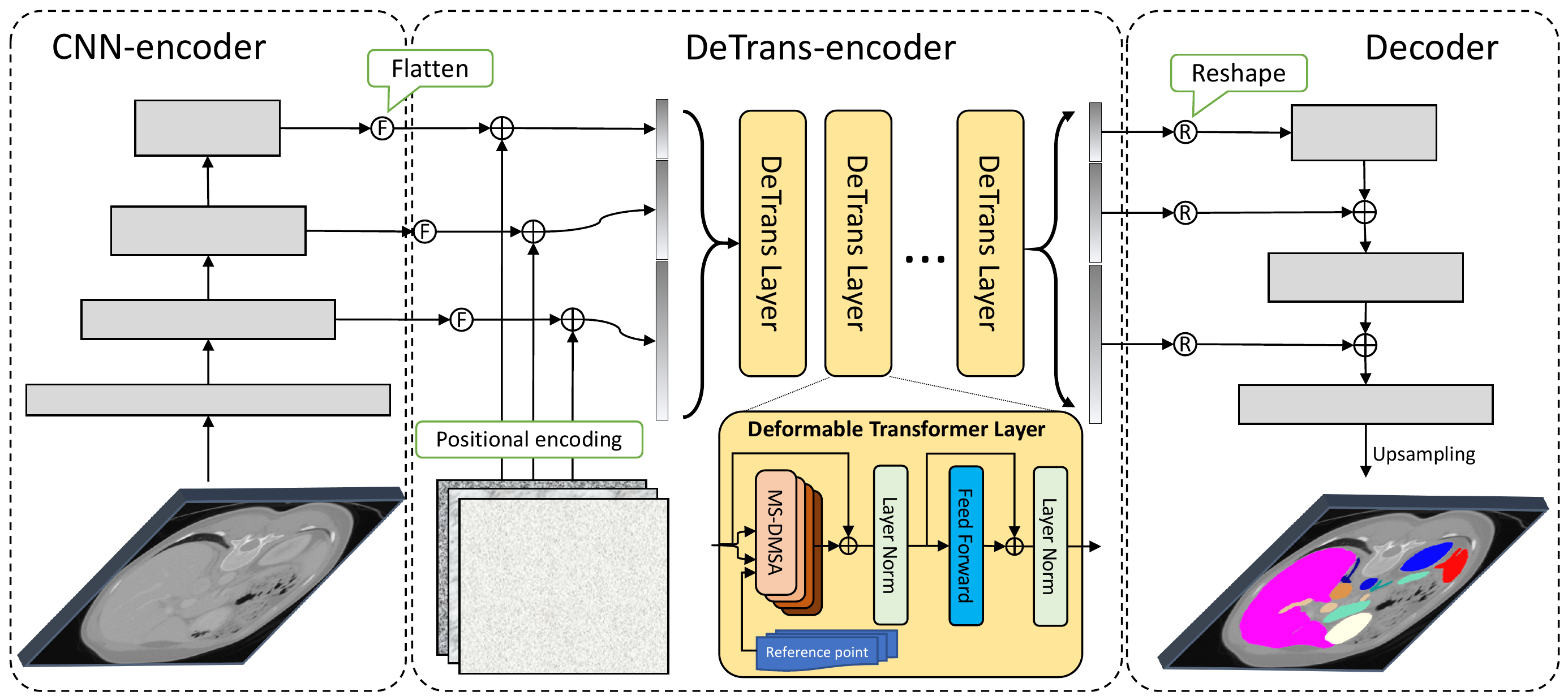}
 \caption{Overview of the CoTr \citep{xie2021cotr} architecture. It is composed of a CNN-encoder, a DeTrans-encoder and a decoder. The CNN-encoder models the local information of the input images and provides the outputs at each stage. The outputs of different resolutions are flattened, fused and passed through the Deformable Transformer Layers along with positional encoding. The decoder reshapes the processed sequences from the DeTrans-encoder and produces the final predictions.}
 \label{fig:cotr structure}
\end{figure}

As shown in \Cref{fig:cotr structure}, the DeTrans-encoder consists of an input-to-sequence layer and multiple DeTrans Layers. The input-to-sequence layer first flattens the feature maps at different resolutions extracted from the CNN-encoder into 1D sequences $\{f_l\}_{l=1}^L$. Then the corresponding 3D positional encoding sequence $p_l$ is supplemented with each of the flattened sequences $f_l$ to complement the spatial information. The combined sequence is fed as the input into the DeTrans Layers. Each of the DeTrans Layers is a composition of an MS-DMSA and a Feed-Forward Network (FFN).
In contrast to the self-attention mechanism which casts attention to all the possible locations, the proposed MS-DMSA layer only attends to a small set of key sampling locations around a reference location. As a result, it can achieve faster convergence and lower computational complexity. The skip connection is utilized after each DeTrans Layer to preserve the low-level details of local information.
The output of the DeTrans-encoder is successively upsampled by the pure CNN encoder to restore the original resolution. Besides, they apply skip connections and a deep supervision strategy to add fine-grained details and auxiliary losses to the prediction outputs.

The results of experiments indicate that CoTr with the hybrid architecture has superior of performance over the models with pure CNN encoder or pure Transformer encoder. It also outperforms other hybrid methods like TransUNet \citep{chen2021transunet} in processing multi-scale 3D medical images with reduced parameters and complexity.

\begin{figure}[h]
 \centering
 \includegraphics[width=0.5\textwidth]{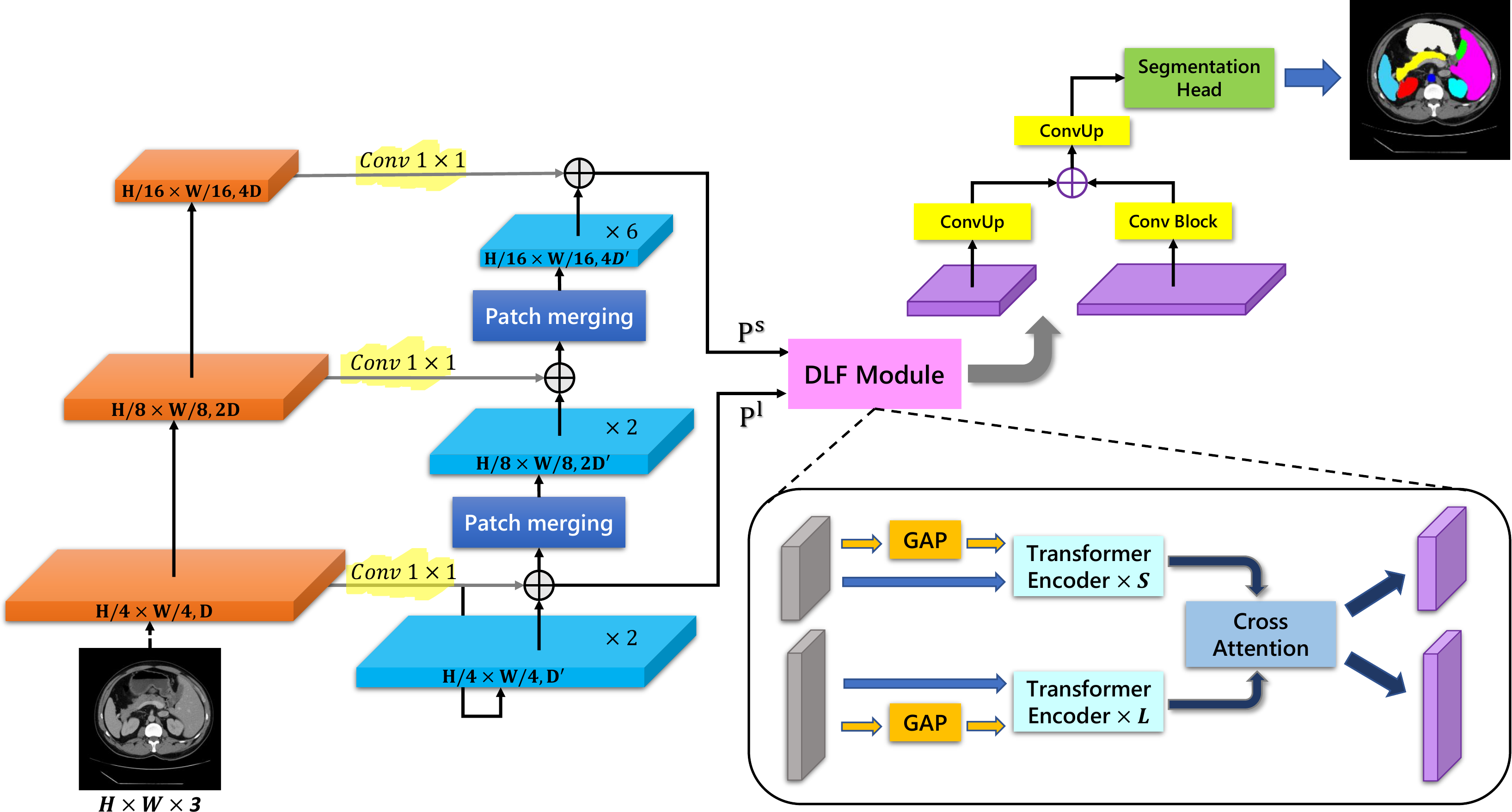}
 \caption{HiFormer comprises the CNN-Transformer encoder, the CNN-based decoder and the Double-Level Fusion Module (DLF). The feature layers of the shallowest level $p^l$ and of the deepest level $p^s$ are fed into the DLF module for the fusion of hierarchical information. Blue blocks and orange blocks refer to Swin Transformer and CNN modules, respectively \citep{heidari2022hiformer}.}
 \label{fig:hiformer structure}
\end{figure}

% \subsubsection{HiFormer}
\textbf{HiFormer} \citep{heidari2022hiformer} is proposed to aggregate a fusion module in the skip connections to learn richer representations.
\Cref{fig:hiformer structure} demonstrates the end-to-end network structure of the strategy that incorporates the global dependencies learned with the Swin Transformer and the detailed local features extracted by the CNN modules.
The encoder is composed of two hierarchical CNN, Swin Transformer modules and the novel Double-Level Fusion module (DLF module).
First, medical images are fed into a CNN module to obtain a local fine-grained semantic representation. After the CNN layer catches the shallow feature layers, HiFormer introduces the Swin Transformer modules to complement the global feature information. The Swin Transformer module employs windows of different sizes to learn the dependencies between multiple scales. To reuse the shallow and deep multi-scale feature information in the encoder, HiFormer designs a novel skip connection module, the DLF module. The deep-level semantic and shallow-level localization information are fed into the DLF module and fused by the cross-attention mechanism. Finally, both generated feature maps are passed into the decoder to produce the final segmentation prediction results.
The experiments conducted on the Synapse dataset \citep{Synapse}, SegPC \cite{gupta2023segpc}, and ISIC 2017 dataset \citep{codella2018skin} demonstrate the superiority of the learning ability of HiFormer. Moreover, the lightweight model with fewer parameters also exceeds CNN-based methods and previous Transformer-based approaches with lower computational complexity.

\subsection{Other Architectures} \label{sec:seg_Other_Architectures}

Most ViT-based models rely on pre-training of large natural image datasets to obtain pre-weights and then solve downstream tasks by transfer learning. Several works explore training in a self-supervised or semi-supervised manner to efficiently utilize medical image datasets of limited size or datasets without manual labels. Furthermore, some approaches apply Transformers to seek the design of architectures that implement medical image segmentation, instead of using the Transformers to act directly on the input image.
% \subsubsection{T-AutoML: Automated Machine Learning for Lesion Segmentation using Transformers in 3D Medical Imaging}

Unlike the previously proposed methods that employ the Transformers to act directly on the medical image for feature extraction, this method \citep{yang2021TAutoML} adopts the AutoML for automatically designing the network architecture without much human heuristics or assumptions, where the Transformer is applied to encode the embedding vector regarding the architecture configurations.
The approach reduces the workload of algorithm design by automatically estimating "almost" all the components of the framework instead of manually designing for the network and training strategies. That improves the model performance of segmentation simultaneously.

The proposed Transformer-based \textbf{T-AutoML} inspired by SpineNet \citep{du2020spinenet} leverages neural architecture search (NAS) with a larger search space to optimize the selection of the network connections. This framework can connect the feature maps at different spatial levels of the network with another one arbitrarily, compared with the previous methods that only search for the encoder-decoder U-shape networks \citep{liu2019autodeep,bae2019resource,kim2019scalable}.
The candidates of different blocks in the network consist of 3D residual blocks, 3D bottleneck blocks, and 3D axial-attention blocks. The residual blocks and bottleneck blocks are effective in alleviating the vanishing gradient. The axial-attention blocks are applied to model the long-range dependency in the 2D medical images. Another upsampling layer (linear interpolation) is utilized at the end of the architecture to produce the results of feature maps at the original volume size.

To search for the optimal architecture and training configuration, the authors first encode the necessary components in the search space to form a one-dimensional vector $v$. The search space contains candidates of different configurations with regard to data augmentation, learning rates, learning rate schedulers, loss function, the optimizer, the number and spatial resolution of blocks, and block types.

After the obtainment of the encoding vector $v$, the proposed new predictor predicts the binary relation of validation accuracy values between $vi$ and $v_j$. The predictor employs the Transformer encoder to encode the vector $v$ of varying lengths into feature maps of a fixed resolution. Then the feature maps are passed through the Multiple FC layers to generate the binary relation predictions denoted as $GT_{v_i,v_j}$. Since the predictor is designed for ranking the vectors with respect to the accuracy values and estimating the relations, the actual values of the predicted accuracy are not necessary to be calculated for each vector. Thus, the new predictor requires less overall training time.

The experiments indicate that the proposed method can achieve the state of the art(SOTA) in lesion segmentation tasks and shows superiority in transferring to different datasets.

% \begin{figure}[h]
%  \centering
%  \includegraphics[width=0.4\textwidth]{images/T-Auto.png}
%  \caption{The figure demonstrates (a) SegNet (b) U-Net (c) U-Net++ and (d) the proposed searched architecture. The different spatial resolutions of the feature maps are presented in different colors. The different shapes illustrate different block types \citep{yang2021TAutoML}.}
%  \label{fig:TAutoML structure}
% \end{figure}

% \subsubsection{CrossTeaching: Semi-Supervised Medical Image Segmentation via Cross Teaching between CNN and Transformer}
Despite the promising results achieved by the CNNs and Transformer-based methods with large-scale images, these approaches require expert labeling at the pixel/voxel level. Expensive time costs and manual annotation limit the size of the medical image dataset. Due to this dilemma, the proposed semi-supervised segmentation \citep{luo2021semi} provides a low-cost and practical scheme, called \textbf{Cross Teaching} between CNN and Transformer, to train effective models using a little amount of correctly labeled data and a large amount of unlabeled or coarsely labeled data.

\begin{figure}[h]
 \centering
 \includegraphics[width=0.5\textwidth]{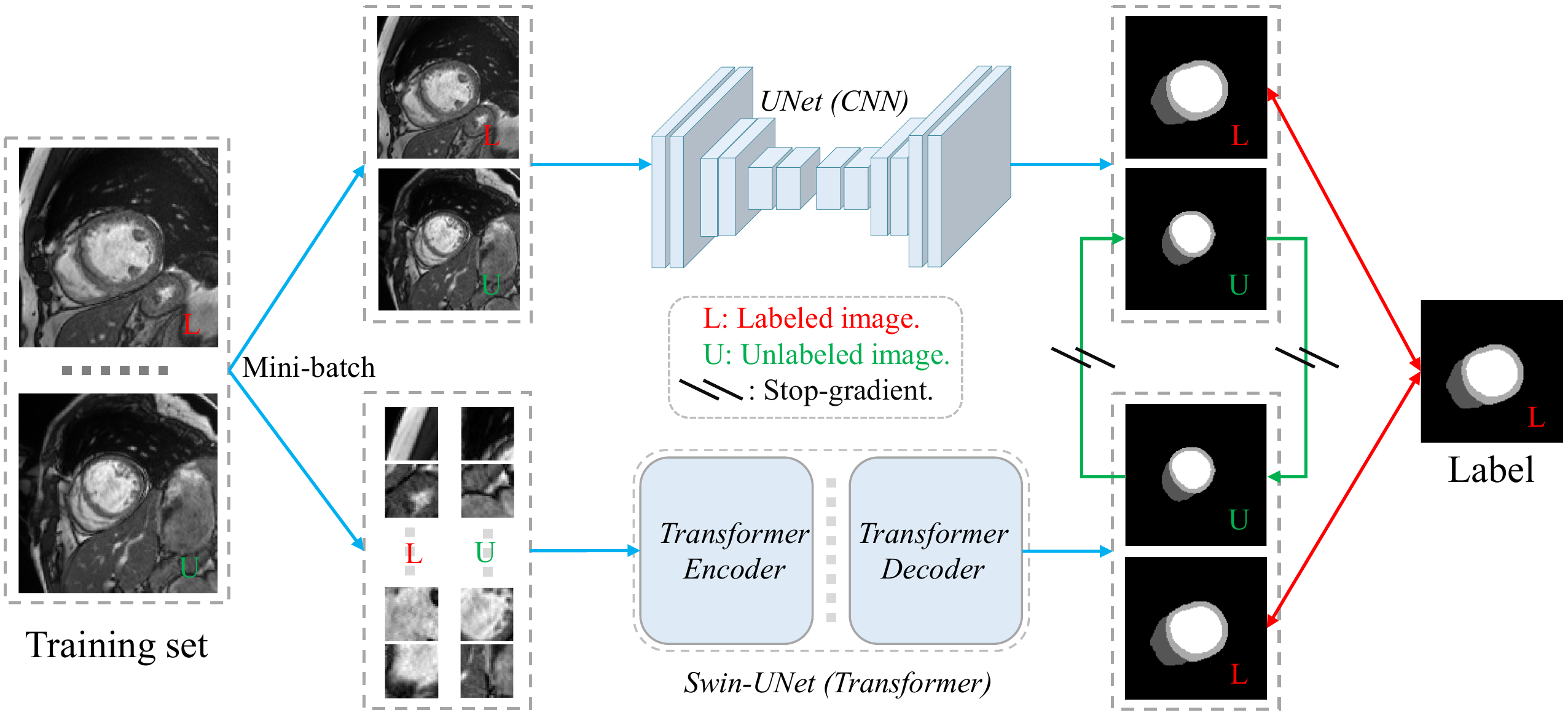}
 \caption{The model performs the semi-supervised medical image segmentation task. The regularization scheme between CNN and Transformer is referred to as Cross Teaching. L denotes the labeled data and U denotes the unlabeled data. The cross-teaching employs a bidirectional loss function: one path is from the CNN branch to the Transformer branch, and the other is from the Transformer to the CNN. A Transformer is applied for complementary training instead of prediction generation \citep{luo2021semi}.}
 \label{fig:CrossTeaching structure}
\end{figure}

Inspired by the existing works \citep{qiao2018cotraining,han2018coteaching,chen2021cropseu} for semi-supervised learning which introduce perturbation at different levels and encourage prediction to be consistent during the training stage, the designed cross teaching introduces the perturbation in both learning paradigm-level and output-level. As shown in \Cref{fig:CrossTeaching structure}, each image within the training set containing labeled and unlabeled images is fed into two different learning paradigms: a CNN and a Transformer respectively. For the unlabeled dataset with raw images, the cross teaching scheme allows the cross supervision between a CNN $(f_{\phi}^{c}(.))$ and a Transformer$(f_{\phi}^{t}(.))$, which aims at integrating the properties of the Transformer modeling the long-range dependency and CNN t0 learn local information in the output level.

The unlabeled data initially passes through a CNN and a Transformer respectively to generate predictions $p_i^c$ and $p_i^t$.
\begin{equation}
p_i^c = f_{\phi}^c(x_i); p_i^t = f_{\phi}^t(x_i);
\end{equation}
Then the pseudo labels $pl_i^c$ and $pl_i^t$ are produced in this manner:
\begin{equation}
   pl_i^c=argmax(p_i^t); 
 pl_i^t=argmax(p_i^c);
\end{equation}
The pseudo label $pl_i^c$ used for the CNN training is generated by the Transformer. Similarly, the CNN model provides pseudo labels for Transformer training.
The cross-teaching loss for the unlabeled data is defined as follows:
\begin{equation}
    L_{ctl}=\underbrace{L_{Dice}(p_i^c,pl_i^c)}_{supervision\; for \;CNNs}+\underbrace{L_{Dice}(p_i^t,pl_i^t)}_{supervision\; for \;Transformers}
\end{equation}
which is a bidirectional loss function. One direction of the data stream is from the CNN to the Transformer, and the other direction is from the Transformer to the CNN.
For the labeled data, the CNN and Transformer are supervised by the ground truth. The commonly-used supervised loss functions, i.e. the cross-entropy loss and Dice loss, are employed to update model parameters.
\begin{equation}
  L_{sup}=L_{ce}(p_i,y_i)+L_{Dice}(p_i,y_i)  
\end{equation}
where $p_i$ , $y_i$ represent the prediction and the label of image $x_i$.
And the overall objective combining the cross-teaching branch and supervised branch is defined as:
\begin{equation}
L_{total}=L_{sup}+\lambda L_{ctl}
\end{equation}

where $\lambda$ is a weight factor, which is defined by time-dependent Gaussian warming up function \citep{yu2019uncertainty, luo2021semi}:
\begin{equation}
\lambda(t)=0.1 \cdot e^{\left(-5\left(1-\frac{t_i}{t_{total}}\right)\right)^2}
\end{equation}

The results of the ablation study indicate that the combination of CNN and Transformer in a cross-teaching way shows superiority over the existing semi-supervised methods. 
Furthermore, the novel method has the potential to reduce the label cost by learning from limited data and large-scale unlabeled data.
However, it is observed that achieving SOTA via semi-supervised approaches remains a significant challenge.

\begin{table*}[ht]
    \centering
    \caption{An overview of the reviewed Transformer-based medical image Segmentation models.}
    \label{tab:segmentation}
    \resizebox{\textwidth}{!}{
    \begin{tabular}{lccccccc}  
    \toprule
    \textbf{Method} & \textbf{Modality} & \textbf{Organ} & \textbf{Type} & \textbf{Pre-trained Module: Type} & \textbf{Datasets} & \textbf{Metrics} & \textbf{Year}\\ 

    \rowcolor{maroon!30}\multicolumn{8}{c}{\textbf{Pure}} \\
    % \multicolumn{8}{c}{{\cellcolor[rgb]{1,0.753,0.478}}\textbf{Pure}} \\
    
     \makecell[l]{Swin-Unet \citep{cao2021swin}} & CT & Multi-organ & 2D &  ViT: Supervised &  \begin{tabular}[c]{@{}c@{}} $^1$ Synapse \citep{Synapse}\\$^2$ ACDC \citep{bernard2018acdc} \end{tabular} & Dice   &  2021 
     \\
     \midrule
     \makecell[l]{nnFormer \citep{zhou2021nnformer} }& \begin{tabular}[c]{@{}c@{}} CT\\ MRI \end{tabular}  & Multi-organ   & 3D     & ViT: Supervised      & \begin{tabular}[c]{@{}c@{}} $^1$ Synapse \citep{Synapse}\\$^2$ ACDC \citep{bernard2018acdc} \end{tabular}  & Dice   & 2021
     \\
     \midrule
      
    \makecell[l]{MISSFormer \citep{huang2021missformer}}        & \begin{tabular}[c]{@{}c@{}} CT\\ MRI \end{tabular}    & Multi-organ    & 2D     & \xmark      & \begin{tabular}[c]{@{}c@{}} $^1$ Synapse \citep{Synapse}\\$^2$ ACDC \citep{bernard2018acdc} \end{tabular}    & \begin{tabular}[c]{@{}c@{}} Dice\\ Hausdorff distance \end{tabular} & 2021 
      \\
      
     \midrule
     \makecell[l]{TransDeepLab \citep{azad2022transdeeplab}} & \begin{tabular}[c]{@{}c@{}} CT\\ Dermoscopy \end{tabular}  & \begin{tabular}[c]{@{}c@{}} Multi-organ \\Skin\end{tabular}   & 2D     & ViT: Supervised      & \begin{tabular}[c]{@{}c@{}c@{}} $^1$ Synapse \citep{Synapse}\\$^2$ ISIC 2017, 2018 \citep{codella2018skin, codella2019skin} \\$^3$ PH2 \citep{Mendonca2013PH2} \end{tabular}  & \begin{tabular}[c]{@{}c@{}} Dice\\ Hausdorff distance \end{tabular}    & 2022
     \\
     
       \rowcolor{maroon!30}\multicolumn{8}{c}{\textbf{Encoder}} \\
    % \multicolumn{8}{c}{{\cellcolor[rgb]{1,0.753,0.478}}\textbf{Encoder}} 

  \makecell[l]{TransUNet \citep{chen2021transunet} }      & \begin{tabular}[c]{@{}c@{}} CT\\ MRI \end{tabular}  & Multi-organ  & 2D     & ViT: Supervised      & \begin{tabular}[c]{@{}c@{}} $^1$ Synapse \citep{Synapse}\\$^2$ ACDC \citep{bernard2018acdc}\end{tabular}  & \begin{tabular}[c]{@{}c@{}} Dice\\ Hausdorff distance \end{tabular} & 2021
     \\
     \midrule
        
  \makecell[l]{TransBTS \citep{wang2021transbts}} & MRI & Brain   & 3D     & \xmark & BraTS 19-20 \citep{menze2014multimodal,bakas2017advancing,bakas2018identifying}& \begin{tabular}[c]{@{}c@{}} Dice\\ Hausdorff distance \end{tabular} & 2021 
     \\
   \midrule
  \makecell[l]{TransFuse \citep{zhang2021transfuse} }& Colonoscopy    & Multi-organ  & \makecell{2D \\ 3D} & ViT: Supervised  & \begin{tabular}[c]{@{}c@{}} $^1$ Synapse \citep{Synapse}\\$^2$ ACDC \citep{bernard2018acdc}\end{tabular} & \begin{tabular}[c]{@{}c@{}} Dice\\ Hausdorff distance \end{tabular} & 2021               
  \\
  \midrule
  \makecell[l]{MedT \citep{valanarasu2021medt} }& \begin{tabular}[c]{@{}c@{}}Microscopy\\Ultrasound \end{tabular}   & Multi-organ  & 2D     & \xmark   &  \begin{tabular}[c]{@{}c@{}} $^1$ Brain US (Private)\\$2$ GLAS \citep{sirinukunwattana2017glas}\\$^3$ MoNuSeg \citep{kumar2017monuseg}\end{tabular}  & F1    & 2021                  
  \\
  \midrule
  \makecell[l]{UNETR \citep{hatamizadeh2021unetr}}  & \begin{tabular}[c]{@{}c@{}} CT\\ MRI \end{tabular}     & Brain, Spleen   & 3D     & \xmark      & \begin{tabular}[c]{@{}c@{}} $^1$ Synapse \citep{Synapse}\\$^2$ MSD \citep{simpson1902msd} \end{tabular}          & \begin{tabular}[c]{@{}c@{}} Dice\\ Hausdorff distance \end{tabular} & 2021                 
  \\
  \midrule
  \makecell[l]{Swin UNETR \citep{hatamizadeh2022swinunetr}}      & MRI      & Brain  & 3D     & ViT: Supervised      & BraTS 21 \citep{baid2021brats}   & \begin{tabular}[c]{@{}c@{}} Dice\\ Hausdorff distance \end{tabular} & 2022 
  \\
  \midrule
  \makecell[l]{Swin UNETR++ \citep{tang2022selfswinunetr}}      & CT      & Multi-organ  & 3D & ViT: Self-supervised  & \makecell{$^1$ BTCV \citep{Synapse} \\ $^2$ MSD \citep{simpson1902msd}} & \makecell{Dice\\ NSD} & 2022 
  \\ 
     \rowcolor{maroon!30}\multicolumn{8}{c}{\textbf{Skip connection}} \\
           
  \makecell[l]{CoTr \citep{xie2021cotr} }  & CT    & Multi-organ  & 3D     & \xmark      & Synapse \citep{Synapse}     & Dice   & 2021       
     \\
\midrule
  \makecell[l]{HiFormer \citep{heidari2022hiformer} }  &\begin{tabular}[c]{@{}c@{}c@{}} MRI \\Dermoscopy\\Microscopic\end{tabular}      & \begin{tabular}[c]{@{}c@{}c@{}} Multi-organ \\Skin\\Cells\end{tabular}   & 2D     & \makecell{ViT: Supervised \\CNN: Supervised}       &\begin{tabular}[c]{@{}c@{}c@{}} $^1$ Synapse \citep{Synapse}\\$^2$ ISIC 2017, 2018 \citep{codella2018skin,codella2019skin}\\$^3$ SegPC 2021 \citep{gupta2018pcseg,gupta2020gcti,gehlot2020ednfc} \end{tabular}    &\begin{tabular}[c]{@{}c@{}} Dice\\ Hausdorff distance \end{tabular}  & 2022       
     \\ 
    \rowcolor{maroon!30}\multicolumn{8}{c}{\textbf{Decoder}} \\

     \makecell[l]{SegTran \citep{li2021segtran}} & \begin{tabular}[c]{@{}c@{}c@{}}Fundus\\ MRI\\X-Colonoscopy \end{tabular} & Multi-organ & \makecell{2D \\ 3D} & CNN: Supervised & \begin{tabular}[c]{@{}c@{}c@{}c@{}}REFUGE 20 \citep{orlando2020refuge}\\$^1$ BraTS 19 \citep{menze2014multimodal,bakas2017advancing,bakas2018identifying}\\$^2$ X-CVC \citep{fan2020xcvc}\\$^3$ KVASIR \citep{jha2020kvasir} \end{tabular}& Dice     & 2021 \\
     \rowcolor{maroon!30}\multicolumn{8}{c}{\textbf{Other architectures}} \\

\makecell[l]{T-AutoML \citep{yang2021TAutoML} }        & CT   & Liver and lung tumor     & 3D     &   \xmark     & MSD 2019 \citep{simpson2019bcvbtcv} & \begin{tabular}[c]{@{}c@{}} Dice\\ NSD \end{tabular}      & 2021  \\

\midrule
\makecell[l]{Cross Teaching \citep{luo2021semi} }    & MRI     & Multi-organ     & 2D     & \xmark       & ACDC \citep{bernard2018acdc}  &\begin{tabular}[c]{@{}c@{}} Dice\\ Hausdorff distance \end{tabular}    & 2022   \\

\midrule
\makecell[l]{Self-pretraining with MAE \citep{zhou2022MAE}} & \begin{tabular}[c]{@{}c@{}c@{}} CT\\ MRI\\X-ray \end{tabular}  & \begin{tabular}[c]{@{}c@{}c@{}} Lung\\ Brain\\Multi-organ \end{tabular} & 3D     & ViT: supervised    & \begin{tabular}[c]{@{}c@{}c@{}}$^1$ ChestX-ray14 \citep{wang2017chestx}\\$^2$ Synapse \citep{simpson2019bcvbtcv}\\$^3$ MSD 2019 \citep{simpson1902msd} \end{tabular} &\begin{tabular}[c]{@{}c@{}} Dice\\ Hausdorff distance \end{tabular}    & 2022
\\
    
    \bottomrule
    \end{tabular}
}
\end{table*}

% \subsection{table2 position}

\begin{table*}[!t]
    \centering
    \caption{A brief description of the reviewed Transformer-based medical image segmentation models. The unreported number of parameters indicates that the value was not mentioned in the paper, and the code was unavailable.}
    \label{tab:segmentation highlight}
    \resizebox{\textwidth}{!}{
    \begin{tabular}{llp{15cm}p{20cm}} 
    \toprule
    % \rowcolor[rgb]{0.976,0.698,1}
    
    \textbf{Method} & \textbf{\# Params} & \textbf{Contributions} & \textbf{Highlights} \\ 
    \rowcolor{maroon!30}\multicolumn{4}{c}{\textbf{Pure}} 
    %---------------line -------------------------------------------------
    \\
     \makecell[l]{Swin-Unet\citep{cao2021swin}} 
     & 
     \makecell[l]{-}
     & 
     $\bullet$ Builds a pure Transformer model with symmetric Encoder-Decoder architecture based on the Swin-Transformer block connected via skip connections. \newline
     $\bullet$ Proposes patch merging layers and patch expanding layers to perform downsampling and upsampling without convolution or interpolation operation.
     &  
     $\bullet$ The results of extensive experiments on multi-organ and multi-modal datasets show the good generalization ability of the model. \newline
     $\bullet$ Pre-trained on ImageNet rather than medical image data, which may result in sub-optimal performance.

     \\
     \midrule
     %---------------line -------------------------------------------------
     \makecell[l]{ nnFormer \citep{zhou2021nnformer}} 
     &
     \makecell[l]{158.92M}
     &
     $\bullet$ Proposes a powerful segmentation model with an interleaved architecture (stem) based on the empirical combination of self-attention and convolution. \newline
     $\bullet$ Proposes a volume-based multi-head self-attention (V-MSA) to reduce computational complexity.
     & 
     $\bullet$ Volume-based operations help to reduce the computational complexity.
     \newline
     $\bullet$ Pre-trained on ImageNet rather than medical image data, which may result in sub-optimal performance.

     \\
     \midrule
     %--------------------------------------------------------------------------
     \makecell[l]{MISSFormer \citep{huang2021missformer}} 
     &
     \makecell[l]{-} 
     &
     $\bullet$  Proposes the Enhanced Transformer Block based on the Enhanced Mix-FFN and the Efficient Self-attention module.
     \newline
     $\bullet$ Proposes the Enhanced Transformer Context Bridge built on the Enhanced Transformer Block to model both the local and global feature representation and fuse multi-scale features.
     &
     $\bullet$ The model can be trained from scratch without the pretraining step on ImageNet.
     \newline
     $\bullet$ Less computational burden due to the novel design of the Efficient Self-attention module. 
      \\
     \midrule
     %--------------------------------------------------------------------------
     \makecell[l]{TransDeepLab \citep{azad2022transdeeplab}} 
     &
     \makecell[l]{21.14M} 
     &
     $\bullet$ Proposes the encoder-decoder DeepLabv3+ architecture based on Swin-Transformer.
     \newline
     $\bullet$ Proposes the cross-contextual attention to adaptively fuse multi-scale representation.
     &
     $\bullet$ The first attempt to combine the Swin-Transformer with DeepLab architecture for medical image segmentation.
     \newline
     $\bullet$ A lightweight model with only 21.14M parameters compared with Swin-Unet\citep{cao2021swin}, the original DeepLab model \citep{chen2017deeplab} and TransUNet\citep{chen2021transunet}.

    %  \midrule
     %-------------------------------------------------------------
     \\
     
      \rowcolor{maroon!30}\multicolumn{4}{c}{\textbf{Encoder}} 
     \\
     \makecell[l]{TransUNet \citep{chen2021transunet}}  
     & 
     96.07M
     & 
     $\bullet$ Proposes the first CNN-Transformer hybrid network for medical image segmentation, which establishes self-attention mechanisms from the perspective of sequence-to-sequence prediction. \newline
     $\bullet$ Proposes a cascaded upsampler (CUP) which comprises several upsampling blocks to generate the prediction results.
     &
     $\bullet$ TransUNet fully exploits the strong global context encoded from the Transformer and local semantics from the CNN module. \newline
     $\bullet$ It presents the generalization ability on multi-modalities. \newline
     $\bullet$ The approach allows the segmentation of 2D and 3D medical images.
  
     \\
     \midrule
     \makecell[l]{TransBTS \citep{wang2021transbts}} 
     &
     \makecell[l]{Moderate TransBTS: 32.99M \\
     Lightweight TransBTS: 15.14M}
     &
     $\bullet$ Proposes a novel encoder-decoder framework TransBTS that integrates Transformer with 3D CNN for MRI Brain Tumor Segmentation.
     &
     $\bullet$ The method can model the long-range dependencies not only in spatial but also in the depth dimension for 3D volumetric segmentation. \newline
     $\bullet$ TransBTS can be trained on the task-specific dataset without the dependence on pre-trained weights. \newline
     $\bullet$ TransBTS is a moderate-size model that outperforms in terms of model complexity with 32.99M parameters and 33G FLOPs. Furthermore, the vanilla Transformer can be replaced with other variants to reduce the computation complexity.
     \\
     \midrule
     \makecell[l]{TransFuse \citep{zhang2021transfuse}} 
     &  
     \makecell[l]{ TransFuse-S: 26.3M}
     &
     $\bullet$ Proposes the first parallel-in-branch architecture — TransFuse to capture both low-level global features and high-level fine-grained semantic details. \newline
     $\bullet$ Proposes BiFusion module in order to fuse the feature representation from the Transformer branch with the CNN branch.
     & 
     $\bullet$ The architecture does not require very deep nets, which alleviates gradient vanishing and feature diminishing reuse problems. \newline
     $\bullet$ It improves performance by reducing parameters and increasing inference speed, allowing deployment on both the cloud and the edge. \newline
     $\bullet$ The CNN branch is flexible to use any off-the-shelf CNN network. \newline
     $^4$ It can be applied to both 2D and 3D medical image segmentation.

     \\
     \midrule
     \makecell[l]{MedT \citep{valanarasu2021medt}} 
     &
     \makecell[l]{1.4M}
     &
     $\bullet$ Proposes a gated axial-attention model that introduces an additional control mechanism to the self-attention module. \newline
     $\bullet$ Proposes a LoGo (Local-Global) training strategy for boosting segmentation performance by simultaneously training a shallow global branch and a deep local branch. 
     &
     $\bullet$ The proposed method does not require pre-training on large-scale datasets compared to other transform-based models. \newline
     $\bullet$ The results of predictions are more precise compared to the full attention model. 
     
     \\
     \midrule
     \makecell[l]{UNETR \citep{hatamizadeh2021unetr}} 
     &
     92.58M
     &
     $\bullet$ Proposes a novel architecture to address the task of 3D volumetric medical image segmentation. \newline
     $\bullet$ Proposes a new architecture where the Transformer-based encoder learns long-range dependencies and the CNN-based decoder utilizes skip connections to merge the outputs of a Transformer at each resolution level with the upsampling part.
     &
     $\bullet$ UNETR shows moderate model complexity while outperforming these Transformer-based and CNN-based models.
     \newline
     $\bullet$ The inference time of UNETR is significantly faster than Transformer-based models.
     \newline
     $\bullet$ They did not use any pre-trained weights for the Transformer backbone.
     \\
     \midrule
     \makecell[l]{Swin UNETR \citep{hatamizadeh2022swinunetr}} 
     & 
     61.98M
     &  
     $\bullet$ Proposes a novel segmentation model, Swin UNETR, based on the design of UNETR and Swin Transformers.
     &
     $\bullet$ The FLOPs of Swin UNETR significantly grow compared to that of UNETR and TransBTS.
     \newline
     $\bullet$ The Swin Transformer is suitable for the downstream tasks.
     \\
     \midrule
     \makecell[l]{Swin UNETR++ \citep{hatamizadeh2022swinunetr}} 
     & 
     61.98M
     &  
     $\bullet$ Presents a new self-supervised learning approach that capitalizes on the Swin UNETR model, harnessing its multi-resolution encoder for efficient pre-training. \newline
     $\bullet$ Introduces a pre-trained robust model using the proposed encoder and proxy tasks on a dataset of 5,050 publicly available CT images, resulting in a powerful feature representation for various medical image analysis tasks.
     &
     $\bullet$ Fine-tuning the pretrained Swin UNETR model leads to improved accuracy, faster convergence, and reduced annotation effort compared to training from randomly initialized weights. \newline
     $\bullet$ Researchers can benefit from the pre-trained encoder in transfer learning for various medical imaging analysis tasks, including classification and detection.
     \\
     \rowcolor{maroon!30}\multicolumn{4}{c}{\textbf{Skip connection}}
     \\
    %  \midrule
     \makecell[l]{CoTr \citep{xie2021cotr}} 
     &
     46.51M
     &
     $\bullet$ Proposes a hybrid framework that bridges a convolutional neural network and a Transformer for accurate 3D medical image segmentation. \newline
     $\bullet$ Proposes the deformable Transformer (DeTrans) that employs the multi-scale deformable self-attention mechanism (MS-DMSA) to model the long-range dependency efficiently.
     &
     $\bullet$ The deformable mechanism in CoTr reduces computational and spatial complexities, allowing the network to model high-resolution and multi-scale feature maps.\\
     
    \midrule
    
     \makecell[l]{HiFormer \citep{heidari2022hiformer}} 
     &
     \makecell[l]{ HiFormer-S: 23.25M\\
     HiFormer-B: 25.51M\\
     HiFormer-L: 29.52M}
     &
     $\bullet$ Proposes an encoder-decoder architecture that bridges a CNN and a Transformer for medical image segmentation.
    \newline
     $\bullet$ Proposes a Double-level Fusion module in the skip connection.
     &
     $\bullet$ Fewer parameters and lower computational cost.
 
     \\
     \rowcolor{maroon!30}\multicolumn{4}{c}{\textbf{Decoder}} 
 
     \\
     \makecell[l]{SegTran \citep{li2021segtran}} 
     &
     86.03M
     &
     $\bullet$ Proposes a novel Transformer design, Squeeze-and-Expansion Transformer, in which a squeezed attention block helps regularize the huge attention matrix, and an expansion block learns diversified representations. \newline
     $\bullet$ Proposes a learnable sinusoidal positional encoding that imposes a continuity inductive bias for the Transformer. 
     &
     $\bullet$ Compared to U-Net and DeepLabV3+, Segtran has the least performance degradation, showing the best cross-domain generalization when evaluated on datasets of drastically different characteristics. 

     \\
  
    \rowcolor{maroon!30}\multicolumn{4}{c}{\textbf{Other architectures}} 
     \\
     \makecell[l]{T-AutoML \citep{yang2021TAutoML}} 
     &
     16.96M
     &
     $\bullet$ Proposes the first automated machine learning algorithm, T-AutoML, which automatically estimates “almost” all components of a deep learning solution for lesion segmentation in 3D medical images. \newline
     $\bullet$ Proposes a new predictor-based search method in a new search space that searches for the best neural architecture and the best combination of hyperparameters and data augmentation strategies simultaneously. 
     &
     $\bullet$ The method is effectively transferable to different datasets.\newline
     $\bullet$ The applied AutoML alleviates the need for the manual design of network structures.\newline
     $\bullet$ The intrinsic limitations of AutoML.

     \\
     \midrule
     \makecell[l]{Cross Teaching \citep{luo2021semi}} 
     &
     \makecell[l]{-}
     &
     $\bullet$ Proposes an efficient regularization scheme for semi-supervised medical image segmentation where the prediction of a network serves as the pseudo label to supervise the other network.\newline
     $\bullet$ The proposed method is the first attempt to apply the Transformer to perform the semi-supervised medical segmentation utilizing the unlabeled data.
     &
     $\bullet$ The training process requires less data cost by semi-supervision. \newline
     $\bullet$ The framework contains components with low complexity and simple training strategies compared to other semi-supervised learning methods. \newline
     $\bullet$ The proposed semi-supervised segmentation still can not achieve the state-of-the-art (SOTA) compared with the fully-supervised approaches.
    
     \\
     \midrule
     \makecell[l]{Self-pretraining with MAE \citep{zhou2022MAE}} 
     & 
     \makecell[l]{-}
     &  
     $\bullet$ Proposes a self-pre-training paradigm with MAE for medical images where the pre-training process of the model uses the same data as the target dataset. 
     &  
     $\bullet$ The proposed paradigm demonstrates its effectiveness in limited data scenarios.
     \\

    \bottomrule
    \end{tabular}
    }
\end{table*}
% \subsubsection{Self Pre-training with Masked Autoencoders for Medical Image Analysis}
\begin{figure}[!th]
 \centering
 \includegraphics[width=0.5\textwidth]{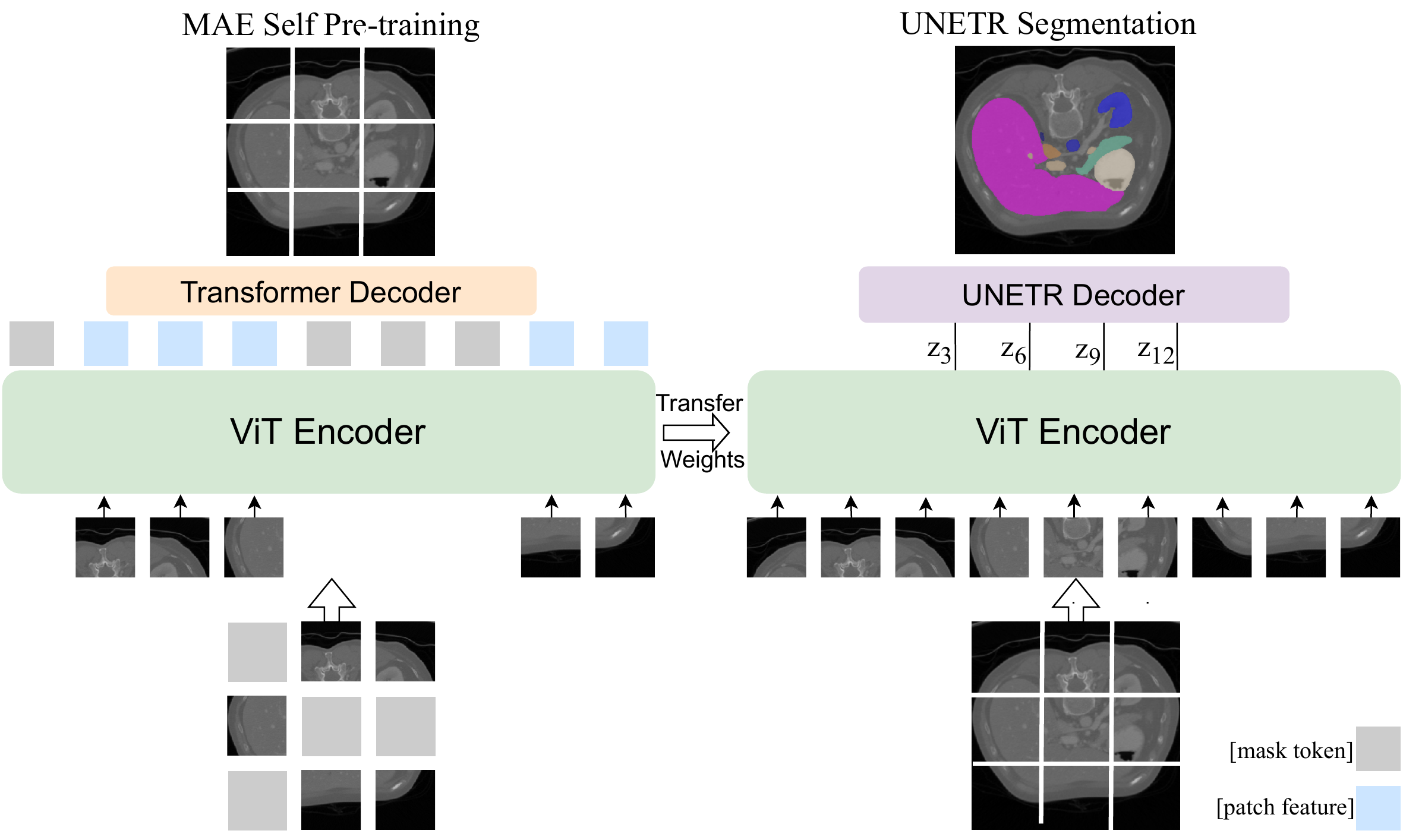}
 \caption{Illustration of MAE self pre-training. First, MAE is pre-trained as an encoder for ViT. The ViT encoder is fed with a random subset of patches and the decoder of the Transformer reconstructs the complete image as shown on the left. Then, the pre-trained ViT weights are transferred to the initialized segmentation encoder, as shown on the right. Finally, the whole segmentation network, such as UNETR, is fine-tuned to perform the segmentation task. \citep{zhou2022MAE}.}
 \label{fig:MAE overview}
\end{figure}

Zhou et al. \citep{zhou2022MAE} hypothesize that the ability to aggregate contextual information is imperative to improve the performance of medical image analysis. Nonetheless, there is no
ImageNet-scale medical image dataset for pre-training. Therefore, they investigate a novel self-pre-training paradigm based on Masked Autoencoder (MAE), \textbf{MAE self pre-training}, for medical image analysis, one of the masked image modeling (MIM) frameworks \citep{bao2022beit} \citep{xie2022simmim} \citep{el2021selfPre} \citep{he2021maeScalable}. MIM encourages the framework to restore the masked target by integrating information from the context, where the main idea of MIM is masking and reconstructing: masking a set of image patches before input into the Transformer and reconstructing these masked patches at the output.

The pipeline for segmentation with MAE self-pre-training contains two stages.
In the first stage (as shown on the left of \Cref{fig:MAE overview}), ViT is pre-trained with MAE as the encoder. The input patches are divided into visible ones and masked ones.
The ViT encoder only acts on the visible patches. Compared to other MIM methods, MAE does not employ mask tokens in the encoder, which saves time and allows for faster pre-training. A lightweight Transformer decoder is appended to reconstruct the full image. The decoder is only an auxiliary part used for pre-training and will not be applied in downstream tasks.

In the second stage (as shown on the right of \Cref{fig:MAE overview}), the pre-trained ViT weights are transferred to initialize the segmentation encoder.
Then, the task-specific heads are appended to perform downstream tasks. The whole segmentation network, e.g., UNETR, is finetuned to perform the segmentation task.

The experiments, including the MAE pre-training and the downstream task, are conducted to evaluate the performance of the proposed method. The results show that MAE can recover the lost information in the masked input patches. MAE pre-training enables the model to improve its classification and segmentation performance on medical image analysis tasks, surpassing the ImageNet pre-trained model to SOTA.

\begin{table*}[!thb]
\centering
\caption{Comparison results of some segmentation methods on the \textit{Synapse} \citep{Synapse} dataset. The methods are ordered based on their dice score.}
\label{tab:segmentation_synapse}
\resizebox{\textwidth}{!}{
\begin{tblr}{
		colspec={|l|cc|cc|*{8}c|},
		row{1}={maroon!30,c},
		column{1} = {font=\bfseries},
		hline{1,13} = {3pt},
		vline{1,14} = {3pt},
		hlines
		}
\textbf{Methods} & \textbf{Params (M)} & \textbf{FLOPs (G)} & \textbf{Dice~$\uparrow$} & \textbf{HD95~$\downarrow$} & \textbf{Aorta} & \textbf{Gallbladder} & \textbf{Kidney(L)} & \textbf{Kidney(R)} & \textbf{Liver} & \textbf{Pancreas} & \textbf{Spleen} & \textbf{Stomach} \\
{TransUNet \citep{chen2021transunet}} & 96.07 & 88.91 & 77.49 & 31.69 & 87.23 & 63.16 & 81.87 & 77.02 & 94.08 & 55.86 & 85.08 & 75.62 \\
{Swin-Unet \citep{cao2021swin}} & 27.17 & \textbf{6.16} & 79.13 & 21.55 & 85.47 & 66.53 & 83.28 & 79.61 & 94.29 & 56.58 & 90.66 & 76.60 \\
{TransDeepLab \citep{azad2022transdeeplab}} & \textbf{21.14} & 16.31 & 80.16 & 21.25 & 86.04 & 69.16 & 84.08 & 79.88 & 93.53 & 61.19 & 89.00 & 78.40\\

{HiFormer \citep{heidari2022hiformer}} & 25.51 & 8.05 & 80.39 & 14.70 & 86.21 & 65.23 & 85.23 & 79.77 & 94.61 & 59.52 & 90.99 & 81.08 \\

{PVT-CASCADE \citep{rahman2023medical}} & 35.28 & 6.40 & 81.06 & 20.23 & 83.01 & 70.59 & 82.23 & 80.37 & 94.08 & 64.43 & 90.10 & 83.69 \\

{MISSFormer \citep{huang2021missformer}} & 42.46 & 9.89 & 81.96 & 18.20 & 86.99 & 68.65 & 85.21 & 82.00 & 94.41 & 65.67 & 91.92 & 80.81\\

{DAEFormer \citep{azad2022dae}} & 48.07 & 27.89 & 82.63 & 16.39 & 87.84 & 71.65 & 87.66 & 82.39 & 95.08 & 63.93 & 91.82 & 80.77 \\

{TransCASCADE \citep{rahman2023medical}} & 123.49 & - & 82.68 & 17.34 & 86.63 & 68.48 & 87.66 & 84.56 & 94.43 & 65.33 & 90.79 & 83.52 \\

{ScaleFormer \citep{huang2022scaleformer}} & 111.60 & 48.93 & 82.86 & 16.81 & \textbf{88.73} & \textbf{74.97} & 86.36 & 83.31 & 95.12 & 64.85 & 89.40 & 80.14 \\

{D-LKA Net \citep{azad2023selfattention}} & 101.64 & 19.92 & 84.27 & 20.04 & 88.34 & 73.79 & \textbf{88.38} & \textbf{84.92} & 94.88 & 67.71 & 91.22 & 84.94 \\

{MERIT \citep{rahman2023multi}} & 147.86 & 33.31 & \textbf{84.90} & \textbf{13.22} & 87.71 & 74.40 & 87.79 & 84.85 & \textbf{95.26} & \textbf{71.81} & \textbf{92.01} & \textbf{85.38} \\

% {TransNorm\\ \citep{azad2022transnorm}} & & & 78.40 & 30.25 & 86.23 & 65.10 & 82.18 & 78.63 & 94.22 & 55.34 & 89.50 & 76.01 \\

% {nnFormer\\ \citep{zhou2021nnformer}} & 149.3 & 240.2 & 86.57 & 10.63 & 92.04 & 70.17 & 86.57 & 86.25 & 96.84 & 83.35 & 90.51 & 86.83 \\

\end{tblr}
}
\end{table*}

\begin{table*}[!thb]
	\centering
	\caption{Performance comparison of several Transformer-based segmentation approaches on \textit{ISIC 2017} \citep{codella2018skin}, \textit{ISIC 2018} \citep{codella2019skin}, and \textit{PH$^2$} \citep{mendoncca2013ph}.}\label{tab:segmentation_skin}
	\resizebox{\textwidth}{!}{
		\begin{tblr}{
				colspec={|l|cc|cccc|cccc|cccc|},
				row{1,2}={maroon!30,c},
				column{1} = {font=\bfseries},
				hline{1,8} = {3pt},
				vline{1,16} = {3pt},
				hlines
			}
			\SetCell[r=2]{c} Methods & \SetCell[r=2]{c} Params (M)& \SetCell[r=2]{c} FLOPs (G) & \SetCell[c=4]{c} \textit{ISIC 2017} & & & & \SetCell[c=4]{c} \textit{ISIC 2018} & & & & \SetCell[c=4]{c} \textit{PH$^2$} & & & \\
			& & & DSC & SE & SP & ACC & DSC & SE & SP & ACC & DSC & SE & SP & ACC \\
			{TransUNet \citep{chen2021transunet}} & 96.07 & 88.91 & 0.8123 & 0.8263 & 0.9577 & 0.9207 & 0.8499 & 0.8578 & 0.9653 & 0.9452 & 0.8840 & 0.9063 & 0.9427 & 0.9200 \\
                {TransNorm \citep{azad2022transnorm}}& 117.98 & 33.41 & 0.8933 & 0.8532 & \textbf{0.9859} & 0.9582 & 0.8951 & 0.8750 & 0.9790 & 0.9580 & 0.9437 & \textbf{0.9438} & \textbf{0.9810} & \textbf{0.9723} \\
			{Swin-Unet \citep{cao2021swin}} & 27.17 & 6.16 & 0.9183    & 0.9142    & 0.9798  &  0.9701 & 0.8946  & 0.9056  & \textbf{0.9798} &  0.9645 & 0.9449  & 0.9410  & 0.9564 &  0.9678 \\
			{D-LKA Net \citep{azad2023selfattention}} & 101.64 & 19.92 & \textbf{0.9254} & \textbf{0.9327} & 0.9793  &  \textbf{0.9705}  & \textbf{0.9177} & \textbf{0.9164} & 0.9773 & \textbf{0.9647} & \textbf{0.9490} & 0.9430 & 0.9775 & 0.9659 \\
   			{HiFormer \citep{heidari2022hiformer}} & 25.51 & 8.05 & 0.9253 & 0.9155 & 0.9840 & 0.9702 & 0.9102 & 0.9119 & 0.9755 & 0.9621 & 0.9460 & 0.9420 & 0.9772 & 0.9661\\
		
		\end{tblr}
	}
\end{table*}

\begin{tcolorbox}[breakable ,colback={maroon!30},title={\subsection{Discussion and Conclusion}},colbacktitle=maroon!30,coltitle=black , left=2pt , right =2pt]

This section comprehensively investigates the overview of around 17 Transformer-based models for medical image segmentation presented in \Cref{sec:seg_pure_Transformer} to \Cref{sec:seg_Other_Architectures}. Some representative self-supervised strategies \citep{tang2022selfswinunetr, zhou2021nnformer} for training transformer modules are also introduced in this section. We provide information on the reviewed segmentation approaches about the architecture type, modality, organ, input size, the pre-trained manner, datasets, metrics, and the year in \Cref{tab:segmentation}. \Cref{tab:segmentation highlight} also lists the methods along with the number of parameters, contributions, and highlights. The comparison results of some SOTA methods on the Synapse dataset are also presented in \Cref{tab:segmentation_synapse}, showcasing their respective performance metrics. Additionally, the evaluation of SOTA approaches on the ISIC 2017, ISIC 2018, and PH$^2$ datasets are demonstrated in \Cref{tab:segmentation_skin}. The tables serve as valuable references for evaluating the relative performance and suitability of different approaches in medical image segmentation. Moreover, inference time and GPU usage are crucial performance factors for model selection. DAE-Former excels with a fast 46.01 ms inference and 6.42 GB GPU memory. MERIT, while effective, takes longer at 93.13 ms and uses 7.78 GB. D-LKA, with 113.18 ms inference, balances GPU memory well at 5.22 GB. TransCASCADE offers a compromise with 53.18 ms inference and 3.75 GB GPU memory. These metrics alongside the parameter numbers and FLOP count guide model selection for efficient medical image segmentation. It's important to note that these assessments were conducted using the same test setup and configurations as those employed in image classification \Cref{class_discuss}.

ViT-based works offer solutions in a broad range of multimodal tasks of 2D or 3D. Most of the approaches demonstrate superior results over CNN-based segmentation models on benchmark medical datasets. Despite the SOTA performance Transformer-based networks have achieved, there are some challenges in deploying the Transformer-based models at present.
The first challenge is the high computational burden due to the relatively large number of parameters of the Transformer-based models \citep{azad2022dae}. The reason is that the time and space complexity of the attention mechanism is quadratic to the sequence length. For example, the CNN-based models such as U-Net \citep{ronneberger2015unet} require 3.7M parameters \citep{valanarasu2021medt} to reach Dice Score 74.68 \citep{chen2021transunet}. However, TransUNet, which achieves Dice Score 77.48 needs 96.07M \citep{hatamizadeh2021unetr} parameters. The researchers have to meet the high demand for GPU resources. Thus, several novel approaches such as Swin Transformer employed in Swin-Unet \citep{cao2021swin}, volume-based Transformer utilized in nnFormer \citep{zhou2021nnformer} and efficient self-attention module in MISSFormer \citep{huang2021missformer} are proposed to simplify the computation of the Transformer models. The direction of facilitating the efficiency of models will play a crucial role in future research. We also note that most existing methods require pre-training strategies on the ImageNet dataset to obtain the pre-trained weights for the following downstream tasks. However, the natural image datasets and medical datasets differ dramatically from one another, which may impact the final performance of extracting the medical features. Meanwhile, pre-training leads to high computational costs, which hinders the training of models in practice. Multiple segmentation networks that can be trained from scratch on the medical dataset are suggested as the solutions, such as MISSFormer \citep{huang2021missformer}. We expect more approaches to exploring more efficient pre-training strategies or without pre-training. Furthermore, considering the limited size of some medical datasets, some approaches propose semi-supervised technologies or self-pre-training paradigms to reduce the dataset burden of training or pre-training. Nevertheless, the performance is still not comparable to that of fully-supervised models. Designing semi-supervised models with improved accuracy in this direction requires more attention.

\end{tcolorbox}
%% End of Jia Section--------------------
%%Ehsan-------------------------------------------
%%-----------------------------------------------

\begin{figure*}[!th]
	\centering
	\begin{subfigure}[t]{0.65\textwidth}
		\centering
		\includegraphics[width=\textwidth]{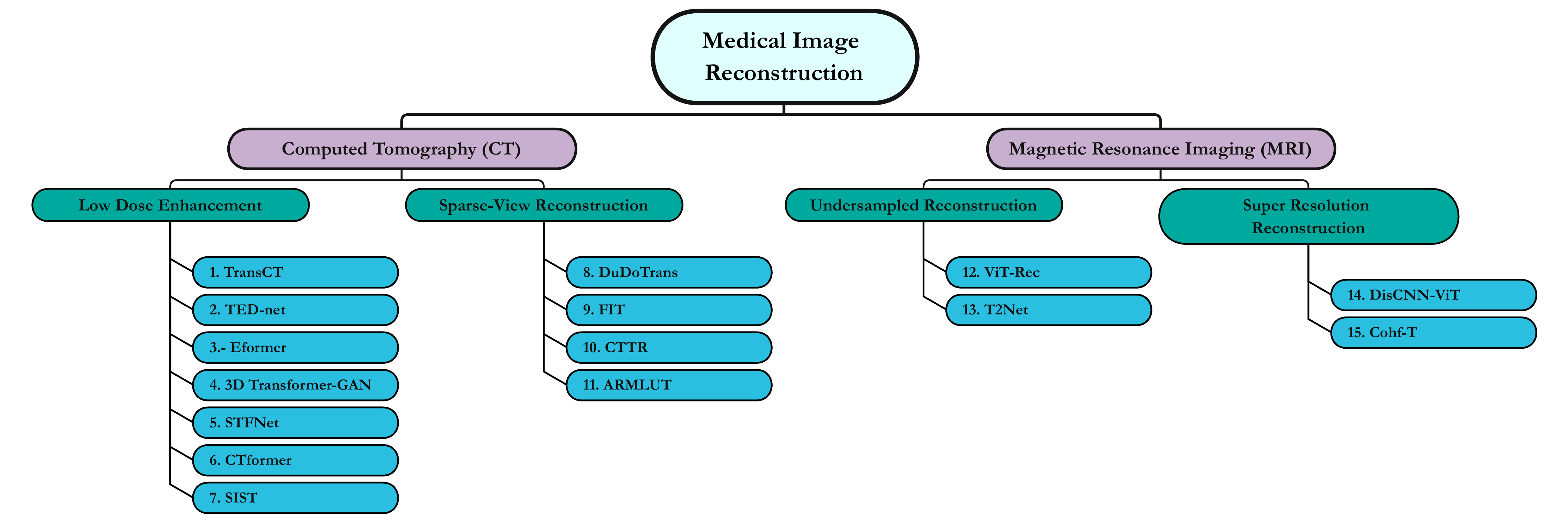}
		\caption{Taxonomy structure for medical image reconstruction. Methods in this field are categorized by their functionality in addressing issues in consensus imaging modalities, not how the Transformer is integrated with the architecture like in the previous sections. The prefix numbers in the paper’s name in ascending order denote the reference for each study as follows: 1. \citep{zhang2021transct}, 2. \citep{wang2021ted}, 3. \citep{luthra2021eformer}, 4. \citep{luo20213d}, 5. \citep{zhang2022spatial}, 6. \citep{wang2023ctformer}, 7. \citep{yang2022low}, 8. \citep{wang2021dudotrans}, 9. \citep{buchholz2022fourier}, 10. \citep{shi2022dual}, 11. \citep{wu2022adaptively}, 12. \citep{lin2021vision}, 13. \citep{feng2021task}, 14. \citep{mahapatra2021mr}, 15. \citep{fang2022cross}, 16.
        \citep{korkmaz2022unsupervised}, 17.
        \citep{zhou2023dsformer}.}
		\label{fig:recon-taxonomy}
	\end{subfigure}
	\hfill
	\begin{subfigure}[t]{0.3\textwidth}
		\centering
		\includegraphics[width=\textwidth]{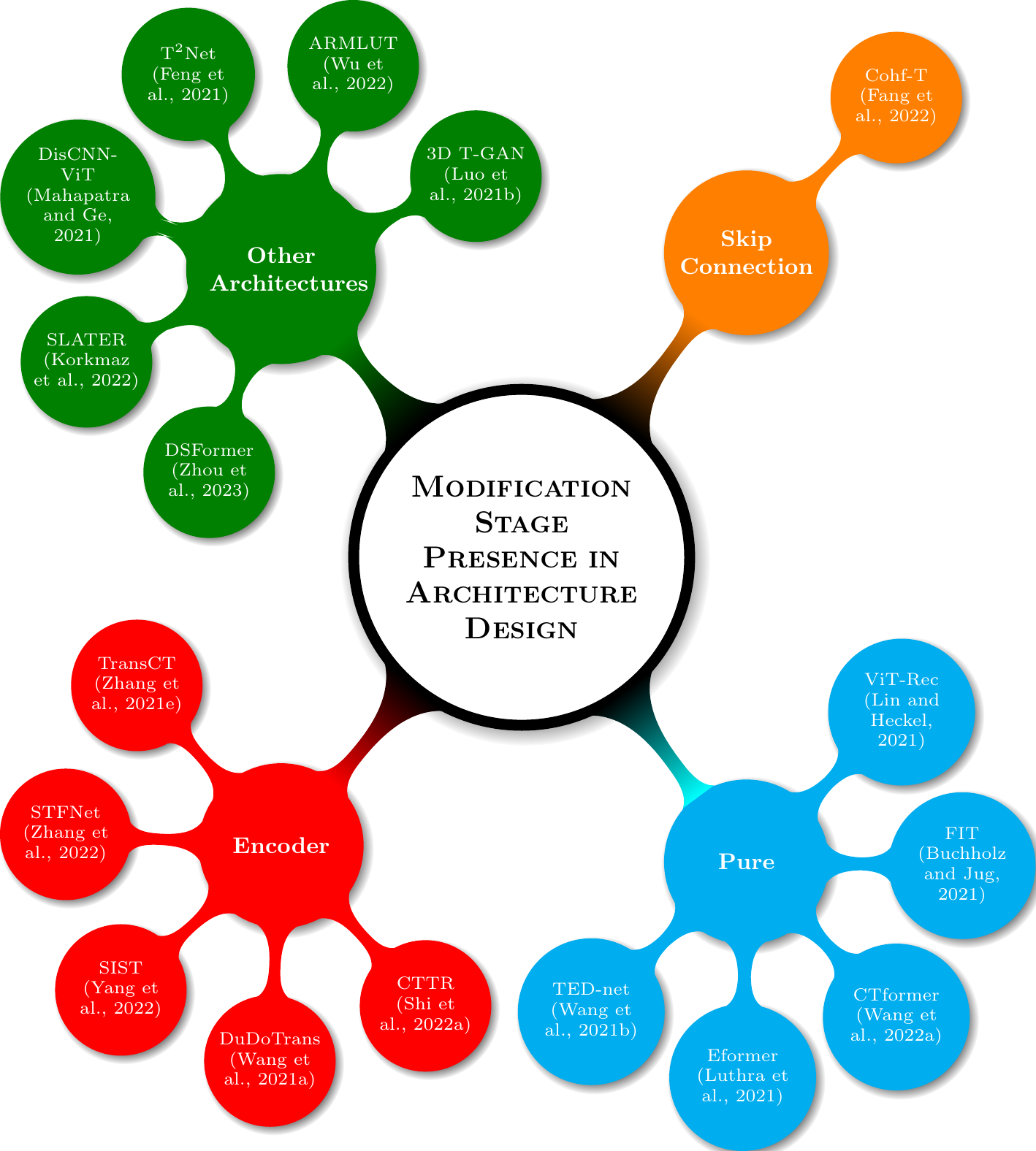}
		\caption{We also presented a second taxonomy due to the presence of the Transformer in the reviewed studies.}
		\label{fig:recon-papers}
	\end{subfigure}
	\caption{An overview of medical image reconstruction taxonomies either as categorizing by the task or the location of using Transformer in an architecture.}
	\label{fig:reconstructiontaxonomy}
\end{figure*}
\section{Medical Image Reconstruction} \label{sec:reconstruction}
3D Medical imaging is a clinical breakthrough and is very popular in medical diagnosis and follow-up after treatment. In Computed Tomography (CT), Single Photon Emission Tomography (SPECT), and Positron Emission Tomography (PET), the imaging process relies on ionizing radiation~\citep{seeram2015computed,mathews2017review}, which implies a potential risk for the patient~\citep{brenner2007computed}. A non-invasive 3D imaging technique is Magnetic Resonance Imaging (MRI), which does not rely on ionizing radiation. However, image acquisition may take longer and confines the patient in a discomforting narrow tube \citep{hyun2018deep}. In order to reconstruct 3D volumetric datasets from the acquired data, Medical image reconstruction is one of the essential components of 3D medical imaging. The primary objective of 3D image reconstruction is to generate high-quality volumetric images for clinical usage at minimal cost and radiation exposure, whilst also addressing potential artifacts inherent to the physical acquisition process. Image reconstruction solves an inverse problem that is generally challenging due to its large-scale and ill-posed nature \citep{zhang2020review}. 

In medical imaging, there are ongoing research efforts to reduce the acquisition time (i.e. to reduce cost and potential movement artifacts) as well as radiation dose. However, lowering the radiation dose results in higher noise levels and reduced contrast, which poses a challenge for 3D image reconstruction.

Vision Transformers (ViTs) have effectively demonstrated possible solutions to address these challenges. We categorize the literature in this domain into \textit{low dose enhancement}, \textit{sparse-view reconstruction}, \textit{undersampled reconstruction}, and \textit{super-resolution reconstruction}. This section will overview some of the SOTA Transformer-based studies that fit into our taxonomy. \Cref{fig:recon-taxonomy} and \Cref{fig:recon-papers} demonstrate our proposed taxonomy for this field of study. \Cref{fig:recon-taxonomy} indicates the diversity of our taxonomy based on the medical imaging modalities we studied in this research. \Cref{fig:recon-papers} endorses the usage of the Transformer within the overviewed studies' pipelines. 

\subsection{Low Dose Enhancement}
Zhang et al. \citep{zhang2021transct} used a very general intuition about image denoising: the noisy image constructed with high-frequency and low-frequency counterparts as $X = X_{H} + X_{L}$ in a study, namely, \textbf{TransCT}. Zhang et al. \citep{zhang2021transct} claim that the noisy image's low counterpart contains two sub-components of main image content and weakened image textures, which are entirely noise-free. They applied a Gaussian filter on the input image to decompose an image into a high-frequency sub-band and a low-frequency sub-band. After this, they extracted $X_{L_c}$ content features and $X_{L_t}$ latent texture features by applying two shallow CNNs on the low-frequency counterpart of the input image. Simultaneously, they applied a sub-pixel layer on a high-frequency counterpart to transform it into a low-resolution image and extracted embedding features ($X_{H_f}$) by applying a shallow CNN. Then the resultant latent texture features ($X_{L_t}$) and corresponding high-frequency representation are fed to the Transformer for noise removal from a high-frequency representation. Ultimately, they reconstruct the high-quality image piecewise. They showed that the latent texture features are beneficial in screening noise from the high-frequency domain.
% \begin{figure}[h]
%  \centering
%  \includegraphics[width=\columnwidth]{./images/TransCT.pdf}
%  \caption{TransCT \citep{zhang2021transct} schematic representation. TransCT utilizes a naive Transformer structure with an encoder and decoder structure. The encoder applies to the low-frequency counterpart of the CT image and the decoder feed with the high-frequency counterpart. For preserving the fine-grained details of the final LDCT image, the output of the Transformer decoder integrates with some of the low-frequency features. 'n16s2' denotes the convolution operation has 16 kernels with stride 2.}
%  \label{fig:TransCT overview}
% \end{figure}

\begin{figure}[h]
 \centering
 \includegraphics[width=\columnwidth]{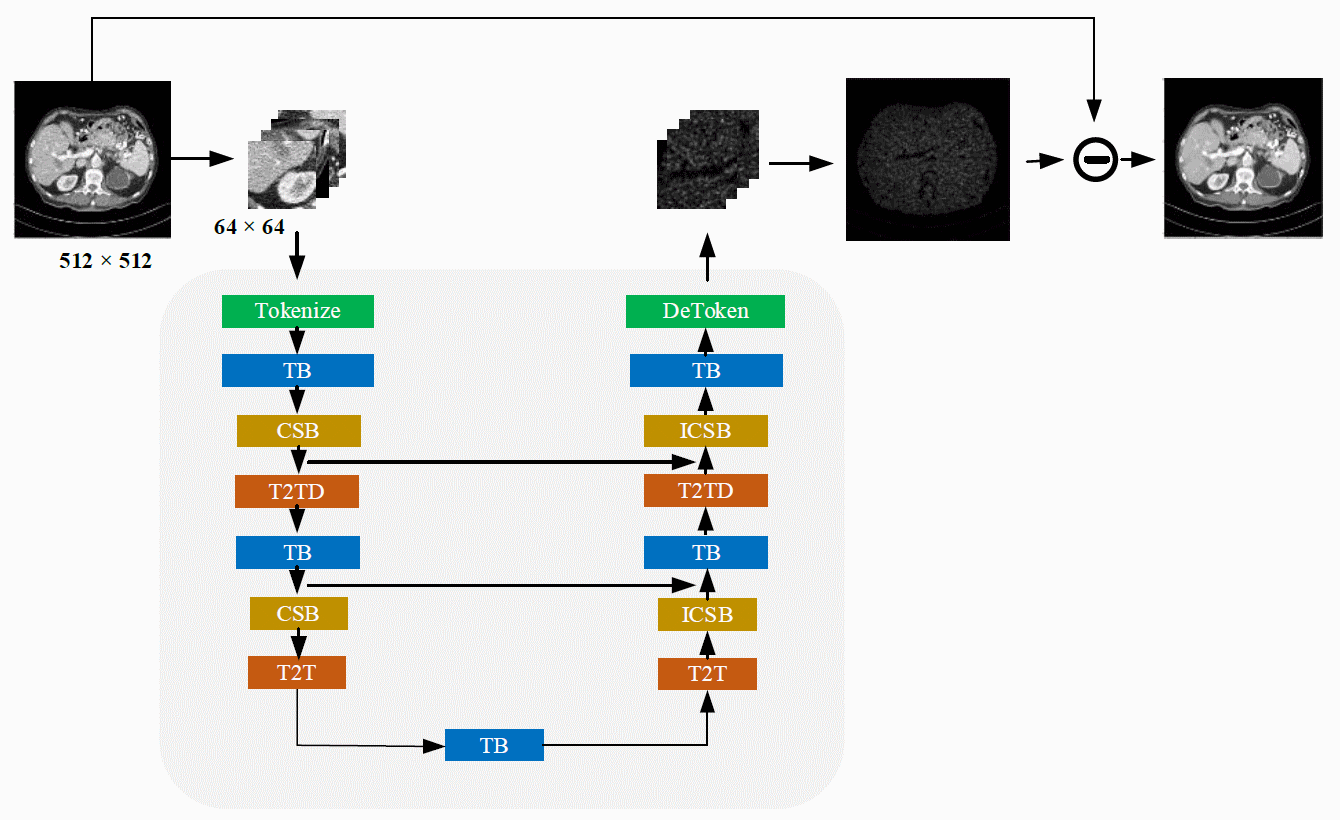}
 \caption{An overview of TED-net \citep{wang2021ted}. Tokenize and DeToken blocks are invertible operations that apply the process of patch embedding and converting patches again to image, respectively. TB represents a standard Transformer block. (I)CSB denotes the (inverse) Cyclic Shift block to modify the feature map, nonetheless, the reverse operation avoids pixel shifts in the final result. T2T block represents the Token-to-Token process \citep{yuan2021tokens} to improve the spatial inductive bias of Transformers by merging the neighboring tokens. The Dilated T2T (T2TD) block is used to refine contextual information further.}
 \label{fig:TED-net}
\end{figure}
Despite the TransCT \citep{zhang2021transct}, Wang et al. proposed a convolution-free Token-to-Token vision Transformer-based Encoder-decoder Dilation network (\textbf{TED-net}) design for CT image denoising \citep{wang2021ted}. Their approach is based on a U-Net encoder-decoder scheme enriched by different modules, i.e., Basic Vision Transformer, Token-to-Token Dilation (T2TD), and (Inverse) Cyclic Shift blocks. Consider $y \in \mathbb{R}^{N \times N}$ a clean natural dose CT image, $x \in \mathbb{R}^{N \times N}$ noisy low dose CT image, and $T: \mathbb{R}^{N \times N} \rightarrow \mathbb{R}^{N \times N}$ is a Transformer-based denoising model. According to the \Cref{fig:TED-net} after tokenization of $x$ and passing through the Vision Transformer block to capture long-range dependencies and alleviate the absence of local inductive bias in Transformers, they employed Token-to-Token serialization \citep{yuan2021tokens}. Also, they utilized feature re-assembling with a Cyclic Shift block (CSB) to integrate more information. Obvious from \Cref{fig:TED-net}, all of these blocks are replicated in a symmetric decoder path, but instead of the CSB, the Inverse Cyclic Shift block (ICSB) is implemented to avoid pixel shifts in the final denoising results ($y = x + T(x)$). They reached SOTA results compared to CNN-based methods and a competitive benchmark with regard to the TransCT \citep{zhang2021transct}.

Luthra et al. \citep{luthra2021eformer} proposed a Transformer-based network, \textbf{Eformer}, to deal with low-dose CT images while concurrently using the edge enhancement paradigm to deliver more accurate and realistic denoised representations. Their architecture builds upon 
% \Cref{fig:Eformer} shows that they proposed an encoder-decoder network heavily built on 
the LeWin (Locally-enhanced Window) Transformer block \citep{wang2022uformer}, which is accompanied by an edge enhancement module. The success of the Swin Transformer  \citep{liu2021swin}  in capturing the long-range dependencies with the window-based self-attention technique makes it a cornerstone in designing new Transformer blocks due to its linear computational complexity. LeWin Transformer is one of these blocks that capture the global contextual information and, due to the presence of a depth-wise block in its structure, could also capture a local context. Eformer's first step is through the Sobel edge enhancement filter. In every encoder-decoder stage, convolutional features pass through the LeWin Transformer block, and downsampling and upsampling procedures are done by convolution and deconvolution layers. Eformer's learning paradigm is a residual learning scheme, meaning it learns the noise representation rather than a denoised image due to the ease of optimization in predicting a residual mapping.

% \begin{figure}[h]
%  \centering
%  \includegraphics[width=\columnwidth]{./images/Eformer.pdf}
%  \caption{Eformer \citep{luthra2021eformer} is a residual learning paradigm that aims to enhance the edge information by utilizing the learnable Sobel-Feldman filter. The design blocks are the successive Lewin Transformer \citep{wang2022uformer}, Concatenation block, Convolution block, and Downsampling or Upsampling blocks --- LC2(D/U). The selection of the up-sampling or down-sampling block depends on the path of the encoder or decoder.}
%  \label{fig:Eformer}
% \end{figure}

Akin to Low Dose CT (LDCT), Low-Dose PET (LDPET) is preferable to avoid the radiation risk, especially for cancer patients with a weakened immune system who require multiple PET scans during their treatment at the cost of sacrificing diagnosis accuracy in Standard-Dose PET (SDPET). Luo et al. \citep{luo20213d} proposed an end-to-end Generative Adversarial Network (GAN) based method integrated with a Transformer block, namely  \textbf{3D Transformer-GAN}, to reconstruct SDPET images from the corresponding LDPET images. To alleviate the inter-slice discontinuity problem of existing 2D methods, they designed their network to work with 3D PET data. Analogous to any GAN network, they used a generator network, encoder-decoder,  with a Transformer placed in the bottleneck of the generator network to capture contextual information. Due to the computational overhead of Transformers, they did not build their proposed method solely on it. Therefore, they were satisfied to place a Transformer counterpart across CNN layers of the generator to guarantee to extract low-level spatial feature extraction and global semantic dependencies. They also introduced adversarial loss term to their voxel-wise estimation error to produce more realistic images.

In contradiction with other works, Zhang et al. \citep{zhang2022spatial} proposed leveraging the PET/MRI data simultaneously for denoising low-count PET images, which is a crucial assessment for cancer treatment. PET scan is an emission Computed Tomography (CT) operating by positron annihilation radiation. Due to the foundation and requirements of PET scans, there is a severe risk of getting infected with secondary cancer by radiotracers. So to degrade the side effects of this imaging process, there are two potential methods: reduction in radiotracer dose and lessening the patient's bedtime duration. The aforementioned approaches, without a doubt, affect the imaging result quality with decreased contrast to noise ratio and bias in texture. The traditional low-count PET denoising approaches are based on Non-Local Means (NLM) \citep{buades2005non}, Block Matching 3D (BM3D) \citep{dabov2006image}, and Iterative methods \citep{wang2014low}, etc., which are firmly in bond with hyperparameter tuning for new data or result in unnatural smoothings over denoised images. Zhang et al. \citep{zhang2022spatial} testify that simultaneous PET/MRI could boost one modality in terms of correct attenuation, motion, and partial volume effects, and also, due to the high contrast among soft tissues in MRI, the denoising process of PET images is preferably straightforward. \textbf{STFNet} \citep{zhang2022spatial} is a U-Net based structure with different medications. They proposed a new Siamese encoder comprising dual input flow for each modality in the encoding path. To obtain sufficient features from different modalities, they used the Spatial Adaptive (SA) block, a dual path in each block with the residual block design, which consists of different consecutive convolutional blocks and deformable convolution with fusion modulation. This module aims to learn more contextual features from each modality. To leverage global attention, they used a Transformer to produce a pixel-to-pixel interaction between the PET and the MRI modality. After this integration, the fused features are input to the two branches based on residual convolution blocks for PET denoising.

\begin{figure}[h]
 \centering
 \includegraphics[width=0.99\columnwidth]{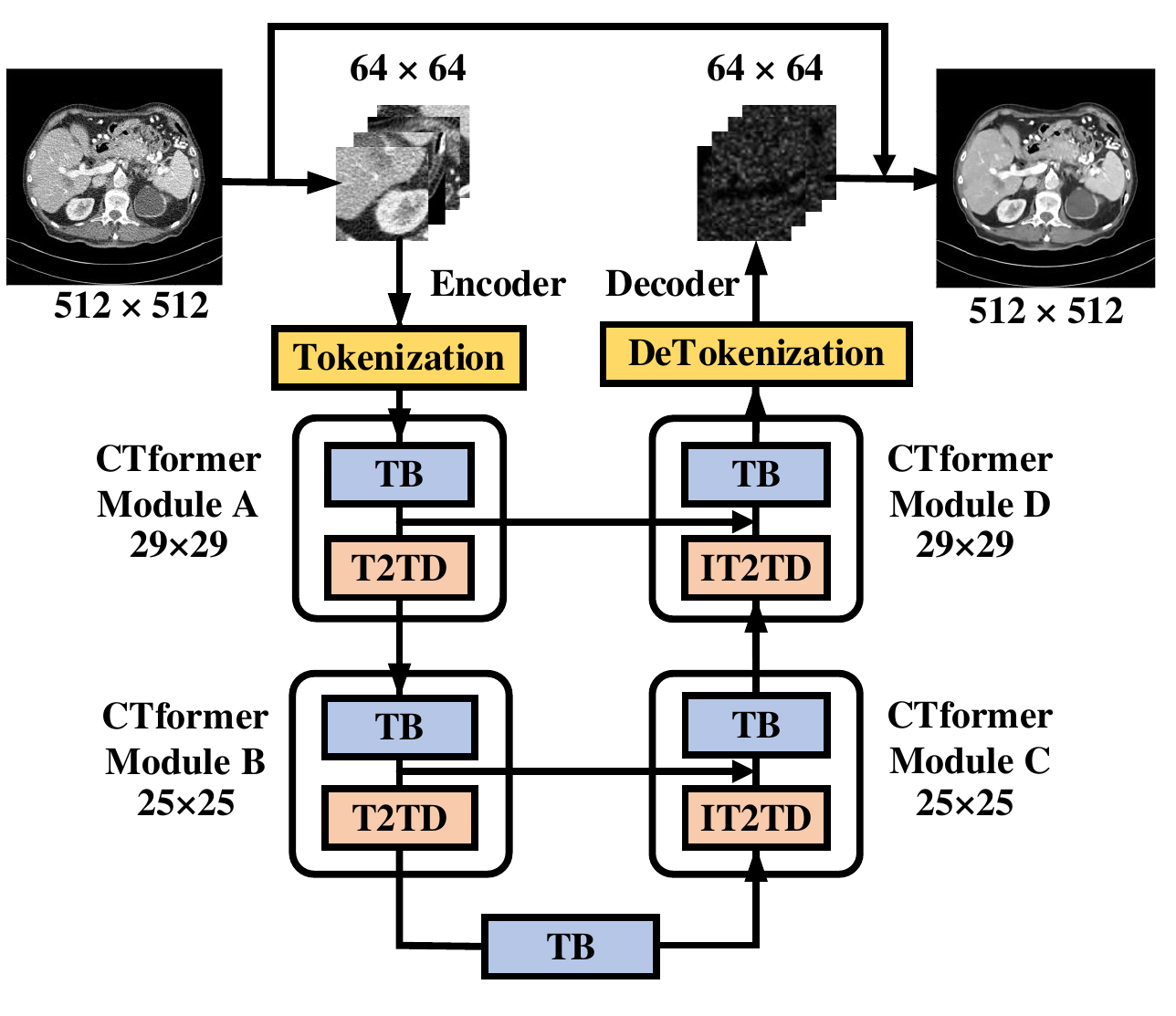}
 \caption{An overview of CTformer \citep{wang2023ctformer}. This structure is analogous to the TED-net \citep{wang2021ted} structure, the previous study by the same authors.}
 \label{fig:CTformer}
\end{figure}

Wang et al. \citep{wang2023ctformer} proposed the enhancement for their previous work, TED-net \citep{wang2021ted} convolution-free, solely Transformer-based network, namely \textbf{CTformer}. From \Cref{fig:CTformer}, it is apparent that their network is an unsupervised residual learning, U-Net-like encoder-decoder structure, rather than direct map learning of LDCT to Natural Dose CT (NDCT). The CTformer tries to compensate for the Transformers' deficiency in capturing path inter-boundary information and spatial inductive bias with token rearrangement, T2T \citep{yuan2021tokens}. To do so, analogously like TED-net, they used dilation and cyclic shift blocks in the Token2Token block to broaden the receptive field to capture more contextual information and not increase the computational cost.

\begin{table}[!thb]
	\centering
	\caption{Comparison result on NIH-AAPM-Mayo \citep{mccollough2017low} dataset in low dose enhancement task. \textit{LDE} indicates the Low Dose Enhancement task.}
	\label{tab:reconstruction_benchmarks}
	\resizebox{\columnwidth}{!}{
		\begin{tblr}{
				colspec={|l|c|cccc|},
				row{1}={LightCyan,c},
				column{1} = {font=\bfseries},
				hline{1,5} = {3pt},
				vline{1,6} = {3pt},
				hlines
			}
			\textbf{Methods} & \textbf{Task} & \textbf{Dataset} & \textbf{SSIM}\;$\uparrow$ & \textbf{RMSE}\;$\downarrow$ \\
			{Eformer \\ \citep{luthra2021eformer}} &LDE & {NIH-AAPM-Mayo \\ \citep{mccollough2017low}} & \textbf{0.9861} & \textbf{0.0067} \\
			{TransCT \\ \citep{zhang2021transct}} &‌LDE & {NIH-AAPM-Mayo \\ \citep{mccollough2017low}} & 0.923 & 22.123 \\
			%			\textcolor{BrickRed}{\faMedal} & { TED-net \\ \citep{wang2021ted}} & LDE & {NIH-AAPM-Mayo \\ \citep{mccollough2017low}} & 0.9144 & 8.7681 \\
			{CTformer \\ \citep{wang2023ctformer}} &LDE &  {NIH-AAPM-Mayo \\ \citep{mccollough2017low}} & 0.9121 & 9.0233 \\
			
		\end{tblr}
	}
\end{table}

\begin{figure}[h]
 \centering
 \includegraphics[width=\columnwidth]{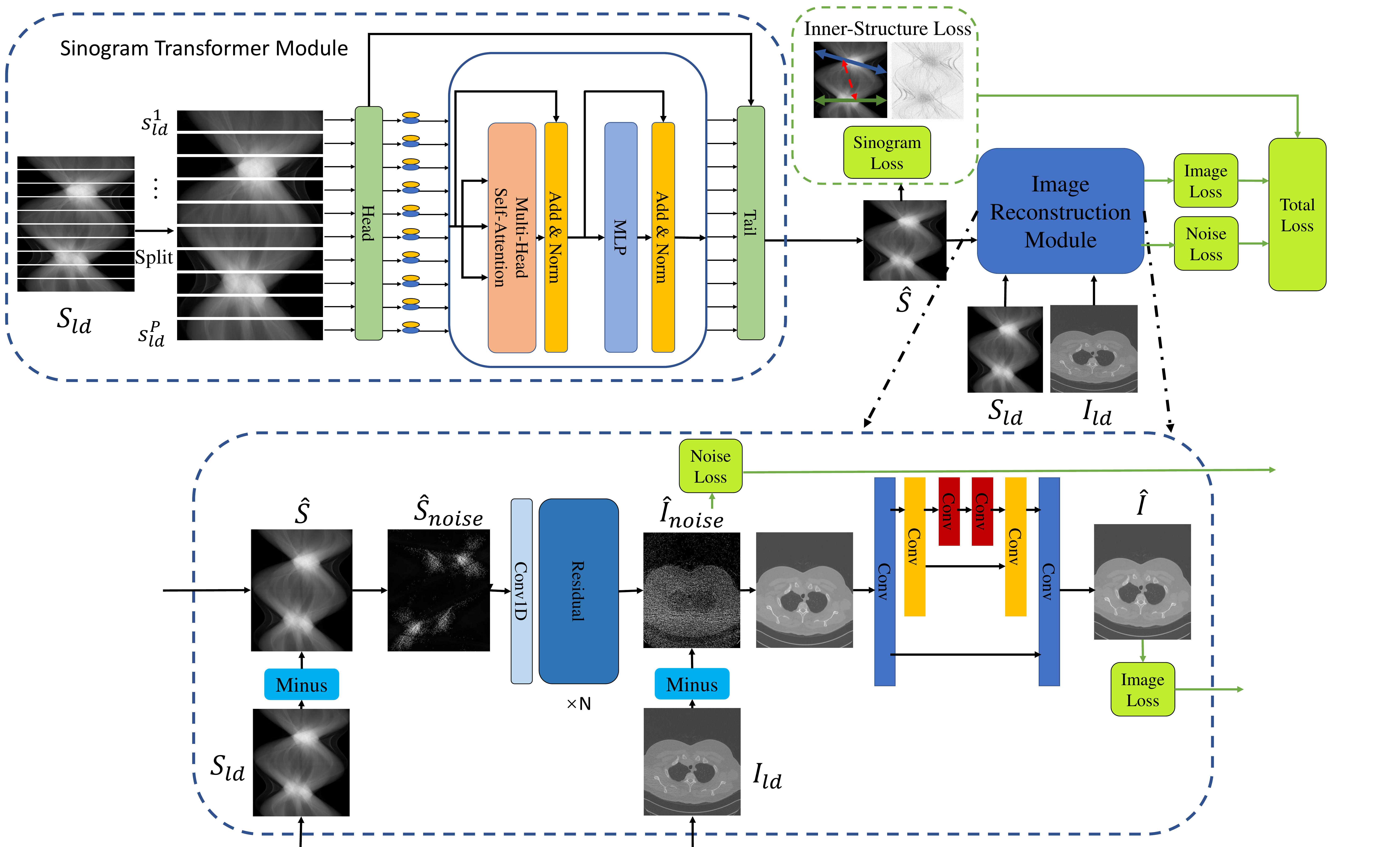}
 \caption{The overall architecture of SIST \citep{yang2022low} pipeline. $S_{ld}$ and $I_{ld}$ are the LDCT sinogram and image, $\hat{S}$ and $\hat{I}$ denote the output sinogram and image,  $\hat{S}_{noise}$ and $\hat{I}_{noise}$ are the sinogram noise and image noise. First, the LDCT sinogram feed to the Transformer for sinogram domain denoising, then the denoised sinogram $\hat{S}$input to the image reconstruction module for image domain denoising. Within the image reconstruction module, the sinogram noise $\hat{S}_{noise}$ with the usage of residual CNN block generates image domain $\hat{I}_{noise}$. NDCT $\hat{I}$, outputs from applying refinement steps on $I_{ld}$ minus $\hat{I}_{noise}$.}
 \label{fig:SIST}
\end{figure}

Yang et al. \citep{yang2022low} were inspired by how sinogram works and proposed Singoram Inner-Structure Transformer (\textbf{SIST}) (\Cref{fig:SIST}). This inner structure of the sinogram contains the unique characteristics of the sinogram domain. To do so, they mimic the global and local characteristics of sinogram in a loss function based on sinogram inner-structure, namely Sinogram Inner-Structure Loss (SISL). The global inner-structure loss utilizes conjugate sampling pairs in CT, and local inner-structure loss considers the second-order sparsity of sinograms. The amalgamation of these two terms could be beneficial in reconstructing NDCT images while retaining the noise. Due to the CT imaging mechanism, each row of the sinogram representation denotes the projection at a certain view. Naturally, this procedure is suitable for leveraging the Transformer block for modeling the interaction between different projections of diverse angles to capture contextual information. Therefore, the SIST module applies to raw sinogram input and captures structural information. Afterward, the unified network reconstructs the high-quality images in a residual policy with the image reconstruction module.

\Cref{tab:reconstruction_benchmarks} represents the benchmark results in the LDCT task over the NIH-AAPM-Mayo \citep{mccollough2017low} dataset respecting SSIM and RMSE metrics on overviewed methods in this study. For clarification, TED-net \citep{wang2021ted} achieved better results than CTformer \citep{wang2023ctformer}, but due to two studies originating from the same authors and the resemblance between architectures, we preferred to mention CTformer to count in the comparison table. This result endorses the capability of the pure Transformer-based Eformer \citep{luthra2021eformer} method in reconstructing natural dose CT images.

\subsection{Sparse-View Reconstruction}
\begin{figure}[h]
 \centering
 \includegraphics[width=\columnwidth]{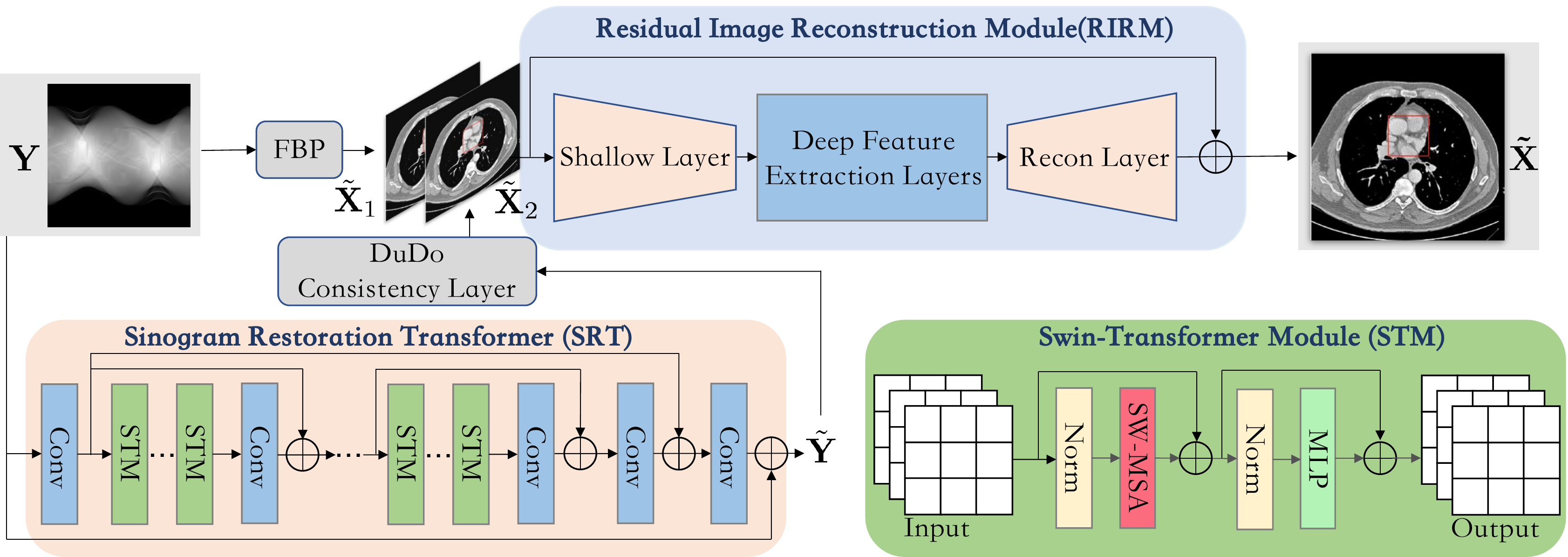}
 \caption{DuDoTrans \citep{wang2021dudotrans} framework for sparse-view CT image reconstruction. First, the sparse-view sinogram $\mathbf{Y}$ maps to a low-quality image $\widetilde {\mathbf{X}}_1$ and other estimation $\widetilde {\mathbf{X}}_2$ generated by SRT module's enhanced sinogram output $\widetilde{\mathbf{Y}}$ followed by DuDo Consistency Layer. Lastly, the predicted estimations are concatenated and fed to the RIRM module that outputs the CT image of $\widetilde{\mathbf{X}}$ in a supervised manner.}
 \label{fig:DuDoTrans}
\end{figure}

Due to the customary usage of CT images in medical diagnosis, another policy to lessen the side effects of X-ray radiation is acquiring fewer projections, known as sparse-view CT, which is a very feasible and effective method rather than manipulating the standard radiation dose \citep{bian2010evaluation,han2018framing}. However, the resultant images from this method suffer from severe artifacts, and decreasing the number of projections demands profound techniques to reconstruct high-quality images. Wang et al. \citep{wang2021dudotrans} is the first paper that inspected the usage of Transformers in this field which was quite successful, namely \textbf{DuDoTrans}. Their intuition was to shed light on the globality nature of the sinogram sampling process, which the previous CNN architectures neglected. DuDoTrans, unlike the conventional iterative methods in this literature, does not provide blocky effects in reconstructed images. This method simultaneously benefits from enhanced and raw sinogram streams to restore informative sinograms via long-range dependency modeling in a supervised policy. DuDoTrans from \Cref{fig:DuDoTrans} is built on three main modules, namely Singoram Restoration Transformer (SRT), the DuDo Consistency layer, and the Residual Image Reconstruction Module (RIRM). SRT block consists of successive hybrid Swin Transformer modules and convolutional layers to model local semantic features and inherent global contextual information in the sinogram to produce the enhanced sinogram.

\begin{figure}[h]
	\centering
	\includegraphics[width=\columnwidth]{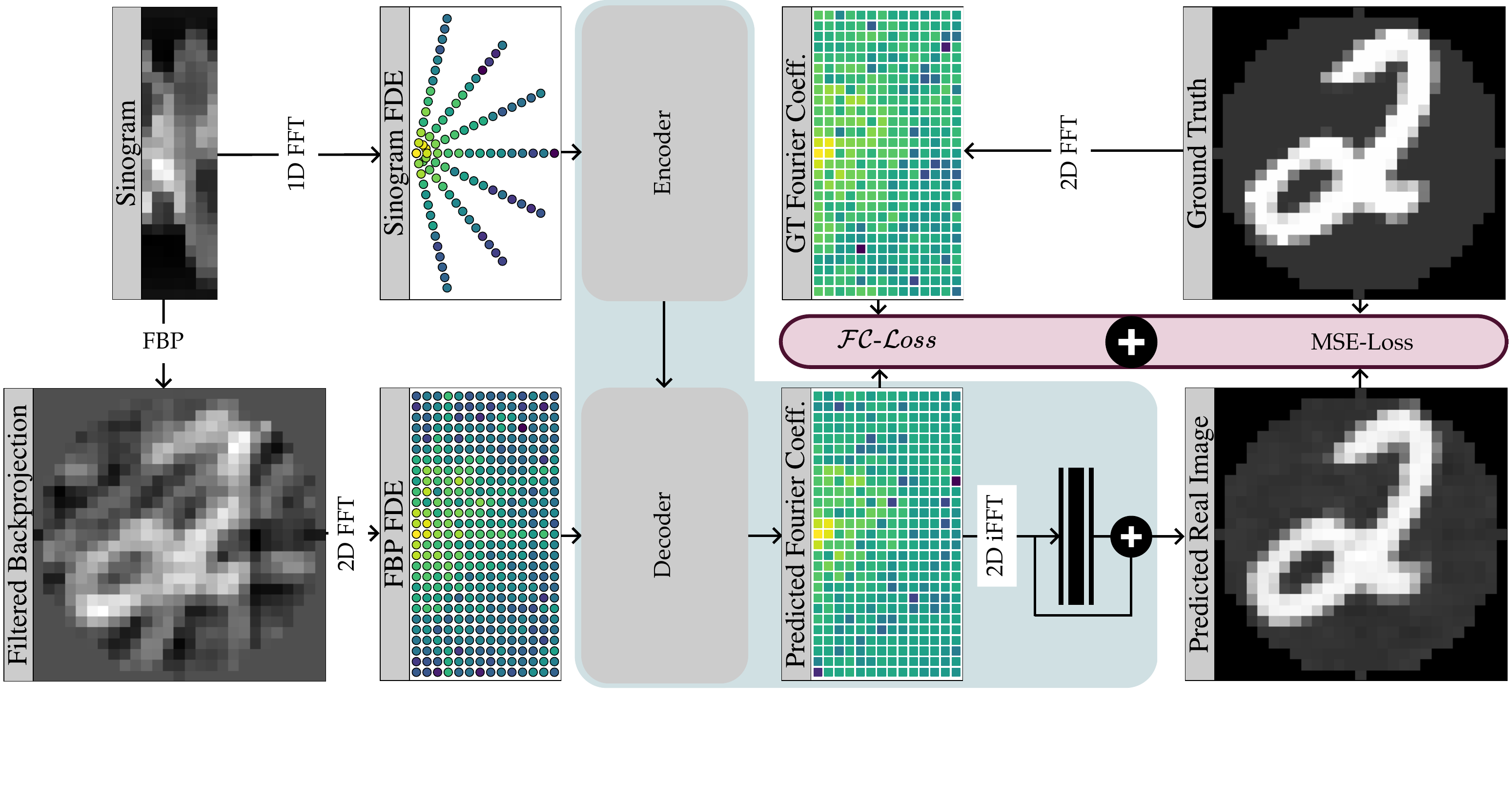}
	\caption{FIT \citep{buchholz2022fourier} framework for sparse-view CT reconstruction. FDE representation of the sinogram calculates that serves as an input to an encoder of Transformer design. The decoder predicts the Fourier coefficients from the encoder's latent space. The Fourier coefficients of applying the FBP \citep{pan2009commercial}  algorithm on sinogram information are fed into a Transformer's decoder to enrich the Fourier query representation. A shallow CNN block applies after inverse FFT to hamper the frequency oscillations.}
	\label{fig:FIT}
\end{figure}

Buchholz et al. \citep{buchholz2022fourier} presented the Fourier Image Transformer (\textbf{FIT}) that operates on the image frequency representation, especially the Fourier description of the image, which in their study is known as Fourier Domain Encoding (FDE), that encodes the entire image at lower resolution. The intuition in their idea is underlying the CT's acquisition process physics. CT utilizes a rotating 1D detector array around the patient body to calculate the Radon transform \citep{kak2001principles} of a 2D object, which leads to a sequence of density measurements at different projection angles, namely sinogram as a 2D image in which each column of this representation corresponds to one 1D measurement. The Filtered Back Projection (FBP) \citep{pan2009commercial,kak2001principles} is a reconstruction method to map sinograms to tangible CT images. FBP is based on the Fourier slice theorem; hence, computing the 1D Fourier transform of 1D projection and rearranging them by their projection angle in Fourier space, followed by an inverse Fourier transformation, results in a reconstructed 2D CT image slice. Limiting the number of projections leads to missing Fourier measurements, which ultimately conduce to reconstruction artifacts. FIT is the first study that uses a Transformer to query arbitrary Fourier coefficients and fill the unobserved Fourier coefficients to conceal or avoid the probable artifacts in reconstruction within sparse-view CT reconstruction literature. From \Cref{fig:FIT} this procedure starts with calculating the FDE of the raw sinogram. To do so, first, the discrete Fourier transform (DFT) of the sinogram will be calculated. Secondly, after dropping half of the coefficients on the Fourier rings of the resultant Fourier representation, it preserves the lower frequency counterparts to recover the lower resolution of the raw sinogram. Afterward, the complex coefficients convert into 1D sequences by unrolling the Fourier rings. These complex values convert to normalized amplitudes and phases. Therefore, each complex coefficient has its own polar representation, which is a normalized real-valued matrix with $N \times 2$ entries ($N$ is equal to half of the DFT coefficients number). A linear layer applies on this tensor to upsample the feature dimensionality to $\frac{F}{2}$. Finally, a 2D positional encoding concatenates to this tensor and produces a 2D FDE image with the size of $N \times F$.

Shi et al. \citep{shi2022dual} presented a CT reconstruction network with Transformers (\textbf{CTTR}) for sparse-view CT reconstruction. In contrast to DuDoTrans \citep{wang2021dudotrans}, CTTR enhances low-quality reconstructions directly from raw sinograms and focuses on global features in a simple policy in an end-to-end architecture. CTTR contains four parts: two CNN-based residual blocks extracting local features from FBP \citep{pan2009commercial} images reconstruction and sinograms, an encoder-decoder Transformer for long-range modeling dependencies, and contextual information between features, and a CNN block to map features to a high-quality reconstruction. 

\begin{figure*}[!th]
	\centering
	\begin{subfigure}[t]{\columnwidth}
		\centering
		\includegraphics[width=\textwidth]{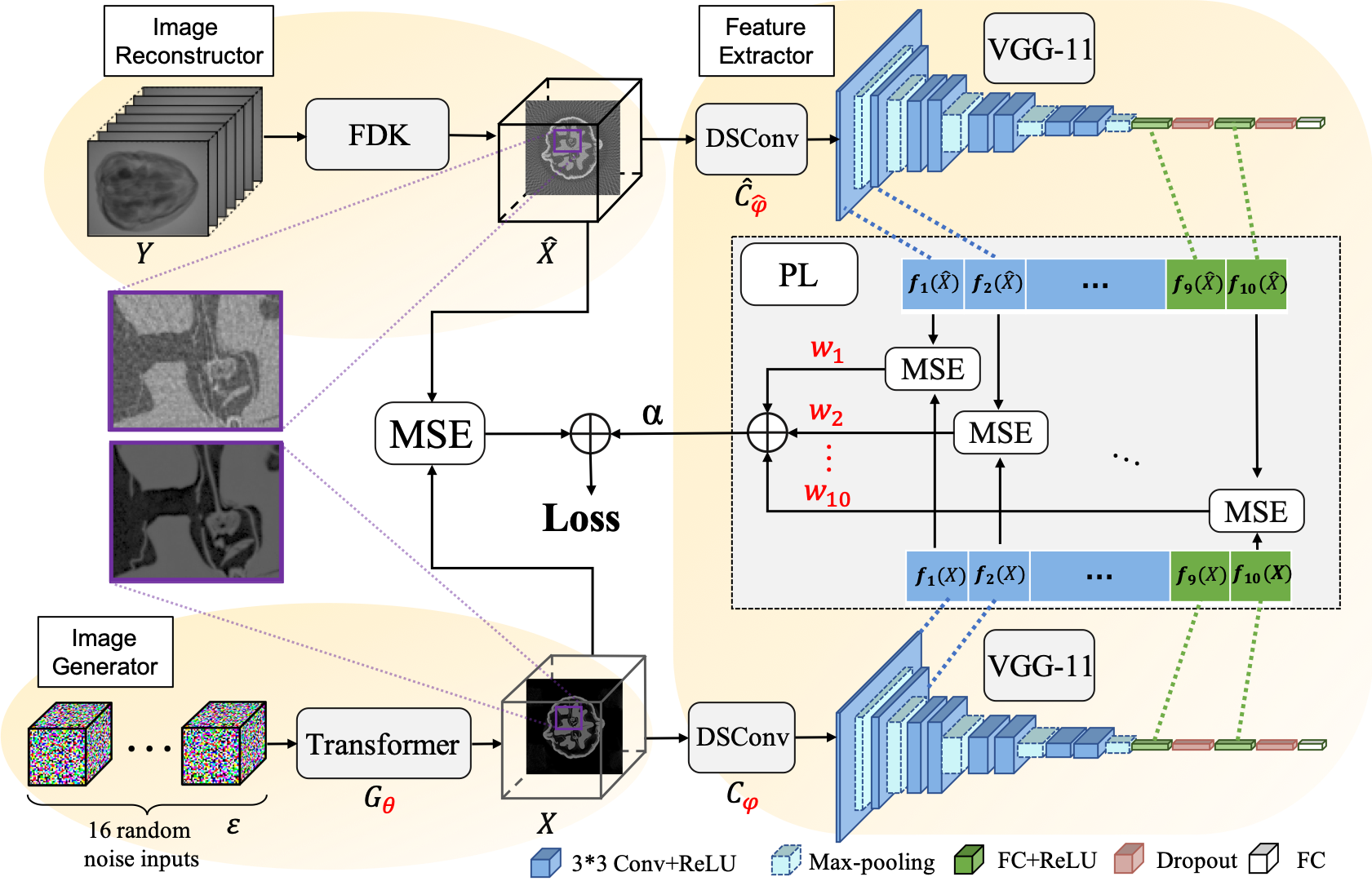}
		\caption{ARMLUT \citep{wu2022adaptively} pipeline. The collaboration of three distinct modules---FDK algorithm \citep{feldkamp1984practical}, prior embedding with Transformer, and VGG-11 network for extracting hierarchical features---in this pipeline generates the reconstructed CBCT image. Red texts in the figure denote the variable weights that contribute to the iterative optimization step.}
		\label{fig:UnifiedARMLUT}
	\end{subfigure}
	\hfill
	\begin{subfigure}[t]{\columnwidth}
		\centering
		\includegraphics[width=\textwidth]{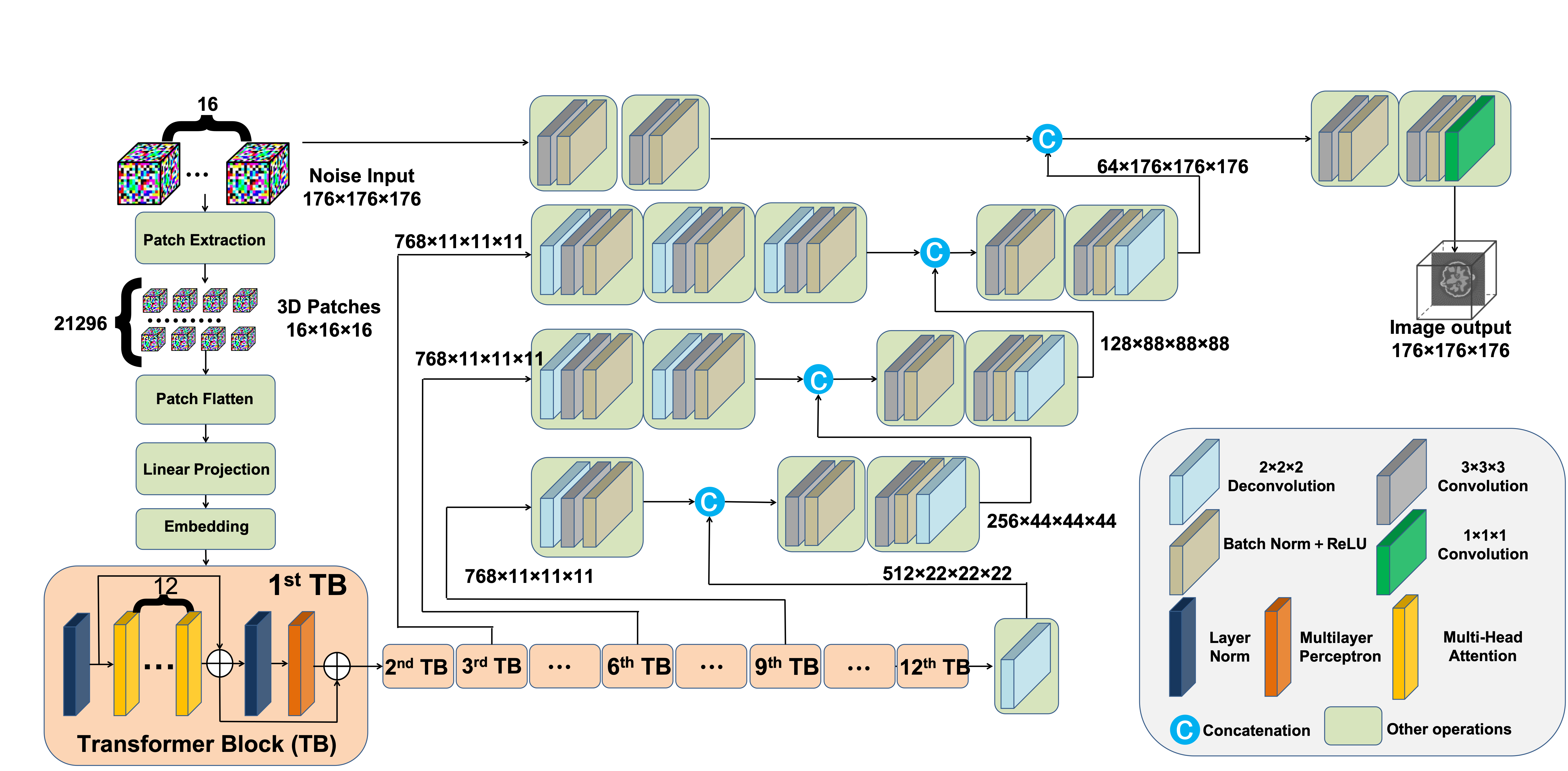}
		\caption{UNETR \citep{hatamizadeh2021unetr} used as an image generator module in the ARMLUT paradigm.}
		\label{fig:UNETR-ARMLUT}
	\end{subfigure}
	\caption{(\subref{fig:UnifiedARMLUT}) represents multi-loss untrained network for sparse-view CBCT reconstruction. (\subref{fig:UNETR-ARMLUT}) architecture of UNETR \citep{hatamizadeh2021unetr}, as a Transformer module in a ARMLUT.}
	\label{fig:ARMLUT}
\end{figure*}

Cone-Beam Computed Tomography (CBCT) is a conventional way of dental and maxillofacial imaging; due to its fast 3D imaging qualifications, its popularity has extended to lung imaging. However, studies approved that its radiation dose is higher than plain radiographs \citep{patel2019cone} hence sparse-view CBCT could be a suitable method to lower radiation dose. Wu et al. \citep{wu2022adaptively} proposed a novel untrained 3D Transformer-based architecture, namely \textbf{ARMLUT}, with a multi-level loss function for CBCT reconstruction. While the Transformer module, especially the UNETR \citep{hatamizadeh2021unetr} in this study, captures long-range contextual information and enhances the resulting image. The intuition behind this strategy is Deep Image Prior (DIP) \citep{ulyanov2018deep} to succeed in the reconstruction field. From \Cref{fig:UnifiedARMLUT}, ARMLUT is an iterative optimization problem between the Image Reconstructor module and Image Generator module to fit a CBCT inverse solver without a large number of data or ground truth images. The multi-level loss function comprises a Mean Squared Error (MSE) and Perceptual Loss (PL) \citep{johnson2016perceptual} to reconstruct smooth and streak artifact-free outputs. The entire framework (\Cref{fig:ARMLUT}) has three main counterparts: Image Reconstructor, Image Generator, and Feature Extractor. Image Reconstructor uses Feldkamp-Davis-Kress (FDK) algorithm \citep{feldkamp1984practical} to produce a low enhanced reconstruction from $M$-view measurements, and the Image generator module maps the noisy voxel inputs to reconstruct a regularised image. The Feature Extractor module applies the VGG-11 pre-trained network on two representations and produces a PL paradigm. To minimize the distance between these two reconstructions, ARMLUT utilizes an adaptively re-weight multi-loss technique to stabilize the convergence of the Transformer in the optimization.

\begin{table*}[!th]
    \centering
    \caption{Medical Image Reconstruction.  \textit{LDE}, \textit{SVR}, \textit{USR}, and \textit{SRR} stand for Low Dose Enhancement, Sparse-View Reconstruction, Undersampled Reconstruction, and Super Resolution Reconstruction, respectively.  $\dag$ indicates that this network uses a pre-trained perceptual loss (loss network).}
    \label{tab:reconstruction}
    \resizebox{\textwidth}{!}{
    \begin{tabular}{lccccccc}  
    \toprule
    \textbf{Method} & \textbf{Task(s)} & \textbf{Modality} & \textbf{Type} & \textbf{Pre-trained Module: Type} & \textbf{Dataset(s)} & \textbf{Metrics} & \textbf{Year} \\ 
    \rowcolor{LightCyan}\multicolumn{8}{c}{\textbf{Pure}} \\
    \makecell[l]{TED-net \citep{wang2021ted}} & LDE & CT &2D&\xmark &NIH-AAPM-Mayo Clinical LDCT \citep{mccollough2017low}&\makecell{SSIM \\ RMSE}&2021\\ \midrule
    \makecell[l]{Eformer \citep{luthra2021eformer}} & LDE & CT & 2D & \xmark$^\dag$ & NIH-AAPM-Mayo Clinical LDCT \citep{mccollough2017low}& \makecell{PSNR, SSIM \\ RMSE}&2021\\ \midrule
    \makecell[l]{CTformer \citep{wang2023ctformer}} & LDE & CT &‌2D & \xmark & NIH-AAPM-Mayo Clinical LDCT \citep{mccollough2017low}& \makecell{SSIM \\ RMSE} & 2023 \\ \midrule
    \makecell[l]{FIT \citep{buchholz2022fourier}} & SVR & CT & 2D & \xmark & LoDoPaB \citep{leuschner2019lodopab} & PSNR & 2021 \\ \midrule
    \makecell[l]{ViT-Rec \citep{lin2021vision}} &USR&MRI&2D&Supervised&fastMRI \citep{zbontar2018fastMRI}&SSIM&2021 \\ \midrule
    
    \rowcolor{LightCyan}\multicolumn{8}{c}{\textbf{Encoder}} \\
    \makecell[l]{TransCT \citep{zhang2021transct}} & LDE & CT & 2D & \xmark & \makecell{$^1$ NIH-AAPM-Mayo Clinical LDCT \citep{mccollough2017low}\\$^2$Private clinical pig‌ head CBCT } & \makecell{RMSE \\ SSIM \\VIF} & 2021 \\ \midrule
    \makecell[l]{STFNet \citep{zhang2022spatial}} &LDE&\makecell{PET\\ MRI}&2D&\xmark&Private Dataset&\makecell{RMSE, PSNR \\ SSIM, PCC}&2022\\ \midrule
    \makecell[l]{SIST \citep{yang2022low}} &LDE &CT&2D&\xmark&\makecell{$^1$LDCT Dataset \citep{moen2021low} \\ $^2$Private dataset}&\makecell{PSNR, SSIM \\ RMSE}&2022\\ \midrule
    \makecell[l]{DuDoTrans \citep{wang2021dudotrans}} & SVR & CT & 2D & \xmark & NIH-AAPM-Mayo Clinical LDCT \citep{mccollough2017low} & \makecell{PSNR, SSIM \\ RMSE} & 2021 \\ \midrule
    \makecell[l]{CTTR \citep{shi2022dual}} &SVR &CT&2D&\xmark&LIDC-IDRI \citep{armato2011lung}&\makecell{RMSE, PSNR \\  SSIM}&2022 \\ \midrule
    
    \rowcolor{LightCyan}\multicolumn{8}{c}{\textbf{Skip Connection}} \\
    \makecell[l]{Cohf-T \citep{fang2022cross}} & SRR & MRI & 2D & \xmark & \makecell{$^1$ BraTS2018 \citep{bakas2018identifying} \\ $^2$ IXI  \citep{ixidataset} }&\makecell{PSNR\\ SSIM}&2022\\ \midrule
    
    % \rowcolor{LightCyan}\multicolumn{8}{c}{\textbf{Decoder}} \\

    \rowcolor{LightCyan}\multicolumn{8}{c}{\textbf{Other Architectures}} \\
    \makecell[l]{3D T-GAN \citep{luo20213d}} & LDE & PET & 3D &\xmark & Private Dataset &\makecell{PSNR, SSIM \\ NMSE}&2021\\ \midrule
    \makecell[l]{ARMLUT \citep{wu2022adaptively}} &SVR&CT&3D&ViT: Supervised $^\dag$&\makecell{$^1$ SPARE Challenge Dataset \citep{shieh2019spare} \\ $^2$ Walnut dataset \citep{der2019cone}}&PSNR, SSIM&2022 \\ \midrule
    \makecell[l]{T$^2$Net \citep{feng2021task}} &\makecell{USR \\ SRR}&MRI&2D&\xmark&\makecell{$^1$ IXI \citep{ixidataset} \\ $^2$ Private Dataset}&\makecell{PSNR, SSIM \\ NMSE}&2021\\ \midrule
    \makecell[l]{DisCNN-ViT \citep{mahapatra2021mr}}&SRR&MRI&3D&ViT: Self-Supervised&\makecell{$^1$fastMRI \citep{zbontar2018fastMRI} \\ $^2$  IXI \citep{ixidataset}}&\makecell{PSNR, SSIM \\ NMSE}&2021 \\ \midrule
    
    \makecell[l]{SLATER} \citep{korkmaz2022unsupervised}& USR & MRI & 2D & ViT: Self-Supervised & \makecell{$^1$fastMRI \citep{zbontar2018fastMRI} \\ $^2$  IXI \citep{ixidataset}}&\makecell{PSNR, SSIM}& 2022 \\ \midrule
    
    \makecell[l]{DSFormer \citep{zhou2023dsformer}} & USR &  MRI  & 2D & ViT: Self-Supervised & \makecell{IXI \citep{ixidataset}}&\makecell{PSNR, SSIM}&2023 \\

\bottomrule
\end{tabular}
}
\end{table*}

\begin{table*}[!th]
	\centering
	\caption{A brief description of the reviewed Transformer cooperated in the medical image reconstruction field.}
	\label{tab:reconstruction_highlight}
	\resizebox{\textwidth}{!}{
		\begin{tblr}{lp{15cm}p{15cm}}
			\toprule
			\textbf{Method}  & \textbf{Contributions} & \textbf{Highlights} \\ 
			\SetRow{LightCyan}\SetCell[c=3]{c} \textbf{Low Dose Enhancement} & & \\
			{TransCT \citep{zhang2021transct}} & {$\bullet$ The proposed prototype was the first successful implementation of a Transformer complement to CNN in the Low Dose CT reconstruction domain by exploring its revenue within high-frequency and low-frequency counterparts.} &{$\bullet$ The Transformer effectively could learn the embedded texture representation from the noisy counterpart. \\ $\bullet$ This paradigm is not convolution-free and uses Transformer as a complement.}\\ \midrule
			{TED-net \citep{wang2021ted}} &{$\bullet$ Convolution-free U-Net based Transformer model.}&{$\bullet$ Introduced Dialted Token-to-Token-based token serialization for an improved receptive field in Transformers. \\ $\bullet$ Using Cyclic Shift block for feature refinement in tokenization.} \\ \midrule
			{Eformer \citep{luthra2021eformer}}& {$\bullet$ Incorporate the learnable Sobel filters into the network for preserving edge reconstruction and improving the overall performance of the network. \\ $\bullet$ Conduct diverse experiments to validate that the residual learning paradigm is much more effective than other learning techniques, such as deterministic learning approaches.} & {$\bullet$ Successfully imposed the Sobel-Feldman generated low-level edge features with intermediate network layers for better performance. \\ $\bullet$ To guarantee the convergence of the network, they used the Multi-scale Perceptual (MSP) loss alongside Mean Squared error (MSE) to hinder the generation of disfavored artifacts.} \\ \midrule
			{3D T-GAN \citep{luo20213d}} & {$\bullet$ It is a 3D-based method rather than conventional 2D methods. \\ $\bullet$ First LDPET enhancement study that leveraged from Transformer to model long-range contextual information.} & {$\bullet$ To produce more reliable images with a generator they used an adversarial loss to make the data distribution the same as real data.} \\ \midrule
			{STFNet \citep{zhang2022spatial}} & {$\bullet$ Proposed a dual input U-Net-based denoising structure for low-count PET images with excessive MRI modality contribution. \\ $\bullet$ Used a Transformer block as a hybrid add-on for feature fusion to make a pixel-to-pixel translation of PET and MRI modalities.} & {$\bullet$ In comparison with the U-Net and residual U-Net structures due to the different training strategy which is roughly named Siamese structure has a low computational burden and simplified network. \\ $\bullet$ This network successfully handled the disparity and nonuniformity of shape and modality of PET and MRI. \\ $\bullet$ The visual results of denoised images testify that the proposed method could recover the detail of texture more clearly than other networks.} & \\ \midrule
			{CTformer \citep{wang2023ctformer}} & {$\bullet$ Convolution-free, computational efficient design. \\ $\bullet$ Introduce a new inference mechanism to address the boundary artifacts. \\ $\bullet$ Proposed interpretability method to follow each path resultant attention map through the model to understand how the model is denoising.} &{$\bullet$ Alleviate receptive filed deficiency with the token rearrangement.} \\ \midrule
			{SIST \citep{yang2022low}} & {$\bullet$ Proposed inner-structure loss to mimic the physics of the functionality of sinogram processing by CT devices to restrain the noise. \\$\bullet$ Extracting the long-range dependencies between distinct sinogram angles of views via Transformer.}& {$\bullet$ Utilizing the image reconstruction module to alleviate the artifacts that could happen in sinogram domain denoising by transferring the sinogram noise into the image domain. \\$\bullet$ Image domain loss back-propagates into the sinogram domain for complementary optimization.} \\ 
			\SetRow{LightCyan}\SetCell[c=3]{c}\textbf{Sparse-View Reconstruction} & & \\
			{DuDoTrans \citep{wang2021dudotrans}} &{$\bullet$ To cope with the global nature of the sinogram sampling process introduced, the SRT module, a hybrid Transformer-CNN to capture long-range dependencies. \\$\bullet$ Utilizing a dual domain model to simultaneously enrich raw sinograms and reconstruct CT images with both enhanced and raw sinograms. \\$\bullet$ To compensate for the drift error between raw and enhanced sinogram representation employs DuDo Consistency Layer.}& {$\bullet$ Utilizing a residual learning paradigm for image-domain reconstruction. \\$\bullet$ Fewer parameters in comparison with other structures, e.g., DuDoNet \citep{lin2019dudonet} with better performance.} \\ \midrule
			{FIT \citep{buchholz2022fourier}} &{$\bullet$ Introduced the Fourier Domain Encoding to encode the image to a lower resolution representation for feeding to the encoder-decoder Transformer for reconstructing the sparse-view CT measurements.}& {$\bullet$ Introduced the Fourier coefficient loss as a multiplicative combination of amplitude loss and phase loss in the complex domain.} \\ \midrule
			{CTTR \citep{shi2022dual}} &{$\bullet$ Introduced the encoder-decoder Transformer pipeline to utilize dual-domain information, raw sinogram, and primary reconstruction of CT via FBP \citep{pan2009commercial} algorithm for sparse-view CT measurement reconstruction.}& {$\bullet$ In contrast to DuDoTrans  \citep{wang2021dudotrans}, CTTR directly utilizes raw sinograms to enhance reconstruction performance.} \\ \midrule
			{ARMLUT \citep{wu2022adaptively}} & {$\bullet$ Proposed a paradigm for CT image reconstruction in a non-trainable manner. \\ $\bullet$ Extending the most DIP research on 2D to 3D medical imaging scenario.}& {$\bullet$ Optimising the large-scale 3D Transformer with only one reference data in an unsupervised manner. \\ $\bullet$ Stabilising the iterative optimization of multi-loss untrained Transformer via re-weighting technique.} \\ 
			\SetRow{LightCyan}\SetCell[c=3]{c} \textbf{Undersampled Reconstruction} & & \\
			{ViT-Rec \citep{lin2021vision}} & {$\bullet$ This study investigated the advantage of pure Transformer framework, ViT, in fastMRI reconstruction problems in comparison with the baseline U-Net. \\$\bullet$ ViT benefits from less inference time and memory consumption compared to the U-Net.}& {$\bullet$ Utilizing pre-training weights, e.g., ImageNet, extensively improves the performance of ViT in the low-data regime for fastMRI reconstruction, a widespread concept in the medical domain. \\$\bullet$ ViTs that accompany pre-training weights demonstrate more robust performance toward anatomy shifts.} \\ \midrule
			{T$^2$Net \citep{feng2021task}} &{$\bullet$ Introduce the first Transformer utilized multi-task learning network in the literature. \\ $\bullet$ Designed the task Transformer for maintaining and feature transforming between branches in the network. \\ $\bullet$ Outperformed the sequentially designed networks for simultaneous MRI reconstruction and super-resolution with T$^2$Net.} &{$\bullet$ Used the same backbone for feature extraction in branches, however, the purpose of the branches is diverse.} \\
			\SetRow{LightCyan}\SetCell[c=3]{c}\textbf{Super Resolution Reconstruction} & & \\
			{DisCNN-ViT \citep{mahapatra2021mr}} & {$\bullet$ Using a Transformer-based network to capture global contextual cues and amalgamate them with CNN’s local information results in the superior quality of high-resolution images in super-resolution literature. \\ $\bullet$ Creating realistic images is just not a burden on an adversarial loss function, in addition, multiple loss functions incorporate extra constraints that preserve anatomical and textural information in the begotten image.} & {$\bullet$ Multi prerequisite steps are required to train the experiments; however, the super-resolution step is a straightforward end-to-end network. \\ $\bullet$ Need fine-tuning steps for two disentanglement and UNETR networks. \\ $\bullet$ The computational burden of UNETR is high and could use new efficient transformer-designed networks.} \\ \midrule
			{Cohf-T \citep{fang2022cross}} & {$\bullet$ Leverage the high-resolution T1WI due to its rich structural information for super-resolving T2-weighted MR images. \\$\bullet$ Introduced the high-frequency structure prior and intra-modality and inter-modality attention paradigms within the Cohf-T framework.} & {$\bullet$ Assess prior knowledge into super-resolution paradigm successfully. \\$\bullet$ Competitive number of FLOPS in reaching SOTA PSNR results in comparison with other attention-based networks. \\$\bullet$ End-to-end pipeline for training the network.} \\

            \SetRow{LightCyan}\SetCell[c=3]
            {c}\textbf{Self-supervised Reconstruction} & & \\
			{SLATER \citep{korkmaz2022unsupervised}} & {$\bullet$ {A Self-supervised MRI reconstruction framework that uses unconditional adversarial network and cross-attention transformers for high-quality zero-shot reconstruction.} \\ $\bullet$ {The first work that proposes an adversarial vision transformer framework designed for the task of MRI reconstruction.} } & {$\bullet$ {SLATER outperforms CNN and self-attention GAN models, showing better invertibility and improved reconstruction performance with cross-attention transformer blocks and unsupervised pretraining.} \\ $\bullet$ {Prior adaptation is computationally burdensome but improves across-domain reconstructions.} } \\ \midrule
			{DSFormer \citep{zhou2023dsformer}} & {$\bullet$ {Proposes a deep learning model for fast MRI reconstruction that uses a deep conditional cascade transformer and self-supervised learning to exploit contrast redundancy and reduce data requirements.} } & {$\bullet$ {Both k-space and image-domain self-supervision independently produce strong reconstruction quality, showing the benefits of self-supervision in both domains.} } \\
			\bottomrule
		\end{tblr}
	}
\end{table*}

\subsection{Undersampled Reconstruction}
Magnetic Resonance Imaging (MRI) is a dominant technique for assistive diagnosis. However, due to the physics behind its operation, the scanning time can take longer and be very tedious, affecting the patient experience and leading to inevitable artifacts in images \citep{plenge2012super}. Hence, reducing the number of MRI measurements can result in faster scan times and artifacts reduction due to the patient's movement at the cost of aliasing artifacts in the image \citep{hyun2018deep}.

Lin et al. \citep{lin2021vision} proposed a comprehensive analytical study to investigate the usage of ViT in a pure (CNN-free modules) and most straightforward Transformer design. This study is evidence of the prominent effect of ViTs in medical image reconstruction. For this work, they adopted the original ViT \citep{dosovitskiy2020image} for image reconstruction by discarding the classification token and replacing the classification head with the reconstruction head, which is comprised of successive Norm and Linear layers to map the Transformer output to a visual image. They performed a complete ablation study with different ViT settings, from the number of stacked Transformers to embedding dimension and the number of heads in Multi-Head Self-Attention (MHSA). Their results were quite effective and proved that trained ViT on sufficient data from natural images like ImageNet or medical data could perform better or achieve on-par reconstruction accuracies compared to CNN baselines such as U-Net \citep{ronneberger2015unet}. The proposed design's distinguished power based on the mean attention distance metric \citep{d2021convit} proves that it effectively mimics the convolutional receptive fields and could concurrently capture local and global dependencies. In addition, they showed that the ViT benefits from two times faster inference times and fewer memory requirements compared to the U-Net.

\begin{figure}[!th]
	\centering
	\begin{subfigure}[t]{\columnwidth}
	\centering
	\includegraphics[width=\textwidth]{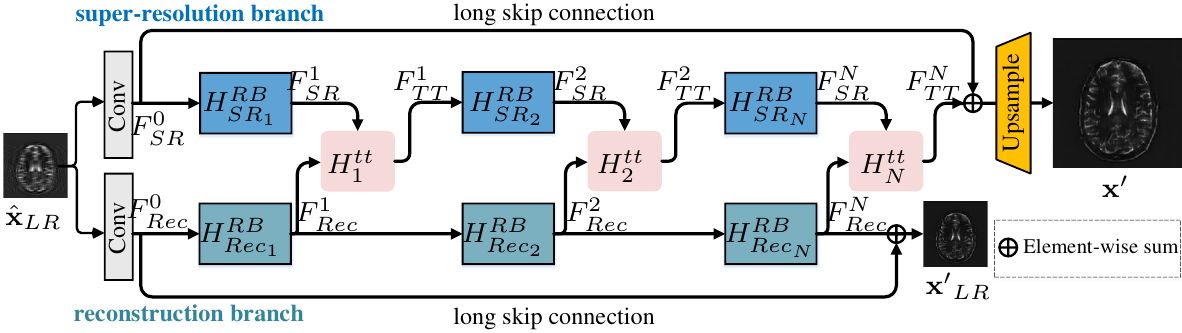}
	\caption{An overview of T$^2$Net \citep{feng2021taskFig}, a multi-task learning framework that consists of a super-resolution branch and reconstruction branch. The reconstruction branch embraces the stronger capability of artifact removal therefore, the task Transformer module is fed with the reconstruction branch.}
	\label{fig:multi-task-T2Net}
	\end{subfigure}
	\hfill
	\begin{subfigure}[t]{\columnwidth}
	\centering
	\includegraphics[width=\textwidth]{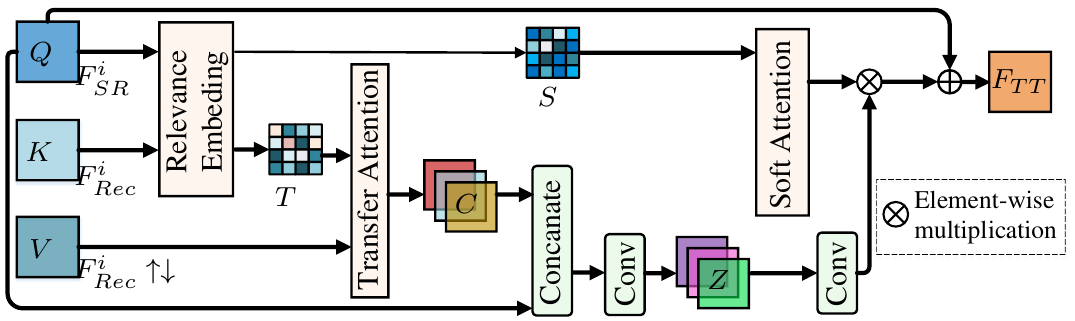}
	\caption{inner design of proposed $H^{tt}$ task Transformer module. Analogous to \Cref{fig:vitbase}, the design of T$^2$ module follows the naive design with some modifications, and in contrast to seminal design, all $Q$, $K$, and $V$ entities do not originate from the same representation---$Q$ comes from the super-resolution branch and the rest from the reconstruction branch.}
	\label{fig:task-Transformer-module}
	\end{subfigure}
	\caption{An overview of T$^2$Net \citep{feng2021taskFig}. (\subref{fig:multi-task-T2Net}) Multi-Task T$^2$Net pipeline and (\subref{fig:task-Transformer-module}) Task Transformer Module---T$^2$ Module}
	\label{fig:T2Net}
\end{figure}

Feng et al. \citep{feng2021task} address the particular issue in this domain by designing an end-to-end multi-task learning paradigm to boost feature learning between two sub-tasks, MRI reconstruction, and super-resolution, which have a high overlap with each other named Task Transformer Network (\textbf{T$^2$Net}) is showed in \Cref{fig:T2Net}. Their network consists of two branches, each for a specific task. T$^2$Net utilizes a Transformer between two branches to share feature representation and transmission. T$^2$Net applies a convolution layer and EDSR \citep{lim2017enhanced} backbone to extract task-specific features in each task branch. To share information between two branches and benefit from the interactions of these two task-specific features concerning the nature of the Transformer's globality, T$^2$Net uses a unique Transformer design to learn a generalized representation. Since the reconstruction branch has more potential in artifact removal capacity than the super-resolution branch, the task Transformer module guides the super-resolution branch into high-quality representation from the reconstruction branch. The Transformer module inherits the query ($Q$: from super-resolution branch), key ($K$: from reconstruction branch), and value ($V$: from reconstruction branch) from each scale's two branches' output. It forms three main concepts: relevance embedding, Transfer attention, and soft attention, which differ from the original Transformer blocks. Relevance embedding tries to enclose the correlated features from the reconstruction branch to the super-resolution branch. Transfer attention aims to transmit the anatomical and global features between two branches, and last but not least, soft attention amalgamates features from the previous two steps. Ultimately, this module lets the whole network transfer and synthesize the representative and anatomical features to produce a high-quality, artifacts-free representation from highly undersampled measurements. The experimental results on two datasets expressed the high potential of this approach rather than conventional algorithms.

\subsection{Super-Resolution Reconstruction}
Improving the resolution of images leads to the more detailed delineation of objects. Increasing the medical image resolution plays a crucial role in computer-aided diagnosis due to its rich anatomical and textural representation. Based on the aforementioned fact and the MRI's pipeline physics during the image acquisition process for having high-resolution images, a patient needs to lie a long time in the MRI tube. Hereupon lower signal-to-noise ratio and more minor spatial coverage drawbacks are inevitable \citep{plenge2012super}. Therefore in this section, we investigate Transformer-utilized algorithms that try to alleviate this problem. Of note, due to the analogy between MRI and super-resolution reconstruction, some studies investigate these two tasks in conjunction with each other.

Mahapatra et al. \citep{mahapatra2021mr} proposed the GAN-based model with structural and textural information preservation done by multiple loss function terms. Their pipeline included two pre-trained modules named feature disentanglement module, a conventional autoencoder, and a Transformer-based feature encoder, UNETR \citep{hatamizadeh2021unetr}. UNETR captures the global and local context of the original low-resolution image and induces the high-resolution image to preserve these contexts too. These two modules fine-tune on a different medical dataset, and afterward, the low-resolution input plus the intermediate generator produced image feed to these modules. The disentanglement network contains two autoencoders to learn two counterparts, latent space, structural and textural space, with fed medical images. In an end-to-end setting, these two pre-trained assistive modules help to generate more realistic and structural and textural preserving high-resolution images by imposing module-related loss terms such as adversarial loss to constrain for producing realistic images and cosine similarity loss for each mentioned module. Results on the IXI dataset proved that Mahapatra et al.'s \citep{mahapatra2021mr} network outperformed a couple of the CNN-based attention mechanism networks and T$^2$Net \citep{feng2021task}.
\begin{figure}[!th]
	\centering
	\includegraphics[width=\columnwidth]{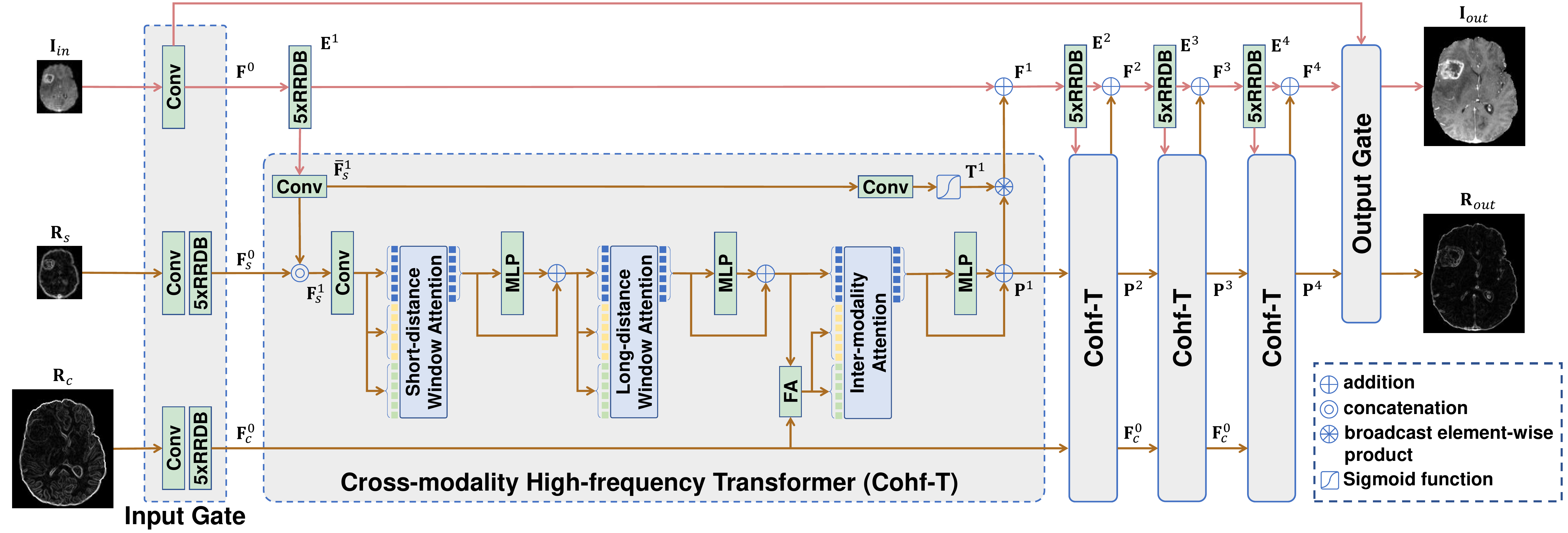}
	\caption{The pipeline of Cohf-T \citep{fang2022cross} consists of three main branches with the corresponding input modalities as follows: $\mathbf{I}_{in}$, $\mathbf{R}_s$, and $\mathbf{R}_c$ denote the low-resolution T2WI, the gradient of low-resolution T2WI and high-resolution T1WI, respectively. A fully-convolutional branch for density-domain super-resolution, a Transformer-based branch for restoring high-frequency signals in the gradient domain, and a guidance branch for extracting priors from the T1 modality. \textit{Conv}, \textit{RRDB} and \textit{MLP} represent a $3\times 3$ convolution operation and residual-in-residual dense block and multi-layer perceptron, respectively.}
	\label{fig:Cohf-T}
\end{figure}
Maintaining structural information during the acquiring high-resolution images plays a crucial role. Hence, the structure information is embedded in an image's high-frequency counterpart, like in an image's gradients. In addition, due to the less time-consuming nature of obtaining MR T1WI (T1 Weighted Image), it is wise to use it as an inter-modality context prior to producing a high-resolution image. Accordingly, Fang et al. \citep{fang2022cross} devised a network (see \Cref{fig:Cohf-T}) to leverage these two concerns in their super-resolution pipeline: Cross-Modality High-Frequency Transformer (\textbf{Cohf-T}). This network is divided into two streams, the first stream is applied on low-resolution T2WI, and the second one manipulates T2WI's gradient and high-resolution T1WI. The Cohf-T module interacts between two streams to embed the prior knowledge in the super-resolution stream's features. The Cohf-T module consists of three different attention modules: short-distance and long-distance window attention and inter-modality attention. The first two attention modules help to model intra-modality dependency. To be precise, the short-distance window helps recover the local discontinuity in boundaries with the help of surrounding structure information, and the long-distance window can capture the textural and structural patterns for enhanced results. Due to the discrepancy in intensity levels between T1WI and T2WI, it is vital to make an alignment between these two domains, and Fang et al. \citep{fang2022cross} presented a Feature Alignment (FA) module to reduce the cross-modality representation gap. They compared their results with T$^2$Net \citep{feng2021task} and MTrans \citep{feng2022multi}, which outperformed both approaches by $\sim 1\%$ in terms of PSNR.

\subsection{Self-supervised Reconstruction}

{Recently, self-supervised reconstruction methods requiring only undersampled k-space data have been proposed for single-contrast MRI reconstruction, enabling the autonomous learning of high-quality image reconstructions without the need for fully sampled data or manual annotations. \textbf{SLATER} \citep{korkmaz2022unsupervised} is introduced as a self-supervised MRI reconstruction method that combines an unconditional adversarial network with cross-attention transformers. It learns a high-quality MRI prior during pre-training and performs zero-shot reconstruction by optimizing the prior to match undersampled data. SLATER improves contextual feature capture and outperforms existing methods in brain MRI. It has potential for other anatomies and imaging modalities, offering reduced supervision requirements and subject-specific adaptation for accelerated MRI. Likewise, \textbf{DSFormer} \citep{zhou2023dsformer} is presented for accelerated multi-contrast MRI reconstruction. It uses a deep conditional cascade transformer (DCCT) with Swin transformer blocks. DSFormer is trained using self-supervised learning in both k-space and image domains. It exploits redundancy between contrasts and models long-range interactions. It also reduces the need for paired and fully-sampled data, making it more cost-effective.}

\begin{tcolorbox}[breakable ,colback={LightCyan},title={\subsection{Discussion and Conclusion}},colbacktitle=LightCyan,coltitle=black , left=2pt , right =2pt]
In this section, we outline different Transformer-based approaches for medical image reconstruction and present a detailed taxonomy of reconstruction approaches. We reviewed {17} studies that profit from the Transformer design to compensate for the deficiency of CNN's limited receptive field, {with two of them incorporating self-supervised strategies \citep{korkmaz2022unsupervised,zhou2023dsformer} for training transformer modules}. We investigate each study in depth and represent \Cref{tab:reconstruction} for detailed information about the dataset, utilized metrics, modality, and objective tasks. In \Cref{tab:reconstruction_highlight}, we provide the main contribution of each study and the prominent highlight of each method, {and in \Cref{tab:reconstruction_benchmarks}, we present a comparison of various methods for the low-dose enhancement task using the NIH-AAPM-Mayo \citep{mccollough2017low} benchmark medical reconstruction dataset, which shows that Eformer \citep{luthra2021eformer} is superior as compared to TransCT \citep{zhang2021transct} and CTformer \citep{wang2023ctformer}, considering its higher SSIM and lower RMSE scores.}

Most of the studies in this domain use the original Transformer as a plug-and-play module in their design and only a limited number of studies utilize hierarchical and efficient Transformers. However, the criteria for using multi-scale and hierarchical architectures are generally important for dense prediction tasks, e.g. image reconstruction, and should be considered further. Also, another direction to follow for future research could be to investigate the influence of using pre-training weights on Transformers due to the need for a large amount of data for better convergence results in Transformers, which contradicts the nature of the medical domain, due to the scarceness of annotated medical data.

In addition, we noticed that most of the studies focus on MRI and CT image reconstruction tasks. So there is a need for evaluating the applicability of these methods on other modalities, too.
\end{tcolorbox}

%% End of Ehsan Section--------------------

\section{Medical Image Synthesis} \label{sec:synthesize}
\begin{figure}[!th]
	\centering
	\includegraphics[width=\columnwidth]{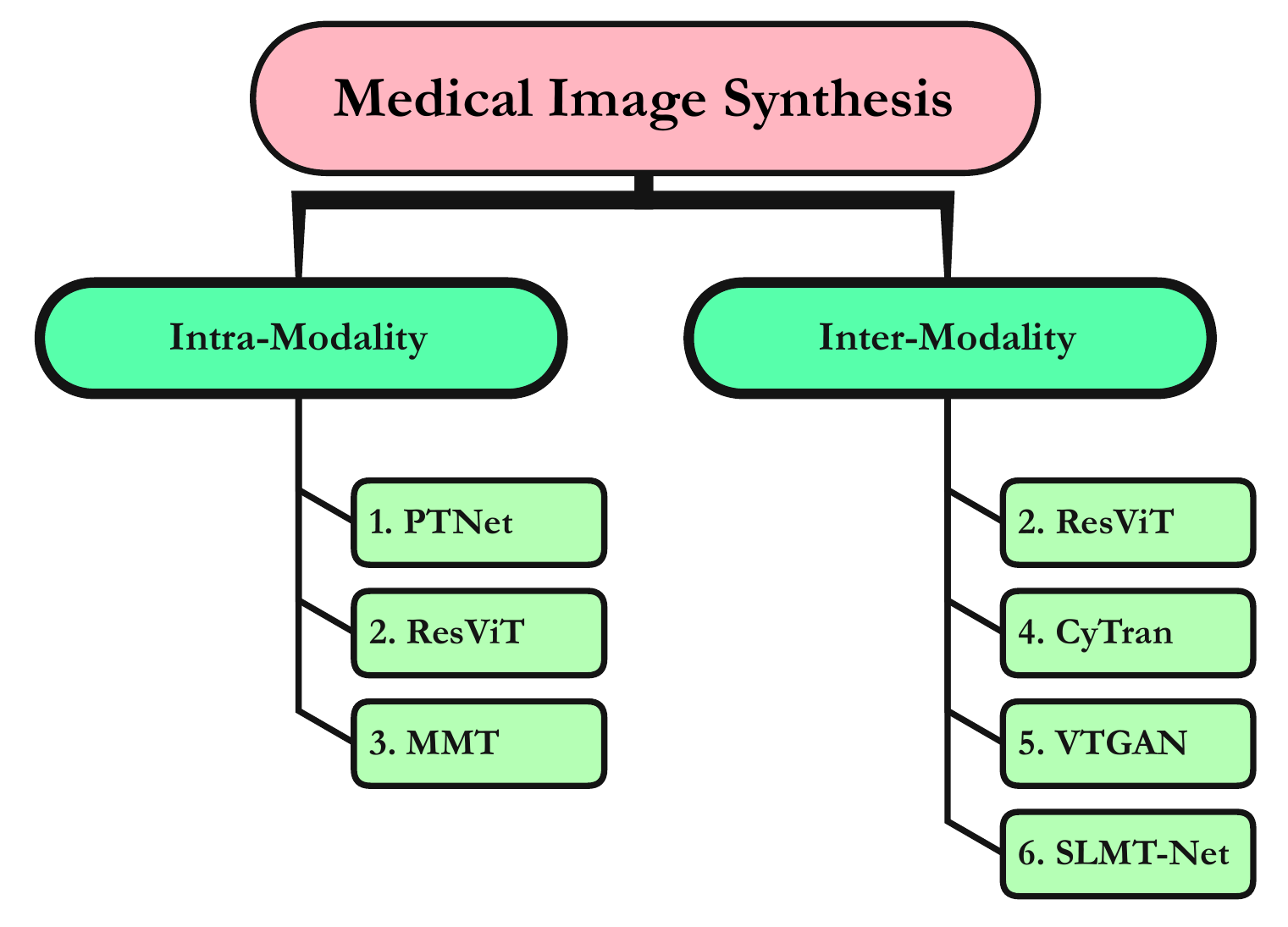}
	\caption{An overview of ViTs in medical image synthesis. Methods are categorized by target and source modality. The prefix numbers in the paper’s name in ascending order denote the reference for each study as follows: 1. \citep{zhang2021ptnet}, 2. \citep{dalmaz2022resvit}, 3. \citep{liu2022one}, 4. \citep{ristea2021cytran}, 5. \citep{kamran2021vtgan}, 6.
    \citep{li2022slmt}.}
	\label{fig:synthesizing-taxonomy}
\end{figure}

In this section, we will overview several instances of Transformers in the medical image synthesis task. The scarcity of medical data and the high cost of acquisition processes make this task very valuable in the medical field. Some studies aim to synthesize missing slices from MRI and CT sequences. In addition, some methods target capturing the structural information in diverse modalities, e.g., CT to MRI image-to-image translation and vice versa. \Cref{fig:synthesizing-taxonomy} shows our taxonomy for the image-synthesized methods. 

\subsection{Intra-Modality}
The main objective of the intra-modality methods is to synthesize high-quality images using low-quality samples from the same modality.
In this respect, several Transformer-based approaches are presented to formulate the synthesis task as a sequence-to-sequence matching problem to generate fine-grained features. In this section, we will briefly present some recent samples \citep{zhang2021ptnet}. 

Brain development monitoring is a de facto standard in predicting later risks; hence it is critical to screen brain biomarkers via available imaging tools from early life stages. Due to this concern and the nature of MRI and subjective infants' restlessness, it is not relatively straightforward to take all the MR modalities during the MRI acquisition. Zhang et al. \citep{zhang2021ptnet} proposed a \textbf{Pyramid Transformer Net (PTNet)} as a tool to reconstruct realistic T1WI images from T2WI. This pipeline is an end-to-end Transformer-based U-Net-like and multi-resolution structure network utilizing an efficient Transformer, Performer \citep{choromanski2020rethinking}, in its encoder (PE) and decoder (PD). Analogously to the original U-Net \citep{ronneberger2015unet}, they used skip connection paths for preserving fine-grained features and accurate localization features for reconstruction. Moreover, the paradigm's two-level pyramidal design helps the network capture local and global information in a multi-resolution fashion. They achieved the SOTA results on the dHCP \citep{makropoulos2018developing} dataset compared with the flagship GAN-based image generation method pix2pix (HD) \citep{isola2017image,wang2018high}.

\begin{figure}[!th]
	\centering
	\includegraphics[width=\columnwidth]{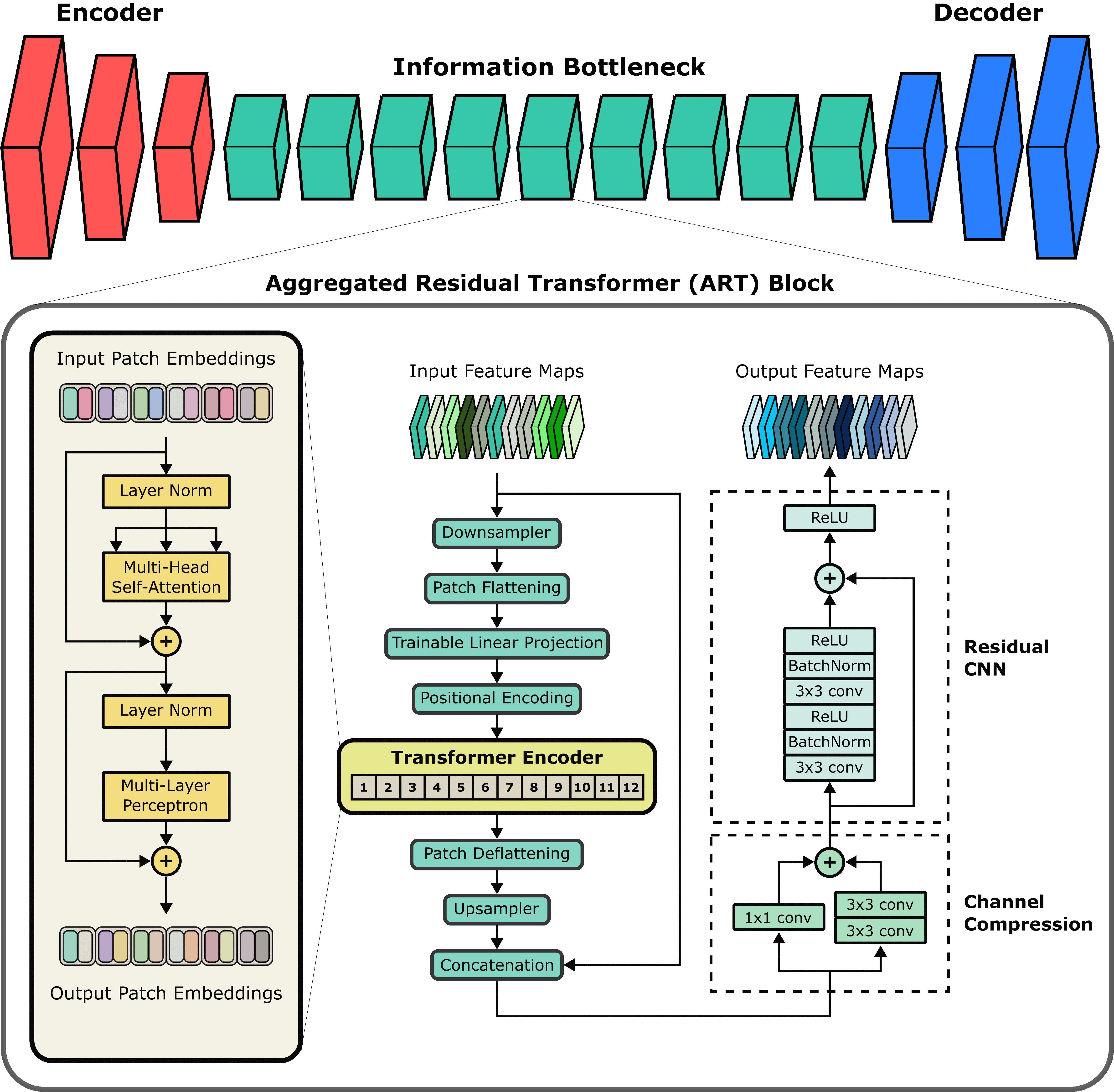}
	\caption{The ResViT \citep{dalmaz2022resvit} framework for multi-modal medical image synthesis. The bottleneck of this encode-decoder comprises successively residual Transformers and residual convolutions layers for synergistically capturing the fine-grained global and local context.}
	\label{fig:ResViT}
\end{figure}

Dalmaz et al. \citep{dalmaz2022resvit} introduced a conditional generative adversarial network based on the cooperation of Transformers and CNN operators, namely \textbf{ResViT}. This paradigm addresses the issue of needing to rebuild separate synthesis models for varying source-target modality settings and represents a unified framework as a single model for elevating its practicality. The ResViT (\Cref{fig:ResViT}) pervasively refers to the generator of its pipeline, whereby it leverages a hybrid pipeline of residual convolutional operations and Transformer blocks that enable effective aggregation of local and long-range dependencies. The discriminator is based on a conditional PatchGAN framework \citep{isola2017image}. Utilizing standalone Transformer architectures (e.g., PTNet \citep{zhang2021ptnet}) in pixel-to-pixel tasks is quite challenging due to the quadratic complexity, which limits its usage to fixed-size patches that hamper its effectiveness. From \Cref{fig:ResViT}, it is evident that residual Transformer blocks stacked successively, known as aggregated residual Transformer (ART) blocks, in the bottleneck of the encoder-decoder design of the generator to extract the hidden contextual information of input features. The primary motivation of ART blocks is to learn an integrated representation that combines contextual, local, and hybrid local-contextual features underhood from the input flow. Channel Compression (CC) module recalibrates the concatenated features from the previous ART block and Transformer module to select the most discriminant representations. Due to the cascade of Transformers in design, to decrease the model complexity and computational burden, ResViT utilizes weight sharing strategy among projection tensors for Query, Key, value, and attention heads besides weight matrices for multi-layer perceptron operation. The superiority of this method has been proved over several MRI datasets in multi-contrast MRI synthesis and MRI to CT experiments with high PSNR and SSIM metrics over the conventional SOTA methods, e.g., pGAN \citep{dar2019image}, SAGAN \citep{zhang2019self}, pix2pix \citep{isola2017image} and PTNet \citep{zhang2021ptnet}.

Likewise, Liu et al. \citep{liu2022one} addressed the issue of missing contrasts in MRI imaging and proposed a multi-contrast multi-scale Transformer (\textbf{MMT}) framework to handle the unavailability of this information by synthesizing the existing contrasts as a means to substitute the missing data. To achieve efficient contrast synthetization, the task is considered as a seq-to-seq problem, in which the model learns to generate missing contrasts by leveraging the existing contrast in the following manner: A Swin multi-contrast Transformer encoder is implemented that creates hierarchical representation from the input MRI image. Then, a Swin Transformer-based architecture decodes the provided representation at multiple scales to perform medical image synthesis. Both the encoder and decoder are composed of two sequential swin blocks that capture contrast dependencies effectively. Conducted experiments on the IXI \citep{ixidataset} and BraTS \citep{baid2021brats} datasets demonstrated MMT's advantage compared to previous methods.

\begin{table}[!thb]
	\centering
	\caption{{Comparison results on BraTS \citep{baid2021brats} dataset for medical image synthesis, focusing on one-to-one and many-to-one tasks with different input to output combinations of T1, T2, and FLAIR modalities. The symbol ($\dagger$) indicates that the values are taken from \citep{li2022slmt}}. The GFLOPs of all
methods are calculated with an input image size of 256$\times$256.}
	\label{tab:synthesis_benchmarks}
	\resizebox{\columnwidth}{!}{
		\begin{tblr}{
				colspec={|l|c|lcc|cc|},
				row{1}={LightPink,c},
				hline{1,5} = {3pt},
				vline{1,8} = {3pt},
				hlines
			}
			\textbf{Methods} & \textbf{Design} & 
                \textbf{Task} &
                \textbf{SSIM}\;$\uparrow$ & \textbf{PSNR}\;$\uparrow$ & \textbf{Params (M)} & \textbf{FLOPs (G)}  \\
                {\textbf{PTNet $\dagger$} \citep{li2022slmt}} 
            & Hybrid & {T1, T2 $\rightarrow$ FLAIR \\ T1, FLAIR $\rightarrow$ T2 \\ T2, FLAIR $\rightarrow$ T1 \\ T2 $\rightarrow$ FLAIR \\ FLAIR $\rightarrow$ T2} & { 0.851 \\ 0.905 \\ 0.920 \\ 0.851 \\ 0.894} &  {23.78 \\ 25.09 \\ 22.19 \\ 23.01 \\ 24.78} & 27.69 & 233.1 \\
            
			{\textbf{MMT} \citep{liu2022one}} & Pure & {T1, T2 $\rightarrow$ FLAIR \\ T1, FLAIR $\rightarrow$ T2 \\ T2, FLAIR $\rightarrow$ T1 \\ T2 $\rightarrow$ FLAIR \\ FLAIR $\rightarrow$ T2} & {0.909 \\ 0.934 \\ 0.922 \\ 0.891 \\ 0.902} & {26.20 \\ 26.74 \\ 25.43 \\ 25.00 \\ 24.84} & 138.2 & 221.5 \\
			{\textbf{ResViT} \citep{dalmaz2022resvit}} &‌ Bottleneck & {T1, T2 $\rightarrow$ FLAIR \\ T1, FLAIR $\rightarrow$ T2 \\ T2, FLAIR $\rightarrow$ T1 \\ T2 $\rightarrow$ FLAIR \\ FLAIR $\rightarrow$ T2} & {0.886 \\ 0.938 \\ 0.924 \\ 0.870 \\ 0.908} & {25.84 \\ 26.90 \\ 26.20 \\ 24.97 \\ 25.78} & 123.4 & 486.1 \\
			
		\end{tblr}
	}
\end{table}

\subsection{Inter-Modality}
Unlike the intra-modality strategies, the inter-modality methods are designed to learn the mapping function between two different modalities. This approach allows the network to convert the samples from the base modality into a new modality and leverage the generated samples in the training process for the sake of performance gain. In this section, we will elaborate on two Transformer-based \citep{ristea2021cytran,kamran2021vtgan} strategies.

\begin{table*}[!th]
\centering
\caption{An overview of the reviewed Transformer-based medical image synthesizing approaches.}
\label{tab:synthesis}
\resizebox{\textwidth}{!}{
\begin{tabular}{lccccccc}
\toprule
\textbf{Method} & \textbf{Concept(s)} & \textbf{Modality} & \textbf{Type} & \textbf{Pre-trained Module: Type} & \textbf{Dataset(s)} & \textbf{Metrics} & \textbf{Year} \\ 
\rowcolor{LightPink}\multicolumn{8}{c}{\textbf{Pure}} \\
\makecell[l]{PTNet \citep{zhang2021ptnet}} & Intra-Modality & MRI &2D&\xmark &dHCP dataset \citep{makropoulos2018developing}&\makecell{SSIM \\ PSNR}&2021\\ \midrule
\makecell[l]{MMT \citep{liu2022one}} & \makecell{Intra-Modality \\ Inter-Modality} & MRI & 2D & \xmark & \makecell{$^1$ IXI \citep{ixidataset} \\$^2$ BraTS \citep{baid2021brats}} & \makecell{SSIM \\PSNR \\LPIPS} & 2022 \\  

\rowcolor{LightPink}\multicolumn{8}{c}{\textbf{Bottleneck}} \\
\makecell[l]{ResViT \citep{dalmaz2022resvit}} & \makecell{Intra-Modality \\ Inter-Modality} & \makecell{CT \\ MRI} & 2D & ViT: Supervised & \makecell{$^1$ IXI \citep{ixidataset} \\$^2$ BraTS \citep{menze2014multimodal,bakas2017advancing,bakas2018identifying} \\$^3$ Multi-modal pelvic MRI-CT \citep{nyholm2018mr}}& \makecell{PSNR \\ SSIM}& 2022\\ \midrule
\makecell[l]{CyTran \citep{ristea2021cytran}} & Inter-Modality & CT &‌ \makecell{2D \\ 3D} & \xmark &  Coltea-Lung-CT-100W \citep{ristea2021cytran} & \makecell{MAE \\ RMSE \\ SSIM} & 2022 \\

\rowcolor{LightPink}\multicolumn{8}{c}{\textbf{Encoder}} \\
\makecell[l]{{SLMT-Net} \citep{li2022slmt}} & {Inter-Modality} & {MRI} & {2D} & {ViT: Self-supervised} & \makecell{{BraTS} \citep{menze2014multimodal,bakas2017advancing,bakas2018identifying} \ $^2$ {ISLES} \citep{maier2017isles}} & \makecell{{SSIM} \\ {PSNR} \\ {NMSE}} & {2022}\\ 

\rowcolor{LightPink}\multicolumn{8}{c}{\textbf{Decoder}} \\
\makecell[l]{VTGAN \citep{kamran2021vtgan}} & Inter-Modality & Angiography & 2D & \xmark & Fundus Fluorescein Angiography \citep{hajeb2012diabetic} & \makecell{Fréchet inception distance \\ Kernel Inception distance}& 2021\\ 
\bottomrule
\end{tabular}
}
\end{table*}

\begin{table*}[!th]
	\centering
	\caption{A brief description of the reviewed Transformer-based medical image synthesizing models.}
	\label{tab:synthesis_highlight}
        \tiny
	\begin{tblr}{p{0.115\textwidth}p{0.405\textwidth}p{0.405 \textwidth}}
		\toprule
		\textbf{Method}  & \textbf{Contributions} & \textbf{Highlights} \\ 
		\SetRow{LightPink}\SetCell[c=3]{c} \textbf{Pure} & & \\
		{PTNet \citep{zhang2021ptnet}} & {$\bullet$ Introduced the pure Transformer-based network with linear computational complexity for image-synthesizing context} &{$\bullet$ Practical inference time around 30 image/s} \\ \midrule
		{MMT \citep{liu2022one}} & {$\bullet$ Proposed a pure Transformer-based architecture that incorporates Swin Transformer blocks to perform missing data imputation by leveraging the existing MRI contrasts. \\ $\bullet$ Conducted experiments on the IXI and BraTS datasets to perform qualitative and quantitative analysis and confirm their model's efficiency.}& {$\bullet$ Since the attention mechanism can be utilized to pinpoint influential features in the model's reasoning and decision-making, the attention scores of the Transformer decoder in MMT make it interpretable by capturing information in different contrasts that play an important role in generating the output sequence. \\ $\bullet$ The framework can be applied to a variety of medical analysis tasks, including image segmentation and cross-modality synthesis.} \\ 
		\SetRow{LightPink}\SetCell[c=3]{c} \textbf{Bottleneck} & &  \\
		{ResViT \citep{dalmaz2022resvit}} & {$\bullet$ First conditional adversarial model for medical image-to-image translation with hybrid CNN-Transformer generator \\ $\bullet$ Introduced a new module, ART block, for simultaneously capturing localization and contextual information}& {$\bullet$ Utilized weight sharing strategy among Transformers to hinder the computational overhead and lessen the model complexity \\ $\bullet$ An end-to-end design for the synthesized model that generalizes through multiple settings of source-target modalities, e.g., one-to-one and many-to-one tasks} \\ \midrule
		{CyTran \citep{ristea2021cytran}} & {$\bullet$ Proposing a generative adversarial convolutional Transformer for two tasks of image translation and image registration. \\ $\bullet$ Introducing a new dataset, named Coltea-Lung-CT-100W, comprised of 100 3D anonymized triphasic lung CT scans of female patients.} & {$\bullet$ The presented method can handle high-resolution images due to its hybrid structure \\ $\bullet$ Utilized style transfer techniques to improve alignment between contrast and non-contrast CT scans} \\ 
		\SetRow{LightPink}\SetCell[c=3]{c} \textbf{Encoder} & &  \\
		{{SLMT-Net} \citep{li2022slmt}} & {$\bullet$ {Proposed a self-supervised method for cross-modality MR image synthesis to generate missing modalities based on available modalities.} \\ $\bullet$ {Addresses the challenge of limited paired data by combining both paired and unpaired data.}} & { $\bullet$ {Despite using only 70\% paired data, MT-Net's performance remains on par with comparable models.} \\ $\bullet$ {Yields Enhanced PSNR scores when utilized for MR image synthesis.} \\ $\bullet$ {The combination of edge-aware pre-training and multi-scale fine-tuning stages contributes to the success of SLMT-Net.}} \\ 
		   
		\SetRow{LightPink}\SetCell[c=3]{c} \textbf{Decoder} & & \\
		{VTGAN \citep{kamran2021vtgan}} & {$\bullet$ Proposed a synthesis model for the task of fundus-to-angiogram that incorporates ViT architecture in the decoder section of the system to concurrently classify retinal abnormalities and synthesize FA images. \\ $\bullet$ Prepared experimental data based on quantitative and qualitative metrics regarding the model's generalization ability under the influence of spatial and radial transformations.} & {$\bullet$ Has the potential to be adopted as a tool for tracking disease progression. \\ $\bullet$ The system is designed to operate on non-invasive and low-cost fundus data to generate FA images.} \\ 

		\bottomrule
	\end{tblr}
\end{table*}

Several medical conditions may prevent patients from receiving intravenous contrast agents while getting CT screening. However, the contrast agent is crucial in assisting medical professionals in identifying some specific lesions. Therefore, \textbf{CyTran} \citep{ristea2021cytran} is proposed as an unsupervised generative adversarial convolutional Transformer for translating between contrast and non-contrast CT scans and image alignment of contrast-enhanced CT scans to non-enhanced. Its unsupervised part is also derived from its cyclic loss. CyTran is composed of three main modules: \textbf{\Romannum{1})} A downsample CNN-based module designed for handling high-resolution images, \textbf{\Romannum{2})} A convolutional Transformer module tailored for incorporating both local and global features, and \textbf{\Romannum{3})} An upsampling module developed to revert the transformation of the downsampling block through transpose-convolution. Additionally, the authors introduce a new dataset, Coltea-Lung-CT-100W, comprised of 100 3D anonymized triphasic lung CT scans of female patients.

Furthermore, Kamran et al. \citep{kamran2021vtgan} trained a ViT-based generative adversarial network (\textbf{VTGAN}) in a semi-supervised fashion on the Fundus Fluorescein Angiography (FFA) dataset provided by \citep{hajeb2012diabetic} via incorporating multiple modules, including residual blocks as generators for coarse and fine image generation, and two ViT architectures consisting of identical transformer encoder blocks for concurrent retinal abnormality classification and FA image synthesis.

{To embed self-supervised training in medical image synthesis, \textbf{SLMT-Net} \citep{li2022slmt} is presented for generating missing modalities in magnetic resonance (MR) images. It tackles the limited paired data challenge by combining paired and unpaired data. The method involves edge-aware pre-training using an Edge-preserving Masked Auto-Encoder (Edge-MAE) model that predicts missing information and estimates the edge map. A patch-wise loss enhances the performance. In the multi-scale fine-tuning stage, a DSF module is integrated to synthesize missing-modality images using multi-scale features. The pre-trained encoder extracts high-level features to ensure consistency between synthesized and ground-truth images during training.}

\begin{tcolorbox}[breakable ,colback={LightPink},title={\subsection{Discussion and Conclusion}},colbacktitle=LightPink,coltitle=black , left=2pt , right =2pt]

This section covers the adoption of ViT architectures in medical image synthesis applications. We explored the proposed methods based on two synthesis approaches: \textbf{(\Romannum{1})} inter-modality, in which the target modality is synthesized in a way that encapsulates crucial diagnostic features from different source images; and \textbf{(\Romannum{2})} intra-modality, with the objective of yielding target images with better quality by integrating information from lower resolution source images. To demonstrate their effectiveness, these approaches usually rely on SSIM, PSNR, and LPIPS as the evaluation metrics, since they are designed to measure the similarity between images. We also reviewed a ViT-based synthesis model \citep{kamran2021vtgan} that operates in a decoder fashion for the task of fundus-to-angiogram translation with different evaluation measurements, including Fréchet Inception Distance (FID) and Kernel Inception Distance (KID). {A combination of self-supervised learning with transformer-based medical synthesis is further reviewed to emphasize the cost-saving benefits of reduced labeling \citep{li2022slmt}.} We have additionally provided the architectural type, modality, input size, training setting, datasets, metrics, and year for every medical registration technique analyzed in \Cref{tab:synthesis}. Furthermore, \Cref{tab:synthesis_highlight} lists the contributions and highlights of the proposed works.
{To facilitate comparison, table \Cref{tab:synthesis_benchmarks} includes performance evaluation of multiple methods on the BraTS dataset \citep{baid2021brats}. In terms of SSIM and PSNR scores, MMT \citep{liu2022one} typically achieves superior performance, while PTNet \citep{li2022slmt} falls slightly behind when compared to the other two methods. Moreover, we have provided information about the model size and computational cost for three distinct approaches: PTNet, MMT, and ResViT. PTNet, with 27.69 million parameters and 233.1 GFLOPs, strikes a balance between model size and computational demand. Meanwhile, MMT offers a larger model with 138.2 million parameters but lower computational requirements at 221.5 GFLOPs, emphasizing efficiency. On the other hand, ResViT provides a mid-sized model with 123.4 million parameters but a notably higher computational load of 486.1 GFLOPs. The choice among these methods should be guided by specific task requirements, weighing the trade-off between model size and computational cost to achieve optimal performance. However, it is important to note that all these models require a significant computational cost.}

With the scarcity of works with ViT implementations and the recent advancement in the medical synthesis field with Transformer models, we believe that these systems require more research effort to be put into them. For example, Transformers have much room for improvement to generate more realism and high-quality synthesized medical images. One way to achieve this is by incorporating more detailed anatomy and physiology features using more efficient and effective attention mechanisms. Additionally, while much of the current research in this area has focused on 2D medical images and CT and MRI modalities, there is potential to apply these techniques to other types of medical images, including 3D and microscopy images.

\end{tcolorbox}

%%%% Beginning of Detection
\section{Medical Image Detection} \label{sec:detection}
\begin{figure*}[!t]
\centering
\includegraphics[width=\textwidth]{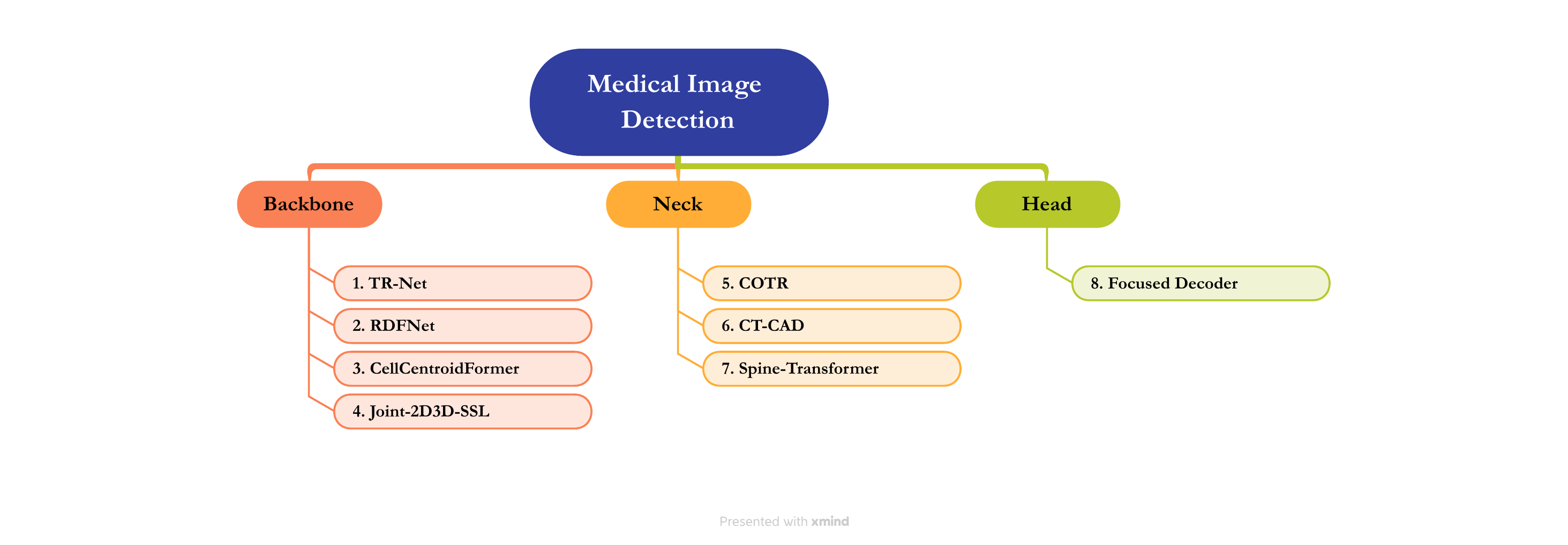}
\caption{An overview of Transformers in medical image detection. Methods are classified into the backbone, neck, and head according to the positions of the Transformers in their architecture. The prefix numbers in the paper’s name in ascending order denote the reference for each study as follows: 1. \citep{trnet}, 2. \citep{rdf}, 3. \citep{wagner123}, {4. \citep{nguyen2022joint},} 5. \citep{cotr2}, 6. \citep{cad1}, 7. \citep{Spine1}, 8. \citep{wittmann2023focused}.}
\label{fig:detectiontax}
\end{figure*}

Object detection remains one of the challenging problems in computer vision, especially detection in the medical image domain has its own challenges. Current SOTA architectures that work on 2D natural images use Vision Transformers. The Vision Transformers used in the detection task can be classified into two Transformer backbones and detection Transformers. In addition, the Transformer module can be used in a hybrid manner. Detection Transformers generally represent an end-to-end detection pipeline with an encoder-decoder structure, while the Transformer backbone solely utilizes the Transformer encoder for feature refinement. In order to increase detection performance, object detectors combine variants of vision Transformers with classical convolutional neural networks (CNNs).

Quite recently, Carion et al. \citep{carion2020end} introduced the concept of DETR, which forms a foundation for Detection Transformers. DETR uses a ResNet backbone to create a lower-resolution representation of the input images. Even though this approach achieves very good 2D detection results, comparable to the R-CNN backbone, high computational complexity is a downside of this method. The deformable DETR \citep{zhu2021deformable} approach has improved DETR's detection performance overcoming the problem of high computational complexity. Many recent approaches have tried to improve DETR's detection concept over time. Efficient DETR \citep{effDETR} eliminated DETR's requirement for iterative refinement. Conditional DETR \citep{conDETR} introduced the concept of a conditional cross-attention module. DN-DETR \citep{dndetr} introduced a denoising strategy, and DINO \citep{zhang2023dino} improved many aspects, such as denoising training, etc. Recently, some studies performed experiments on 2D medical data such as \citep{data1}, \citep{cotr2}, etc. However, only very few attempts tried to adapt it to 3D. Spine-Transformer was proposed by Tao et al. \citep{Spine1} for sphere-based vertebrae detection. Another approach in 3D detection was proposed by Ma et al. \citep{trnet}, which introduced a novel Transformer that combines convolutional layers and Transformer encoders for automatically detecting coronary artery stenosis in Coronary CT angiography (CCTA). An approach to better extract complex tooth decay features was proposed by \citep{rdf}.
For end-to-end polyp detection, Shen et al.~\citep{cotr2} proposed an approach that was based on the DETR model. Kong et al. have proposed the approach CT-CAD \citep{cad1}, context-aware Transformers for end-to-end chest abnormality detection. As illustrated in \Cref{fig:detectiontax}, we have broadly classified these methods based on the role ViT plays in the detection's commonly accepted pipeline. \Cref{tab:b} indicates details on modalities, organs, datasets, metrics, etc. The highlights of different approaches are summarized in \Cref{tab:a}. Some of the aforementioned detection papers in the medical image domain are reviewed in detail in this section.

\subsection{Backbone}  
This section explains Transformer networks using only the Transformer encoder layers for object detection.
The work proposed by Ma et al. \citep{trnet} uses a Transformer network \textbf{(TR-Net)} for identifying stenosis. A leading threat to the lives of cardiovascular disease patients globally is Coronary Artery Disease (CAD). Hence, the automatic detection of CAD is quite significant and is considered a challenging task in clinical medicine. The complexity of coronary artery plaques, which result in CAD, makes the detection of coronary artery stenosis in Coronary CT angiography (CCTA) challenging.

\begin{figure}[h]
	\centering
	\includegraphics[width=0.5\textwidth]{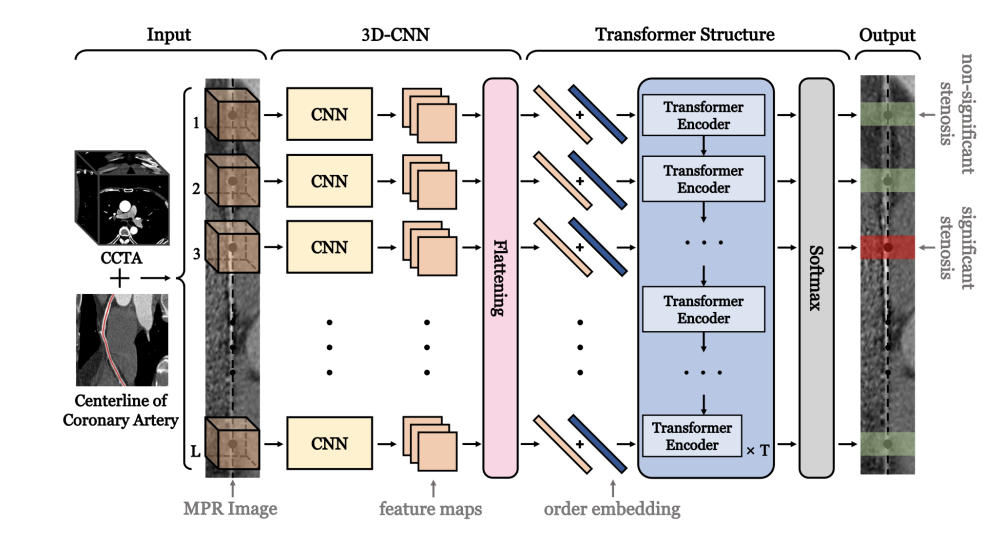}	\caption{Proposed architecture of TR-Net model \citep{trnet}.}
        \label{fig:trnet1}
\end{figure}

The architecture introduces a Transformer and combines the feature extraction capability of convolutional layers and Transformer encoders. TR-Net can easily analyze the semantic information of the sequences and can generate the relationship between image information in each position of a multiplayer reformatted (MPR) image. This model can effectively detect stenosis based on both local and global features. The CNN easily extracts the local semantic information from images, and the Transformer captures global semantic details more easily. A 3D-CNN is employed to capture the local semantic features from each position in an MPR image. After this step, the Transformer encoders are mainly used to analyze feature sequences. The main advantage here is that this helps in mining the dependency of local stenosis on each position of the coronary artery. The architecture of the TR-Net is given in \Cref{fig:trnet1}. One part of the figure indicates the 3D-CNN. This module extracts the local features. The other part indicates the Transformer encoder structure. This module associates the local feature maps of each position and helps in analyzing the dependency between different positions, which in turn is helpful for classifying the significant stenosis at each position. The CNN part mainly has two main advantages: it prevents the overfitting of semantic information and improves the model's efficiency. The input to the network architecture is the coronary artery MPR image.

The 3D-CNN module has four sequentially connected substructures, which consist of a convolutional kernel of size $3 \times 3 \times 3$, a non-linear ReLU layer and a $2 \times 2 \times 2$ max-pooling layer. The number of filters is $16$ in the first part, and in subsequent parts, the number of filters is double the number in the previous part. Since Transformers have 1D vector sequences as input, the feature maps are flattened. The Transformer in the proposed architecture consists of $12$ Transformer encoders. Each Transformer encoder mainly consists of two sub-blocks - multi-head self-attention (MSA) and the feed-forward network (FFN), which are connected sequentially. Layer normal (LN) and residual connections are employed before and after two sub-blocks. In order to ensure the consistency of the encoders, the size of the input is the same as the size of the output. The output of the previous encoder is given as input to the next encoder. In the final layer, the embeddings are fed into softmax classifiers to detect significant stenosis.

\textbf{RDFNet} approach proposed by Jiang et al. \citep{rdf} basically incorporates the Transformer mechanism in order to better extract the complex tooth decay features. The incorporation of the Transformer has improved the detection accuracy. The main three modules of the network are the backbone, neck, and prediction modules. The backbone module is mainly used to extract the features from caries images. In the backbone module, the focus operation is a slicing operation that could easily replace the convolution operation and reduce the loss of feature information. The C3Modified layer is a convolution module activated by the FReLU function, which extracts complex visual-spatial information of the caries images.
% \begin{figure}[h]
%  \centering
%  \includegraphics[width=0.9\columnwidth]{images/RDFNet}
%  \caption{An illustration of RDFNet architecture \citep{rdf}.}
%  \label{fig:rdfnet}
% \end{figure}
SPP \citep{rdf20} module has a spatial pyramid structure that could expand the perceptual field, fusing the local and global features and enhancing the feature maps. After the SPP structure, RDFNet appends an improved Transformer-encoder module to improve the feature extraction capability. The main functionality of the neck module is to fuse the feature maps of different sizes and extract high-level semantic structures. This module mainly uses the structure of the feature pyramid network (FPN) proposed in \citep{rdf21}, and path aggregation network (PAN) proposed in \citep{rdf22}. The FPN approach is employed in a top-down fashion, and  PAN is performed in a bottom-up fashion to generate the feature pyramids. In order to prevent information loss, feature fusion is performed using both bottom-up and top-down approaches. An improved C3Modified convolutional module is adopted into the neck module to better extract the semantic features of caries images. The high-level features generated by the neck module are used by the prediction module, which in turn is used to classify and regress the location and class of the objects. To overcome the problems of the single-stage detection method, which has quite a low detection accuracy, it mainly has three detection heads for detecting large, medium, and small objects. As Transformers have proved to have strong feature extraction capability, in order to extract complex features, they utilized the Transformer model.
% \begin{figure}[h]
%  \centering
%  \includegraphics[width=0.6\columnwidth]{images/Feature_extractor}
%  \caption{Feature extraction module integrated with Transformer \citep{rdf}.}
%  \label{fig:fe_edfnet}
% \end{figure}
To better extract the features, three Transformer encoders were stacked together. To simplify the model, the authors removed the original normalization layer from the Transformer encoder. In order to extract the deep features, the feature map was fed into this structure. For each head, the attention values were calculated independently and later concatenated.

\begin{figure}[h]
	\centering
	\includegraphics[width=0.5\textwidth]{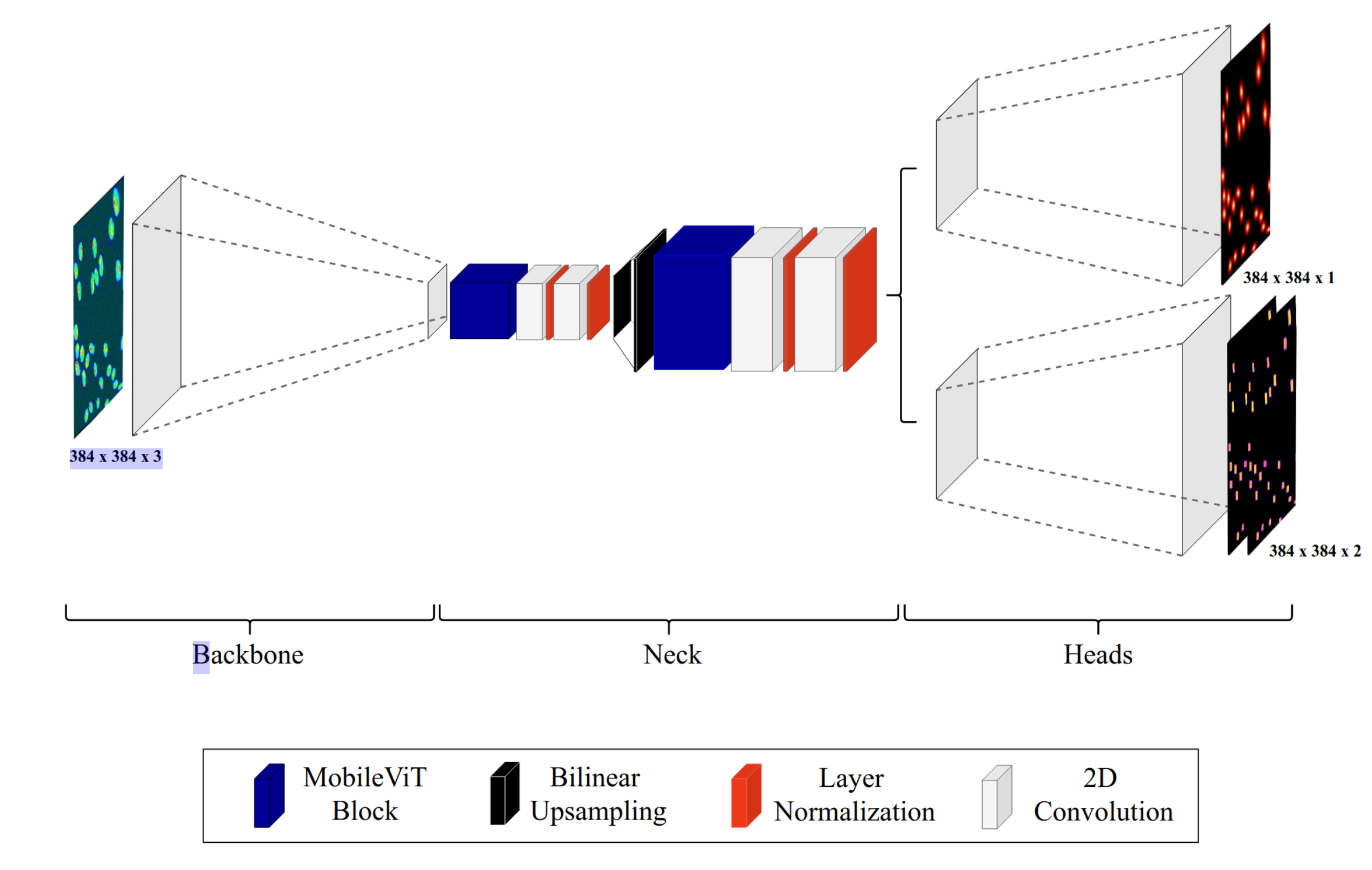}	
    \caption{Proposed architecture of CellCentroidFormer model \citep{wagner123}.}
\label{fig:cellcen}
\end{figure}

Wagner et al. \citep{wagner123} proposed a novel hybrid cell detection approach \textbf{(CellCentroidFormer)} in microscopic images that combines the advantages of vision Transformers (ViTs) and convolutional neural networks (CNNs). Authors show that the combined use of convolutional and Transformer layers is advantageous as the convolutional layers can focus on the local information (cell centroids), and the Transformer layers can focus on the global information ( overall shapes of a cell). The proposed centroid-based approach represents the cells as ellipses and is trainable in an end-to-end fashion. Four different 2D microscopic datasets were used for experimental evaluations, and the results outperformed the fully convolutional architecture-based methods.
\Cref{fig:cellcen} shows the architecture.
The encoder is then folded into a 3D tensor, which is afterward concatenated with the input tensor. The MobileViT block is a lightweight alternative to the actual encoder-decoder approach using a Transformer \citep{vaswani2017attention}. Due to the multi-head self-attention layers, the MobileViT block causes a much higher computational complexity than convolutional layers. To not increase the computational complexity excessively, the MobileViT blocks are combined in the neck part of the proposed model. Layer normalization is added for regularization and to allow higher learning rates. The backbone module of the proposed model is the EfficientNetV2S \citep{wag16} CNN model. This block mainly consists of six high-level blocks, out of which five blocks are used to extract image features. To use the advantage of transfer learning, the backbone module is initialized with weights learned from training on ImageNet. This, in turn, reduces the amount of required training data. The EfficientNetV2S  \citep{wag16} CNN models are generally optimized for a fixed input size. Therefore the input images need to be resized to this input size. The cells are represented mainly by the centroid, width, and height parameters. Mainly, two fully convolutional heads are used to predict these cell parameters in the paper. These heads contain 2D convolution, batch normalization, and bilinear upsampling layers. More MobileViT blocks are not used as it will increase the computational complexity. Later convolutional layers have a bigger receptive field which helps in capturing the global information \citep{wag14} effectively. The first convolutional head predicts a heatmap for detecting the cell centroids, and the second head is used for predicting the cell dimensions. The output dimensions of this model are $384 \times 384$. The authors use one decoder of the Dual U-Net to predict the centroid heatmap, and the second branch predicts the dimensions of the detected cells. The shapes of the cells are focused on by the Transformer layers in the network.

{Nguyen et al. \citep{nguyen2022joint} present a novel self-supervised learning (SSL) framework, \textbf{Joint-2D3D-SSL}, designed to address the scarcity of labeled training data for joint learning on 2D and 3D medical data modalities. The framework constructs an SSL task based on a 2D contrastive clustering problem for distinct classes using 2D images or 2D slices extracted from 3D volumes. In particular, the framework leverages 3D volumes by computing feature embeddings for each slice and then constructing a comprehensive feature representation through a deformable self-attention mechanism. This approach enables the capture of correlations among local slices, resulting in a holistic understanding of the data. The global embedded features derived from this transformer are subsequently utilized to define an agreement clustering for 3D volumes and a masked encoding feature prediction. By employing this methodology, the framework enables the learning of feature extractors at both the local and global levels, ensuring consistency and enhancing performance in downstream tasks. Experimental results across various downstream tasks, including abnormal chest X-ray detection, lung nodule classification, 3D brain segmentation, and 3D heart structures segmentation, illustrate the efficacy of this joint 2D and 3D SSL approach, surpassing plain 2D and 3D SSL approaches, as well as improving upon SOTA baselines. Additionally, this method overcomes limitations associated with fine-tuning pre-trained models with different dimensionality, providing versatile pre-trained weights suitable for both 2D and 3D applications.}

\subsection{Head}
Detection Transformers based on Transformer encoder-decoder architecture require a large amount of training data to deliver the highest performance. However, this is not feasible in the medical domain, where access to labeled data is limited. To address this problem, for the detection of 3D anatomical structures from the human body, Wittmann et al. \citep{wittmann2023focused} proposed a detection Transformer network with a \textbf{focused decoder}. This network considers the relative position of the anatomical structures and thus requires less training data. The focused decoder uses an anatomical region atlas to deploy query anchors to focus on the relevant anatomical structures. The proposed network omits the Transformer encoder network and consists of only Transformer decoder blocks. The authors show that in 3D datasets, avoiding the encoder can reduce the complexity of modeling relations with a self-attention module.

The model architecture contains a backbone network for feature extraction, a focus decoder network for providing well-defined detection results, a classification network to predict the classes, and a bounding box regression network to output the best possible bounding box. The feature extraction backbone network is a feature pyramid network (FPN) inspired by the RetinaNet \citep{schoppe2020deep}. Features from the second layer (P2) are flattened before being given as input to the focus decoder. A specific anatomical region atlas \citep{hohne1992volume}  containing regions of interest (RoI) is determined for each dataset. Then to each RoI, uniformly spaced query anchors are placed, and a dedicated object query is assigned to each. Such an object query will restrict the focus decoder network to predict solely within their respective RoI.

The focused decoder network contains a self-attention module, a focused cross-attention module, and a feedforward network (FFN). The self-attention module encodes strong positional inter-dependencies among object queries. The focused cross-attention module matches the input sequence to object queries to regulate the individual feature map for prediction via attention. The FFN network then enables richer feature representation. Also, residual skip connections and normalizations are used to increase gradient flow. The classification network consists of a single fully-connected layer, and the bounding box regression network consists of three layers. The bounding box predictions are combined with query anchors to get the bounding box together with class-specific confidence scores. The network is trained to predict 27 candidates predictions per class. Dynamic labeling with the help of generalized intersection over union (GIoU) is created during training to get 27 predictions. During inference, the prediction with the highest confidence score indicates the best candidate. The model is trained end-to-end with the above GIoU loss, binary cross-entropy loss for the classification network, and L1 loss for the bounding box predictions.

\subsection{Neck}
Detection methods using region-based approaches need to generate anchor boxes to encode their prior knowledge and use a non-maximum suppression to filter the resulting bounding boxes after prediction. These pre-and post-processing steps remarkably reduce the detection performance. To bypass these surrogate tasks, Carion et al. \citep{carion2020end} proposed Detection Transformer (DETR), which views the object detection task as a direct set prediction problem using an encoder-decoder architecture using Transformers. The self-attention mechanism of the Transformers, which explicitly models all pairwise interactions between elements in a sequence, helps to predict the set of detections with absolute prediction boxes directly from the image rather than using an anchor.
For the end-to-end detection of polyp lesions, Shen et al. \citep{cotr2} proposed a convolution in Transformer \textbf{(COTR)} network based on the DETR model. COTR consists of 4 main layers: 1) a CNN backbone network used for extracting features, 2) Transformer encoder layers embedded with convolutional layers used for feature encoding and reconstruction, 3) Transformer decoder layers used for object querying, and 4) a feed-forward network used for detecting prediction. Embedding convolutional layers into the Transformer encoder layer leads to convergence acceleration compared to the slow convergence of the DETR model.
\begin{table*}[!th]
    \centering
    \caption{An overview of the reviewed Transformer-based medical image detection approaches.}
    \label{tab:b}
    \resizebox{\textwidth}{!}{
    \begin{tabular}{lccccccc}  
    \toprule
    \textbf{Method} & \textbf{Modality} & \textbf{Organ} & \textbf{Type} & \textbf{Pre-trained Module: Type} & \textbf{Datasets} & \textbf{Metrics} & \textbf{Year}\\ 

    \rowcolor{Plum2}\multicolumn{8}{c}{\textbf{Backbone}} \\
      \makecell[l]{TR-Net \citep{trnet} } & \begin{tabular}[c]{@{}c@{}}  MPR \end{tabular} & Heart (Coronary Artery)  & 3D & Supervised &  \makecell{Private dataset}  &  \makecell{Accuracy,\\ Sensitivity,\\ Specificity,\\ PPV, NPV, \\F1-score} & 2021
     \\
     \midrule
    \makecell[l]{RDFNet \citep{rdf}}  & \begin{tabular}[c]{@{}c@{}} Dental Caries \end{tabular} & Teeth & 2D & Supervised & Private dataset  & \makecell{Precision,\\Recall,\\mAP@0:5}  &  2021 
     \\
    \midrule
    \makecell[l]{CellCentroidFormer \citep{wagner123} } & Microscopy & Cells  & 2D & Supervised &  \makecell{ $^1$ Fluo-N2DL-HeLa (HeLa) \citep{cell_challenge} ,\\ $^2$ Fluo-N2DH-SIM+ (SIM+)  \citep{cell_challenge}, \\
     $^3$ Fluo-N2DH-GOWT1 (GOWT1)  \citep{cell_challenge} , \\
     $^4$ PhC-C2DH-U373 (U373)  \citep{cell_challenge}.}  &  \makecell{Mean-IoU, \\ SSIM} & 2022   
     \\
    \midrule
    {\makecell[l]{Joint-2D3D-SSL \citep{nguyen2022joint}}} &  {X-ray} & {Chest}  & {\makecell{2D \\ 3D}} & {Self-supervised} & {VinDr-CXR} \citep{nguyen2022vindr}  &  {mAP@0.5} & {2022} 
     \\
     \rowcolor{Plum2}\multicolumn{8}{c}{\textbf{Head}} \\

         \makecell[l]{Focused decoder \citep{wittmann2023focused}} & \begin{tabular}[c]{@{}c@{}}    \end{tabular} CT  &  Multi-organ  & 3D & Semi-Supervised & \makecell{ $^1$ VISCERAL anatomy benchmark \citep{jimenez2016cloud}, \\ $^2$ AMOS22 challenge \citep{ji2022amos}.} & \makecell{mAP,\\AP50,\\AP75}  & 2023 

     \\
     \rowcolor{Plum2}\multicolumn{8}{c}{\textbf{Neck}} \\
     \makecell[l]{COTR \citep{carion2020end}}  & \begin{tabular}[c]{@{}c@{}} Colonoscopy \end{tabular} & Colon & 2D & Supervised &  \makecell{$^1$ CVC-ClinicDB \citep{cotr7},\\ $^2$ ETIS-LARIB \citep{cotr8},\\ $^3$ CVC-ColonDB \citep{cotr9}. }  &  \makecell{Precision,\\Sensitivity,\\F1-score}& 2021
     \\
    %\multicolumn{8}{c}{{\cellcolor[rgb]{1,0.753,0.478}}\textbf{Pure}} \\
    \midrule
    \makecell[l]{CT-CAD \citep{cad1}}  & \begin{tabular}[c]{@{}c@{}}  X-ray \end{tabular} & Chest & 2D & Supervised &  \makecell{$^1$ Vinbig Chest X-Ray dataset \citep{vind} \\ $^2$ ChestXDet-10 dataset \citep{chestxray}}  &  AP50 & 2021
     \\
    \midrule
     %\multicolumn{8}{c}{{\cellcolor[rgb]{1,0.753,0.478}}\textbf{Skip-connection}} \\
     \makecell[l]{Spine-Transformer \citep{Spine1}} & \begin{tabular}[c]{@{}c@{}} CT \end{tabular} & Vertebra & 3D & Supervised &  \makecell{$^1$ VerSe 2019 \citep{Spine5}, \\ $^2$ MICCAI-CSI 2014  \citep{Spine6},  \\ $^3$ Private dataset } & \makecell{Id-Rate,\\ L-Error}   &  2021     
     \\
     \bottomrule
    \end{tabular}
    }
\end{table*}

 \begin{table*}[!t]
    \centering
    \caption{A brief description of the reviewed Transformer-based medical image detection models. The unreported number of parameters indicates that the value was not mentioned in the paper.}
    \label{tab:a}
    \resizebox{\textwidth}{!}{
    \begin{tabular}{llp{11cm}p{10cm}} 
    \toprule
    % \rowcolor[rgb]{0.976,0.698,1}
    \textbf{Method} & \textbf{\# Params} & \textbf{Contributions} & \textbf{Highlights} \\ 
    \rowcolor{Plum2}\multicolumn{4}{c}{\textbf{Backbone}} \\
     \makecell[l]{TR-Net \citep{trnet}} 
     &
     -
     &
     $\bullet$ This work is the first attempt to detect coronary artery stenosis more accurately by employing Transformers. \newline
     $\bullet$ To detect significant stenosis, local and global features are effectively integrated into this approach, which has resulted in more accurate results.
     & 
     $\bullet$ While detecting significant stenosis, the TR-Net architecture is capable of combining the information of local areas near stenoses and the global information of coronary artery branches.\newline
     $\bullet$ Compared to state-of-the-art methods, the TR-Net model has better results on multiple indicators. \newline
     $\bullet$ The shallow CNN layer prevents the overfitting of semantic information and improves the overall efficiency. \newline
     $\bullet$ The gain in performance comes with a trade-off in the number of parameters, which affects the computational complexity.
    
     \\
     \midrule
     \makecell[l]{RDFNet \citep{rdf}} 
     &
     -
     &
     $\bullet$ An image dataset of caries is created, which is annotated by professional dentists. \newline
     $\bullet$ For better extraction of the complex features of dental caries, the Transformer mechanism is incorporated. \newline
     $\bullet$ In order to increase the inference speed significantly, the FReLU activation function is adopted.
     & 
     $\bullet$ Compared with existing approaches, the accuracy and speed of caries detection are better.\newline
     $\bullet$ Method is applicable to portable devices.\newline
     $\bullet$ The method does not work really well when the illumination of the oral image is insufficient. \newline
     $\bullet$ Even though detection accuracy and speed are improved compared to the original approach, the detection speed is not the fastest.
     
     \\
      \midrule
     \makecell[l]{CellCentroidFormer \citep{wagner123}} 
     &
     11.5M
     &
     $\bullet$ A novel deep learning approach that combines the self-attention of Transformers and the convolution operation of convolutional neural networks is proposed. \newline
     $\bullet$ A centroid-based cell detection method, denoting the cells as ellipses is proposed.
     &
     $\bullet$ Pseudocoloring in combination with pre-trained backbones shows improved cell detection performance. \newline
     $\bullet$ The model outperforms other state-of-the-art fully convolutional one-stage detectors on four microscopy datasets, despite having a lower number of parameters.\newline
     $\bullet$ Larger output strides worsen the performance.
     \\ 
           \midrule
     {\makecell[l]{Joint-2D3D-SSL \citep{nguyen2022joint}}} 
     &
     {31.16M}
     &
     {$\bullet$ Introduces a self-supervised learning framework capable of leveraging both 2D and 3D data for downstream applications. \newline
     $\bullet$ Proposes a deformable self-Attention mechanism that captures flexible correlations between 2D slices, leading to powerful global feature representations. \newline
     $\bullet$ Extends the SSL tasks by proposing a novel 3D agreement clustering and masking embedding prediction}
     &
     {$\bullet$ Overcomes limitations in fine-tuning pre-trained models with different dimensionality. \newline
     $\bullet$ Produces versatile pre-trained weights for both 2D and 3D applications.} \\
     
     \rowcolor{Plum2}\multicolumn{4}{c}{\textbf{Head}} \\
     \makecell[l]{Focused Decoder \citep{wittmann2023focused}} 
     &
     \makecell[l]{VISCERAL Dataset - 41.8M \\ AMOS22 Dataset - 42.6M}
     &
     $\bullet$ First detection Transformer model for 3D anatomical structure detection. \newline
     $\bullet$ Introduced a focused decoder to focus the predictions on RoI.
     & 
     $\bullet$ Better results compared to existing detection models using a Transformer network like DETR~\citep{carion2020end} and deformable DETR~\citep{zhu2021deformable}. \newline
     $\bullet$ Comparable results to the RetinaNet~\citep{schoppe2020deep}.\newline
     $\bullet$ Varying anatomical fields of view (FoVs) can affect the robustness of the model.
     
     \\ 
     \rowcolor{Plum2}\multicolumn{4}{c}{\textbf{Neck}} \\
     \makecell[l]{COTR \citep{xie2021cotr}} 
     &
     -
     &
     $\bullet$ Proposed a convolution layer embedded into the Transformer encoder for better feature reconstruction and faster convergence compared to DETR. 
     &
     $\bullet$ COTR has comparable results with state-of-the-art methods like Mask R-CNN \citep{cotr5} and MDeNet-plus \citep{cotr6}.\newline
     $\bullet$ This approach produces low confidence for a particular type of lesion. 
     
     \\
     \midrule
     \makecell[l]{CT-CAD \citep{cad1}} 
     &
     -
     &
     $\bullet$ Proposed a context-aware feature extractor, which enhances the receptive fields to encode multi-scale context-relevant information. \newline
     $\bullet$ Proposed a deformable Transformer detector that attends to a small set of key sampling locations and then the Transformers can focus to feature subspace and accelerate the convergence speed.
     &
     $\bullet$ CT-CAD outperforms the existing methods in Cascade R-CNN \citep{cad6}, YoLo \citep{redmon2016you}, and DETR \citep{zhu2021deformable}. \newline
     $\bullet$ CT-CAD is capable to detect hard cases, such as nodules that are ignored by Faster R-CNN.\newline
     $\bullet$ Compared to the ChestXDet-10 dataset, this model has a lower performance on the Vinbig Chest X-Ray dataset which has higher categories of abnormalities with more complex patterns.
     
     \\
    \midrule
     \makecell[l]{Spine-Transformers \citep{Spine1}} 
     & 
     - 
     & 
     $\bullet$ Proposed a 3D object detection model based on the Transformer's architecture. \newline
     $\bullet$ Proposed a one-to-one set global loss that enforces unique prediction and preserves the sequential order of vertebrae.  \newline
     $\bullet$ Proposed a Sphere-based bounding box to enforce rotational invariance.
     &
     $\bullet$ Obtained better results for all datasets compared to state-of-the-art methods.\newline
     $\bullet$ The model has a higher Id-Rate on both the datasets, but a higher L-Error compared to the benchmark by \citep{Spine5}. 
     
     \\
     
    \bottomrule
    \end{tabular}
    }
\end{table*}

The CNN backbone uses a pre-trained model with ResNet18 \citep{resnet} architecture for feature extraction. This layer converts input medical images to a high-level feature map. The authors then use a $1 \times 1$ convolution to reduce the channel dimensions. In the Transformer encoder layers, they used six convolution-in-Transformer encoders to collapse this spatial structure into a sequence. Then they use a convolution layer to reconstruct the sequential layer back to the spatial one. In the encoder layer, each Transformer has a standard architecture with a multi-head self-attention module and a feed-forward network. To the input of each attention layer, a positional embedding \citep{vaswani2017attention} is also introduced. In the Transformer decoder layers, they used six decoders which follow the standard architecture of the Transformer except that it also decodes object queries in parallel. Each object query will correspond to a particular object in the image. The decoders take these object queries with position embeddings as well as output embeddings from the encoder network and convert them into decoded embeddings. Then they used a feed-forward network with two fully connected layers for converting these decoded embeddings into object predictions. The first fully connected layer is a box regression layer to predict object location, and the second one is a box-classification layer to predict object scores. Therefore, the object queries are independently decoded into box coordinates and classes by the feed-forward network, which results in final predictions, including object and no object (background) predictions. This model transforms the object detection problem into a direct set prediction problem by training end-to-end by calculating bipartite matching loss (Hungarian algorithm) between predictions and ground truth for each query. If the number of queries exceeds the number of objects in the image, the remaining boxes are annotated as no object class. Thus, the model is trained to predict output for each query as an object or no object detection. For the class prediction, they used negative log-likelihood loss, and for the bounding box localization, they used an L1 loss with generalized intersection over union (GIOU) \citep{rezatofighi2019generalized} loss. The experiments demonstrated that the proposed model achieved comparable performance against state-of-the-art methods.

Many deep learning detection methods lack using context-relevant information for improved accuracy, and they also generally suffer from slower convergence issues and high computational costs. The proposed \textbf{CT-CAD} \citep{cad1}, context-aware Transformers for end-to-end chest abnormality detection, address these problems. The model consists of two main modules: 1) a context-aware feature extractor module for enhancing the features, and 2) a deformable Transformer detector module for detection prediction and to accelerate the convergence speed. The context-aware feature extractor network uses a ResNet50 backbone, dilated context encoding (DCE) blocks, and positional encoding structure. The deformable Transformer detector contains a Transformer encoder-decoder architecture and a feed-forward network. 
% The block diagram of CT-CAD is shown in \Cref{fig:cad}.
% \begin{figure}[h]
% 	\centering
% 	\includegraphics[width=0.5\textwidth]{images/cotr1.png}	\caption{Proposed architecture of CT-CAD model \citep{cad1}.}
%         \label{fig:cad}
% \end{figure}
The proposed design of the context-aware feature extractor is inspired by the feature fusion scheme from DetectoRS \citep{cad2} which is based on the Feature Pyramid Networks (FPN) \citep{cad3}. The feature fusion scheme iteratively enhances the features of the FPN to powerful feature representations. Likewise, the DCE blocks enhance the features extracted from the ResNet50 backbone by expanding the receptive fields to fuse multiscale context information using dilated convolution filters of different sizes. This powerful feature map benefits in detecting objects across various scales. Inspired by YOLOF \citep{cad4} the DCE block uses dilated convolution and skip connections to achieve a larger receptive field and acquire more local context information. Finally, all the features from different DCE blocks computed at different scales are summed up to get the feature map for the output.
% \par
% \begin{figure}[h]
% 	\centering
% 	\includegraphics[width=0.45\textwidth]{images/cotr2.png}
% 	\caption{Illustration of the proposed DCE block, where the receptive field is enlarged, which enables the acquisition of features with multi-scale context information \citep{cad1}.}
%         \label{dce}
% \end{figure}

The proposed design of the deformable Transformer detector contains single-scale and multi-head attention properties. The deformable attention block attends to a small set of key sampling points, thus allowing the Transformer to focus on the feature space and accelerate the convergence. The authors used six encoder and decoder layers with positional encoding to obtain the decoder outputs. The outputs from the decoder are the number of abnormalities detected and the dimension of the decoder layers. Finally, a feed-forward network is used to output the category classification and location regression results. The model is trained end-to-end with a combination of bounding box loss and classification (cross-entropy) loss. The authors adopted GIoU \citep{cad5} to balance the loss between large and small object bounding boxes.

The attention module in the detection Transformers computes similarity scores between elements of each input data to identify complex dependencies within these data. Calculating similarities of all possible positional pairs in the input data scales quadratically with the number of positions and thus becomes computationally very expensive. For this reason, the Transformer-based object detection model from 3D images has never been applied. Tao et al. \citep{Spine1} proposed a novel Transformer-based 3D object detection model as a one-to-one set prediction problem for the automatic detection of vertebrae in arbitrary Field-Of-View (FOV) scans, called the \textbf{Spine-Transformers}. Here the authors used a one-to-one set-based global loss that compels a unique prediction for preserving the sequential order of different levels of vertebrae and eliminated bipartite matching between ground truth and prediction. The main modules of the Spine-Transformer are (1) a backbone network to extract features, (2) a light-weighted Transformer encoder-decoder network using positional embeddings and a skip connection, and (3) two feed-forward networks for detection prediction. 
% The schematic illustration of the Spine-Transformer is shown in \Cref{fig:spine}.
% \begin{figure}[h]
%  \centering
%  \includegraphics[width=\columnwidth]{images/Spine1}
%  \caption{The vertebral semantic labeling network uses a combination of CNN and Transformer network to predict the location of each joint \citep{Spine1}.}
%  \label{fig:spine}
% \end{figure}
The authors used a ResNet50 \citep{resnet} architecture without the last SoftMax layer as the backbone network to extract high-level features. These features are passed through a $1 \times 1 \times 1$ convolutional layer to reduce the channel dimensions and then flattened to get a feature sequence to feed as the input for the Transformer network. The light-weighted Transformer encoder-decoder network contains only a two-layer encoder and two-layer decoder to balance between feature resolution and memory constraint. In both the encoder and decoder layers of the network, learnable positional embeddings are used. The authors found that using a skip connection across the Transformer encoder-decoder network will help in the propagation of context and gradient information during training and thus improves performance. The two feed-forward networks are then used to predict the existence of the objects and regress their coordinates. The authors also proposed a sphere-based bounding box detector to replace the rectangular-based bounding box to introduce rotational invariance called InSphere detector. The Spine-Transformer is trained end-to-end with fixed-size patch images to predict all the vertebrae objects in parallel by forcing one-to-one matching. Binary cross-entropy loss is used as classification loss, and to enforce the order of the predicted vertebrae objects, an edge loss is introduced, which is an L1 distance loss introduced between the centers of the top and bottom neighborhood vertebrae objects. For better localization accuracy of the bounding sphere detection, the authors used generalized inception-over-union (GIoU) \citep{rezatofighi2019generalized} loss. The results of this model showed superior results to all the state-of-the-art methods. The authors also claim that by using a 3D CNN-based landmark regression \citep{cciccek20163d}, the localization accuracy can be further improved.

\begin{tcolorbox}[breakable ,colback={Plum2},title={\subsection{Discussion and Conclusion}},colbacktitle=Plum2,coltitle=black , left=2pt , right =2pt]
	In this chapter, several well-known Transformer architectures are analyzed to address the automatic detection challenge. Based on the Transformer model contribution to the network structure, we grouped the set of literature work into the backbone, neck, or head strategies and for each category, we provided sample works. We also described the details regarding self-supervised learning in \citep{nguyen2022joint}. In this respect, the core idea behind each network design along with the pros and cons of the strategies are highlighted in the summary tables. Vision Transformers have been shown to make more accurate diagnoses compared to traditional methods of analyzing medical images. These deep learning models can be trained on large datasets, such as ImageNet, and fine-tuned on medical image datasets to improve their performance in detecting abnormalities in X-rays, CT scans, and MRIs. By incorporating information from multiple modalities, Transformers can further enhance their ability to identify and detect rare or subtle abnormalities in medical images. Many medical images are often taken over time, and incorporating temporal information into the model can improve its performance. For example, the model can be designed to take into account the temporal evolution of diseases or conditions. Overall, Transformers have demonstrated their capabilities to significantly improve the accuracy and efficiency of medical image analysis, leading to advances in healthcare. 
\end{tcolorbox}

%%%%%%%%% Medical Image Registration %%%%%%%%% 

\section{Medical Image Registration}
\label{sec:registration}

\begin{figure}[h]
	\centering
	\includegraphics[width=\columnwidth]{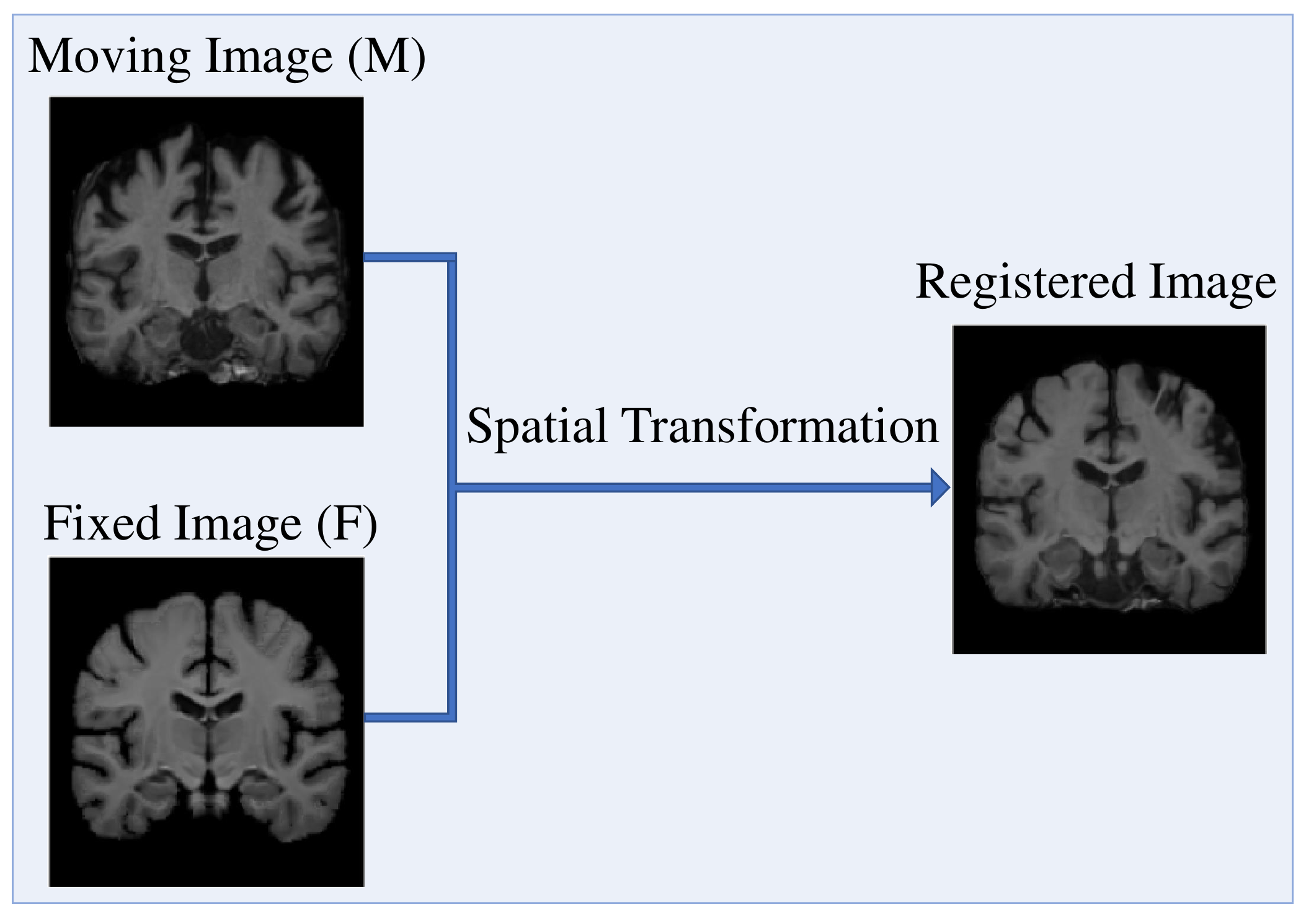}
	\caption{An example of pair-wise medical image registration. The goal of image registration is to geometrically align the moving image with the target or fixed image by performing the spatial transformation.}
	\label{fig:CTregistration}
\end{figure}

Medical image registration is the task of transforming a set of two or more images of an organ or a biological process taken with different poses, time stamps, or modalities (e.g., CT and MRI) into a geometrically aligned and spatially corresponding image that can be utilized for medical analysis. The transformation can be discovered by solving an optimization problem that maximizes the similarity between the images to be registered \citep{haskins2020deep}. A pair-wise registration of two MRI brain scans is shown in \Cref{fig:CTregistration} for illustration.

Despite remarkable advancements in the quality of medical imaging techniques that aid professionals in better visualization and analysis of image data, a prominent challenge prevails in developing a system capable of effective integration of visual data that captures useful information from original images with high precision. Most registration procedures take into account the whole image as input by utilizing global information for spatial transformation, which leads to inefficient and slow integration of data. Furthermore, the collection process of medical images for training is slow and toilsome, performance degrades due to the presence of outliers, and local maxima entail negative effects on performance during optimization \citep{alam2018medical,alam2019challenges}. 
The emergence of deep learning methods alleviated these problems by automatic extraction of features utilizing convolutional neural networks (CNN), optimizing a global function, and improving registration accuracy. For instance, Balakrishnan et al. \citep{balakrishnan2018unsupervised} utilized a CNN to achieve unsupervised deformable registration by treating it as a parametric function to be optimized during training. Furthermore, Chen et al. \citep{chen2020generatingreg} presented an unsupervised CNN-based registration algorithm to produce anthropomorphic phantoms. However, there are still limitations in capturing long-range spatial correspondence in CNN-based frameworks \citep{chen2102Transformers,matsoukas2021time}.

Fueled by the strong ability of Transformers to model long-range dependencies and detect global information \citep{lin2021survey,khan2022Transformers,xu2022svort}, they have gained the attention of researchers in the medical image registration domain in recent years. In this section, we review Transformer-based methods in medical image registration that ameliorate the aforementioned shortcomings of previous systems by utilizing the self-attention mechanism. We have organized the relevant approaches based on their type of registration:
\begin{enumerate}[(a)]
\item Deformable registration, which employs an optimization algorithm to tune the transformation model, is a way that maximizes the similarity measure function for the images of interest \citep{rong2021rigid};
\item Rigid registration, which achieves correspondence by maintaining the relative distance between each pair of points between the patient's anatomy images \citep{rong2021rigid}.
\item Affine registration, which contains the same operations as rigid registration plus non-isometric scaling.
\end{enumerate}

\begin{figure}[t]
 \centering
 \includegraphics[width=\columnwidth]{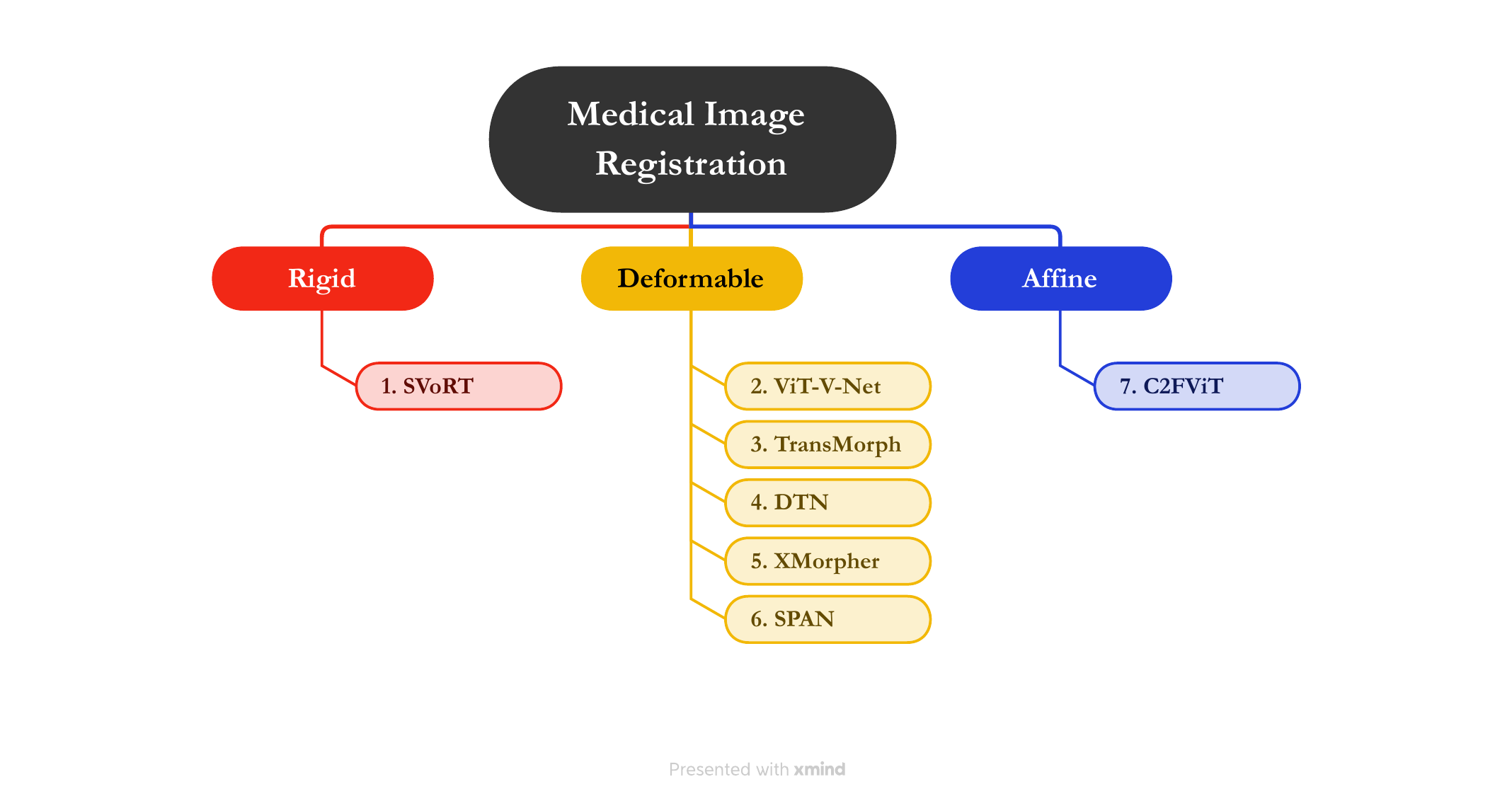}
 \caption{Taxonomy of Transformer-based image registration based on their transformation type. We use the prefix numbers in the figure in ascending order and reference the corresponding paper as follows: 
 1. \citep{xu2022svort}, 
 2. \citep{chen2021vit}, 
 3. \citep{chen2022transmorph}, 
 4. \citep{zhang2021learning}, 
 5. \citep{shi2022xmorpher}, 
 {6. \citep{miao2022prior}},
 7. \citep{mok2022affine}.}
 \label{fig:Registration Taxonomy}
\end{figure}

\subsection{Deformable Registration}
\begin{figure}[h]
	\centering
	\includegraphics[width=0.48\textwidth]{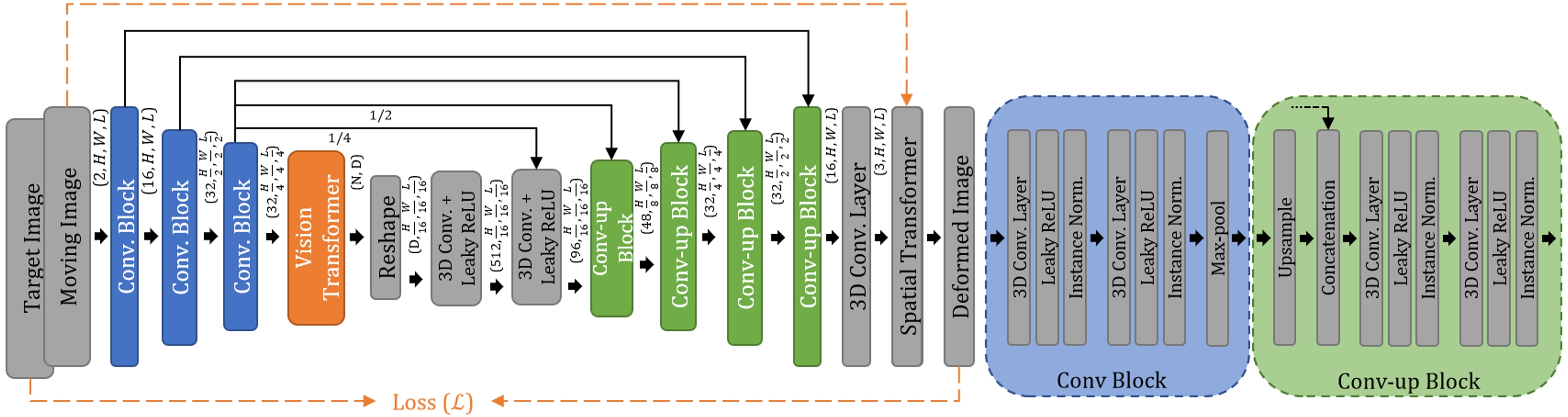}
	\caption{Overview of ViT-V-Net. Multiple convolution blocks encode images into high-level features, which the Vit block splits into patches. These patches are then mapped to D-dimensional patch embeddings that get integrated with learnable positional encodings to retain positional information. Next, these patches are passed into the Transformer encoder block, followed by multiple skip connections to retain localization information, and decoded using a V-Net style decoder. Using the network's final output, a spatial Transformer warps the moving image. Figure taken from \citep{chen2021vit}.}
	\label{fig:ViT-V}
\end{figure}

Most existing Transformer-based algorithms focus on deformable transformation to perform medical image registration.
\textbf{Vit-V-Net} \citep{chen2021vit} is the earliest work that incorporates Transformers to perform medical image registration in a self-supervised fashion. It is inspired by the integration of vision Transformer-based segmentation methods with convolutional neural networks to enhance the localization information recovered from the images. Unlike previous research that employed 2D images for spatial correspondence, Vit-V-net stepped towards utilizing ViT \citep{dosovitskiy2020image} as the first study for volumetric medical image registration (i.e., 3D image registration). As illustrated in \Cref{fig:ViT-V}, the images are first encoded into high-level feature representations by implementing multiple convolution blocks; then, these features get split into \emph{P} patches in the ViT block. Next, the patches are mapped to a D-dimensional embedding space to provide patch embeddings, which are then integrated with learnable positional encodings to retain positional information. Next, these patches are passed into the encoder block of the Transformer, followed by multiple skip connections to retain localization information, and then decoded employing a V-Net style decoder \citep{milletari2016v}. Finally, a spatial Transformer \citep{jaderberg2015spatial} warps the moving image by utilizing the final output of the network.
TransMorph \citep{chen2022transmorph} extended ViT-V-Net and proposed a hybrid Transformer ConvNet framework that utilizes the Swin Transformer \citep{liu2021swin} as the encoder and a ConvNet as the decoder to provide a dense displacement field. Like ViT-V-Net, it employed long skip connections to retain the flow of localization information that may enhance registration accuracy. The output of the network, which is a nonlinear warping function, gets applied to the moving image with the deformation field utilizing the spatial transformation function proposed in \citep{jaderberg2015spatial}. An affine transformation Transformer network is incorporated to align the moving image with the fixed image before feeding it to the deformable registration network. This work also proposed two variants of TransMorph: diffeomorphic TransMorph (TransMorph-diff) to facilitate topology-preserving deformations and Bayesian TransMorph (TransMorph-Bayes) to promote a well-calibrated registration uncertainty estimate. 

Likewise, Zhang et al. \citep{zhang2021learning} introduced the dual Transformer network (\textbf{DTN}) framework to perform diffeomorphic registration. It is composed of a CNN-based 3D U-Net encoder \citep{cciccek20163d} for the embedding of separate and concatenated volumetric images and a dual Transformer to capture the cross-volume dependencies. One of the Transformers is responsible for modeling the inter- and intra-image dependencies, and the other one handles the modeling of the global dependencies by employing the self-attention mechanism. The concatenation of the generated features from these Transformers results in enhanced feature embeddings, which are utilized by the CNN-based decoder to provide a diffeomorphic deformation field. The evaluation of the framework was conducted on the brain MRI scans of the OASIS dataset \citep{marcus2007open}, which substantiates their improvements in diffeomorphic registration compared to the existing deep-learning-based approaches.

Furthermore, \textbf{XMorpher} \citep{shi2022xmorpher} put emphasis on the significance of backbone architectures in feature extraction and match of pair-wise images, and proposed a novel full Transformer network as the backbone, which consists of two parallel U-Net structures \citep{ronneberger2015unet} as the sub-networks with their convolutions replaced by the introduced Cross Attention Transformer for feature extraction of moving and fixed images, and cross-attention-based fusion modules that utilize these features for generating the feature representation of moving-fixed correspondence and fine-grained multi-level semantic information that contributes to a fine registration.

{The latest developments in self-supervised representation learning have primarily aimed at eliminating inductive biases in training processes. Nonetheless, these biases can still serve a purpose in scenarios with scarce data or when they offer further understanding of the underlying data distribution. To mitigate this problem, \citep{miao2022prior} introduces spatial prior attention (\textbf{SPAN}), a framework that leverages the consistent spatial and semantic structure found in unlabeled image datasets to guide the attention mechanism of vision Transformers. \textbf{SPAN} integrates domain expertise to enhance the effectiveness and interpretability of self-supervised pretraining specifically for medical images using image registration and alignment methodologies. Specifically, they exploit image registration to align the attention maps with an inductive bias corresponding to a salient region so only a single representative sample is required. When utilizing deformable image registration templates, \textbf{SPAN} achieves the highest mAUC score, followed by the predicted global templates and the triangular spatial heuristic resulting in improved performance in downstream tasks.}

\definecolor{emeraldbetter}{rgb}{0.588, 0.871, 0.659 }

\begin{table*}[ht]
    \centering
    \caption{An overview of the reviewed Transformer-based medical image registration approaches.}
    \label{tab:Registration}
    \resizebox{\textwidth}{!}{
	\begin{tabular}{lcccccc}  
	\toprule
	\textbf{Method} & \textbf{Modality} & \textbf{Organ} & \textbf{Type} & \textbf{Datasets} & \textbf{Metrics} & \textbf{Year}\\ 
    
    \rowcolor{emeraldbetter}\multicolumn{7}{c}{\textbf{Deformable}} \\

    \makecell[l]{ViT-V-Net \citep{chen2021vit}} & MRI & Brain & 3D &  Private Dataset & Dice &  2021
    \\
    \midrule
    
    \makecell[l]{TransMorph \citep{chen2022transmorph}}  & \begin{tabular}[c]{@{}c@{}} MRI \\ CT \end{tabular} & \begin{tabular}[c]{@{}c@{}} Brain \\ Chest-Abdomen-Pelvis region \end{tabular} & 3D & \makecell{ $^1$ IXI \citep{ixidataset} \\ $^2$ T1-weighted brain MRI scans from Johns Hopkins University \\ $^3$ Chest-Abdomen-Pelvis CT \citep{segars2013population} }  & \makecell{ Dice \\ $\%$ of $|J_\varPhi|\leq$ 0 \\ SSIM } &  2022
    \\
    \midrule

    \makecell[l]{DTN \citep{zhang2021learning}}  & MRI & Brain & 3D &  OASIS \citep{marcus2007open} & \makecell{Dice \\ $|J_\varPhi|\leq$ 0 }&  2021
    \\
    \midrule
    
    \makecell[l]{XMorpher \citep{shi2022xmorpher}}  & \makecell{CT \\ MRI} & Heart & 3D &  \makecell{$^1$ MM-WHS 2017 \citep{zhuang2016multi} \\ $^2$ ASOCA \citep{ramtin_gharleghi_2020_3819799} }& \makecell{ Dice \\ $\%$ of $|J_\varPhi|\leq$ 0 }&  2022
    \\
    \midrule

    {\makecell[l]{SPAN \citep{miao2022prior}}}  & {\makecell{CT \\ MRI}} & {Lung} & {3D} &  {\makecell{CheXpert \citep{irvin2019chexpert} \\  JSRT \citep{shiraishi2000development}}} 
        & {mAUC} &  {2022}
    \\
    \midrule
    
    \rowcolor{emeraldbetter}\multicolumn{7}{c}{\textbf{Affine}} \\
    
    \makecell[l]{C2FViT \citep{mok2022affine}}  & MRI & \makecell{Brain} & 3D &  \begin{tabular}[c]{@{}c@{}} $^1$ OASIS \citep{marcus2007open} \\ $^2$ LPBA \citep{shattuck2008construction} \end{tabular} & \begin{tabular}[c]{@{}c@{}} Dice \\Hausdorff distance\end{tabular} &  2022
    \\
    \midrule

    \rowcolor{emeraldbetter}\multicolumn{7}{c}{\textbf{Rigid}} \\
    
    \makecell[l]{SVoRT \citep{xu2022svort}} &  MRI & Brain & 3D &  FeTA \citep{payette2021automatic}  & \makecell{PSNR \\ SSIM  } &  2022
    \\
    \bottomrule
    \end{tabular}
}
\end{table*}

\begin{table*}[!t]
    \centering
    \caption{A brief description of the reviewed Transformer-based medical image registration techniques.}
    \label{tab:Registration highlight}
    \resizebox{\textwidth}{!}{
    \begin{tabular}{lp{15cm}p{20cm}}
    \toprule
    % \rowcolor[rgb]{0.976,0.698,1}
    \textbf{Method} & \textbf{Contributions} & \textbf{Highlights} \\ 
    
    \rowcolor{emeraldbetter}\multicolumn{3}{c}{\textbf{Deformable}} \\

     \makecell[l]{ViT-V-Net \citep{chen2021vit}} 
     & 
     $\bullet$ Contributed to the medical image registration domain as the first work to exploit ViTs to develop a volumetric (3D) registration. \newline
     $\bullet$ Integrated Transformers with CNNs to build a hybrid architecture for self-supervised brain MRI registration
     &  
     $\bullet$ Employed a hybrid architecture to incorporate long-range and local information in the registration process. \newline
     $\bullet$ Attempted to preserve the localization data with the help of long skip connections between the encoder and decoder stages.
     \\
     \midrule
     
     \makecell[l]{TransMorph \citep{chen2022transmorph}} 
     &
     $\bullet$ Proposed a Transformer-based unsupervised registration approach for affine and deformable objectives. \newline
     $\bullet$ Conducted experiments on two brain MRI datasets and in a phantom-to-CT registration task to demonstrate their superior performance compared to traditional approaches.
     &
     $\bullet$ They additionally proposed two distinguishable versions of their model: a diffeomorphic variant to facilitate the topology-preserving deformations and a Bayesian variant to promote a well-calibrated registration uncertainty estimate. \newline 
     $\bullet$ Studied the effect of receptive fields by comparing TransMorph with CNNs and addressed that while the receptive field of ConvNets only increases with the layer depth, their presented model takes into account the whole image at each due to the self-attention mechanism.
     \\
     \midrule
     
     \makecell[l]{DTN \citep{zhang2021learning}} 
     & 
     $\bullet$ Proposed a dual Transformer architecture to capture semantic correspondence of anatomical structures. \newline
     $\bullet$ The suggested DTN demonstrated remarkable results in diffeomorphic registration and atlas-based segmentation of multi-class anatomical structures.
     &  
     $\bullet$ The Dual Transformer is capable of reducing the negative Jacobian determinant while preserving the atlas-based registration quality. \newline
     $\bullet$ The qualitative and quantitative analysis of their method on the OASIS dataset indicates that diffeomorphic registration fields are effective.
     \\
     \midrule
     
     \makecell[l]{XMorpher \citep{shi2022xmorpher}} 
     &
     $\bullet$ Devised a deformable registration system consisting of dual parallel feature extraction networks which facilitate the association of representative features between moving and fixed images. \newline
     $\bullet$ Proposed cross-attention Transformer that establishes spatial correspondences through computation of bilateral information in the attention mechanism.
     &
     $\bullet$ Promotes visual superiority by presenting fine-grained visual results in terms of boundary smoothness, adjacent regions' resolution quality, and deformation grid polishness. \newline 
     $\bullet$ Demonstrated the model's great diagnostic potential by conducting experiments with different training regimes.
     \\
     \midrule

     \makecell[l]{{SPAN} \citep{miao2022prior}}
     &
     $\bullet$ {Developed a self-supervised transformer-based registration method that leverages consistent spatial and semantic structure in unlabeled image datasets to guide the attention mechanism of vision Transformers.}
     &
     $\bullet$ {Integrates domain expertise and uses image registration and alignment methodologies to enhance the effectiveness and interpretability of self-supervised pretraining, particularly for medical images.}
     \\
     \midrule
     
     \rowcolor{emeraldbetter}\multicolumn{3}{c}{\textbf{Affine}} \\
     
     \makecell[l]{C2FViT \citep{mok2022affine}} 
     &
     $\bullet$ Presented a method in order to learn the global affine registration by taking advantage of the strong long-range dependency recognition and locality of the hybrid Transformer and the multi-resolution strategy.
     &
     $\bullet$ The suggested training framework can be extended to a number of parametric-based registration approaches by removing or scaling the geometrical transformation matrices.
     \\
     
     \rowcolor{emeraldbetter}\multicolumn{3}{c}{\textbf{Rigid}} \\
    
    \makecell[l]{SVoRT \citep{xu2022svort}} 
     &
     $\bullet$ Devised an approach for the task of Volumetric reconstruction of fetal brains based on Transformer architectures. \newline
     $\bullet$ Employed a Transformer network trained on artificially sampled 2D MR slices that estimate the underlying 3D volume from the input slices to more accurately predict transformation.
     &
     $\bullet$ Experimental procedures on the FeTA dataset \citep{payette2021automatic} represented the model's ability in high-quality volumetric reconstruction. \newline 
     $\bullet$ The volumetric reconstruction associated with the transformations of the proposed method displays higher visual quality.
     \\
     
    \bottomrule
    \end{tabular}
    }
\end{table*}

\subsection{Affine Registration}

\begin{figure}[!th]
	\centering
	\includegraphics[width=0.48\textwidth]{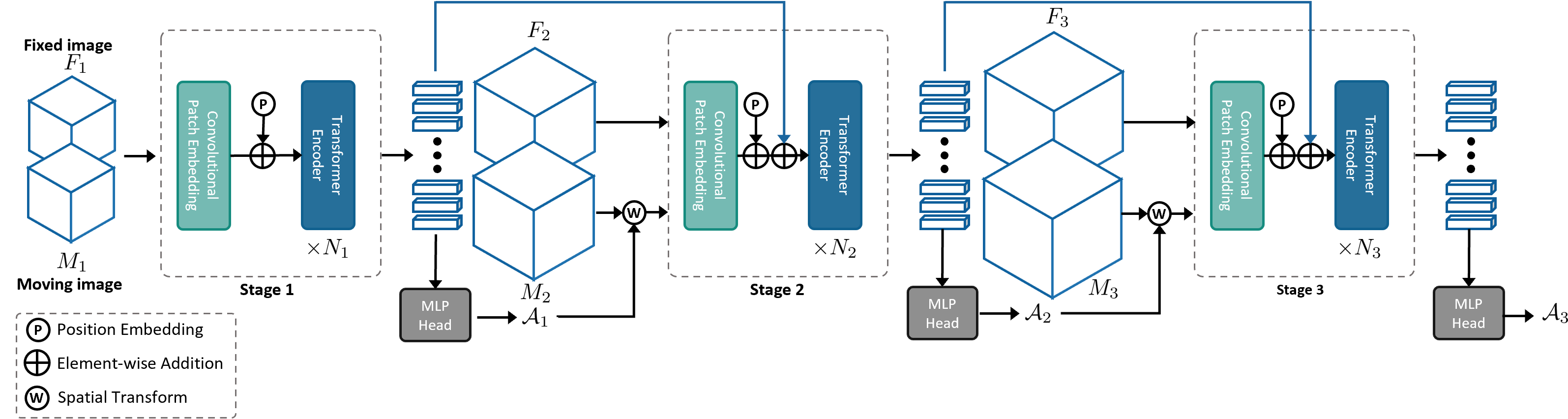}
	\caption{The model has \emph{L} stages with convolutional patch embedding layers and $N$ Transformer encoder blocks to learn the optimal affine registration matrix. In each stage, fixed and moving images are downsampled and concatenated, then passed to the convolutional patch embedding layer to produce image patch embeddings. The Transformer then produces the input feature embedding from the embeddings \citep{mok2022affine}.}
	\label{fig:C2FViT}
\end{figure}

To perform affine medical image registration with Transformers, Mok et al. \citep{mok2022affine} proposed \textbf{C2FViT}, a coarse-to-fine vision Transformer that performs affine registration, a geometric transformation that preserves points, straight lines, and planes while registering 3D medical images. Former studies have relied on CNN-based affine registration that focuses on local misalignment or global orientation \citep{de2019deep,zhao2019unsupervised}, which limits the modeling of long-range dependencies and hinders high generalizability. C2FVit, as the first work that takes into account the non-local dependencies between medical images, leverages vision Transformers instead of CNNs for 3D registration. As depicted in \Cref{fig:C2FViT}, the model is split into \emph{L} stages, each containing a convolutional patch embedding layer and $N_i$ Transformer encoder blocks (i indicates the stage number), intending to learn the optimal affine registration matrix. In each stage, the fixed and moving images are downsampled and concatenated with each other, then the new representation gets passed to the convolutional patch embedding layer to produce image patch embeddings. Next, the Transformer receives the embeddings and produces the feature embedding of the input. Conducted experiments on OASIS \citep{marcus2007open} and LPBA \citep{shattuck2008construction} demonstrated their superior performance compared to existing CNN-based affine registration techniques in terms of registration accuracy, robustness, and generalization ability.

\subsection{Rigid Registration}

\textbf{SVoRT} \citep{xu2022svort} addressed the necessity of slice-to-volume registration before volumetric reconstruction for the task of volumetric fetal brains reconstruction, and employed a Transformer network trained on artificially sampled 2D MR slices that learn to predict slice transformation based on the information gained from other slices. The model also estimates the underlying 3D volume from the input slices to promote higher accuracy in transformation prediction. The superiority of their proposed method in terms of registration accuracy and reconstruction based on the evaluation of synthetic data and their experiments on real-world MRI scans demonstrated the ability of the model in high-quality volumetric reconstruction.

\begin{tcolorbox}[breakable ,colback={emeraldbetter},title={\subsection{Discussion and Conclusion}},colbacktitle=emeraldbetter,coltitle=black , left=2pt , right =2pt]    

According to the research discussed in this section, vision Transformers are prominent tools in image registration tasks due to their training capability on large-scale data, which is made feasible by parallel computing and self-attention mechanisms. Leveraging Transformers to encourage better global dependency identification improves registration in terms of dice scores and Jacobian matrix determinants compared to CNNs.

To mitigate the burden of quadratic complexity when processing images at high resolution and modeling local relationships, reviewed studies usually employ CNNs to provide feature maps or dense displacement fields \citep{chen2021vit,chen2022transmorph,zhang2021learning}. C2FViT \citep{mok2022affine} disregarded convolutional networks and implemented convolutional patch embeddings to promote locality. However, in deformably registering medical content, XMorpher recently demonstrated the power of cross-attention in better capturing spatial relevancy without a CNN implementation \citep{shi2022xmorpher}, and SVoRT purely utilized Transformers to perform rigid registration \citep{xu2022svort}. {Regarding the challenges caused by the data-hungry nature of deep neural networks and the difficulties in gathering annotated training data, studies in medical image registration have also focused on integrating self-supervised learning with transformer models \citep{miao2022prior}.}

The notable experimental attempts on brain MRI scan data, such as OASIS \citep{marcus2007open} and FeTA \citep{payette2021automatic}, show the importance of accurate automatic registration for neuroimaging data. One particular work \citep{shi2022xmorpher} proposed to evaluate their registration on images of cardiac region datasets including MM-WHO-2017 \citep{zhuang2016multi} and ASOCA \citep{ramtin_gharleghi_2020_3819799}. To further clarify the modality type used in the aforementioned proposed methods, all works conducted their evaluations on 3D or volumetric imaging modalities.

Based on the brief review of Transformer-based medical image registration research, we believe that other regions of interest (ROI) such as neurons, retina, and neck area are worth exploring to facilitate diagnostic operations in different domains with more precise registration models.

We have also specified the architectural type, modality, organ, data size, training paradigm, datasets, metrics, and year for each medical registration technique reviewed in \Cref{tab:Registration}. Furthermore, \Cref{tab:Registration highlight} provides a list of the contributions and highlights of the proposed works. We also elaborate on the use cases of self-supervision in medical image registration.
\end{tcolorbox}

\definecolor{aliceblue}{rgb}{0.643, 0.831, 1.0}

%%%%%%%%% Medical Report Generation %%%%%%%%% 
\section{Medical Report Generation} \label{sec:report}
\begin{figure*}[t]
 \centering
 \includegraphics[width=0.99\textwidth]{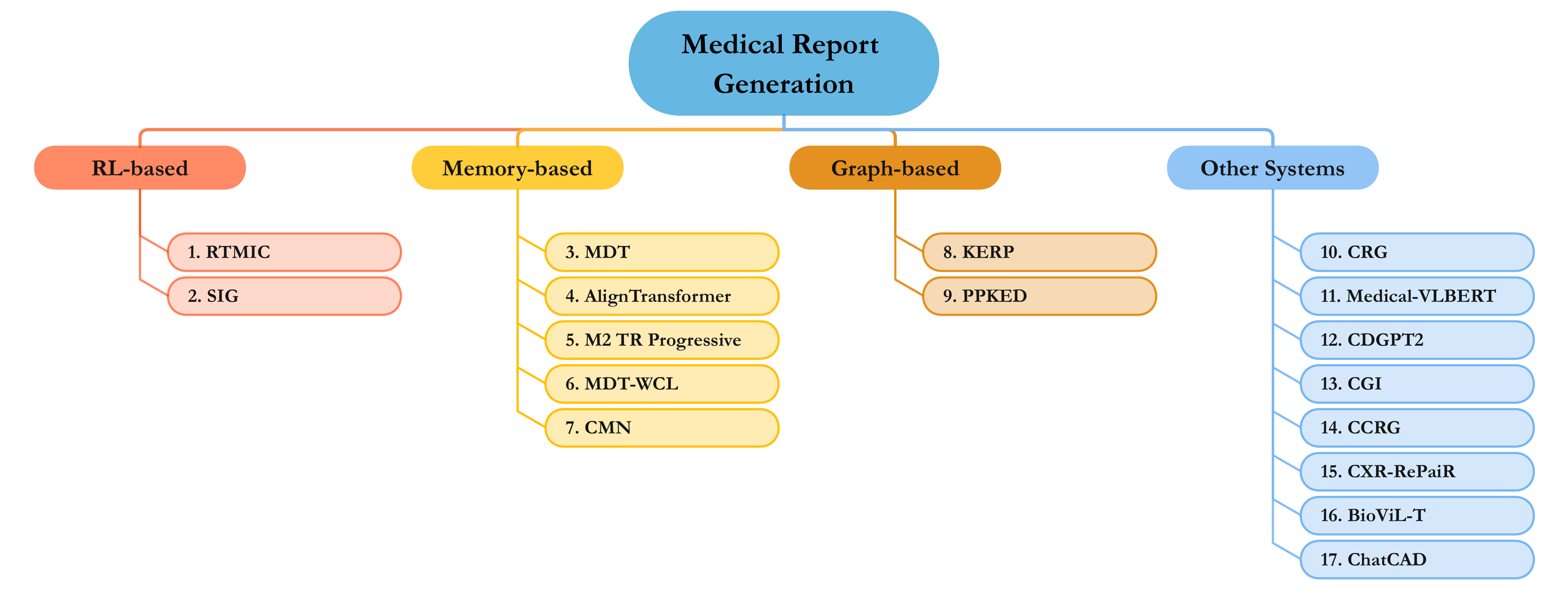}
 \caption{Taxonomy of Transformer-based medical report generation approaches based on the mechanism by which they generate clinical reports. We reference the papers in ascending order corresponding to their prefix number:
 1. \citep{xiong2019reinforced}, 
 2. \citep{zhang2021surgical}, 
 3. \citep{chen2020generating}, 
 4. \citep{you2021alignTransformer}, 
 5. \citep{nooralahzadeh2021progressive}, 
 6. \citep{yan2021weakly}.
 7. \citep{chen2022cross}, 
 8. \citep{li2019knowledge}, 
 9. \citep{liu2021exploring}, 
 10. \citep{lovelace2020learning}, 
 11. \citep{liu2021medical}, 
 12. \citep{alfarghaly2021automated},
 13. \citep{nguyen2021automated}, 
 14. \citep{wang2021confidence},
 15. \citep{endo2021retrieval},
 16. \citep{bannur2023learning},
 17. \citep{wang2023chatcad}.}
 \label{fig:Generation Taxonomy}
\end{figure*}
Medical report generation focuses on producing comprehensive captions and descriptions pivoting on medical images for diagnostic purposes. Designing automatic methods capable of performing this task can alleviate tedious and time-consuming work in producing medical reports and promote medical automation \citep{jing2017automatic}. Recently, advancements in deep learning have brought the attention of researchers to employing an intelligent system capable of understanding the visual content of an image and describing its comprehension in natural language format \citep{stefanini2022show}. Research efforts in improving this area can be employed in medical imaging by implementing systems capable of providing descriptions and captions (i.e., generating medical reports) concerning medical images. 
These captioning systems usually utilize encoder-decoder models that encode medical images and decode their understandings to provide diagnostic information in a natural language format.

Despite the success of deep learning, limitations including reliability on an immense amount of data, unbalanced data in radiology datasets (e.g., IU X-ray chest X-Ray \citep{demner2016preparing}), and the black box nature of DL models entail challenges in medical report generation \citep{monshi2020deep}. 
The success of Transformer models in many vision-and-language tasks has drawn the attention of researchers in the medical report generation domain to the employment of this architecture. In this section, we discuss approaches that utilize Transformers to promote effective capture of long-range context dependencies and better report generation. As illustrated in \Cref{fig:Generation Taxonomy}, the following is our taxonomy of these systems according to the mechanism by which they produce accurate and reliable clinical reports:

\begin{enumerate}[(I)]
\item \emph{Reinforcement Learning-based.} 
 The ultimate goal of a medical report generation system is to provide clinically accurate and reliable reports. In reinforcement learning, the MRG system is considered an agent with the objective of maximizing clinical accuracy based on the feedback given by the reward signal, which is directly calculated by the evaluation metric score (e.g., CIDEr \citep{vedantam2015cider}).

\item \emph{Graph-based.} Radiology reports are typically composed of a long finding section with multiple sentences that make report generation a challenging task. Therefore, the inclusion of prior information is beneficial for facilitating the generation of long narratives from visual data. Knowledge graphs, which are powerful models that can capture domain-specific information in a structured manner, can be used to exploit prior information for medical report generation \citep{li2018hybrid,li2019knowledge,zhang2020radiology}.

\item \emph{Memory-based.}
Memory is a resource through which important information is recorded. In designing a proper MRG system, it is crucial to store vital and diagnostic information that can benefit the generation process by incorporating prior knowledge and experience. Hence, configuring a memory mechanism with Transformers as a report generation framework facilitates longer and more coherent text generation by sharing information gained through the process \citep{chen2020generating,chen2022cross}.

\item \emph{Other Systems.}
Systems that introduce different ideas from previous categories to improve clinical accuracy, such as curriculum learning, contrastive learning, and alternate learning, belong to this group.
\end{enumerate}

\begin{table*}[!th]
    \centering
    \caption{An overview of the reviewed Transformer-based Medical Report Generation approaches.}
    \label{tab:Generation}
    \resizebox{\textwidth}{!}{
    \begin{tabular}{lccccccc}  
    \toprule
    \textbf{Method} & \textbf{Modality} & \textbf{Organ} & \textbf{Type} &
    \textbf{Visual Backbone} &
    \textbf{Datasets} & \textbf{Metrics} & \textbf{Year}\\

    \rowcolor{aliceblue}\multicolumn{8}{c}{\textbf{Reinforcement Learning}} \\
    
     \makecell[l]{RTMIC \citep{xiong2019reinforced}} & X-ray & Lung & 2D  & DenseNet-121 \citep{huang2017densely} &  \begin{tabular}[c]{@{}c@{}} IU Chest X-ray \citep{demner2016preparing} \end{tabular} & 
     \begin{tabular}[c]{@{}c@{}} BLEU \citep{papineni2002bleu} \\CIDEr \citep{vedantam2015cider} \end{tabular}   
     &  2019
     \\
     \midrule
    
    \makecell[l]{SIG \citep{zhang2021surgical}}  & \begin{tabular}[c]{@{}c@{}} Ultrasound \\ Colonoscopy \end{tabular}  & Multi-organ   & 3D   & ResNet-101 \citep{resnet} & DAISI \citep{rojas2020daisi} & \begin{tabular}[c]{@{}c@{}} BLUE \citep{papineni2002bleu} \\ Meteor \citep{banerjee2005meteor} \\ CIDEr \citep{vedantam2015cider} \\ ROUGE \citep{lin2004rouge} \\ SPICE \citep{anderson2016spice} \end{tabular} & 2021
    \\
    \midrule

    \rowcolor{aliceblue}\multicolumn{8}{c}{\textbf{Graph}} \\
    
    \makecell[l]{KERP \citep{li2019knowledge}}  & X-ray & Lung & 2D & DenseNet-121 \citep{huang2017densely} & \begin{tabular}[c]{@{}c@{}} $^1$ IU Chest X-ray \citep{demner2016preparing} \\$^2$ CX-CHR (private dataset) \end{tabular} & \begin{tabular}[c]{@{}c@{}} BLEU \citep{papineni2002bleu}\\ CIDEr \citep{vedantam2015cider} \\ ROUGE \citep{lin2004rouge} \end{tabular} &  2019
    \\
    \midrule
    
    \makecell[l]{PPKED \citep{liu2021exploring}}  & X-ray  & Lung   & 2D  & ResNet-152 \citep{resnet} & \begin{tabular}[c]{@{}c@{}} $^1$ IU Chest X-ray \citep{demner2016preparing} \\$^2$ MIMIC-CXR \citep{johnson2019mimic}   \end{tabular}  & \begin{tabular}[c]{@{}c@{}} BLUE \citep{papineni2002bleu} \\ Meteor \citep{banerjee2005meteor} \\ CIDEr \citep{vedantam2015cider} \\ ROUGE \citep{lin2004rouge} \end{tabular} & 2021 
    \\ 
    \midrule 
    
    \rowcolor{aliceblue}\multicolumn{8}{c}{\textbf{Memory}} \\

    \makecell[l]{MDT \citep{chen2020generating}}  & X-ray & Lung & 2D & ResNet-121 \citep{resnet} & \begin{tabular}[c]{@{}c@{}} $^1$ IU Chest X-ray \citep{demner2016preparing} \\$^2$ MIMIC-CXR \citep{johnson2019mimic} \end{tabular} & \begin{tabular}[c]{@{}c@{}} BLEU \citep{papineni2002bleu}\\ Meteor \citep{banerjee2005meteor} \\ ROUGE \citep{lin2004rouge} \end{tabular} &  2020
    \\
    \midrule
    
    \makecell[l]{AlignTransformer \citep{you2021alignTransformer}}  & X-ray & Lung & 2D & ResNet-50 \citep{resnet} & \begin{tabular}[c]{@{}c@{}} $^1$ IU Chest X-ray \citep{demner2016preparing} \\$^2$ MIMIC-CXR \citep{johnson2019mimic} \end{tabular} & \begin{tabular}[c]{@{}c@{}} BLEU \citep{papineni2002bleu}\\ Meteor \citep{banerjee2005meteor} \\ ROUGE \citep{lin2004rouge} \end{tabular} &  2021
    \\
    \midrule
    
    \makecell[l]{M$^2$ TR. progressive \citep{nooralahzadeh2021progressive}} & X-ray & Lung & 2D & DenseNet-121 \citep{huang2017densely} &  \begin{tabular}[c]{@{}c@{}} $^1$ IU Chest X-ray \citep{demner2016preparing} \\$^2$ MIMIC-CXR \citep{johnson2019mimic} \end{tabular} & \begin{tabular}[c]{@{}c@{}} BLEU \citep{papineni2002bleu}\\ Meteor \citep{banerjee2005meteor} \\ ROUGE \citep{lin2004rouge} \end{tabular} &  2021
    \\
    \midrule
    
    \makecell[l]{MDT-WCL \citep{yan2021weakly}}   & X-ray  & Lung   & 2D & ResNet \citep{resnet} & \begin{tabular}[c]{@{}c@{}} $^1$ MIMIC-ABN \citep{ni2020learning} \\$^2$ MIMIC-CXR \citep{johnson2019mimic} \end{tabular}  & \begin{tabular}[c]{@{}c@{}} BLUE \citep{papineni2002bleu}\\ Meteor \citep{banerjee2005meteor} \\ ROUGE \citep{lin2004rouge} \end{tabular} & 2021 
    \\ 
    \midrule 
    
    \makecell[l]{CMN \citep{chen2022cross}}  & X-ray & Lung & 2D & ResNet-101 \citep{resnet} & \begin{tabular}[c]{@{}c@{}} $^1$ IU Chest X-ray \citep{demner2016preparing} \\$^2$ MIMIC-CXR \citep{johnson2019mimic} \end{tabular} & \begin{tabular}[c]{@{}c@{}} BLEU \citep{papineni2002bleu}\\ Meteor \citep{banerjee2005meteor} \\ ROUGE \citep{lin2004rouge} \end{tabular} &  2022
    \\
    \midrule
    
    \rowcolor{aliceblue}\multicolumn{8}{c}{\textbf{Other}} \\

    \makecell[l]{CRG \citep{lovelace2020learning}}  & X-ray & Lung & 2D & DenseNet-121 \citep{huang2017densely} & MIMIC-CXR \citep{johnson2019mimic} & \begin{tabular}[c]{@{}c@{}} BLUE \citep{papineni2002bleu}\\ Meteor \citep{banerjee2005meteor} \\ CIDEr \citep{vedantam2015cider} \\ ROUGE \citep{lin2004rouge} \end{tabular} &  2020
    \\
    \midrule
    
    \makecell[l]{Medical-VLBERT \citep{liu2021medical}}   & \begin{tabular}[c]{@{}c@{}} CT\\ X-ray \end{tabular}    & Lung   & 2D & DenseNet-121 \citep{huang2017densely} & \begin{tabular}[c]{@{}c@{}} $^1$ Chinese Covid-19 CT \citep{Covid-19CT}\\$^2$ CX-CHR (private dataset)\end{tabular}    & \begin{tabular}[c]{@{}c@{}} BLUE \citep{papineni2002bleu}\\ CIDEr \citep{vedantam2015cider} \\ ROUGE \citep{lin2004rouge} \end{tabular} & 2021 
    \\
    \midrule

    \makecell[l]{CDGPT2 \citep{alfarghaly2021automated}}   & X-ray  & Lung   & 2D & DenseNet-121 \citep{huang2017densely} & IU chest X-ray \citep{demner2016preparing} & \begin{tabular}[c]{@{}c@{}} BLUE \citep{papineni2002bleu}\\ Meteor \citep{banerjee2005meteor} \\ CIDEr \citep{vedantam2015cider} \\ ROUGE \citep{lin2004rouge} \end{tabular} & 2021 
    \\ 
    \midrule 

    \makecell[l]{CGI \citep{nguyen2021automated}}   & X-ray  & Lung   & 2D & DenseNet-121 \citep{huang2017densely} &  \begin{tabular}[c]{@{}c@{}} $^1$ MIMIC-CXR  \citep{johnson2019mimic} \\$^2$ IU chest X-ray \citep{demner2016preparing} \end{tabular}  & \begin{tabular}[c]{@{}c@{}} BLUE \citep{papineni2002bleu}\\ Meteor \citep{banerjee2005meteor} \\  ROUGE \citep{lin2004rouge} \end{tabular} & 2021 
    \\ 
    \midrule

    \makecell[l]{CGRG \citep{wang2021confidence}}  & X-ray  & Lung   & 2D & ResNet-101 \citep{resnet} & \begin{tabular}[c]{@{}c@{}} $^1$ IU Chest X-ray  \citep{demner2016preparing} \\$^2$ COV-CTR \citep{li2022auxiliary} \end{tabular}  & \begin{tabular}[c]{@{}c@{}} BLUE \citep{papineni2002bleu}\\ Meteor \citep{banerjee2005meteor} \\ ROUGE \citep{lin2004rouge} \end{tabular} & 2021 
    \\ 

    \midrule 

    \makecell[l]{{CXR-RePaiR} \citep{endo2021retrieval}}   & {X-ray}  & {Lung}   & {2D} & \xmark &  \begin{tabular}[c]{@{}c@{}} {MIMIC-CXR}  \citep{johnson2019mimic} \end{tabular}  & \begin{tabular}[c]{@{}c@{}} {BLUE} \citep{papineni2002bleu} \\ {F1} \end{tabular} & {2021} 
    \\ 

    \midrule 

    \makecell[l]{{BioViL-T} \citep{bannur2023learning} }   & {X-ray}  & {Lung}   & {2D} & {ResNet-50} \citep{resnet} &  \begin{tabular}[c]{@{}c@{}} $^1$ {MIMIC-CXR}  \citep{johnson2019mimic} \\$^2$ {MS-CXR-T} \citep{bannur2023learning} \end{tabular}  & \begin{tabular}[c]{@{}c@{}} {BLUE} \citep{papineni2002bleu} \\  {ROUGE} \citep{lin2004rouge} \\ {CHEXBERT} \citep{smit2020chexbert} \end{tabular} & {2023} 
    \\ 

    \midrule
    \makecell[l]{{ChatCAD} \citep{wang2023chatcad}}   & {X-ray}  & {Lung}   & {2D} & \xmark &  \begin{tabular}[c]{@{}c@{}} {MIMIC-CXR}  \citep{johnson2019mimic} \end{tabular}  & \begin{tabular}[c]{@{}c@{}} {Precision}, {Recall}, {F1}\end{tabular} & {2023}
    \\ 
    \bottomrule
    \end{tabular}
}
\end{table*} 

\begin{table*}[!t]
    \centering
    \caption{A brief summary of the reviewed Transformer-based medical report generation methods.}
    \label{tab:Generation highlight}
    \resizebox{\textwidth}{!}{
    \begin{tabular}{lp{15cm}p{20cm}}
    \toprule
    % \rowcolor[rgb]{0.976,0.698,1}
    \textbf{Method} & \textbf{Contributions} & \textbf{Highlights} \\ 

    \rowcolor{aliceblue}\multicolumn{3}{c}{\textbf{Reinforcement Learning}} 
    \\
     \makecell[l]{RTMIC \citep{xiong2019reinforced}} 
     & 
     $\bullet$ Presented a novel Hierarchical Reinforced Transformer for producing comprehensible, informative medical reports by training through reinforcement learning-based training. \newline
     $\bullet$ The initial attempt at incorporating Transformers to develop a medical report generation system.
     &  
     $\bullet$ Utilized reinforcement learning to ameliorate the exposure bias problem. \newline
     $\bullet$ Enhanced clinical report coherence by employing Transformers to capture long-range dependencies. \newline
     $\bullet$ The selected metric (CIDEr) as a reward signal is not designed for the medical domain \citep{messina2022survey}.\\
     \midrule

     \makecell[l]{SIG \citep{zhang2021surgical}} 
     &
     $\bullet$ Generated surgical instructions from multiple clinical domains by utilizing a Transformer-based Encoder-decoder architecture.
     &
     $\bullet$ The proposed method is able to produce multimodal dependencies, form pixel-wise patterns, and develop textual associations for the masked self-attention decoder. \newline
     $\bullet$ Utilizing self-critical reinforcement learning to perform optimization increased the performance of surgical instruction generation.  \newline
     $\bullet$ The selected metric (CIDEr) as a reward signal is not designed for the medical domain \citep{messina2022survey}. \\
     \midrule
     
     \rowcolor{aliceblue}\multicolumn{3}{c}{\textbf{Graph}} 
    \\
    
    \makecell[l]{KERP \citep{li2019knowledge}} 
     &
     $\bullet$ Developed an MRG system using a hybrid retrieval-generation technique that unifies standard retrieval-based and recent visual text generation methods. \newline
     $\bullet$ Introduced Graph Transformer (GTR) as the first research to employ an attention mechanism to convert different data types formulated as a graph.
     &
     $\bullet$ Aligning the generated reports with abnormality attention maps by providing location reference facilitates medical diagnosis. \newline
     $\bullet$ Since KERP is designed based on abnormality detection, it may disregard other valuable information \citep{nguyen2021automated}.
     \\
     \midrule
     
     \makecell[l]{PPKED \citep{liu2021exploring}} 
     &
     $\bullet$ Proposed a three-module system that mimics the working habits of radiologists by extracting abnormal regions, encoding prior information, and distilling the useful knowledge to generate accurate reports.
     &
     $\bullet$ Provides abnormal descriptions and locations to facilitate medical diagnosis. \newline    
     $\bullet$ Capable of extracting relevant information from the explored posterior and prior multi-domain knowledge. \newline    
     $\bullet$ Some mistakes, such as duplicate reports and inaccurate descriptions, are present in the generated reports. \citep{liu2020layer}
     \\
     \midrule
     
    \rowcolor{aliceblue}\multicolumn{3}{c}{\textbf{Memory}} 
    \\ 
     
    \makecell[l]{MDT \citep{chen2020generating}} 
     &
     $\bullet$ Introduced Memory-Driven Transformer for radiology report generation \newline
     $\bullet$ Developed a relational memory to retain essential knowledge gathered through the previous generations.
     &
     $\bullet$ To facilitate medical diagnosis, visual-textual attention mappings were incorporated to capture correspondence with essential medical terms. \newline
     $\bullet$ Dataset imbalance with dominating normal findings hinders the model's generalizability.
     \\
     \midrule
    
    \makecell[l]{AlignTransformer \citep{you2021alignTransformer}} 
     &
     $\bullet$ Introduced an MRG framework that mitigates the problem of data bias by hierarchically aligning visual abnormality regions and illness tags in an iterative fashion.
     &
     $\bullet$ Conducted experiments on MIMIC-CXR and IU-Xray datasets and demonstrated the capability of the model in ameliorating the data bias problem  
     \\
     
     \midrule
    
     \makecell[l]{M$^2$ TR. progressive \citep{nooralahzadeh2021progressive}} 
     &
     $\bullet$ Developed a progressive text generation model for medical report generation by incorporating high-level concepts into the process of generation.
     &
     $\bullet$ The division of report generation into two steps enhanced the performance in terms of language generation and clinical efficacy metrics. \newline
     $\bullet$ The progressive generation process increases the false positive rate by including abnormality mentions in negation mode. \citep{messina2022survey}. 
     \\
     \midrule
     
     \makecell[l]{MDT-WCL \citep{yan2021weakly}} 
     &
     $\bullet$ Introduced the contrastive learning technique into chest X-ray report generation by proposing a weakly supervised approach that contrasts report samples against each other to better identify abnormal findings.
     &
     $\bullet$ Optimization with contrastive loss facilitates generalizability in comparison to contrastive retrieval-based methods.
     \\
     
     \midrule
     
     \makecell[l]{CMN \citep{chen2022cross}} 
     &
     $\bullet$ Cross-modal memory networks were introduced to improve report generation based on encoder-decoder architectures by incorporating a shared memory to capture multi-modal alignment. 
     &
     $\bullet$ Capable of properly aligning data from radiological images and texts to aid in the preparation of more precise reports in terms of clinical accuracy.\\
    \midrule
    
    \rowcolor{aliceblue}\multicolumn{3}{c}{\textbf{Other}} 
    \\ 

    \makecell[l]{CRG \citep{lovelace2020learning}} 
     &
     $\bullet$ Formulated the problem in two steps: (1) a report generation phase incorporating a standard language generation objective to train a Transformer model, and (2) a sampling phase that includes sampling a report from the model and extracting clinical observations from it. 
     &
     $\bullet$ Transformers' ability to provide more coherent and fluent reports was demonstrated. \newline
     $\bullet$ Due to the biased nature of the dataset caused by dominant normal findings, the algorithm tends to generate reports that lack essential descriptions of abnormal sections \citep{liu2022competence}.\\
    \midrule
    
    \makecell[l]{Medical-VLBERT \citep{liu2021medical}} 
     &
     $\bullet$ Proposed a framework as the first work that generates medical reports for the COVID-19 CT scans
     \newline
     $\bullet$ Devised an alternate learning strategy to minimize the inconsistencies between the visual and textual data. 
     &
     $\bullet$ Alleviated the shortage of COVID-19 data by employing the transfer learning strategy. \newline
     $\bullet$ Capable of effective terminology prediction \newline
     $\bullet$ Overreliance on predetermined terminologies undermines robustness and generalizability.\\
     \midrule

     \makecell[l]{CDGPT2 \citep{alfarghaly2021automated}} 
     &
     $\bullet$ Presented a conditioning mechanism to improve radiology report generation in terms of word-overlap metrics and time complexity.
     \newline
     $\bullet$ Utilized a pre-trained GPT2 conditioned on visual and weighted semantic features to promote faster training, eliminate vocabulary selection, and handle punctuation. \newline
     $\bullet$ The first study to employ semantic similarity metrics to quantitatively analyze medical report generation results.
     &
     $\bullet$ Conditioning mechanism tackled punctuations, vocabulary collection, and reduced training duration. \newline
     $\bullet$ The architecture does not require modification to be trained on distinct data sets. \newline
     $\bullet$ Incorporating semantic similarity in addition to word overlap metrics improved medical report evaluation. \newline
     $^4$ The model's generalization ability and robustness against over-fitting are both hindered when the size of the dataset is small.\\
     \midrule
     
     \makecell[l]{CGI \citep{nguyen2021automated}} 
     &
     $\bullet$ Provides cohesive and precise X-ray reports in a fully differentiable manner by dividing the report generation system into a classifier, generator, and interpreter. \newline
     $\bullet$ Their conducted experiments revealed that incorporating additional scans besides clinical history can be beneficial in providing higher-quality X-ray reports.
     &
     $\bullet$ Flexibility in processing additional input data, such as clinical documents and extra scans, which also contributes to performance improvement. \newline  
     $\bullet$ The model doesn't provide vital information, such as illness orientation and time-series correlations, which facilitates more reliable reports.\\
     
     \midrule

     \makecell[l]{CGRG \citep{wang2021confidence}} 
     &
     $\bullet$ Presented a Transformer-based method that estimates report uncertainty to develop a more reliable MRG system and facilitate diagnostic decision-making. \newline
     $\bullet$ Introduced the Sentence Matched Adjusted Semantic Similarity (SMAS) to capture vital and relevant features in radiology report generation more effectively. 
     &
     $\bullet$ Assessing visual and textual uncertainties leads to more reliable reports in medical diagnosis. \newline
     $\bullet$ Measuring uncertainties can properly provide correlated confidence between various reports, which is beneficial to aiding radiologists in clinical report generation \citep{liu2022competence}.\\
     
     \midrule

     \makecell[l]{{CXR-RePaiR} \citep{endo2021retrieval}} 
     &
     $\bullet$ {Employs contrastive language image pre-training (CLIP) to generate chest X-ray reports.}\newline
     $\bullet$ {The utilization of representations pairs of X-ray reports aids the process of retrieving unstructured radiology reports written in natural language.}
     &
     $\bullet$ {Accuracy is evaluated by comparing it with the ground truth using a labeler, and performance scores are calculated accordingly.} \newline
     $\bullet$ {Their experiments indicate that despite generating precise diagnostic explanations, it may not consistently employ identical wording as the initial report.}\\
     
     \midrule

     \makecell[l]{{BioViL-T} \citep{bannur2023learning}} 
     &
     $\bullet$ {Provides a hybrid design that aligns X-ray report pairs using temporal information and prior knowledge.}\newline
     $\bullet$ {Combines CNN and transformer encoders to extract image representations and matches them with text representations using a chest X-ray domain-specific transformer model.}\newline
     $\bullet$ {Introduced MS-CXR-T, a benchmark dataset that facilitates the evaluation of vision-language representations in capturing temporal semantics.}
     &
     $\bullet$ {Prior report is more crucial for optimal performance compared to prior image as it outlines the image content and enhances the signal clarity.} \newline
     $\bullet$ {A significant factor in effective language supervision during the pre-training phase is the utilization of contrastive loss.}\\
     
     \midrule
     
     \makecell[l]{{ChatCAD} \citep{wang2023chatcad}}
     &
     $\bullet$ {Integrates MRG with large language models such as ChatGPT to improve diagnosis accuracy} \newline
     $\bullet$ {The LLM summarizes and corrects errors in the generated reports via the concatenation of textual representations from CAD networks.}
     &
     $\bullet$  {The experiments show that model size and complexity have a significant impact on generation accuracy, as larger models generally result in improved F1-scores.} \newline
     $\bullet$ {Due to the reports being less similar to human-written text, the system's BLEU score is lower compared to previous methods.}\\
     
    \bottomrule
    \end{tabular}
    }
\end{table*}

\begingroup
\def\arraystretch{1.2}%
\begin{table*}[!t]
	\fontsize{8}{10}\selectfont
	\centering
	\caption{{Comparison of Transformer-based medical report generation systems in terms of NLG performance metrics. The methods are ordered by BL-4 score, which captures more precise phrase and sentence structure.}}
	\label{tab:performance_generation}
	\resizebox{\textwidth}{!}
	{
		\begin{tabular}{V{4}l|c|c|c|c|c|c|cV{4}}
			\Xhline{3\arrayrulewidth}
			\rowcolor{aliceblue}
			\multicolumn{8}{|c|}{\textbf{IU Chest X-ray \citep{demner2016preparing}}} \\
			\textbf{Method} & \textbf{BL-1} & \textbf{BL-2} & \textbf{BL-3} & \textbf{BL-4} & \textbf{METEOR} & \textbf{ROUGE} & \textbf{CIDEr}\\
			\Xhline{3\arrayrulewidth}
			\textbf{RTMIC} \citep{xiong2019reinforced} & 0.350 & 0.234 & 0.143 & 0.096 & - &  - &  0.323 \\ \hline
               \textbf{CDGPT2}
            \citep{alfarghaly2021automated} & 0.387 & 0.245 & 0.166 & 0.111 & 0.164 & 0.289 & 0.257 \\ \hline
			\textbf{KERP} \citep{li2019knowledge} & 0.482 & 0.325 & 0.226 & 0.162 & - & 0.339 & 0.280 \\ \hline
			\textbf{MDT} \citep{chen2020generating} & 0.470 & 0.304 & 0.219 & 0.165 &  0.187 & 0.371 & - \\ \hline
               \textbf{PPKED}
            \citep{liu2021exploring} & 0.483 & 0.315 & 0.224 & 0.168 & 0.190 & 0.376 & \textbf{0.351} \\ \hline
                \textbf{CMN} 
            \citep{chen2022cross} & 0.475  & 0.309  &  0.222 & 0.170  & 0.191 &  0.375 &  - \\ \hline
			\textbf{AlignTransformer} \citep{you2021alignTransformer} & 0.484 &  0.313 & 0.225  &  0.173 & 0.204 & 0.379  & -  \\ \hline
			\textbf{M$^2$ TR. progressive} \citep{nooralahzadeh2021progressive} & 0.486 & 0.317 & 0.232 & 0.173 & 0.192 & 0.390 & - \\ \hline
			\textbf{CGRG} \citep{wang2021confidence} &  \textbf{0.497} & \textbf{0.357}  &  \textbf{0.279} & \textbf{0.225}  & \textbf{0.217} &  \textbf{0.408} &  - \\ 
			\hlineB{4}

			\rowcolor{aliceblue}
			\multicolumn{8}{|c|}{\textbf{MIMIC-CXR \citep{johnson2019mimic}}} \\
  
			\textbf{Method} & \textbf{BL-1} & \textbf{BL-2} & \textbf{BL-3} & \textbf{BL-4} & \textbf{METEOR} & \textbf{ROUGE} & \textbf{CIDEr}\\
			\Xhline{3\arrayrulewidth}
			\textbf{MDT} \citep{chen2020generating} & 0.353 & 0.218 & 0.145 & 0.103 &  0.142 & 0.277 & - \\ \hline
                \textbf{PPKED} 
            \citep{liu2021exploring} & 0.360 & 0.224 & 0.149 & 0.106 & 0.149 & 0.284 & 0.237 \\ \hline
   			\textbf{CMN} \citep{chen2022cross} & 0.353  & 0.218  &  0.148 & 0.106  & 0.142 &  0.278 &  - \\ \hline
               \textbf{M$^2$ TR. progressive}
            \citep{nooralahzadeh2021progressive} & 0.378 & 0.232 & 0.154 & 0.107 & 0.145 & 0.272 & - \\ \hline
			\textbf{AlignTransformer} \citep{you2021alignTransformer} & 0.378  & 0.235  & 0.156  & 0.112  & 0.158 & 0.283  &  - \\ \hline
               \textbf{CRG} 
            \citep{lovelace2020learning} & 0.415 & 0.272 & 0.193 & 0.146 & 0.159 & 0.318 & - \\ \hline
			\textbf{CGI} \citep{nguyen2021automated} & \textbf{0.495} & \textbf{0.360} & \textbf{0.278} & \textbf{0.224} & \textbf{0.222} & \textbf{0.390} & - \\
			\hlineB{4}
		\end{tabular}
	}
\end{table*}
\endgroup

\subsection{Reinforcement Learning-based Systems}
The first work to implement a Transformer architecture for medical report generation is RTMIC \citep{xiong2019reinforced}. It used the reinforcement learning strategy in training to mitigate the problem of exposure bias prevailing in Seq2Seq models \citep{zhang2020minimize}. In their approach, the original images are fed into a DenseNet \citep{huang2017densely} as the region detector to extract bottom-up visual features. These features are then passed into a visual encoder to generate visual representations from the detected regions, which the captioning detector then utilizes to generate captions for the specified regions. The proposed method was experimented on the IU X-Ray dataset \citep{demner2016preparing} and achieved state-of-the-art results. Integration of RL and Transformers was also applied in surgical instruction generation since the joint understanding of surgical activity along with modeling relations linking visual and textual data is a challenging task. Zhang et al. \citep{zhang2021surgical} employed a Transformer-backboned encoder-decoder architecture and applied the self-critical reinforcement learning \citep{rennie2017self} approach to optimize the CIDEr score \citep{vedantam2015cider} as the reward. Their approach surpasses existing models in performance on the DAISI dataset \citep{rojas2020daisi} with caption evaluation metrics applied to the model. This work's key difference from others is that their model is proposed to generate instructions instead of descriptions.

% \begin{figure}[h]
% 	\centering
% 	\includegraphics[width=0.48\textwidth]{images/RTMIC.jpg}
% 	\caption{Overview of the proposed RTMIC. To extract bottom-up visual features, the original images are passed into a DenseNet region detector. These features are then fed to a Transformer-based visual encoder to provide visual representations that the captioning detector uses to generate captions for the specified regions \citep{xiong2019reinforced}.}
% 	\label{fig:RTMIC}
% \end{figure}

\subsection{Graph-based Systems}

\begin{figure}[h]
	\centering
	\includegraphics[width=0.48\textwidth]{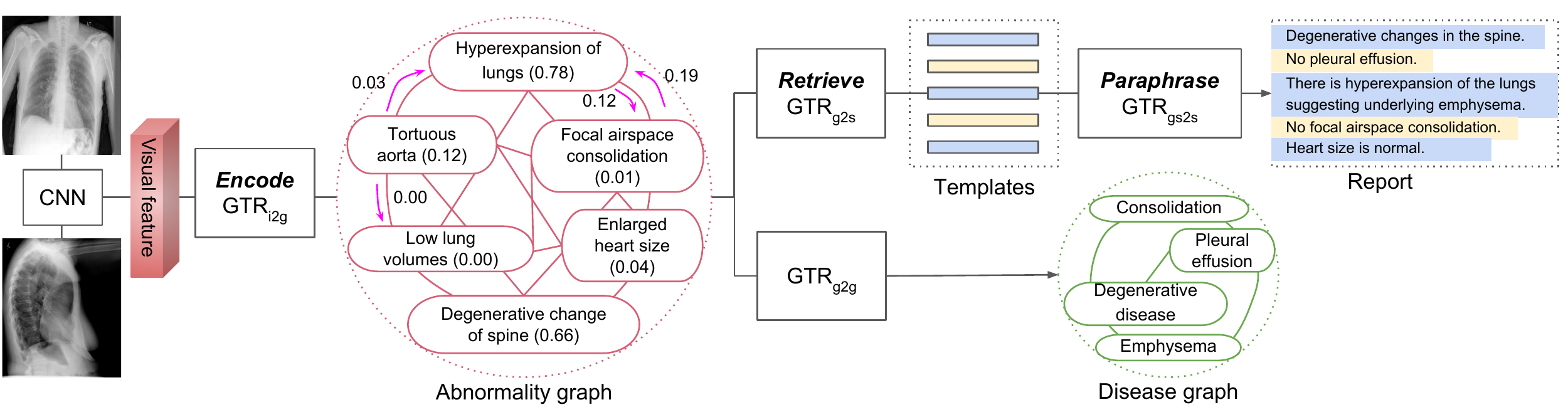}
	\caption{Using an encoder module, KERP creates an abnormality graph from the extracted visual features. Then, a retrieval module retrieves a sequence of templates based on detected abnormalities. Next, the paraphrase module paraphrases the templates' terms into a report \citep{li2019knowledge}.}
	\label{fig:KERP}
\end{figure}

In graph-based medical report generation, Li et al. \citep{li2019knowledge} proposed \textbf{KERP}, a Graph Transformer implementation to generate robust graph structures from visual features that are extracted by a DenseNet \citep{huang2017densely}  backbone. This approach is composed of three modules: Encode, Retrieve and Paraphrase. First, it constructs an abnormality graph by converting the visual features extracted from the medical images via an encoder module. Next, a sequence of templates is retrieved considering the detected abnormalities by utilizing a retrieve module. Subsequently, the terms of the produced templates are paraphrased into a report by employing the paraphrase module. The KERP's workflow is illustrated in \Cref{fig:KERP}.

Additionally, Liu et al. \citep{liu2021exploring} addressed the visual and textual data biases and their consequences in generating radiology reports and proposed the \textbf{PPKED} framework to alleviate these challenges. Their work introduced three modules to perform report generation: (1) Prior Knowledge Explorer (PrKE), which obtains relevant prior information for the input images; (2) Posterior Knowledge Explorer (PoKE), which extracts the posterior information, including the abnormal regions of the medical image; and (3) Multi-domain Knowledge Distiller (MKD), which distills the obtained information from the previous modules to perform the final report generation. PPKED then formulated the problem by employing the presented modules in the following manner: PoKE first extracts the image features corresponding to the relevant disease topics by taking the visual features extracted by ResNet-152 \citep{resnet} from the input image and abnormal topic word embeddings as the input. Next, the PrKE module filters the prior knowledge from the introduced prior working experience (a BERT encoder) and prior medical knowledge component that is relevant to the abnormal regions of the input image by utilizing the output of the PoKE module. Next, the MKD module generates the final medical report by using this obtained information, which is implemented based on the decoder part of the Transformers equipped with Adaptive Distilling Attention.

\subsection{Memory-based Systems}

Concerning the development of systems that rely on a memory mechanism to generate medical reports, Chen et al. \citep{chen2020generating} presented a Memory-Driven Transformer (\textbf{MDT}), a model suitable for the generation of long informative reports and one of the first works on the MIMIC-CXR dataset \citep{johnson2019mimic}. MDT employs a relational memory to exploit characteristics prevailing in reports of similar images, and then the memory is incorporated into the decoder section of the Transformer by implementing a memory-driven conditional layer normalization (MCLN).

\begin{figure}[h]
	\centering
	\includegraphics[width=0.48\textwidth]{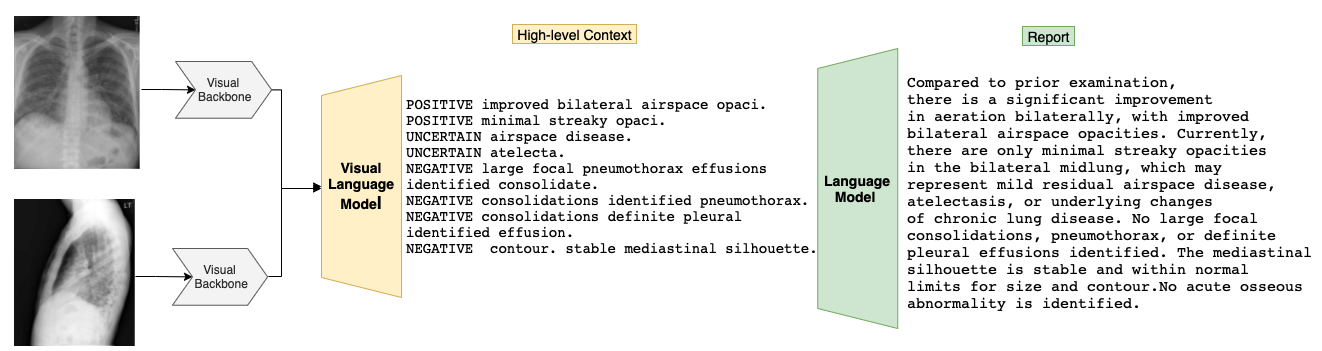}
	\caption{Workflow of the $M^2$ Tr. Progressive framework. The task is accomplished in two stages: First, the Meshed-Memory Transformer (M$^2$ TR.) receives visual features extracted by a DenseNet \citep{huang2017densely} backbone and generates high-level global context. Second, the BART \citep{lewis2019bart} architecture encodes contexts with a bidirectional encoder and decodes them with a left-to-right decoder to produce coherent reports \citep{nooralahzadeh2021progressive}.}
	\label{fig:M2-Tr}
\end{figure}

Likewise, Nooralahzadeh et al. \citep{nooralahzadeh2021progressive} introduced \textbf{M$^2$ TR. progressive}, a report generation approach that utilizes curriculum learning, which is a strategy of training machine learning models by starting with easy samples and gradually increasing the samples' difficulty \citep{wang2021survey}. Instead of directly generating full reports from medical images, their work formulates the problem into two steps: first, the Meshed-Memory Transformer (M$^2$ TR.) \citep{cornia2020meshed}, as a powerful image captioning model, receives the visual features extracted by a DenseNet \citep{huang2017densely} backbone and generates high-level global context. Second, BART \citep{lewis2019bart}, as a Transformer-based architecture, encodes these contexts with a bidirectional encoder and decodes its output using a left-to-right decoder into coherent reports. The overview of the process is depicted in \Cref{fig:M2-Tr}.

Additionally, You et al. \citep{you2021alignTransformer} proposed \textbf{AlignTransformer}, a framework composed of two modules: Align Hierarchical Attention (AHA) and Multi-Grained Transformer (MGT). In their approach, first visual features and disease tags are extracted from the medical image by an image encoder, then they get aligned hierarchically to obtain multi-grained disease-grounded visual features in the AHA module. The obtained grounded features are capable of tackling the data bias problem by promoting a better representation of abnormal sections. Next, these grounded visual features are exploited by an adaptive exploiting attention (AEA) \citep{cornia2020meshed} mechanism in the MGT module for the generation of the medical reports. They also justified their model's efficiency through the manual evaluation of clinical radiologists.

In \textbf{MDT-WCL} \citep{yan2021weakly}, the problem is approached with a weakly supervised contrastive loss, which lends more weight to the reports that are semantically close to the target reports, and a memory-driven Transformer is adopted as the backbone model to store key information in its memory module. To aid the contrastive learning during training, after clustering the reports into groups with the K-Means algorithm, each report is assigned a label corresponding to its cluster, and the semantically closed ones are considered to be in the same cluster.

Although previous approaches have achieved promising results, they lack the ability to generate mappings between images and texts to align visual-textual information and assist medical diagnosis. In order to facilitate visual-textual alignment, the Cross-modal Memory Network (\textbf{CMN}) \citep{chen2022cross} extended encoder-decoder methods by utilizing a shared memory for better alignment of information between images and texts. It uses a pre-trained ResNet \citep{resnet} as the visual extractor to output visual features, then passes them to the cross-modal memory network that utilizes a matrix to store information where each row represents the embedding of information linking images and texts. To access the stored information aligning the modalities, memory querying and responding are implemented in a multi-threaded manner.

\subsection{Other Systems}
\label{sec:other_report}
Other MRG systems focus on solving the problem with different ideas.  Lovelace et al. \citep{lovelace2020learning} proposed a generation framework composed of two stages. In the first stage, a Transformer model is adopted to map the input image features extracted by a DenseNet-121 \citep{huang2017densely} to contextual annotations and learn report generation. In the second stage, a procedure is introduced to differentiably sample a clinical report from the Transformer decoder and obtain observational clinical information from the sample. This differentiability is further employed to fine-tune the model for improving clinical coherence by applying their differentiable CheXpert to the sampled reports. 
Fueled by recent progress in explainable artificial intelligence and the introduction of algorithms that attempt to provide interpretable prediction in DL-based systems, Likewise, in \textbf{CDGPT2} \citep{alfarghaly2021automated}, the medical image is passed into a Chexnet \citep{rajpurkar2017chexnet} to provide localizations of 14 types of diseases from the images as visual features. To implement better semantic features, the model was fine-tuned as a multi-label classification problem to extract manual tags from the IU-Xray dataset \citep{demner2016preparing} by replacing the final layer of the model with a layer containing 105 neurons to produce 105 tags. The vector representation of the tags is then fed into a pre-trained distilGPT2 \citep{radford2019language} as the decoder to generate medical reports.
Moreover, Wang et al. \citep{wang2021confidence} presented a confidence-guided report generation (\textbf{CGRG}) approach to support reliability in report generation by quantifying visual and textual uncertainties. It's comprised of an auto-encoder that reconstructs images, a Transformer encoder that encodes the input visual feature extracted by ResNet-101 \citep{resnet}, and a Transformer decoder for report generation. Visual uncertainty is obtained by the AutoEncoder, which acts as a guide for the visual feature extractor, and textual uncertainty is quantified based on the introduced Sentence Matched Adjusted Semantic Similarity (SMAS) which captures the similarity between the generated reports. These uncertainties are further utilized to aid the model optimization process.

The recent outbreak of COVID-19, one of the deadliest pandemics, has influenced the research community to alleviate the tedious and time-consuming work of producing medical reports. VL-BERT, \citep{su2019vl} as an extension of BERT, \citep{devlin2018bert} can be employed as an intelligent medical report generation system to expedite the diagnosis process. \textbf{Medical-VLBERT} \citep{liu2021medical} introduced VL-BERT to the medical report generation domain. It defines the problem as a two-step procedure: First, it utilizes two distinct VL-BERTs as terminology encoders to produce terminology-related features (textual and visual), and then these features are fed into a shared language decoder to produce medical textbooks and reports. The proposed method takes into account predefined terminology word embeddings that represent medical domain knowledge. These embeddings are paired distinctly with two other embeddings as an input to the encoders: textbook embeddings, which are generated by employing a lookup table, and spatial feature embeddings (termed "visual context") that are extracted from medical images by implementing DenseNet-121 \citep{huang2017densely}. The encoders then integrate this pairwise information separately to produce textual and visual terminological features. Subsequently, a shared language decoder is trained by utilizing an alternate approach to properly exchange the knowledge captured by the encoders. 

Furthermore, in the work of Nguyen et al. \citep{nguyen2021automated}, a classification, generation, and interpretation framework (\textbf{CGI}) is proposed to address clinical accuracy. Each term of the framework's name represents a different module to perform the task. The classification module learns how to discover diseases and generate their embeddings, which consist of an image and text encoder to extract the global visual features from medical images and obtain text-summarized embeddings from clinical documents. The generation module is a Transformer model that takes the disease embeddings as input and generates medical reports from them. The interpretation module then takes these reports for evaluation and fine-tuning.

{Deep learning methods in medical report analysis require extensive amounts of annotated training data, which requires a significant amount of time and effort. To this end, self-supervised report generation approaches use a large amount of unlabeled text data without explicit supervision. For instance, \textbf{CXR-RePaiR} \citep{endo2021retrieval} utilizes self-supervised contrastive language image pre-training. It encodes reports with the CheXbert transformer network\citep{smit2020chexbert}, and leverages the X-ray report pair representations to retrieve unstructured radiology reports with a contrastive scoring function trained on chest X-ray-report pairs to rank the similarity between a test dataset of X-rays and a large corpus of reports. Likewise, \textbf{BioViL-T} \citep{bannur2023learning} is a hybrid design that improves the alignment of X-ray report pairs by incorporating temporal information and prior knowledge to support complementary self-supervision and exploit the strengths of each modality. It compares prior images with given reports to exploit temporal correlations and enhance representation learning. BioViL-T uses a multi-image encoder to handle the absence of prior images and spatial misalignment, benefiting both image and text models. Image representations are extracted utilizing a hybrid CNN and transformer encoder, which are then matched with corresponding text representations obtained with CXR-BERT \citep{ramesh2022improving} by training with contrastive objectives.}

%Likewise, Meng et al. \citep{meng2020self} provided the contextual representation via pre-training a BERT \citep{devlin2018bert} on vast samples of unlabeled radiology reports, then incorporated the learned representation into a text classifier to predict if a radiology report requires immediate communication to referring physicians based on its impression section.
 
{
Recently, large language models (LLMs), which are AI systems that learn the patterns of human language from massive amounts of text data, have revolutionized human-like text generation and natural language understanding. Consequently, they hold great potential in clinical applications, such as understanding medical texts and report generation \citep{zhou2023skingpt,ma2023impressiongpt}. One of the successful large language models based on the GPT-3.5 architecture is ChatGPT \citep{OpenAI2023ChatGPT}, with extensive knowledge and abilities to perform a wide range of tasks such as question answering, text summarization, and generating textual instructions. Wang et al. \citep{wang2023chatcad} presented a scheme called \textbf{ChatCAD} that integrates LLMs such as GPT-3 \citep{brown2020language} and ChatGPT with Computer-Aided Diagnosis (CAD) models to perform medical report generation. CAD systems are computer programs that assist doctors in interpreting medical images, such as X-rays or MRI scans. To generate reports via ChatCAD, the medical image is passed to a segmentation, image classification, and a report generation network. Since the outputs of segmentation and classification networks are a mask and a vector, respectively, they are transformed into textual format to be understandable by LLMs. These textual representations are then concatenated and presented to the LLM, which then summarizes the results from all the CAD networks and exploits them to correct  the errors in the generated report. Their experiments indicate that LLMs can be more effective than conventional CAD systems in enhancing medical report quality.}

\begin{tcolorbox}[breakable ,colback={aliceblue},title={\subsection{Discussion and Conclusion}},colbacktitle=aliceblue,coltitle=black , left=2pt , right =2pt]    

This section offers a systematic review of the Transformer architectures configured for medical report generation. Compared to previous sections that reviewed ViT-based frameworks to tackle different medical tasks and problems, this section focuses mostly on using standard Transformers as the core of a medical report generation system. A common theme prevailing in these systems is to solve the problem with an encoder-decoder architecture supported by a CNN-based visual backbone. 
As mentioned in previous sections, the self-attention mechanism undermines the representation of low-level details. On the other hand, since medical reports consist of long and multiple sentences, Transformers are of great significance to model long-term dependencies, which assists clinically accurate report generation \citep{lin2022survey,nguyen2021automated}. To exploit the power of both CNNs and Transformers simultaneously, state-of-the-art MRG systems usually embed CNNs along with Transformers in their frameworks \citep{xiong2019reinforced,alfarghaly2021automated,wang2021confidence}. We have provided information in \Cref{tab:Generation} on the reviewed report generation methods concerning their architectural type, modality, organ, pre-trained strategy, datasets, metrics, and year. \Cref{tab:Generation highlight} contains summarized information about the methodologies, including their contributions and highlights. In addition, it should be noted that several survey publications have been published in this field of medicine \citep{monshi2020deep,messina2022survey,pavlopoulos2022diagnostic}, and the most recent one provided a technical overview of Transformer-based clinical report generation \citep{shamshad2022Transformers}. We approach our review differently by distinguishing the proposed methods based on the mechanism they used to support the prevailing concerns such as long and coherent text generation, reliability, and visual-textual biases. 

The ultimate goal of these frameworks is to increase clinical accuracy to expedite the diagnosis process and reduce the workloads in radiology professions \citep{lovelace2020learning,nguyen2021automated}. Numerous works have attempted to facilitate diagnostic decision-making by aligning correlated sections of medical image and textual report that provide valuable information for detecting abnormalities \citep{you2021alignTransformer,chen2022cross}. Also, multiple studies emphasized the importance of universal knowledge, and designed a system to incorporate prior information for detecting disease \citep{li2019knowledge,chen2020generating}. Some research effort was also put into better representation learning by contrasting normal and abnormal samples against each other in representation space by utilizing a contrastive loss as the objective \citep{yan2021weakly}. One recent work was inspired by curriculum learning to imitate the order of the human learning process \citep{nooralahzadeh2021progressive}. 

{As for the cumbersomeness of accessing annotated textual training data, research in MRG also focused on consolidating self-supervised learning and transformer architectures for report generation, which are discussed in \Cref{sec:other_report} through the analysis of two specific works \cite{meng2020self,endo2021retrieval}.} {ChatCAD \citep{wang2023chatcad} was also introduced recently to integrate MRG with large language models like ChatGPT to improve diagnosis accuracy, but it achieves a lower BLEU score as it generates reports that are less similar to human-written text.}
    
{The \Cref{tab:performance_generation} includes performance values of BLEU, METEOR, ROUGE, and CIDEr for each method. Among the methods evaluated on the IU Chest X-ray, KERP and CGRG perform the best across most metrics, while RTMIC and CDGPT2 are the weakest. MDT has the highest METEOR score, CGRG achieves the highest ROUGE score, and PPKED demonstrates the highest CIDEr score. In contrast, for the MIMIC-CXR dataset, CRG and CGI demonstrate the best performance, while MDT shows the weakest overall performance.}

Additionally, \Cref{tab:performance_generation} compares the performance of the reviewed methods Natural Language Generation (NLG) metrics. The methods are presented in separate sections based on the dataset they were evaluated on, with one section dedicated to the IU Chest X-ray \citep{demner2016preparing}, and the other to the MIMIC-CXR \citep{johnson2019mimic}. 

Overall, we believe that MRG systems need more research and progression to be robustly incorporated in a practical setting. 

\end{tcolorbox}    

\section{Open Challenges and Future Perspectives} \label{sec:challenges}

So far, we discussed the application of Transformers (especially vision Transformers) and reviewed state-of-the-art models in medical image analysis. Even though their effectiveness is exemplified in previous sections by delicately presenting their ideas and analyzing the significant aspects that were addressed in their proposed methods, there is still room for improvement in many areas to devise a more practical and medically accurate system by leveraging Transformers. Consequently, we discuss the challenges and future directions hoping to help researchers gain insight into the limitations and develop more convenient automatic medical systems based on Transformers.

\subsection{Explainability}

Fueled by recent progress in XAI (explainable artificial intelligence) and the introduction of algorithms that attempt to provide interpretable prediction in DL-based systems, researchers are putting effort into incorporating XAI methods into constructing Transformer-based models to promote a more reliable and understandable system in different areas, including medical analysis \citep{alicioglu2022survey,singh2020explainable}. Existing approaches usually highlight important regions of the medical image that contribute to the model prediction by employing attention maps \citep{hou2021ratchet,mondal2021xvitcos}. Furthermore, Vision Transformers (ViTs) have the ability to provide attention maps that indicate the relevant correlations between the regions of the input and the prediction. However, the challenge of numerical instabilities in using propagation-based XAI methods such as LRP \citep{binder2016layer} and the vagueness of the attention maps, which leads to inaccurate token associations \citep{chefer2021Transformer,kim2022vit}, makes interpretable ViTs an open research opportunity in computer vision, especially in medical image analysis. We believe that including interpretable vision Transformers, such as ViT-NeT \citep{kim2022vit}, in various medical applications can promote user-friendly predictions and facilitate decision-making in the diagnosis of medical conditions, and is a promising direction in medical research problems.

\subsection{Richer Feature Representation}
An effective and suitable representation space is substantially influential in building medical analysis systems. Transformers have demonstrated their efficiency in obtaining global information and capturing long-term dependencies in many areas, such as Natural Language Processing (NLP), Computer Vision, and Speech Recognition \citep{lin2021survey}, and CNNs have proven to be effective in extracting local context from visual data \citep{li2021survey}. However, this locality usually enables these networks to capture rich local texture representation and lacks model global dependency. As a result, many approaches stack Transformers along with CNNs to leverage both local and global information simultaneously in clinical applications (e.g., medical report generation) \citep{you2021alignTransformer,lovelace2020learning,tanzi2022vision}. Recent studies stated that the single-scale representation of ViTs hinders improvement in dense prediction tasks, so a multi-scaled feature representation is implemented which achieves better performance in computer vision tasks, including image classification, object detection, and image segmentation \citep{gu2022multi,lee2022mpvit}. Generalizing this idea to medical applications of ViTs to facilitate devising a clinically suitable system can be considered as future work.

\subsection{Video-based analysis}

There has been an increasing interest in the vision community in extending ViT architectures to video recognition tasks. Recently, a handful of papers have integrated standard Transformers with their models in AI-assisted dynamic clinical tasks \citep{reynaud2021ultrasound,long2021dssr,czempiel2021opera,zhao2022trasetr}. However, the scarcity of the proposed approaches puts video-based medical analysis in an infancy stage and open for future investigations. Another potential research direction is to explore the power of video vision Transformer variants, such as Video Swin Transformer \citep{liu2022video}, in clinical video understanding and to facilitate automatic robotic surgery.

\subsection{High Computational Complexity}
The robustness of Transformer models in layouts that implement large numbers of parameters is one of their strengths. While this is a beneficial trait that makes it possible to train models of enormous scale, it leads to the requirement of large resources for training and inferencing \citep{khan2022Transformers}. Particularly disadvantageous to medical image analysis is that expanding the use of ViTs for pretraining in new tasks and datasets comes with substantial expenses and burdens. Additionally, gathering medical samples can be difficult and the dataset scale is often limited. For instance, according to empirical studies in \citep{dosovitskiy2020image}, pretraining a ViT-L/16 model on the large-scale dataset of ImageNet takes approximately 30 days employing a standard cloud TPUv3 with 8 cores. As a result, a notable number of papers utilized the pre-trained weights of ViT models to exploit the transfer learning strategy to alleviate training load \citep{shome2021covid,chen2021transunet,gheflati2022vision}, but in some cases, such as dealing with volumetric medical images, where transfer learning doesn't demonstrate any improvements \citep{hatamizadeh2021unetr,chen2021vit}, the pretraining process is necessary to capture domain-specific features for generalization and better performance. 

{In addition, the standard self-attention mechanism suffers from a significant drawback: its computational complexity grows quadratically with the number of tokens ($n$), making it impractical for tasks where input images can be millions of pixels in size or when dealing with volumetric data. To address this issue, several approaches have been proposed to reduce the computational burden of self-attention. They either reduce the number of tokens by windowing \citep{ding2022davit,liu2021swin}, shift the calculation to the channel dimension \citep{ali2021xcit,maaz2022edgenext}, or change the order of multiplying query, key, and value \citep{shen2021efficient}. In the context of 3D biomedical image segmentation, Zhang et al. \citep{zhang2022dynamic} introduced a dynamic linear Transformer algorithm that achieves linear complexity by focusing computations only on the region of interest (ROI). This significantly reduces the overall computational requirements. Shen et al. \citep{shen2021efficient} proposed an efficient attention mechanism that avoids the pairwise similarity computations of dot-product attention. Instead, it normalizes the keys and queries first, performs multiplication between the keys and values, and then multiplies the resulting global context vectors with the queries. This approach reduces the computational complexity of self-attention from $O(dn^2)$ to $O(d^2n)$, where $d$ is the embedding dimension. XCiT, \citep{ali2021xcit} another approach, addresses the complexity challenge by introducing cross-covariance attention, which has a complexity of $O(\frac{d^2n}{h})$, where $h$ is the number of attention heads. EdgeNext, an architecture proposed by Maaz et al. \citep{maaz2022edgenext} for edge devices, optimizes the number of Multiplication-Addition (MAdd) operations required. It employs a variation of the attention mechanism used in XCiT called split depth-wise transpose attention (SDTA). Huang et al. \citep{huang2021missformer} introduce efficient self-attention (ESA), specifically for the spatial reduction in 3D data. ESA reduces the number of tokens by a spatial reduction ratio $R$ while expanding the channel dimension by the same ratio. As a result, the complexity of ESA is reduced to $O(\frac{n^2}{R})$.}

{Despite progress in making self-attention more efficient for computer vision tasks, further work is needed to develop efficient Transformer models tailored to the medical domain, particularly for handling challenging 3D data or large-scale images (e.g., WSI). Moreover, the ultimate goal is to deploy these models on low-cost edge devices or in resource-constrained environments. Therefore, the focus should be on designing Transformer systems that strike a balance between computational complexity and clinical accuracy while ensuring robustness. This area of research holds great promise and should be pursued further.}

\subsection{Transformer-based Registration}
As reviewed in \Cref{sec:registration}, the idea of employing Transformers to support efficient medical image registration has become popular in recent years. The ability of the self-attention mechanism assists the learning of long-term visual correlations since their unlimited receptive field promotes a more accurate understanding of the spatial relationship between moving and fixed images \citep{chen2021vit,chen2022transmorph}. However, registration systems composed of Transformer architectures are still in their infancy and require more research effort to be put into them.

\subsection{Data-Driven Predictions}
With supervised learning as a popular fashion in building intelligent systems, the model learns features based on the provided annotations that are suitable to accomplish a specific task, which hinders generalizability. In other words, supervised learning modifies the bias-variance trade-off in favor of the strong inductive biases that lead to making assumptions as a means to aid the model in learning a particular task quicker and with higher sample efficiency. However, these hard assumptions sacrifice adaptability to other settings and unseen datasets, and the model learns to accomplish its task without having an innate understanding of the data. To tackle this issue, unsupervised regimes enable the algorithms to act as general descriptors and capture features that will assist them in performing efficiently in a wide range of tasks. Similarly, in medical image analysis, adopting Transformer networks with unsupervised learning algorithms promotes robustness and generalizability to other datasets and tasks.

\subsection{Medical Software Ecosystems}
A future direction for advancing in the automatic medical analysis is to provide an open-source environment that contains libraries suitable for solving multiple medical tasks and challenges with Transformer architectures. Developers can further contribute to the ecosystem by updating and adding additional tasks, bringing novelty, and proposing ideas to enhance performance and accuracy \citep{kazerouni2022diffusion}. Companies and organizations can support the system by preparing the necessary computational resources and hardware requirements. Sample of software prototypes in this direction are nnU-Net \citep{isensee2021nnu}, Ivadomed \citep{gros2020ivadomed}, and preliminary works such as \citep{azad2022medicalunet}, which provides an end-to-end pipeline for implementing deep models on medical data.

\section{Discussion and Conclusion} \label{sec:discussion}
In this paper, we presented a comprehensive encyclopedic review of the applications of Transformers in medical imaging. First, we provided preliminary information regarding the Transformer structures and the idea behind the self-attention mechanism in the introduction and background sections. Starting from \Cref{sec:classification}, we reviewed the literature on Transformer architecture in diverse medical imaging tasks, namely, classification, segmentation, detection, reconstruction, synthesis, registration, and clinical report generation. For each application, we provided a taxonomy and high-level abstraction of the core techniques employed in these models along with the SOTA approaches. We also provided comparison tables to highlight the pros and cons, network parameters, type of imaging modality they are considering, organ, and the metrics they are using. Finally, we outlined possible avenues for future research directions.

\noindent\textbf{Acknowledgments}
This work was funded by the German Research Foundation (Deutsche Forschungsgemeinschaft, DFG) under project number 191948804. We thank Johannes Stegmaier for his contribution to the proofreading of this document.

%%Harvard
% \bibliographystyle{model2-names.bst}
% \biboptions{authoryear}
\bibliographystyle{elsarticle-num.bst}
% \biboptions{number,nonatbib}
\bibliography{refs}

\end{document}